%% file: apc_sk.tex
\documentclass[journal]{IEEEtran}
\usepackage{graphicx,multirow,amsmath,soul,mathtools,amssymb,cancel,float,mathrsfs,amsfonts,lineno}
\usepackage[dvipsnames,usenames]{color}
\usepackage{hyperref}
\usepackage[flushleft]{threeparttable}
\usepackage[loose]{subfigure}

\usepackage{tikz}
\usepackage{bm,times}
\usetikzlibrary{shapes.geometric, arrows} 
\usetikzlibrary{shapes,arrows,positioning, calc, matrix}
\usepackage{verbatim}
\usepackage{varwidth}
\usepackage{fancyhdr}
\usepackage{datetime}
\usepackage{enumerate}
\usepackage{hyperref}
\usepackage{url}
\usepackage{breakurl}
\usepackage{algpseudocode,algorithm,algorithmicx}
\usepackage{booktabs}
\usepackage{float}
\usepackage[font=scriptsize]{caption}
 \usepackage{colortbl, tabularx}

\graphicspath{ {images/} {./fig/} {./fig/rcnn_fig/}}

\input{./fig/FlowchartStyle.tex}

\pgfdeclarelayer{background}
\pgfdeclarelayer{foreground}
\pgfsetlayers{background,main,foreground}

\usepackage{array}
\newcolumntype{L}[1]{>{\raggedright\let\newline\\\arraybackslash\hspace{0pt}}m{#1}}
\newcolumntype{C}[1]{>{\centering\let\newline\\\arraybackslash\hspace{0pt}}m{#1}}
\newcolumntype{R}[1]{>{\raggedleft\let\newline\\\arraybackslash\hspace{0pt}}m{#1}}

\begin{document}

\title{Design and Development of an automated Robotic Pick \& Stow System for an e-Commerce Warehouse}
%
%
%

\author{ Swagat Kumar$^{\dagger\ddagger}$, Anima Majumder$^*$, Samrat Dutta$^*$, Sharath
  Jotawar$^\dagger$, Ashish Kumar$^*$, Manish Soni$^\dagger$, Venkat Raju$^\dagger$, Olyvia
Kundu$^\dagger$ , Ehtesham Hassan$^\dagger$, 
Laxmidhar Behera$^*$, K. S. Venkatesh$^*$ and Rajesh Sinha$^\dagger$
\thanks{$^*$The authors are associated with Indian Institute of Technology Kanpur, Uttar Pradesh, India.}
\thanks{$^\dagger$The authors are associated with Tata Consultancy services, New Delhi, India.}
\thanks{Email: \texttt{\{animam,samratd,lbehera,venkats\}@iitk.ac.in}; 
\texttt{\{swagat.kumar,sharath.jotawar,manish.soni,venkat.raju,
  olyvia.kundu, ehtesham.hassan, rajesh.sinha\}@tcs.com.}}
  \thanks{$\ddagger$The corresponding author for this paper}
}

\maketitle


\begin{abstract}
In this paper, we provide details of a robotic system that can
automate the task of picking and stowing objects from and to a rack in
an e-commerce fulfillment warehouse.  The system
primarily comprises of four main modules: (1) Perception module
responsible for recognizing query objects and localizing them in the
3-dimensional robot workspace; (2) Planning module generates necessary
paths that the robot end-effector has to take for reaching the objects
in the rack or in the tote; (3) Calibration module that defines the
physical workspace for the robot visible through the on-board vision
system; and (4) Gripping and suction system for picking and stowing
different kinds of objects. The perception module uses a faster
region-based Convolutional Neural Network (R-CNN) to recognize
objects. We designed a novel two finger gripper that incorporates
pneumatic valve based suction effect to enhance its ability to pick
different kinds of objects. The system was developed by IITK-TCS team
for participation in the Amazon Picking Challenge 2016 event. The team
secured a fifth place in the stowing task in the event. The purpose of
this article is to share our experiences with students and practicing
engineers and enable them to build similar systems. The overall
efficacy of the system is demonstrated through several simulation as
well as real-world experiments with actual robots. 

\end{abstract}


\section{Introduction} \label{intro}

\begin{figure}[!t]
  \centering
  \includegraphics[scale=0.3]{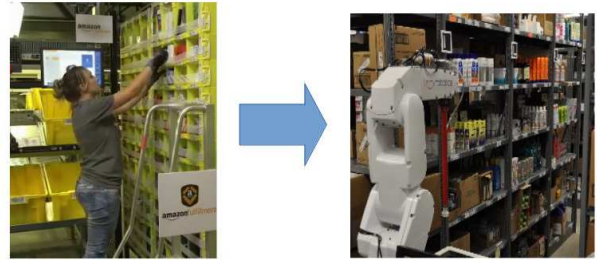}
  \caption{Amazon plans to employ robots to pick and stow things from
  racks in retail warehouses.}
  \label{fig:hr_pick}
\end{figure}

Warehouses are important links in the supply chain
  between the manufacturers and the end consumers. People have been
  increasingly adopting automation to increase the efficiency of
  managing and moving goods through warehouses
  \cite{luo2001integrated}. This is becoming even more important for
  e-commerce industries like Amazon \cite{amazon} that ships millions
  of items to its customers worldwide through its network of
  fulfillment centres. These fulfillment centres are sometimes are big
  as nine football pitches \cite{o2013amazon} employing thousands of
  people for managing inventories. While these warehouses employ IoT
  and IT infrastracture \cite{reaidy2015bottom, ding2013study} to keep
  track of goods moving in and out of the facility, it still requires
  the staffs to travel several miles each day in order to pick or
  stow products from or to different racks \cite{o2013amazon}.  The
  problem related to the goods movement was  solved by the
  introduction of mobile platforms like KIVA systems
  \cite{wurman2008coordinating}  that could carry these racks
  autonomously to human `pickers' who would then, pick things from
  these racks while standing at one place. These mobile platforms
  could then be programmed \cite{mathew2015planning}
  \cite{digani2015ensemble} to follow desired paths demarcated using
  visual \cite{truc2016navigation} or magnetic markers
  \cite{xu2010magnetic}. However, it still needs people to pick or
  stow items from or to these racks. Amazon hires several
  hundred people during holiday seasons, like Christmas or New Year,
  to meet this increased order demands.  Given the slimmer operating
  margins, e-commerce industries can greatly benefit from deploying
  robotic `pickers' that can replace these humans. This transition is
  illustrated in Figure \ref{fig:hr_pick}.  The left hand side of this
  figures shows the current state of affairs where a human picks or
  stows items from or to the racks, which are brought to the station
  by mobile platforms.  The right hand side of this figure shows the
  future where robots will be able to do this task autonomously. In
  the later case, it won't be required to bring the racks to a picking
  station anymore if the robot arm is itself mounted on a mobile
  platform \cite{muis2005eye}.   However, building such robots that
  can pick / stow items from / to these racks with the accuracy,
  dexterity and agility of a human picker is still far too
  challenging. In order to spur the advancement of research and
  development in this direction, Amazon organizes annual competition
  known as `Amazon Picking Challenge' \cite{wurman2016amazon} every
  year since 2015. In this competition, the participants are presented
  with a simplified version of the problem where they are required to
  design robots that can pick and stow items autonomously from or to a
  given rack.

The picking task involves moving items from a rack and place them
into a tote while the stowing task involves moving items from the tote
to the rack. The objects to be picked or stowed are general household
items that varies greatly in size, shape, appearance, hardness and
weight. Since there is no constraint on how the products are organized
on the rack or the tote, there are several possibilities of
configuration one might encounter during the actual operation.  This
uncertainty that may arise due to factors like occlusion, variation in
illumination, pose, viewing angle etc. makes the problem of autonomous
picking and stowing extremely challenging.

This paper provides the details of the proposed system that can
accomplish this task and share our experiences of participating in the
APC 2016 event held in Leipzig, Germany. The proposed system primarily
consists of three main modules: (1) Calibration, (2) Perception, (3)
Motion Planning as shown in Figure \ref{fig:scheme}. Some of the
distinctive features of our implementation are as follows. In contrast
to other participants, we took a minimalistic approach making use of
minimum number of sensors necessary to accomplish the task. These
sensors were mounted on the robot itself and the operation did not
require putting any sensor in the environment.  Our motivation has
been to develop robotic systems that can work in any environment
without requiring any modification to the existing infrastructure. The
second distinctive feature of our approach was our lightweight object
recognition system that could run on a moderate GPU laptop. The object
recognition system uses a trained Faster RCNN based deep network
\cite{ren2015faster} to recognize objects in an image. Deep network
requires large number of training examples for higher recognition
accuracy. The training examples are generated and annotated through a
laborious manual process requiring considerable amount of time and
effort. Moreover, larger training set requires larger time for
training the network for a given GPU configuration.  In a deviation to
the usual trend, a hybrid method is proposed to reduce the number of
training examples required for obtaining a given detection accuracy.
Higher detection accuracy corresponds to tighter bounding box around
the target object while lesser training examples will result in bigger
bounding box around the target object. The exact location for making
contact with the object within this bounding box is computed using an
algorithm that uses surface normals and depth curvatures to segment
the target object from its background and finds suitable graspable
affordance to facilitate its picking. In other words, the limitations
of having smaller training set is overcome by an additional step which
uses depth information to localize the targets within the bigger
bounding box obtained from the RCNN network. This is another step
which helps us in maintaining our minimalistic approach towards
solving the problem.  This approach allowed us to achieve accuracy of
about $90\pm5\%$ in object recognition by training the RCNN network
using only 5000 images as opposed to other participants who used more
than 20,000 images and high end GPU machines. The third distinctive
feature of this paper is the details that has been put in to explain
the system integration process which, we believe, would be useful for
students, researchers and practicing engineers in reproducing and replicating
similar systems for other applications.

In short, the major contributions made in this paper could be
summarized as follows: (1) a novel hybrid perception method is
proposed where depth information is used to compensate for the lesser
size of dataset required for a training a deep network based object
recognition system.  (2) The proposed system uses minimal resources to
accomplish the complete task. It essentially uses only one Kinect
sensor in a Eye-in-hand configuration for all perception task in
contrast to others \cite{hernandez2016team} who used expensive camera
like Ensenso \cite{ensenso}.  (3) An innovative gripper design is
provided that combines both suction as well as gripping action. (4) A
detailed description of the system implementation is provided which
will be useful for students, researchers and practicing engineers.
The performance of the proposed system is demonstrated through
rigorous simulation and experiments with actual systems. The current
system can achieve a pick rate of approximately 2-3 objects per
minute.

The rest of this paper is organized as follows. An overview of various
related work is provided in the next section. A formal definition of
the problem to be solved in this paper is described in Section
\ref{sec:pd}. The system architecture and schematic of the system that
is developed to solve this problem is described in Section
\ref{sec:sysarch}. The details of methods for each of these modules
are described in detail in Section \ref{sec:methods}. The system
performance as well as the results of various experiments are provided
in Section \ref{sec:expt}. The conclusion and direction for future
work is provided in Section \ref{sec:conc}.


\section{Problem definition} \label{sec:pd}

As described before the objective of this work is to replace humans
for picking and stowing tasks in an e-commerce warehouse as shown in
Figure \ref{fig:hr_pick}. The schematic block diagram of our proposed
system which can accomplish this objective is shown in Figure
\ref{fig:scheme}. The list of items to be picked or stowed is provided
in the form of a JSON file. The system comprises of a rack, a tote and
a 6 DOF robotic arm  with appropriate vision system and end-effector
for picking items from the rack or the tote.

The task is to develop a robotic system that can automatically pick
items from a rack and put them in a tote and vice-versa. The reverse
task is called the stowing task. The information about the rack as
well as the objects to be picked or stowed are known apriori. The rack
specified by APC 2016 guidelines had 12 bins arranged in a 4$\times$3
grid. There were about 40 objects in total which were provided to each
of the participating teams.

In the pick task, the robot is expected to move items from the shelves
of a rack to a tote. A subset of the 40 objects  (known apriori) were
randomly distributed in these 12 bins.  Each bin would contain minimum
of one and maximum of 10 items and the list of items at individual
bins are known. Multiple copies of the same item could be placed in
the same bin or in different bin.  The bins may contain items which
are partially occluded or in contact with other items or the wall of
the bin. In other words, there is no constraint on how the objects
would be placed in these bins.  A \textit{json} file is given prior to
start the task which contain the details about which item is in which
bin and what items are to be picked up from these bins.  The task is
to pick 12 specified items, only one from each of the bin in any
sequence and put it into the tote. 
 
In the stow task, the robot is supposed to move items from a tote and
place them into bins on the shelf. The tote contained 12 different
items, which are placed in such a way that some items are fully
occluded or partially occluded by other items.  The rest of the items
are placed in the bins  so that each bin can have minimum one item and
maximum 10 items. The task is to stow 12 items from the tote one by
one in any sequence and put them into any bin. 

The challenge was to get the robot to pick or stow autonomously as
many items as it could within 15 minutes.  Different objects carried
different reward points if they were picked or stowed successfully.
Penalty was imposed on making mistakes such as picking or stowing
wrong items, dropping them midway or damaging the items or the rack
during robot operation etc.


\section{System Architecture} \label{sec:sysarch}

\begin{figure}[!t]
  \centering
  \includegraphics[scale=0.3]{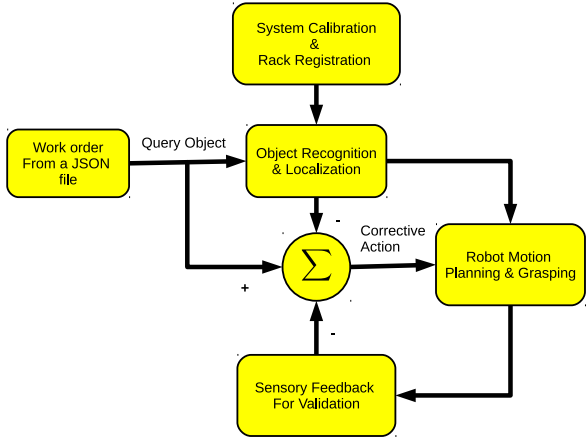}
  \caption{Schematic showing the important blocks of the system}
  \label{fig:scheme}
\end{figure}

The schematic block diagram of the complete system is shown in Figure
\ref{fig:scheme}. The system reads the query items one by one from a
JSON file. The JSON file also provides the bin location for each of
these queried items. The robot has to pick these items from their
respective bins. Since there could be several other objects in the
bin, robot has to identify and localize the target object inside these
bins. The system consists of the following three main components: (1)
Calibration module, (2) Perception module, and (3) Motion planning
module. The calibration module is used for defining the workspace of
the robot with respect to the rack and the tote. It computes the
necessary transformations needed for converting image features into
physical real world coordinates.  The perception module is responsible
for recognizing queried items, localize them in the bin and find the
respective physical coordinates which can be used by robot for motion
planning.  The motion planning module generates necessary robot
configuration trajectories and motion to reach the object, pick it
using suction or gripping action and move it to a tote. This module
makes use of several sensors to detect the completion of the task.
Once the task is completed, the system moves to the next item in the
JSON query list.  

\begin{figure*}[htbp]
  \centering
  \scalebox{0.3}{\input{./fig/apc_ros_arch2.pstex_t}}
  \caption{ROS Architecture for Pick and Place application. Various
  nodes and topics run on three different computers (PC1, PC2 and
PC3). The solid arrows indicate the topics which are either published
or subscribed by a node. The dotted line represent service calls. }
  \label{fig:apc_ros_arch} 
\end{figure*}
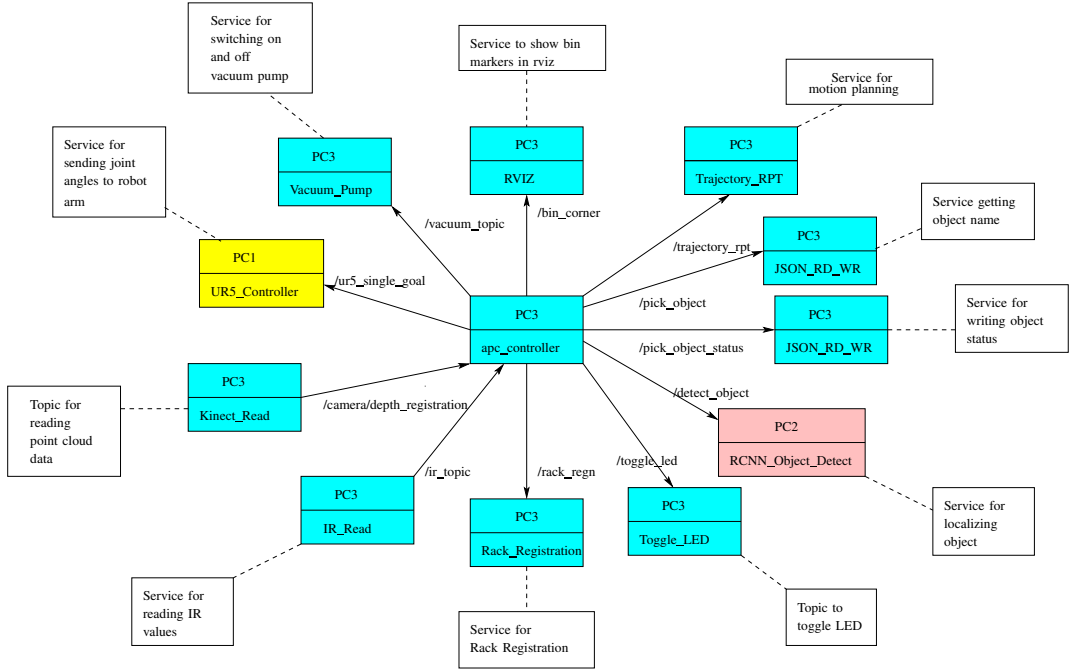

The system is implemented using Robot Operating
  System (ROS) framework \cite{quigley2009ros}. The operation of the
  complete system is divided into different modules each performing a
  specific task. Each of these modules are made available as a
  \emph{node} which are the basic computation units in a ROS
  environment. These nodes communicate with each other using
  \emph{topics}, \emph{services} and \emph{parameter servers}. Readers
  are advised to go through basic ROS tutorials available
  online\footnote{\burl{http://wiki.ros.org/ROS/Tutorials}} in order
  to understand these concepts before proceeding further. Topics are
  unidirectional streaming communication channels where data is
  continuously published by the generating node and other nodes can
  access this data by subscribing to this topic. In case, nodes are
  required to receive a response from other nodes, it can be done
  through \emph{services}. The complete set of modules which are
  required for building the complete system is shown in Figure
  \ref{fig:apc_ros_arch}. These modules or \emph{nodes} run on
  different computing machines which are connected to a common LAN.
  The dotted line indicate service calls which execute a particular
  task on demand basis. All these modules are controlled by a central
node named ``apc\_controller''.  Simulation environment and RVIZ
visualizer is also part of this system and is made available as an
independent node.  

\section{The Methods} \label{sec:methods}

In this section, we provide the details of underlying methods for each
of the modules described in the previous section.

\subsection{System Calibration} \label{sec:calib}

\begin{figure}[!h]
  \centering
  \includegraphics[height=5cm, width = 4cm]{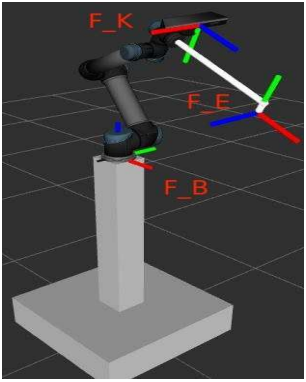}
  \caption{Cartesian Coordinate Frames for the robotic system. The
    transformation matrix between the robot base frame $F_b$ and
    the end-effector frame $F_e$ is known through robot forward kinematics.
    The transformation matrix between the Kinect frame $F_k$ and the end
  effector frame $F_e$ is estimated in the calibration step. }
  \label{fig:rob_frames}
\end{figure}

The calibration step is needed to define the
  workspace of the robot as seen through a camera so that the robot
  can reach any visible location in the workspace. The calibration is
  an important step in all robotic systems that use camera as a sensor
  to perceive the environment. The purpose is to the transform the
  points visible in the camera plane to the physical Cartesian plane.
  A number of methods have been devised for calibration the normal RGB
  cameras \cite{tsai1987versatile} \cite{zhang2000flexible} which try
  to estimate the camera parameters so that the required
  transformation from pixel coordinates to 3D Cartesian coordinates
  could be carried out. The depth estimation has been simplified with
  the advent of RGBD camera such as Kinect \cite{zhang2012microsoft}
  \cite{andersen2015kinect} which provides depth value for each RGB
pixel of the image frame.

In this work, a Kinect RGBD camera is used in
  eye-in-hand configuration to detect as well as find the Cartesian
  coordinate of a query object with respect to its frame $F_K$. These
  coordinates are required to be transformed into robot base frame
  coordinate $F_R$ so that it can be reached by the robot.  In order
  to do this, it is necessary to know the transformation between the
  Kinect camera frame $F_K$ and the robot end-effector frame $F_E$.
  The corresponding frames are shown in Figure \ref{fig:rob_frames}.
The transformation between the frames $F_E$ and $F_R$ is known through
the forward kinematics of the robot manipulator. Hence the calibration
step aims at finding this transformation between the robot end effector
frame $F_e$ and the Kinect frame $F_K$ as explained below.


Let us consider a set of points  $\lbrace P_K^i,\ i = 1,2,
\dots, N \rbrace $ which are recorded with respect to the Kinect frame $F_K$.
The same set of points as recorded with respect to the robot base
frame $F_B$ is represented by $\lbrace P_B^i,\ i = 1,2,\dots,N \rbrace$.
 These later points are obtained by moving the robot so that the
robot end-effector touches these points which are visible in the
Kinect camera frame.  Since these two sets refer to the same set of
physical locations, the relation between them may be written as
\begin{equation}
  P_B^i = RP_K^i + \mathbf{t}
\label{eq:trans}
\end{equation} where $\{R,\mathbf{t}\}$ denotes the corresponding
Rotation and translation needed for the transformation between the
coordinate frames. These equations are solved for $\{R,\mathbf{t}\}$
using least square method based on Singular Value Decomposition (SVD)
\cite{lawson1995solving} \cite{arun1987least} as described below.

The centroid of these points is given by
\begin{eqnarray*}
  \bar{P}_K &=&  \frac{1}{N} \Sigma_{i=1}^N P_K^i  \\
  \bar{P}_B &=&  \frac{1}{N} \Sigma_{i=1}^N P_B^i
\end{eqnarray*}
and the corresponding Covariance matrix is given by
\begin{equation}
  C = \sum_{i=1}^N (P_K^i - \bar{P}_K)(P_B^i - \bar{P}_B)^T
  \label{eq:cov}
\end{equation}

Given SVD of covariance matrix $C=USV^T$, the rotation matrix $R$ and
translation vector $\mathbf{t}$ are given by
\begin{eqnarray}
R &=& VU^T   \\
\mathbf{t} &=& -R \bar{P}_K + \bar{P}_B
\end{eqnarray}

\begin{figure}[h]
\centering
\includegraphics[scale=0.5]{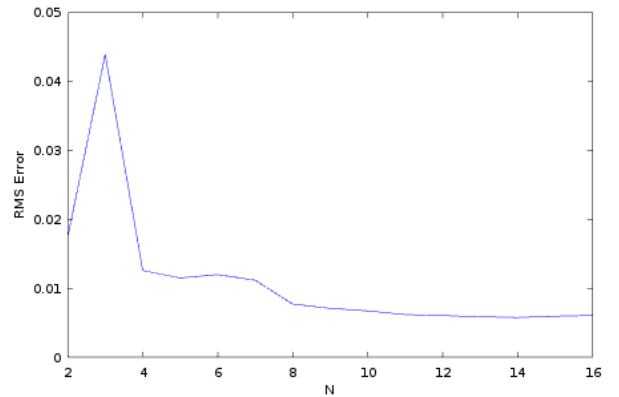}
\caption{Plot of Average RMS error (in meters) with the sample size $N$.}
\label{fig:rms_error}
\end{figure}

\begin{figure}[h]
\centering
\includegraphics[scale=0.25]{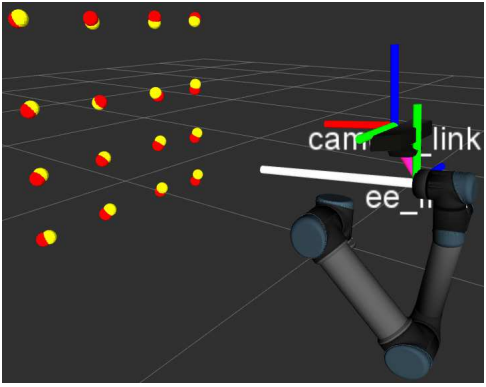}
\caption{Checking the accuracy of Calibration. Points in red color are
  the robot end-effector points collected prior to calibration. The
  yellow points are the points in the Kinect frame which are
  transformed into the robot base frame using estimated
  $\{R,\mathbf{t}\}$.}
\label{fig:est_points}
\end{figure}

The RMS error between the actual points and the points computed using
estimated $\{R,\mathbf{t}\}$ is shown in Figure \ref{fig:rms_error}
and the corresponding points are shown in Figure \ref{fig:est_points}.
The points are shown with respect to the robot base coordinate frame.
The red points are the actual points and the yellow points are
computed using estimated values of $\{R,\mathbf{t}\}$. It is possible
to obtain an RMS error of 1 cm with as small as 8 points. 

\subsection{Rack detection}\label{sec:rack_detect}

\begin{figure}[h]
  \centering
  \begin{tabular}{cc}
    \subfigure[Line detection]{\label{f:line_det}\includegraphics[scale=0.2]{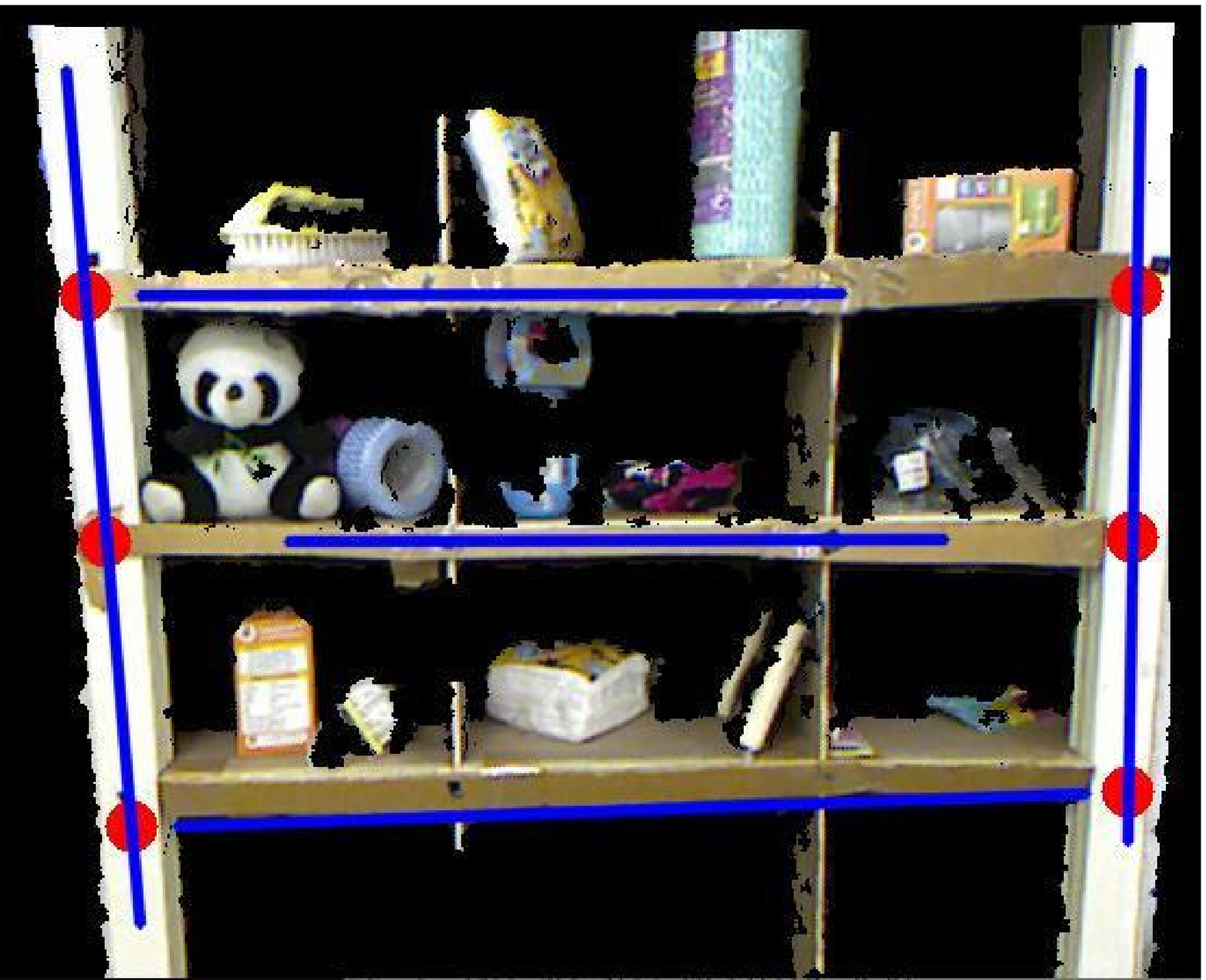}} & 
    \subfigure[Bin Centres]{\label{f:bin_cen}\includegraphics[scale=0.3]{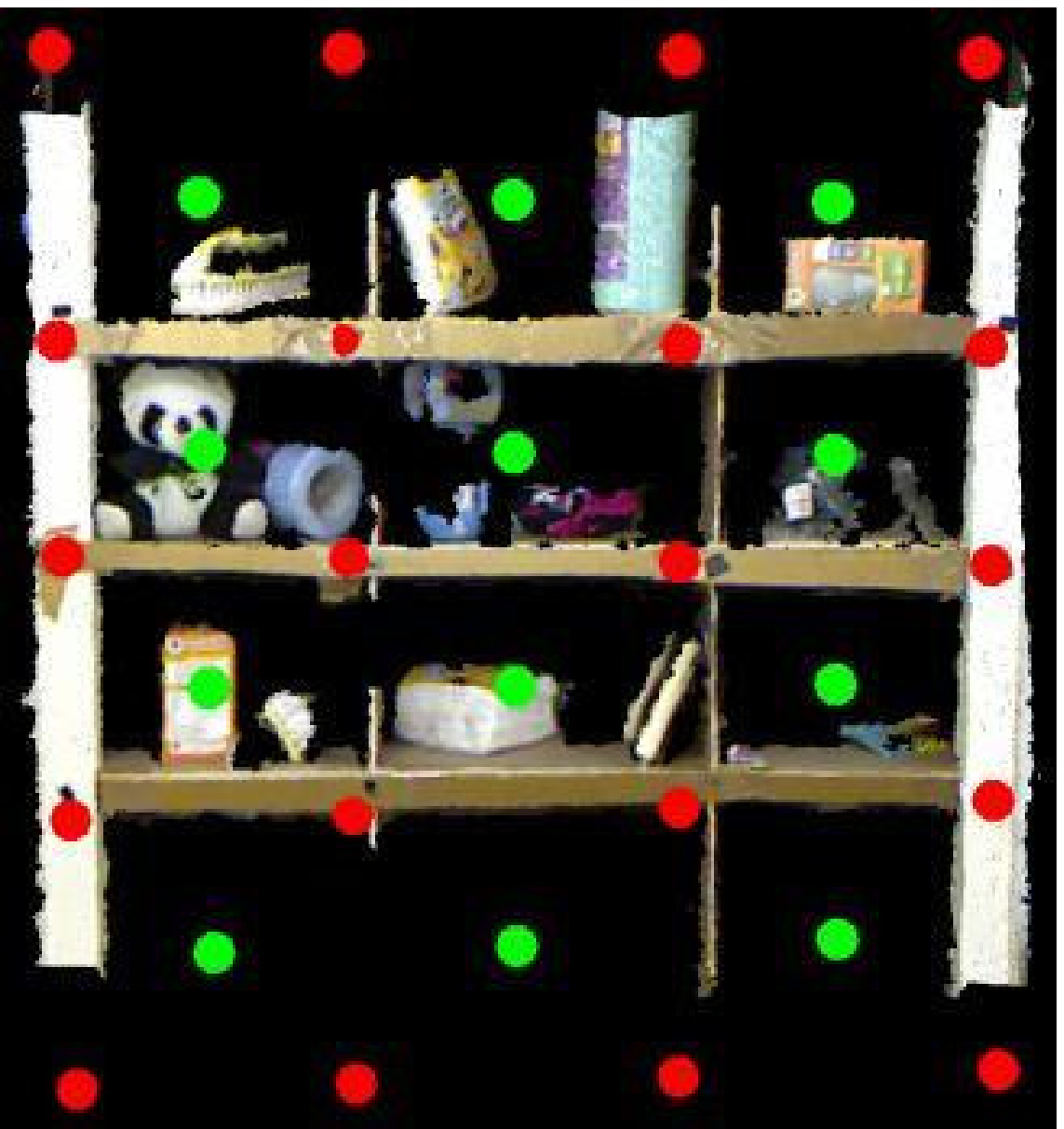}}
  \end{tabular}
  \caption{Rack Detection from the RGBD point cloud. (a) Vertical and
    horizontal lines are detected using Hough line transform.
    Intersection of these vertical and horizontal lines provide
    corners for bins. (b) The bin centres are computed as the mean of
  bin corners.}  
    \label{fig:rack_detect} 
  \end{figure}

Rack detection involves finding the corners of the rack and the bin
centers automatically from an RGBD image recorded by the on-board
Kinect camera. The bin corners information is useful for defining
region of interest (ROI) for identifying objects within the bin. The
bin corners and centres are also useful for planning motion to and
inside the bins for picking objects. The bins in the rack are
in form of a grid structure consisting of 4 vertical and 5 horizontal
lines, and hence the bin corners can be identified by the
intersection of vertical and horizontal lines.  The vertical and the horizontal lines on
the rack are detected using Hough line transform \cite{matas2000}.  If
$(x_1^v,y_1^v), (x_2^v,y_2^v)$ are end points of a vertical line and
$(x_1^h,y_1^h), (x_2^h,y_2^h)$ are end points of a horizontal line
then the
equation\footnote{\burl{https://en.wikipedia.org/wiki/Line-line_intersection}}
to compute the intersection $(x_i,y_i)$ of the two lines  is given by
\begin{eqnarray*}
x_i &=
\frac{(x_1^vy_2^v-y_1^vx_2^v) (x_1^h-x_2^h)-(x_1^v-x_2^v)
(x_1^hy_2^h-y_1^hx_2^h)}
{(x_1^v-x_2^v)(y_1^h-y_2^h)-(y_1^v-y_2^v)(x_1^h-x_2^h)} \nonumber \\
y_i &= \frac{(x_1^vy_2^v-y_1^vx_2^v) (y_1^h-y_2^h)-(y_1^v-y_2^v)
(x_1^hy_2^h-y_1^hx_2^h)} {(x_1^v-x_2^v)
(y_1^h-y_2^h)-(y_1^v-y_2^v)(x_1^h-x_2^h)} 
\label{eq:intsecpt}
\end{eqnarray*} 
Once the corners are known, the bin centre can be computed as the mean
of its centres. The Figure \ref{f:line_det} shows the vertical and horizontal
lines detected using an OpenCV \cite{bradski2000opencv} implementation
for Hough transform \footnote{
  \burl{http://docs.opencv.org/2.4/doc/tutorials/imgproc/imgtrans/hough_lines/hough_lines.html}}.
  The intersection points computed using above equations are shown in
  Figure \ref{f:bin_cen} where the bin corners are shown in red
  while the bin centres are shown in green. Note that only three
  middle horizontal lines and two outer vertical lines are required to
be detected. Rest of the points can be estimated using the prior
knowledge of rack geometry.

\subsection{Object Recognition} \label{sec:obj_recog}

Recognition and localization of an object in an image have been a
fundamental and challenging problems in computer vision since decades
\cite{jain1997object, belongie2002shape, lowe2004distinctive,
dalal2005histograms}.  In the era of deep learning, CNN has been
widely used for object recognition task and it has shown outstanding
performance \cite{girshick2016region, zhang2015improving,
ren2015faster, liu2016ssd} as compared to the conventional
hand-crafted feature based object recognition techniques
\cite{lowe2004distinctive, dalal2005histograms}.  Techniques, like
deformable parts models (DPM) \cite{felzenszwalb2008discriminatively}
uses a sliding window method where at every evenly spaced spatial
location the classifier is trained.  The approach hence fails to
progress further due to huge computational complexity.  Eventually, in
$2014$ R-CNN was introduced by Girshick et al.
\cite{girshick2014rich}, which use region proposal methods to generate
potential bounding boxes at the first stage. Then the classifier is
trained on each of these proposed boxes.  The bounding boxes are
fine-tunned by post-processing followed by eliminating duplicate
detection and re-evaluating the box based on objects in the scene.
There are other variants of R-CNN  with improved recognition accuracy
and faster execution time. Some of theses are presented in
\cite{girshick2015fast,ren2015faster, redmon2016you}.     

In a recent work Redmon et al. proposes \emph{you only look once}
(YOLO) \cite{redmon2016you}, where the object detection is transformed
to a single regression problem.  The approach improves the performance
in terms of computational cost, however, the recognition accuracy is
slightly inferior as compared to the Faster RCNN \cite{ren2015faster}.
We use Faster RCNN as a base for our object recognition and
localization task, as it localizes the objects in an image in
real-time with very high recognition accuracy.  
%

 In APC 2016, object detection is considered to be a challenging
 problem due to varying illumination conditions, placement of the
 objects in different orientation and depths inside the rack. In case
 of stowing, the objects in the tote can be fully or partially
 occluded by other objects.  These, resulted in a very complex object
 recognition task. 

We have combined the deep learning approach and standard image processing techniques 
for robust object detection. We are using Faster RCNN based deep neural network
to find the bounding box of the target object. A second step
verification of target object in the bounding box provided by RCNN is
performed using random forest classifier. We have done fine tuning of pretrained
object detection model with our own dataset. The details of the data preparation,
training and verification step are given in the below sections.

\textbf{Annotation:} We have prepared two different training datasets for picking 
 and stowing task. We have annotated 150 RGB images per object with different orientations
 and backgrounds for each task. A total of 6000 images were annotated for each task. 

\textbf{Training models:}
To do object detection task, which includes classification and localisation,
we are using VGG-16 layered classification network in combination with Region proposal Networks. 
RPN are basically fully covolutional network which takes an image as input and outputs a set of
rectangular object proposals, each with an objectness score. It is a 16 layered 
classification network which consists of 13 convolution layers and 3 fully connected layers.
These RPN share convolutional layers with object detection networks due to
which it does not add significant computation at run time(~10ms per image).
We have fine tuned VGG-16 pretrained model of faster RCNN for our own dataset of 6000
images for 40 different objects.

\textbf{Object Verification:}
We have added an additional step in object detection pipeline to verify object in the window proposed
by RCNN. It uses shape and color information to verify the presence of the 
object. Both shape and color informations are incorporated as a feature vector and a random
forest is used to classify each pixel inside the object box. After finding the most probable region inside
the window using random forest, we apply a meanshift algorithm to obtain the suction point
for that object. The details of the feature (shape and color) and classifier used
are explained below:

\textbf{Shape and color information as a feature:} As we know, any 3D
surface of the object is characterized by the surface normals at each
point in the pointcloud. The angle between neighboring normals at
interest points  on any suface can be used to the shape of any object.
A shape histogram is created for each object model which is used as a
shape feature vector.  Similarly, we are incorporating color
information in the feature using color and grayscale histogram of the
objects.

\textbf{Random Forest:}
After computing histograms, all three histograms are concatenated as a
37 dimensional feature vector for each object.  The training data is
prepared by extracting features from pointcloud and RGB data for each
object.  A Random forest classifier is trained for each object with
One vs all strategy. In one vs all, the target object features are
trained as positive class and rest all features are considered as
negative class.  The number of trees and depth of the trees used in
the random forest are 100 and 30 respectively.

\begin{figure}[!t]
  \centering
  \begin{tabular}{ccccc}
     \includegraphics[scale=0.07]{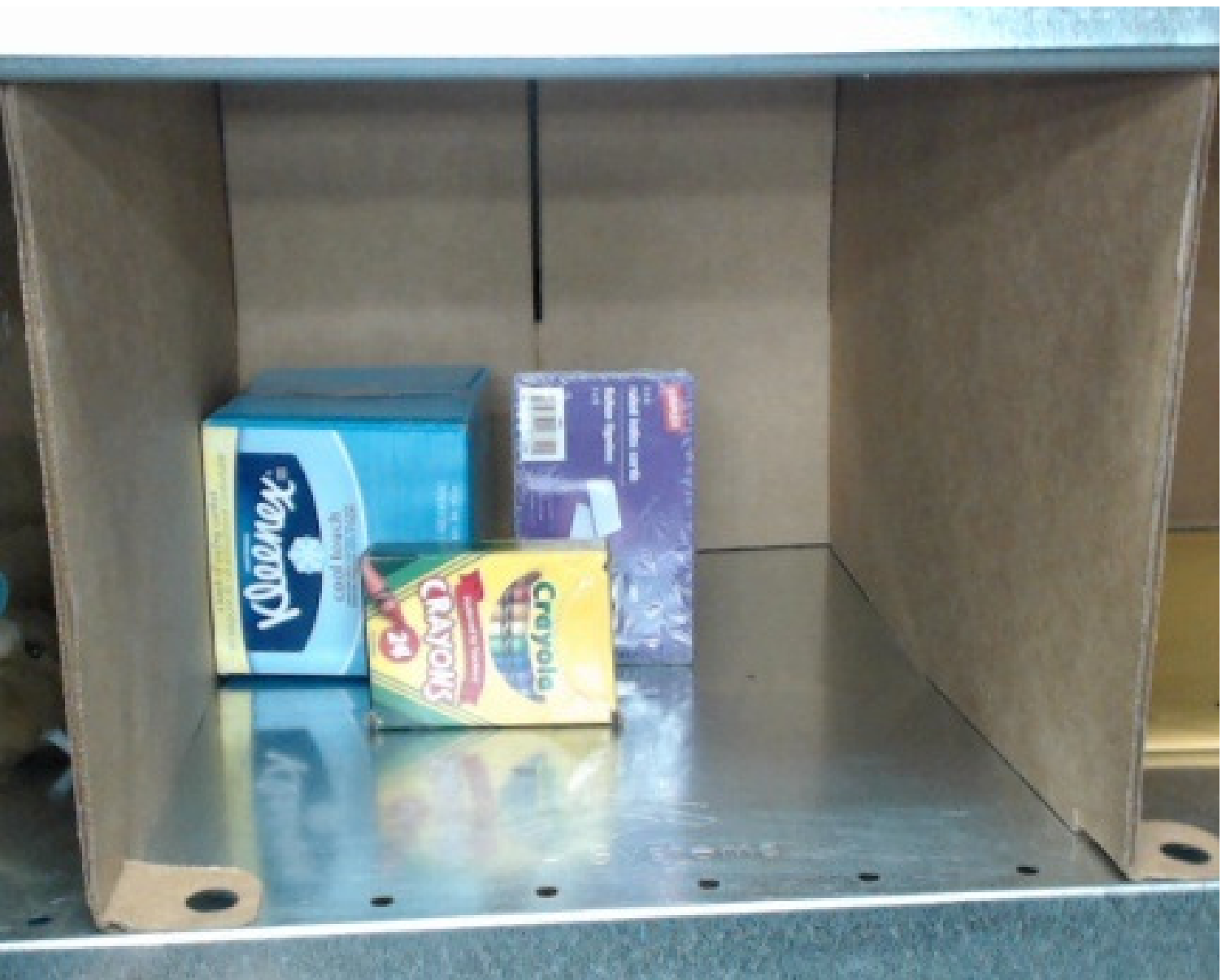}&
  \includegraphics[scale=0.05]{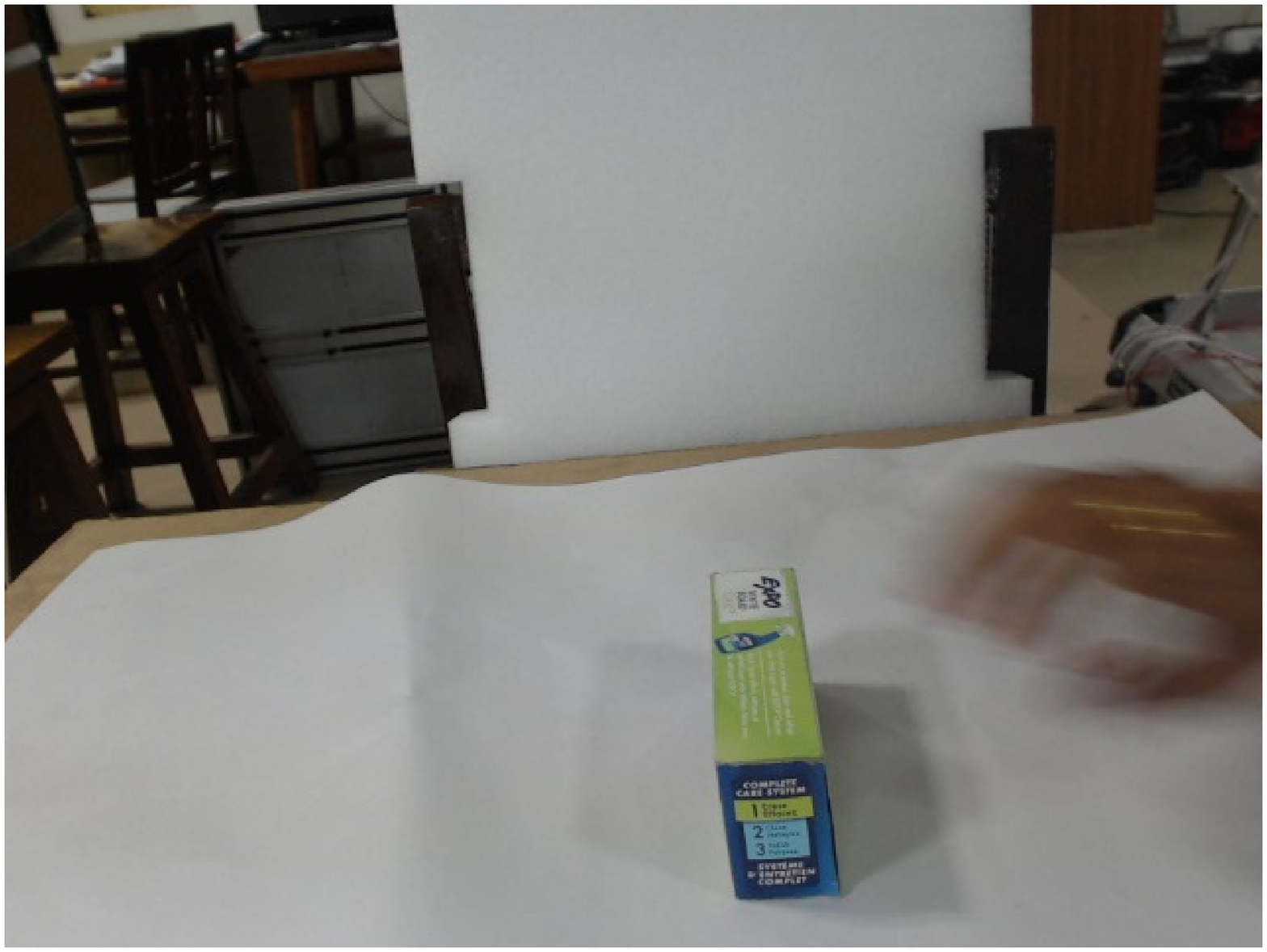} & 
    \includegraphics[scale=0.07]{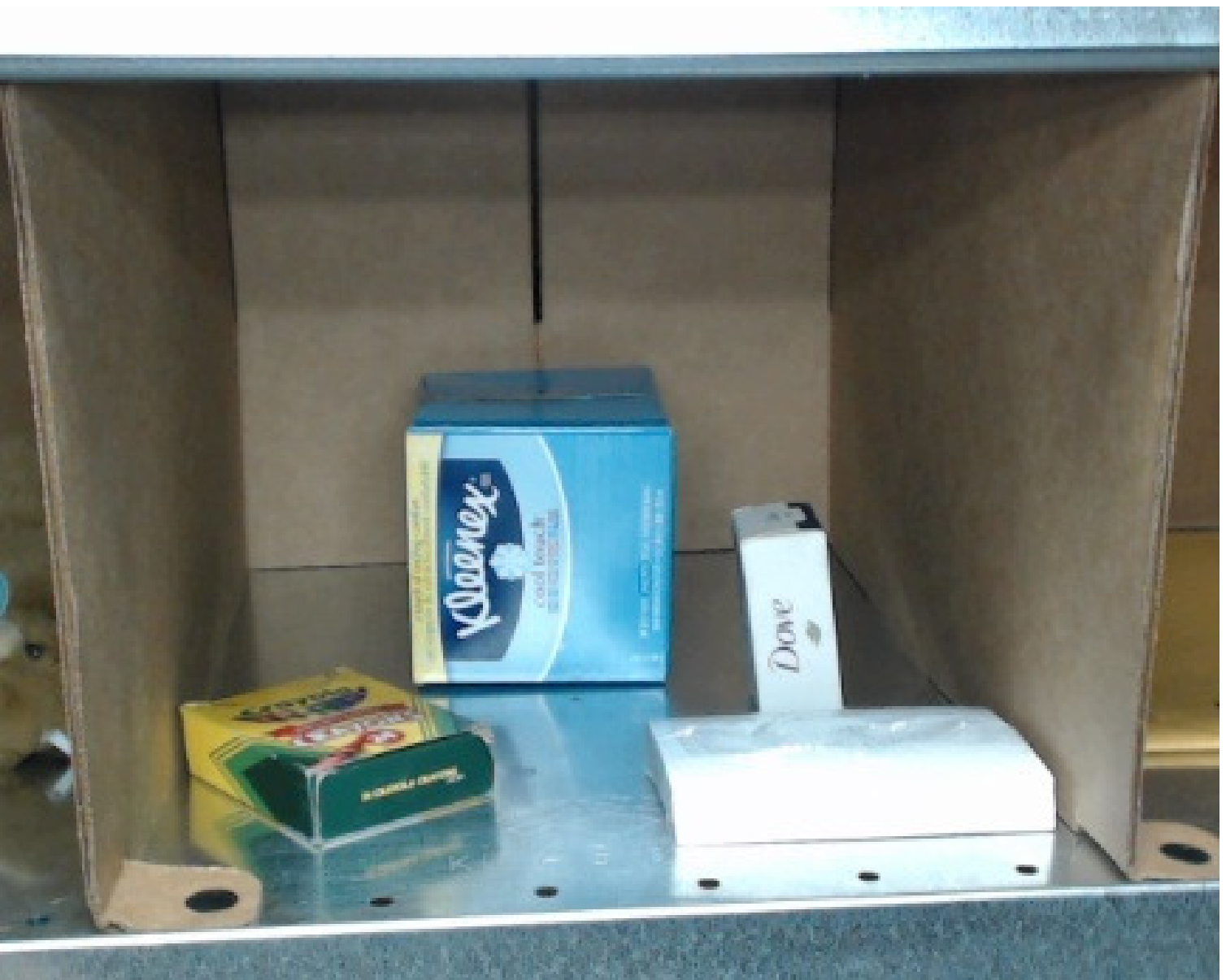} & 
    \includegraphics[scale=0.05]{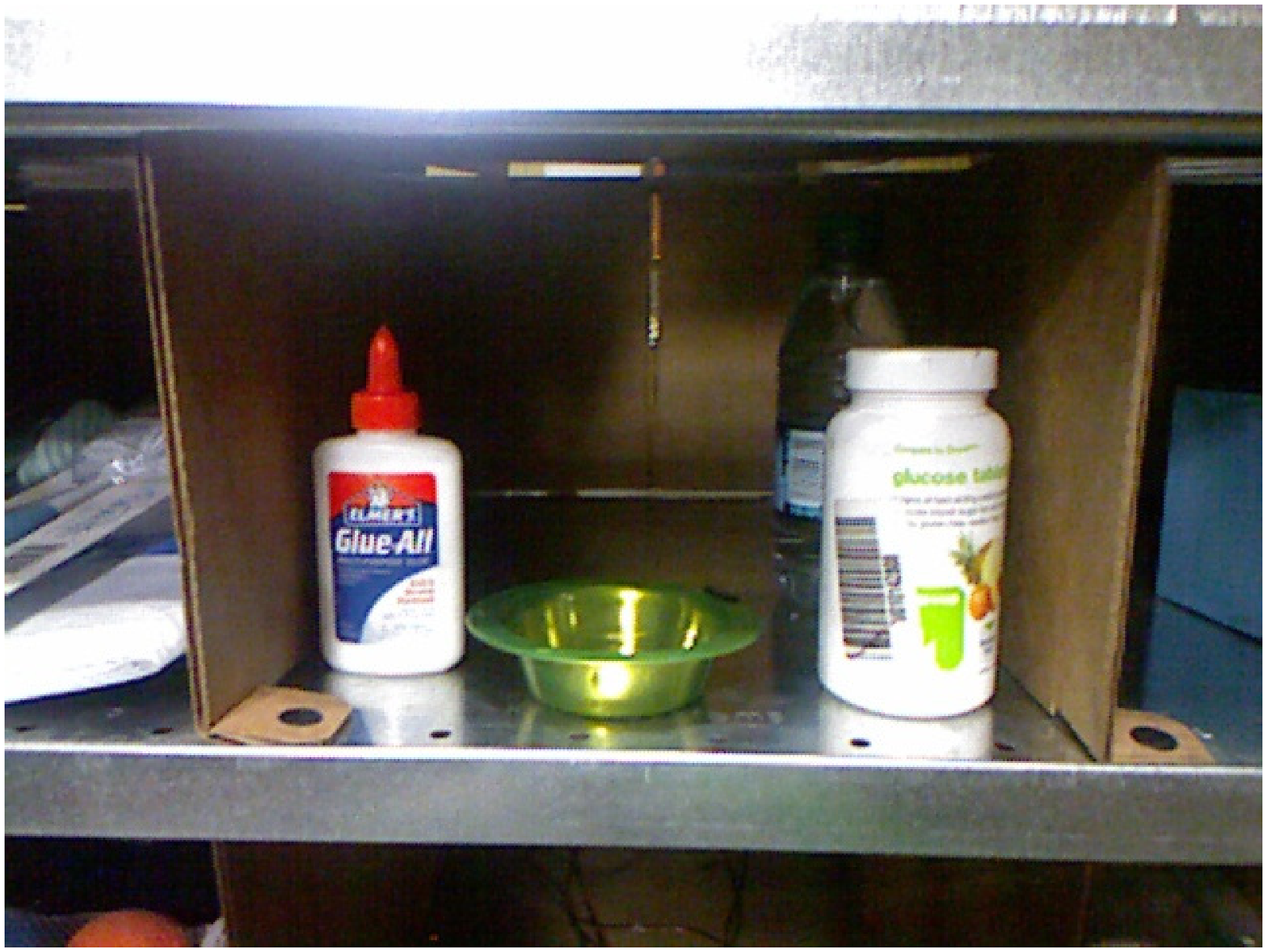} & 
    \includegraphics[scale=0.07]{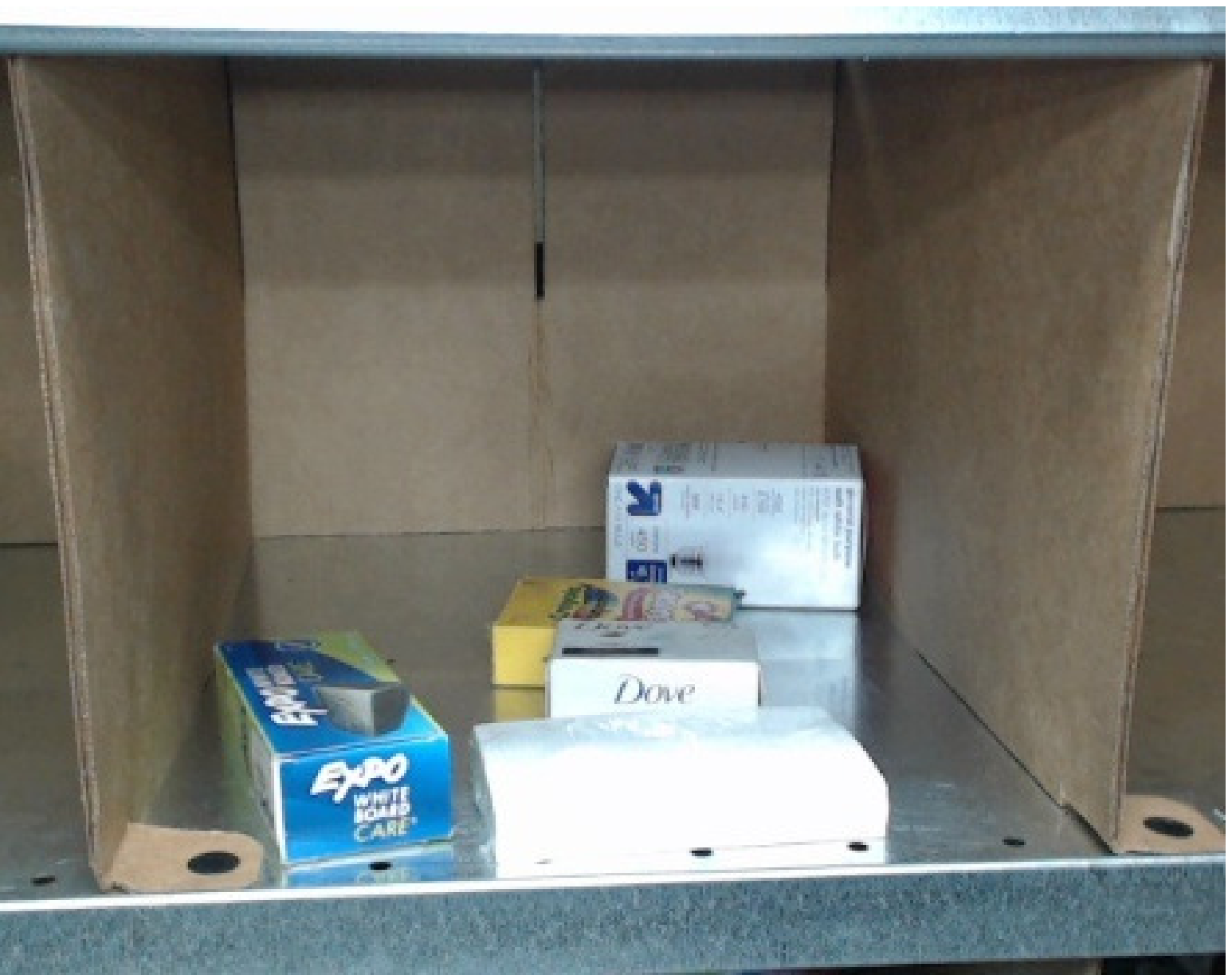} \\ 
    \includegraphics[scale=0.1]{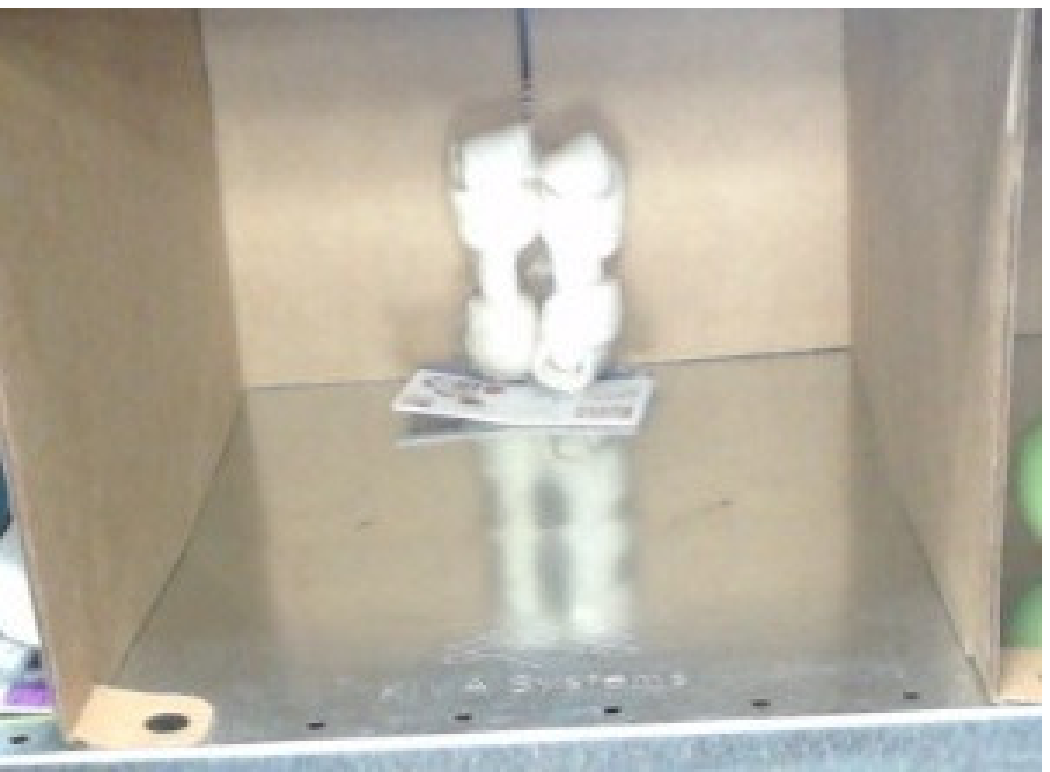} & 
    \includegraphics[scale=0.1]{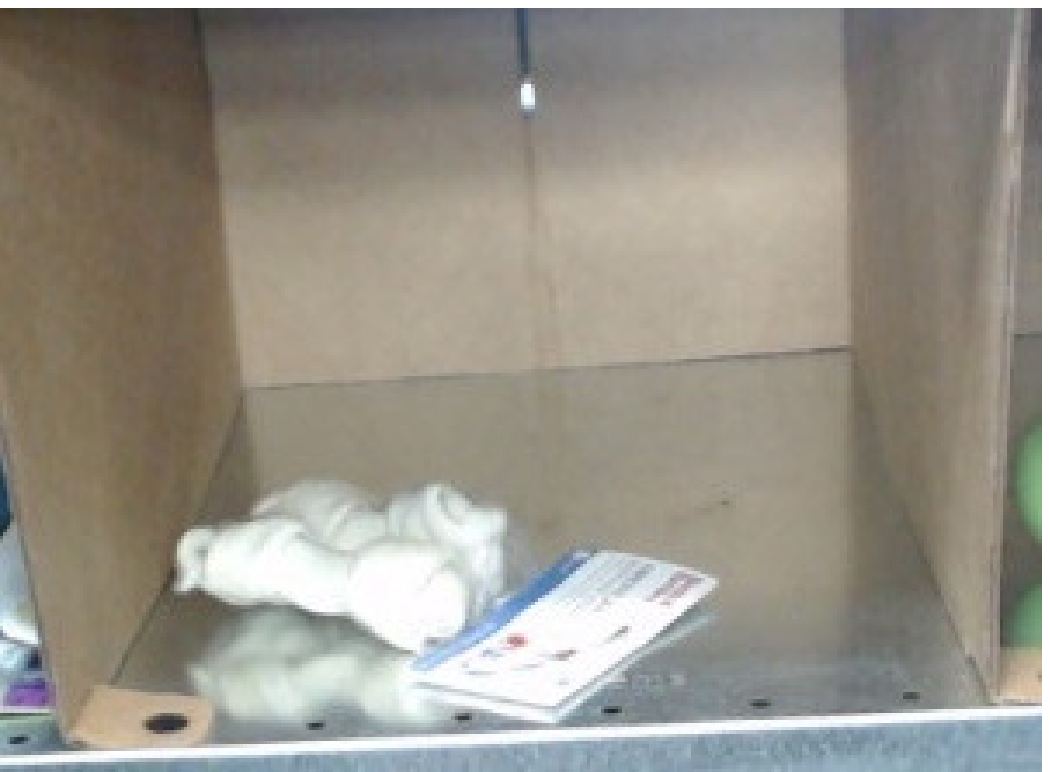} & 
    \includegraphics[scale=0.05]{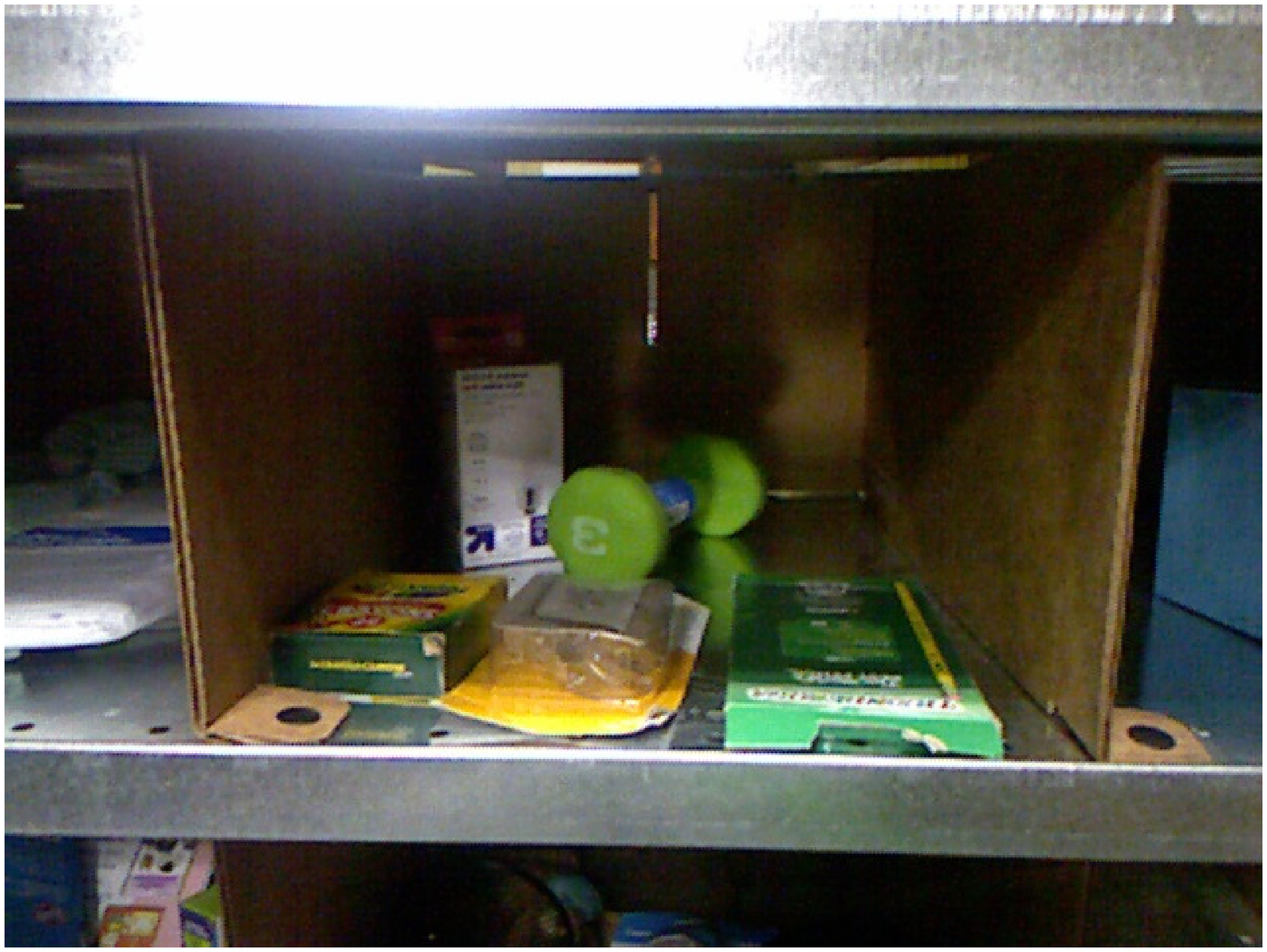} & 
    \includegraphics[scale=0.07]{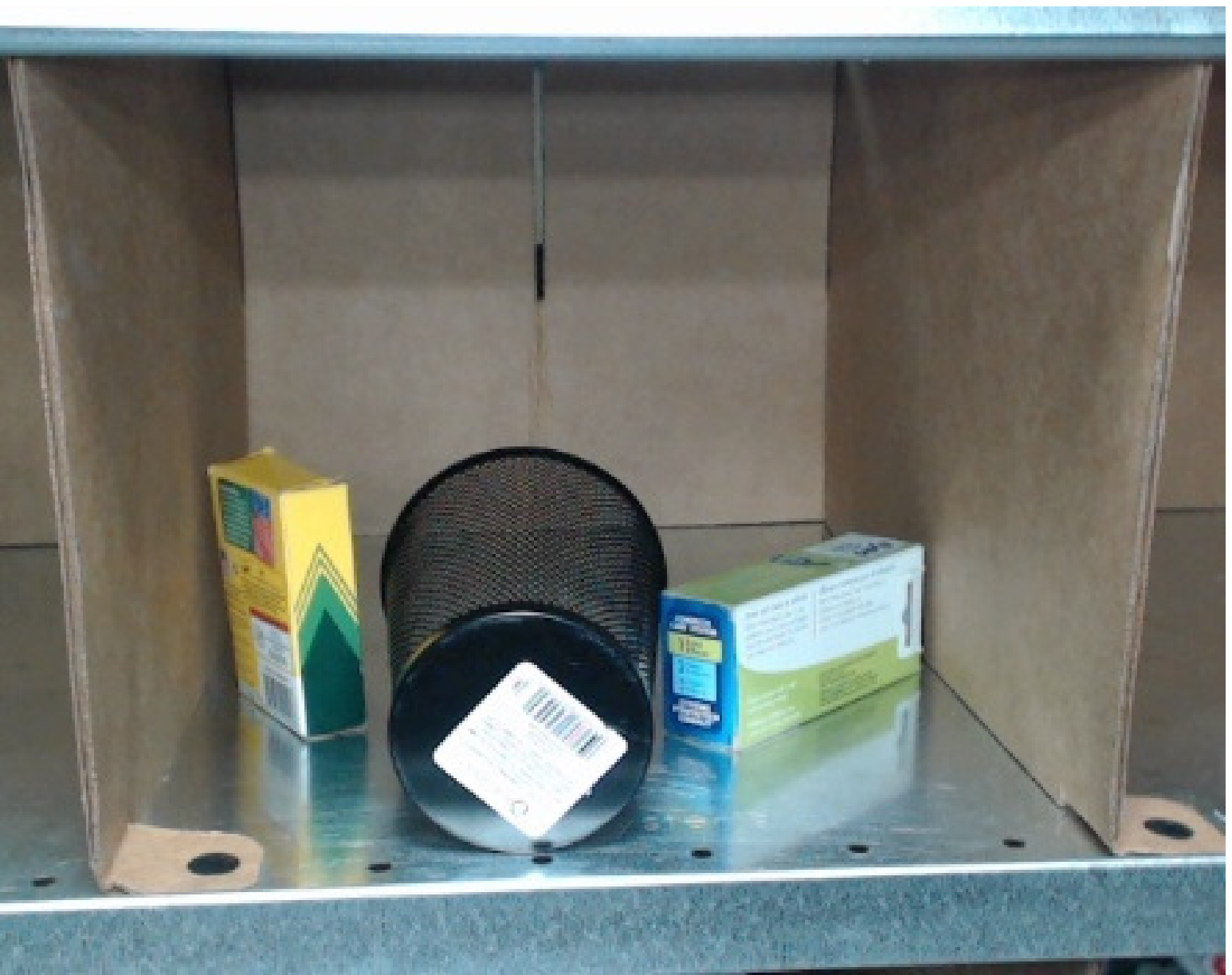} & 
    \includegraphics[scale=0.07]{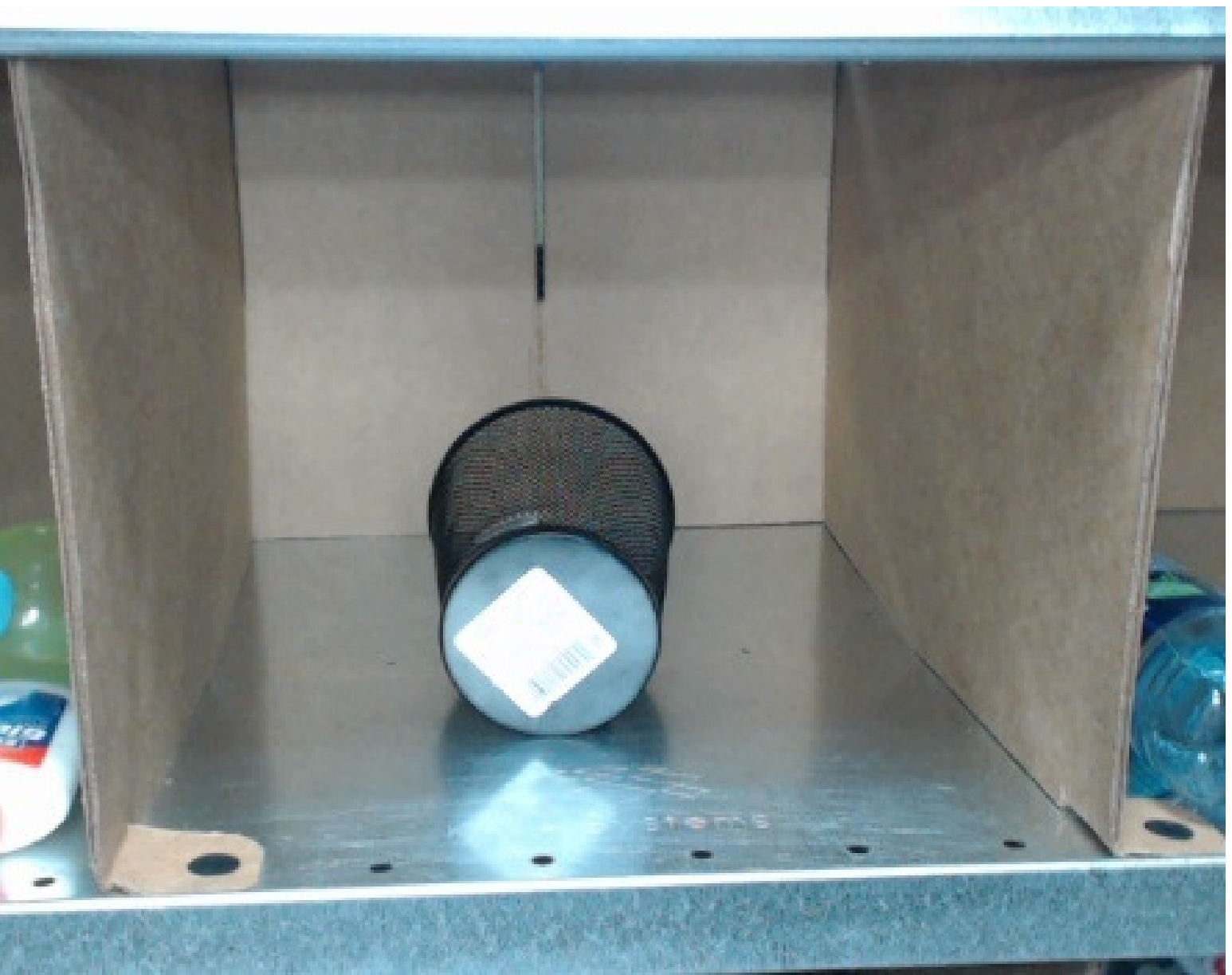} \\ 
    \includegraphics[scale=0.07]{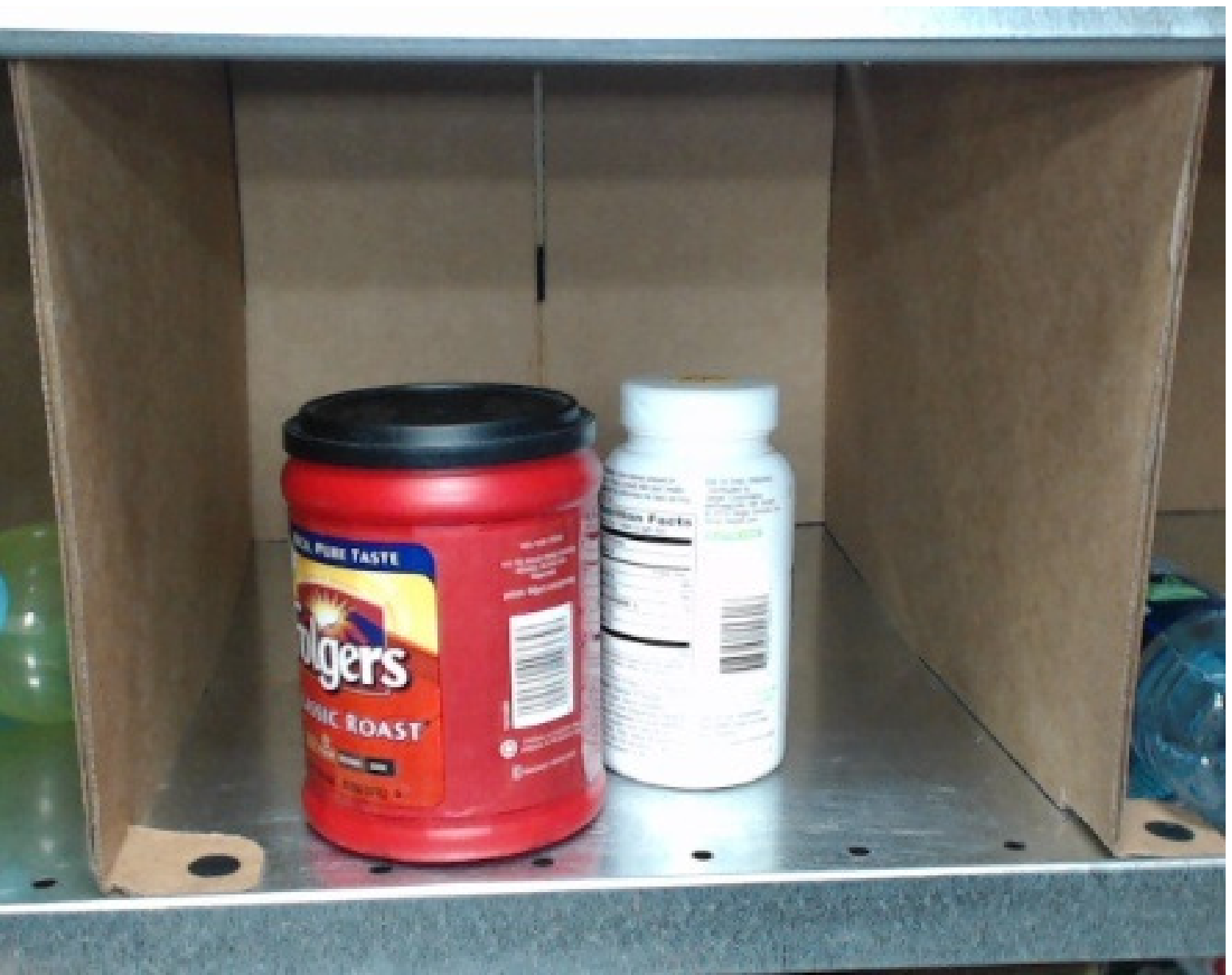} & 
    \includegraphics[scale=0.07]{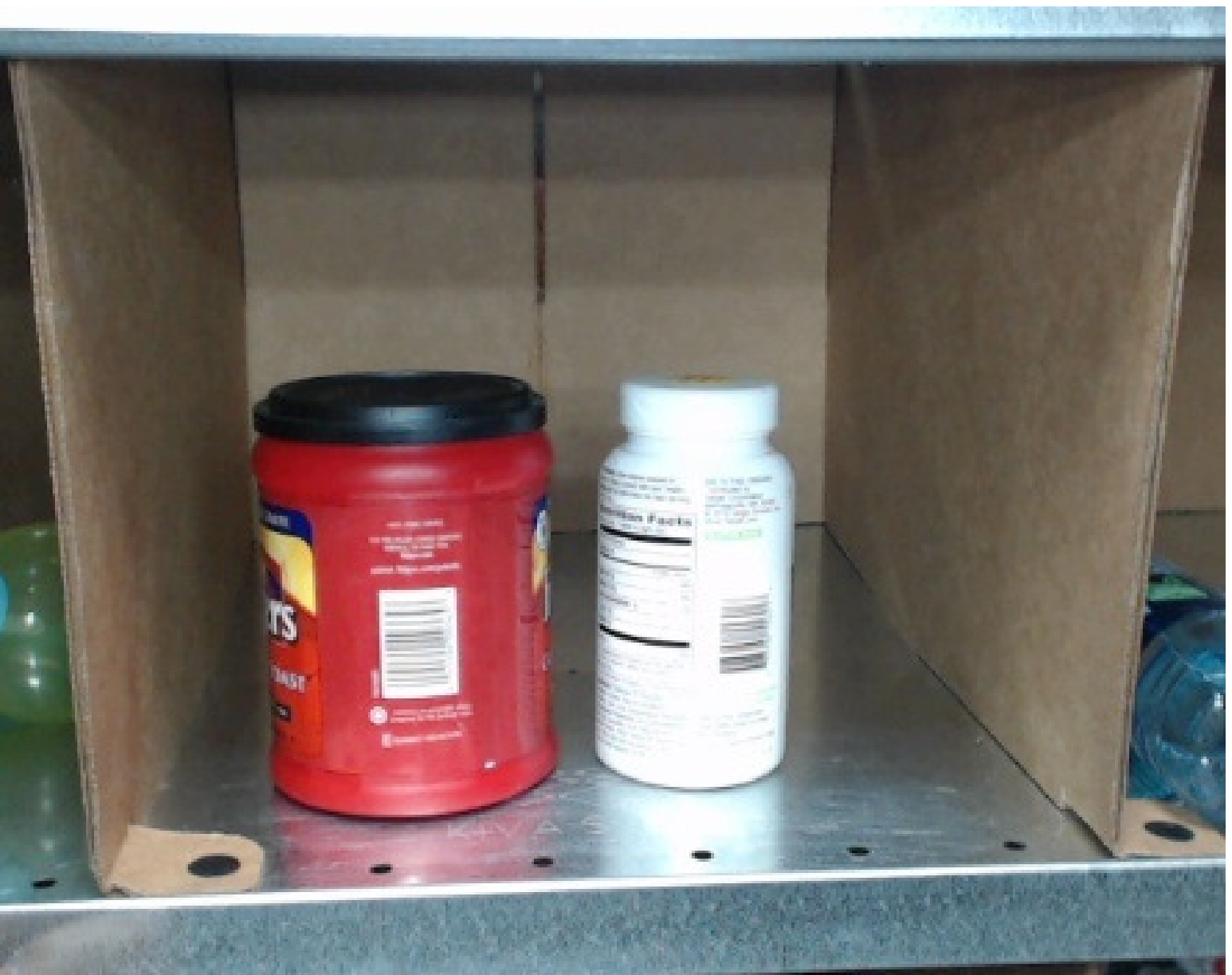} & 
    \includegraphics[scale=0.07]{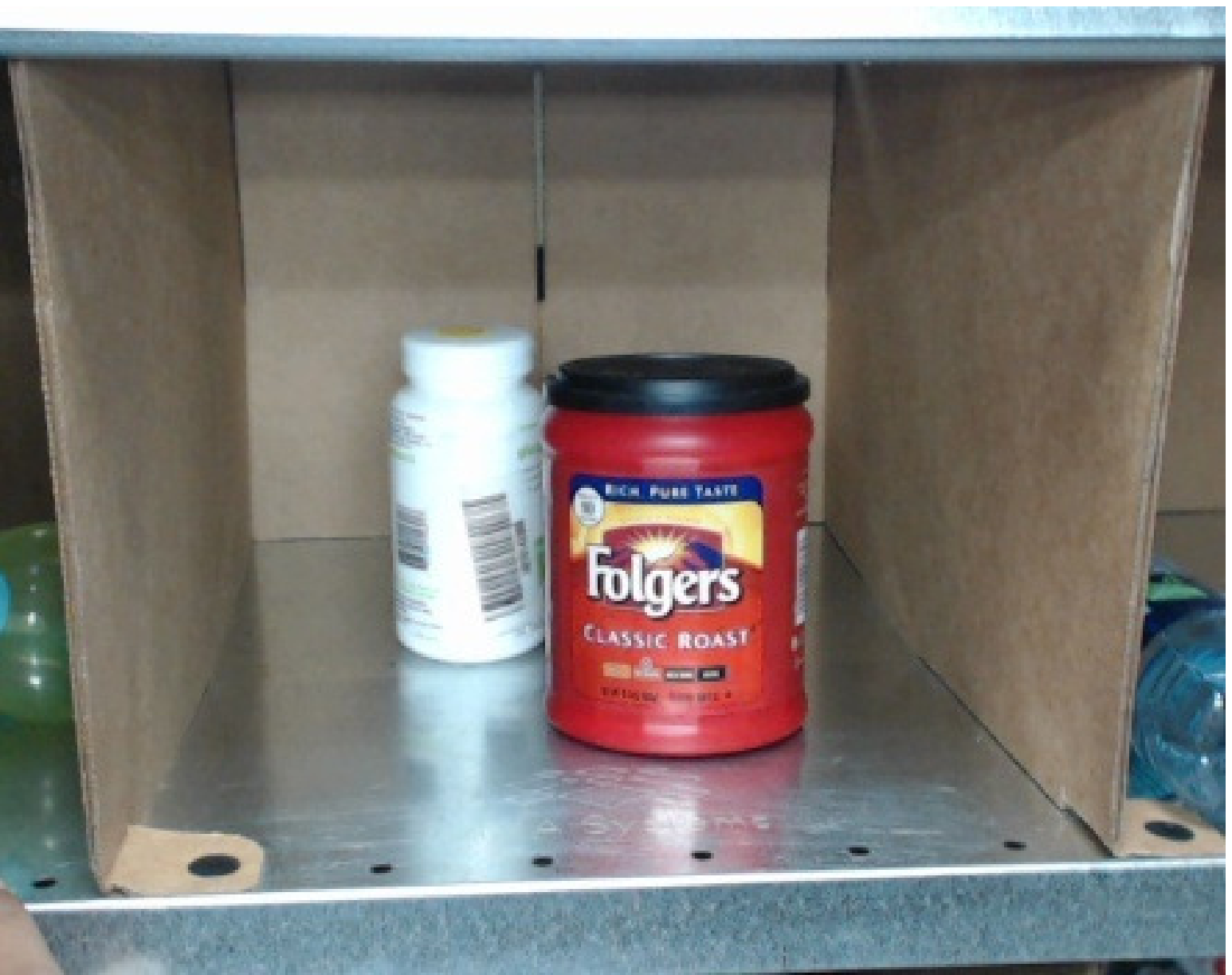} & 
    \includegraphics[scale=0.07]{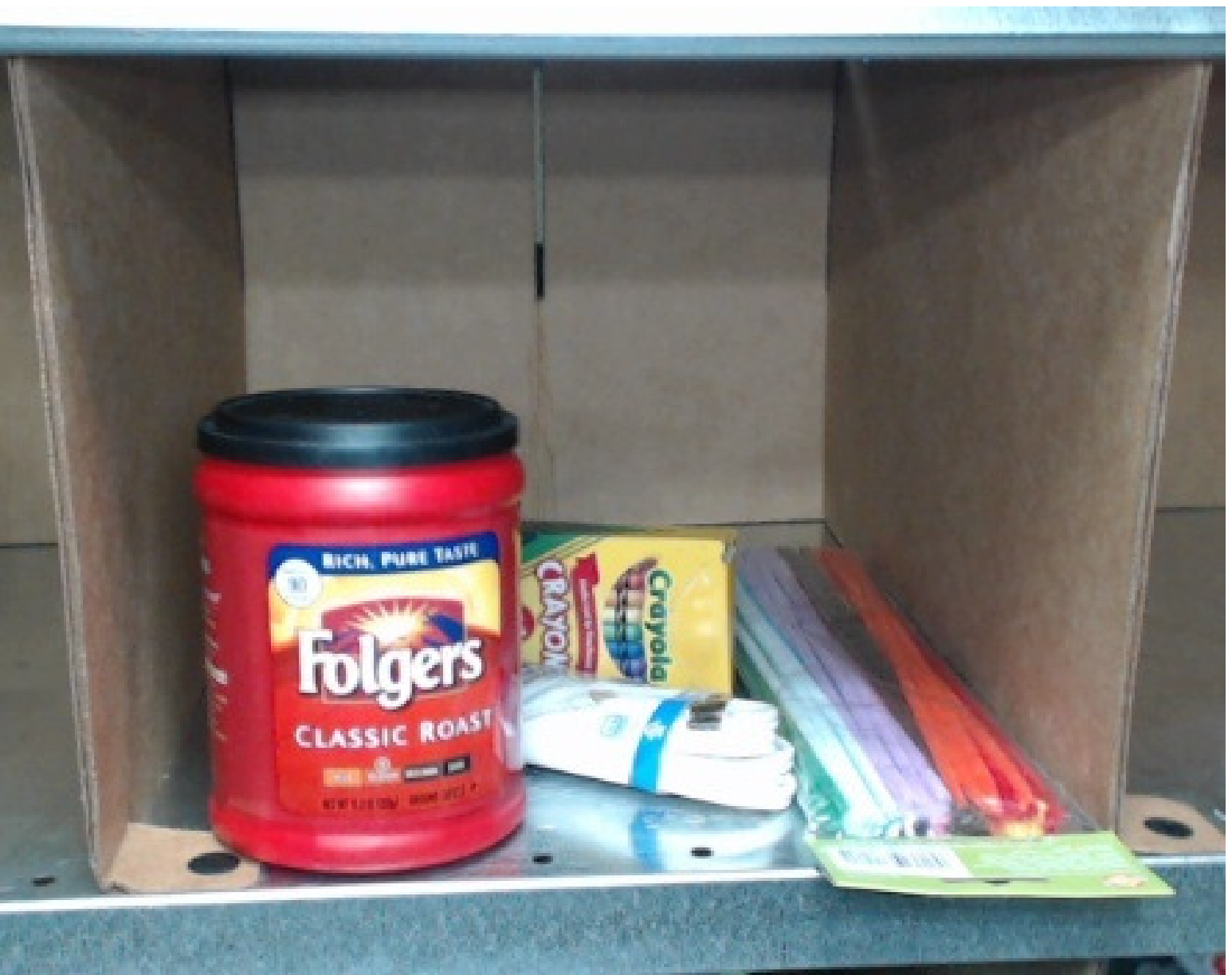} & 
    \includegraphics[scale=0.07]{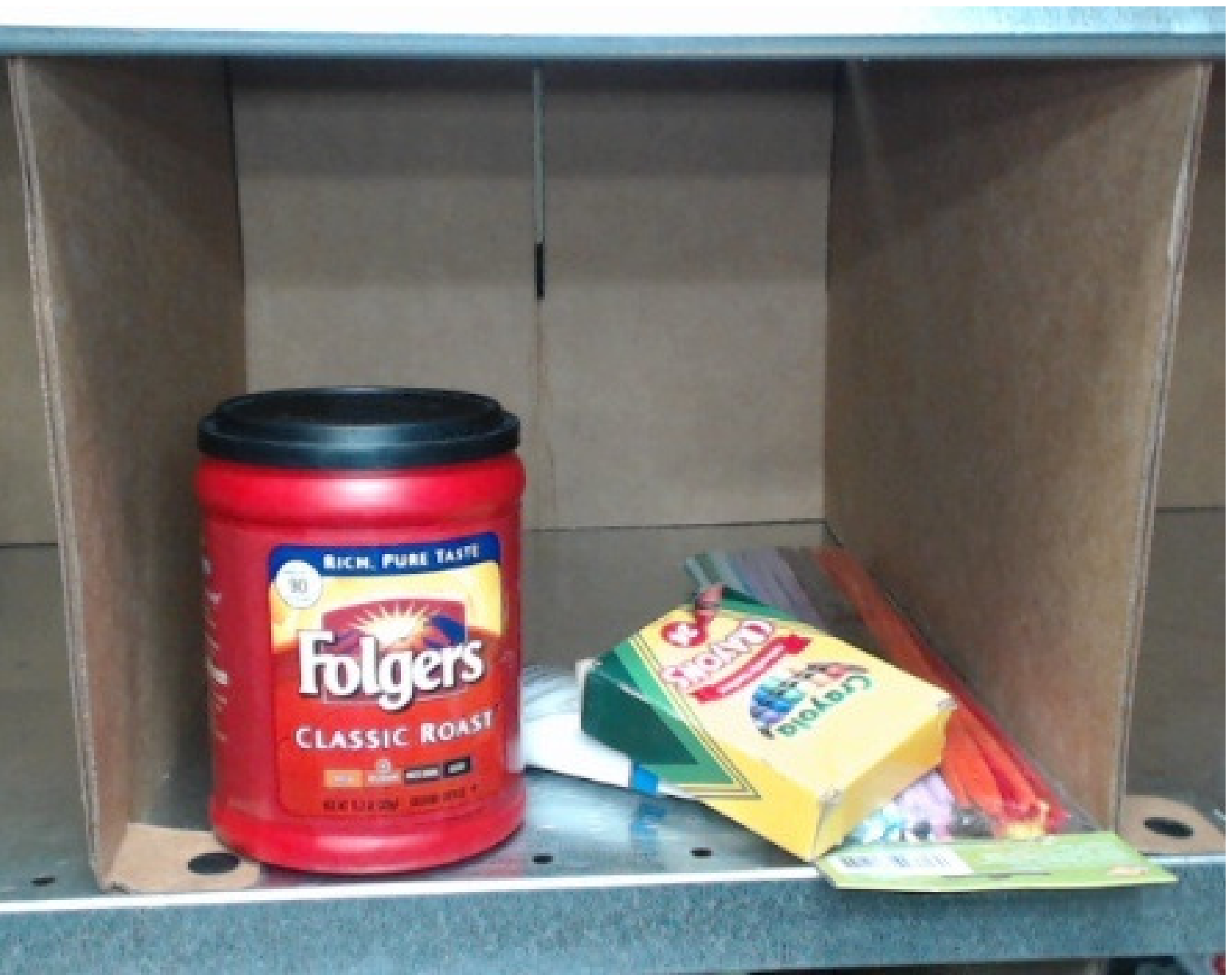} \\ 
    \includegraphics[scale=0.04]{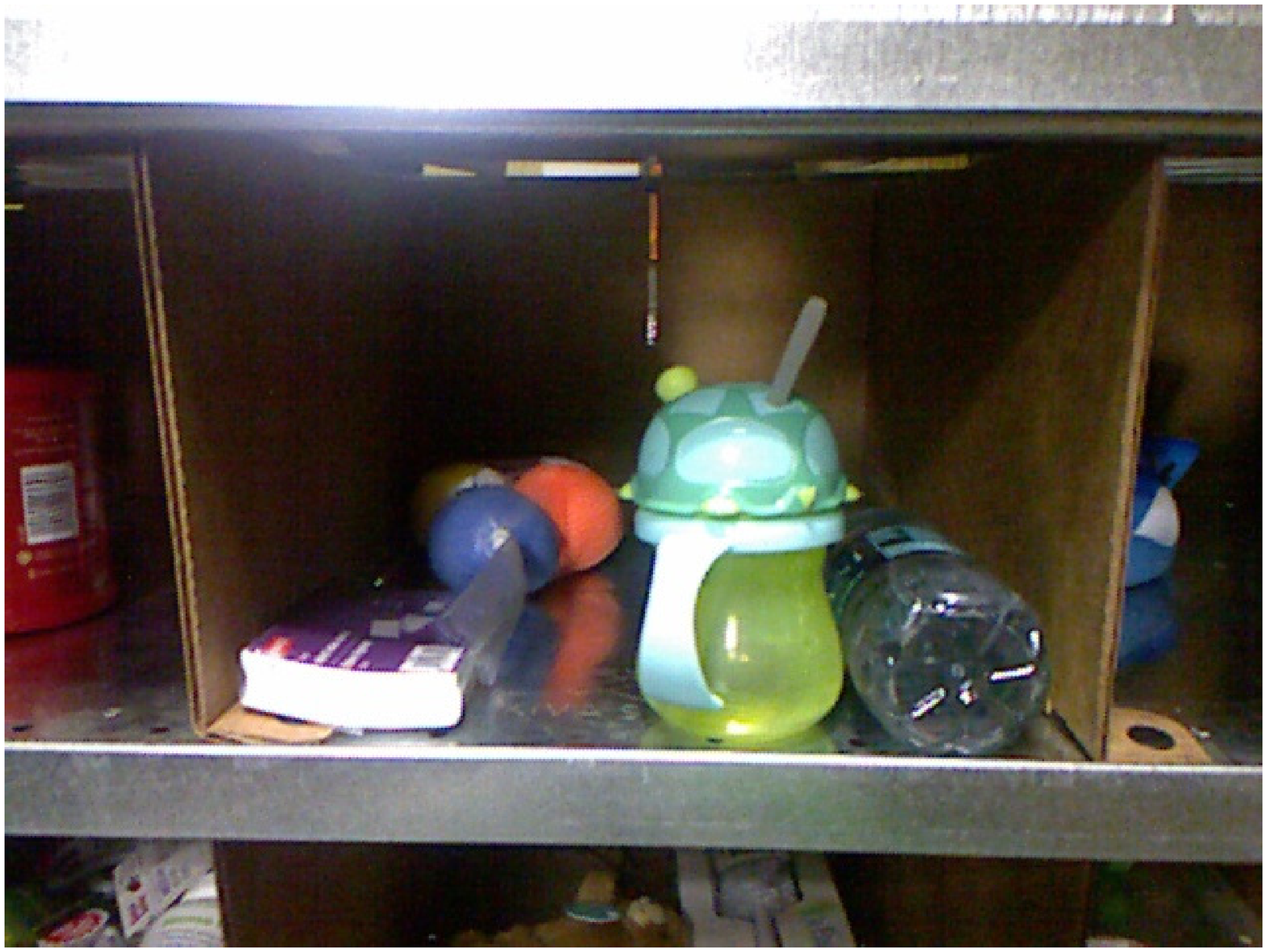} & 
    \includegraphics[scale=0.04]{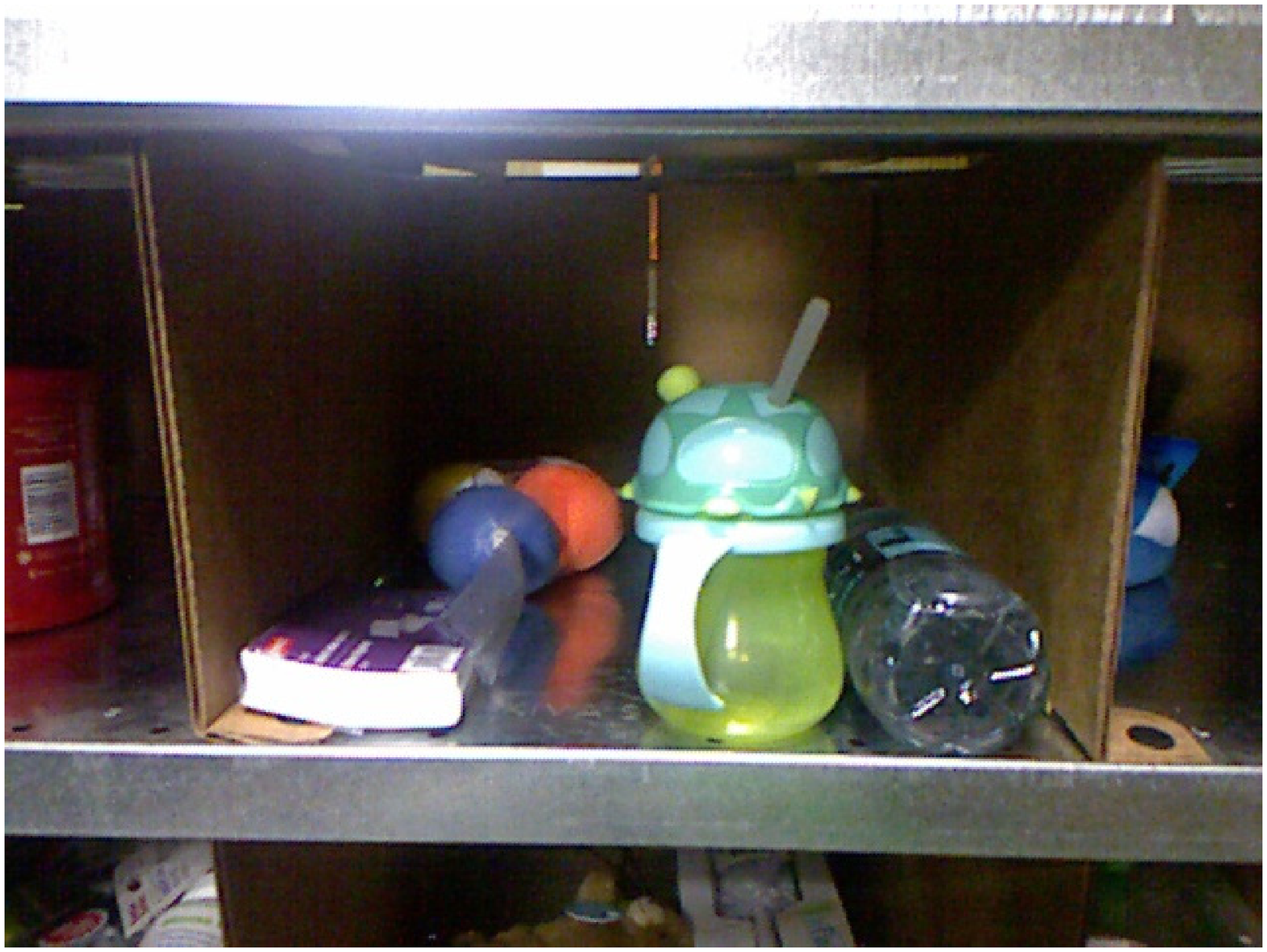} & 
    \includegraphics[scale=0.04]{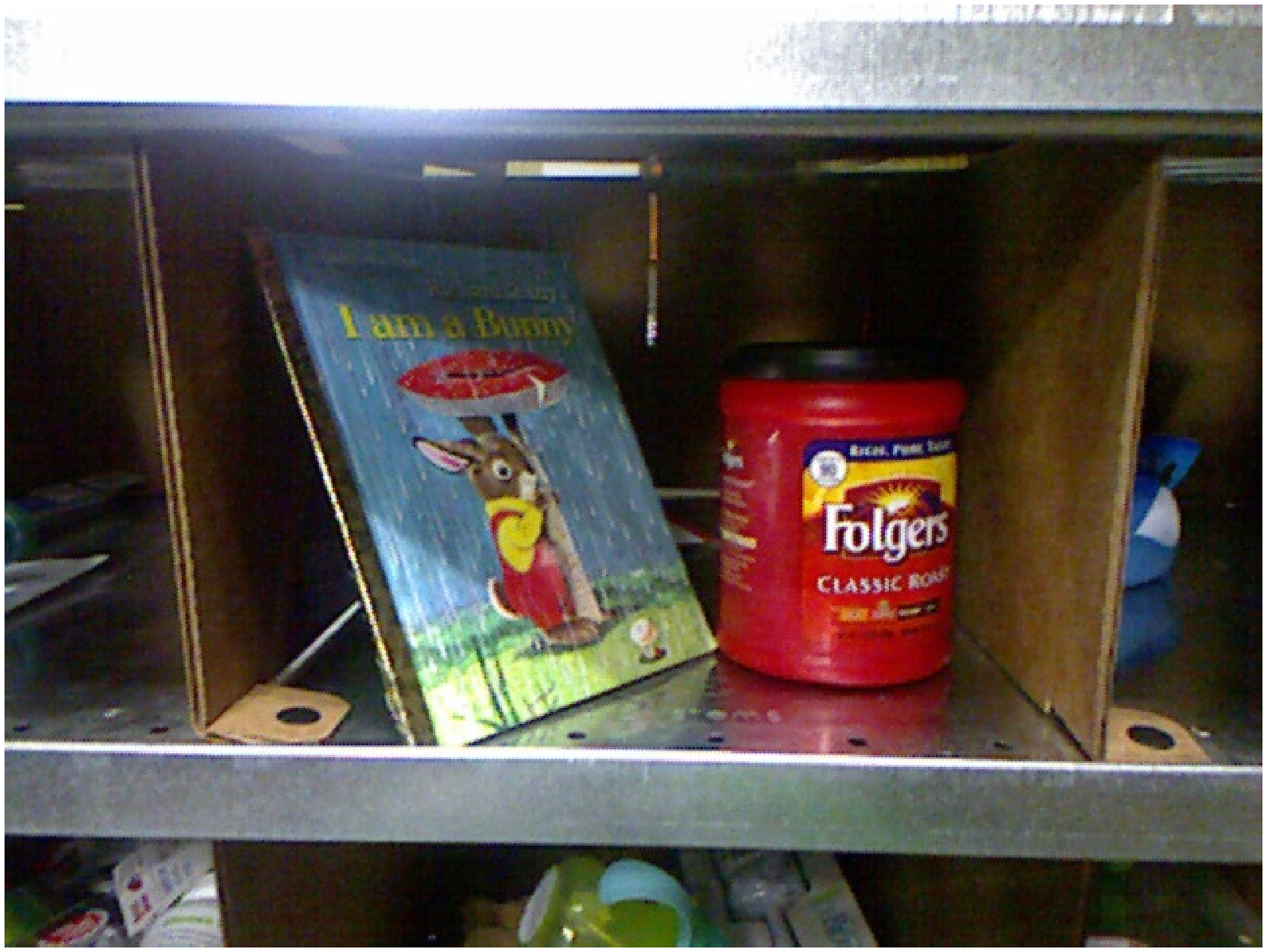} & 
    \includegraphics[scale=0.04]{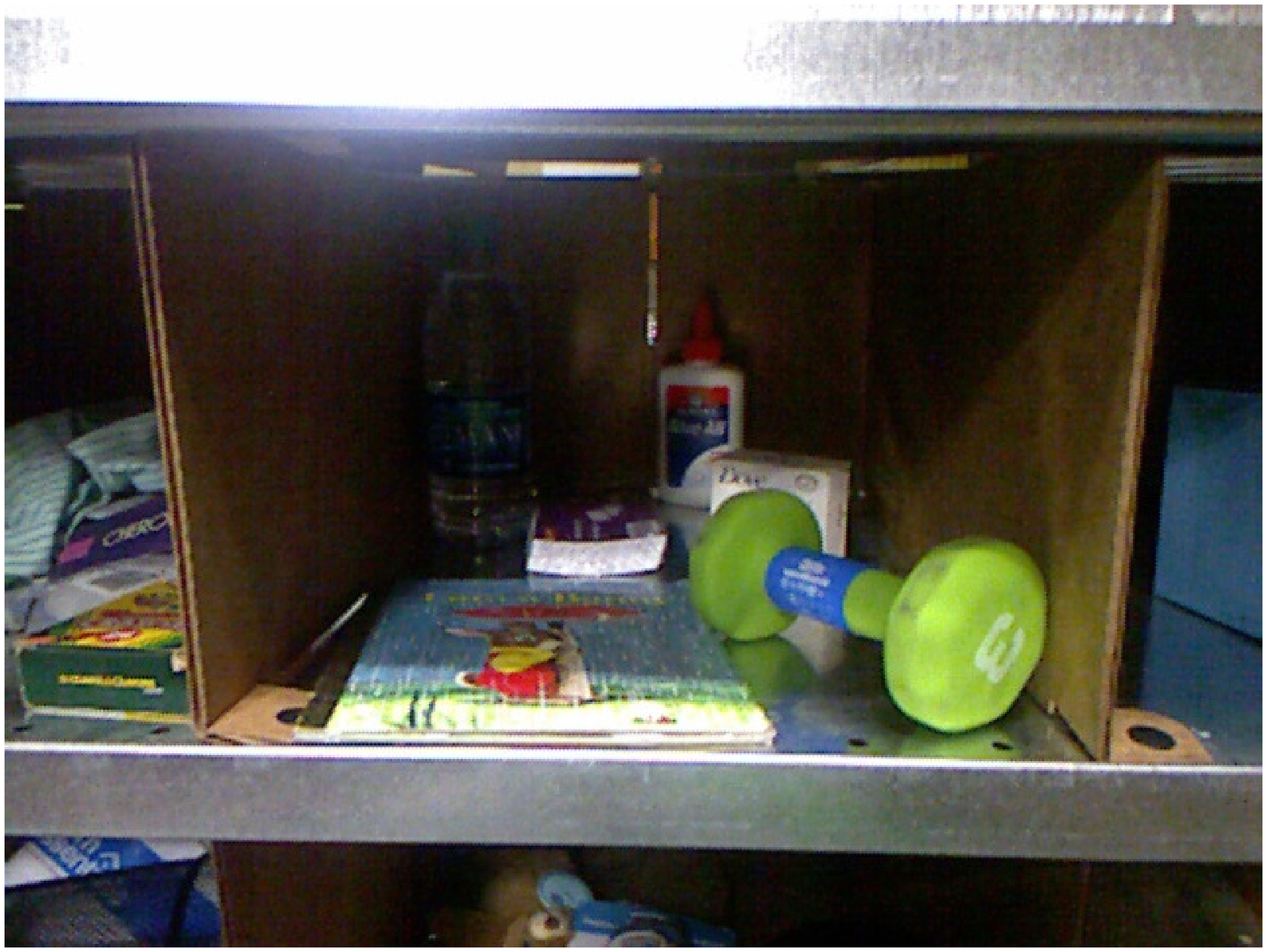} &
    \includegraphics[scale=0.04]{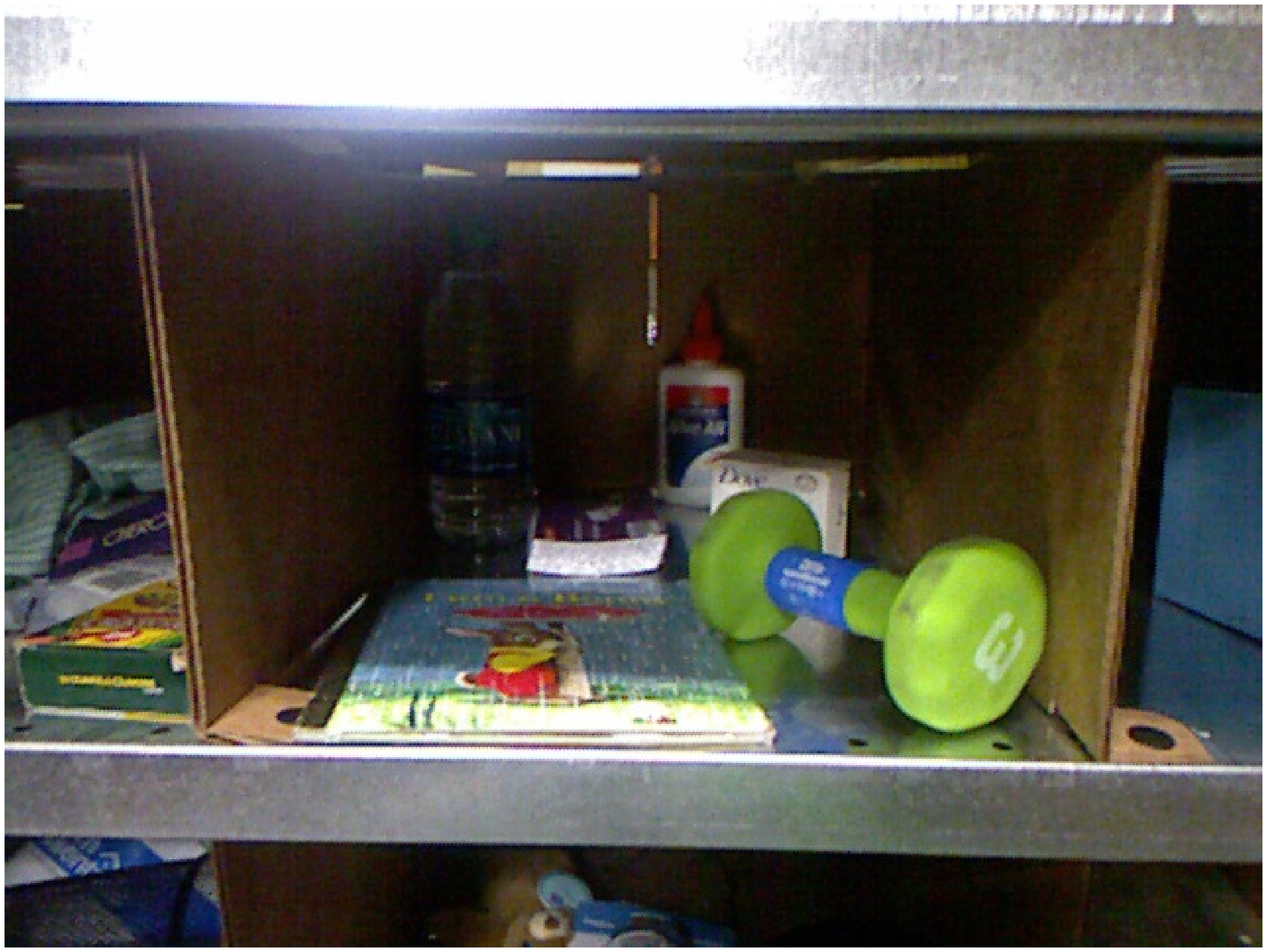} \\
    \includegraphics[scale=0.07]{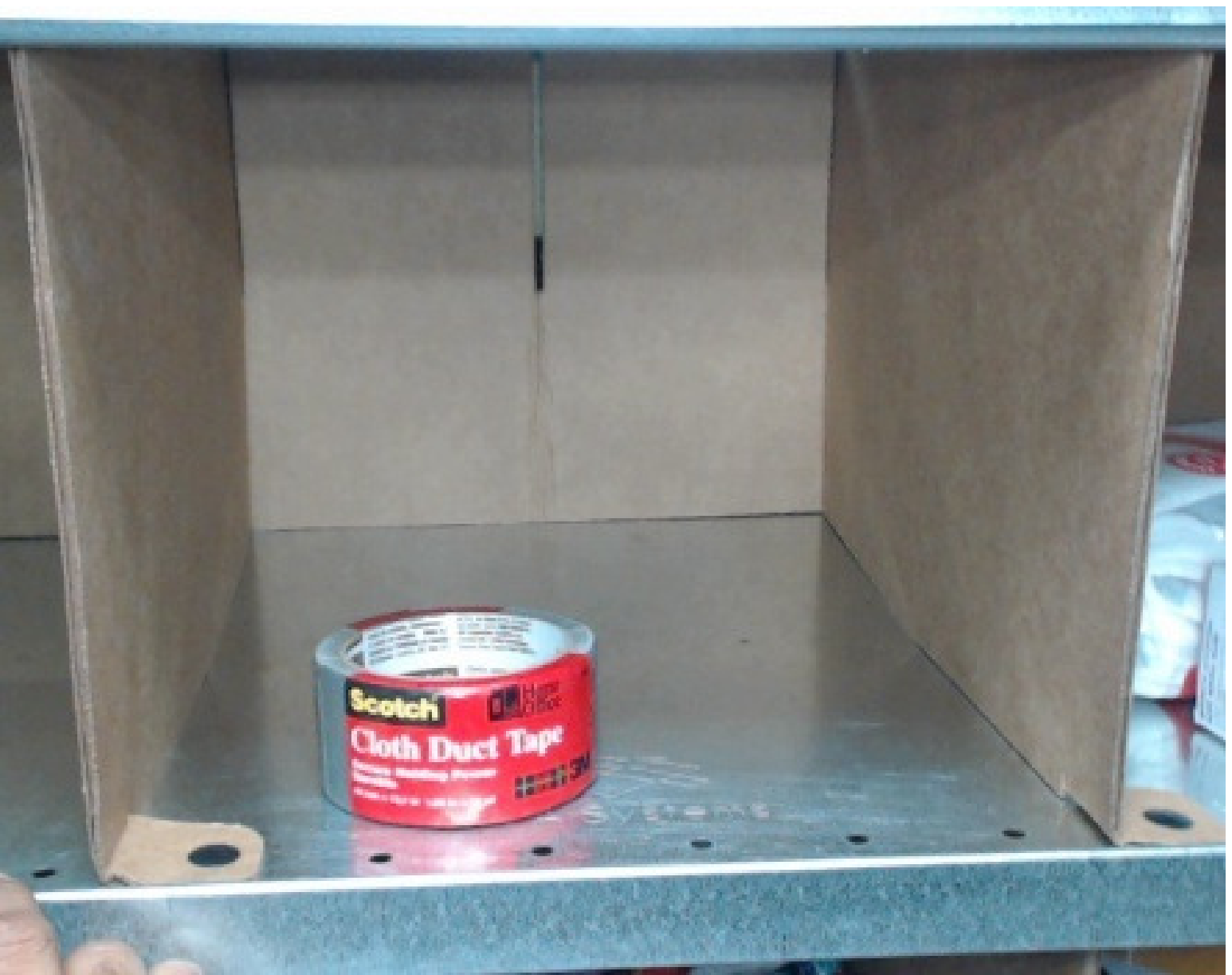} & 
    \includegraphics[scale=0.07]{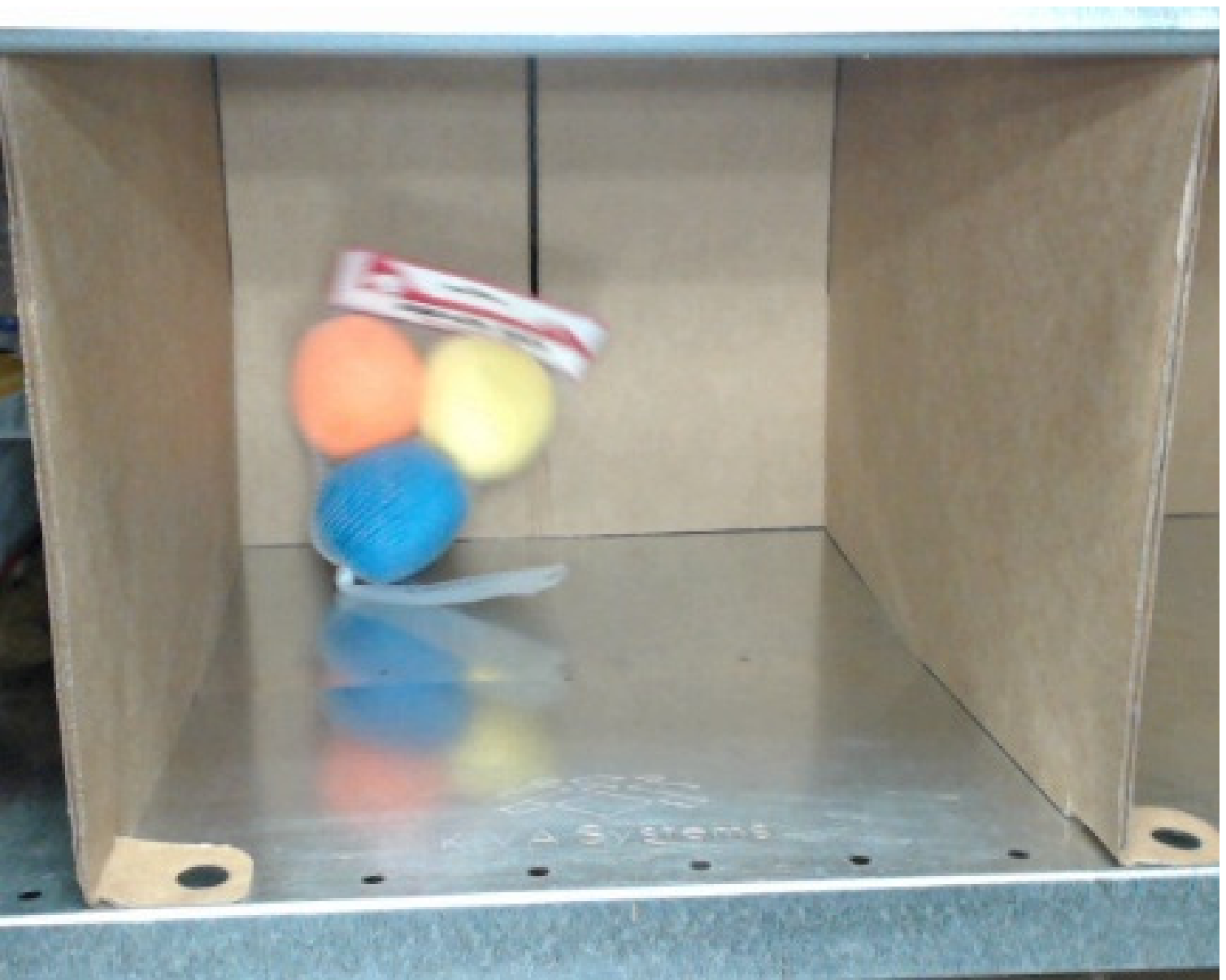} & 
    \includegraphics[scale=0.05]{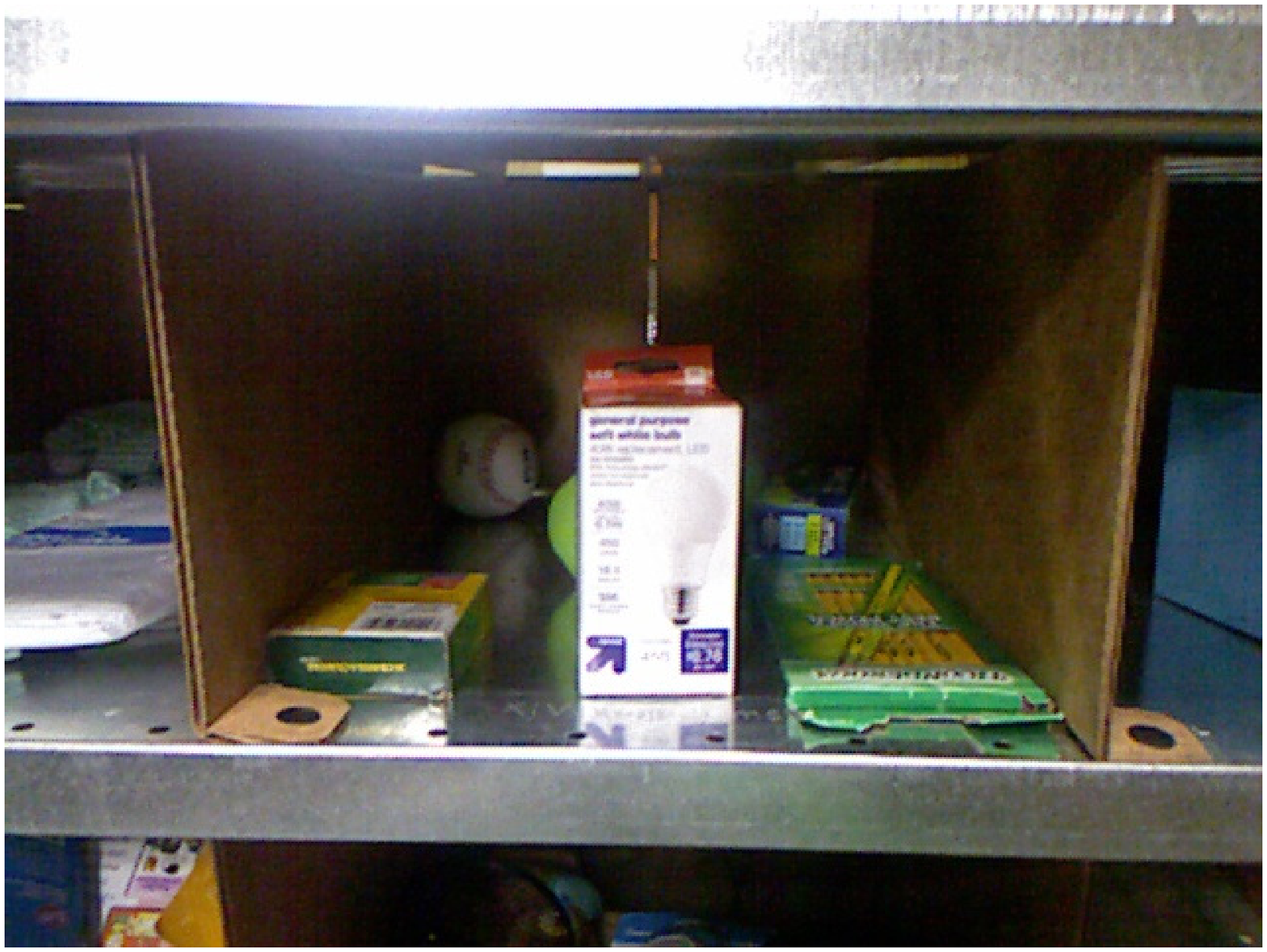} & 
    \includegraphics[scale=0.07]{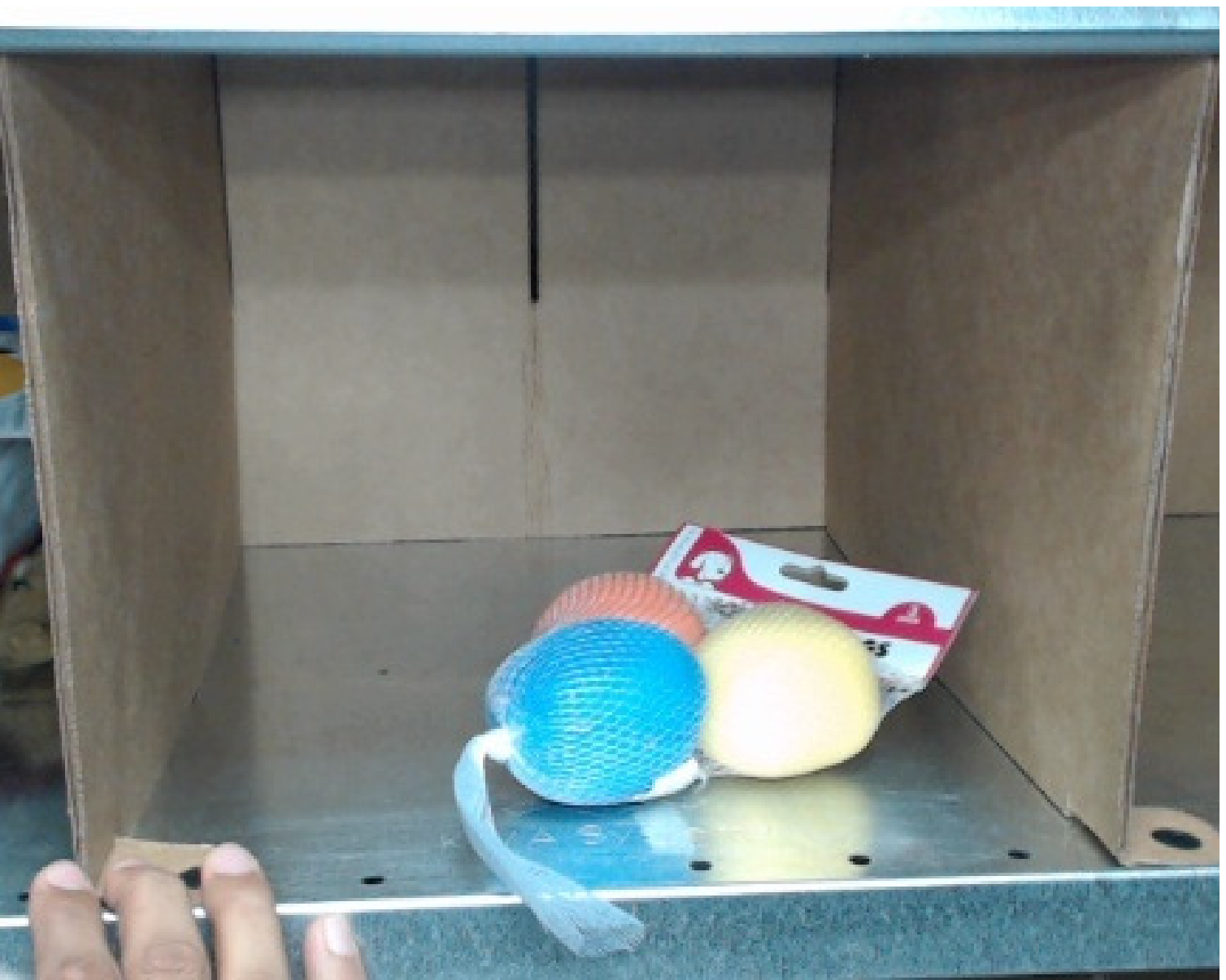} &
    \includegraphics[scale=0.07]{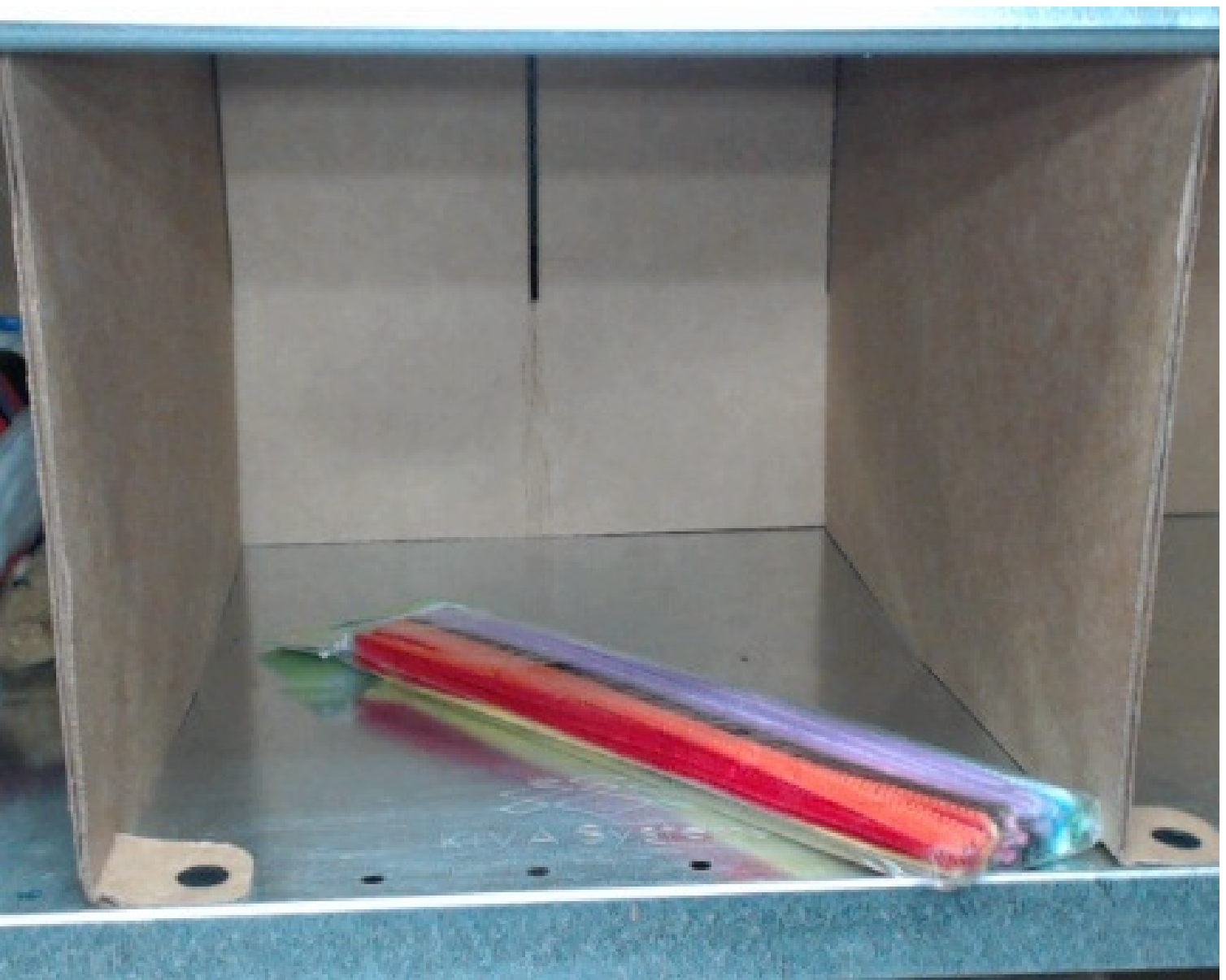} 
\end{tabular}
  \caption{Snapshot of examples used for training the RCNN network.}
  \label{fig:train}
\end{figure}

%

\begin{figure}[!h]
  \begin{center}
    \includegraphics[scale=0.3]{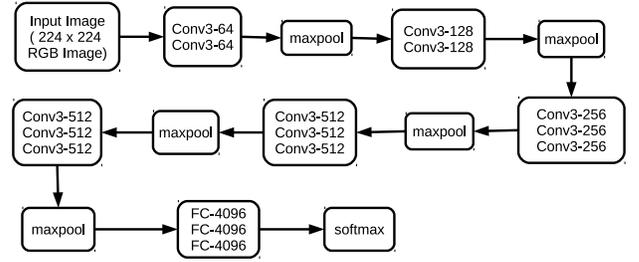}
    \end{center}
  \caption{RCNN layer Architecture used for object detection}
  \label{fig:rcnn-arch}
\end{figure}

\begin{algorithm}
  \caption{Algorithm for object detection technique}
  \label{alg:face_reg}
  \footnotesize{
    \begin{algorithmic}[1]
      \State Calibrate and get the rack transformation matrix using kinect. 

      \For {each object $i$ in the JSON file $i \gets 1 \textrm{ to } N$}

      \State Read JSON file. Get bin number and object identifier. 

      \State Take RGB image of the bin and corresponding 3d Point Cloud
      according to transformation matrix. 

      \State Using trained Faster R-CNN model get the ROI of the object in the RGB input image. 

      \State Select the object ROI with the highest (score) probability 

      \State Apply color and shape backprojection technique in the resultant object ROI using corresponding 3d Point Cloud.

      \State Classify each pixel inside the object ROI using Random Forest 
      classifier based on combined shape and color information. 
      \State Apply adaptive meanshift to find the most probable suction point. 
      \State Find normal at the suction point and the centroid of the object to be picked. 

      \State Instruct motion planner to move to the given position. 

      \State Robot controller 

      \EndFor
    \end{algorithmic}}
  \end{algorithm}

\subsection{Grasping} \label{sec:grasp}

\begin{figure}[!h]
  \centering
  \includegraphics[scale=0.3]{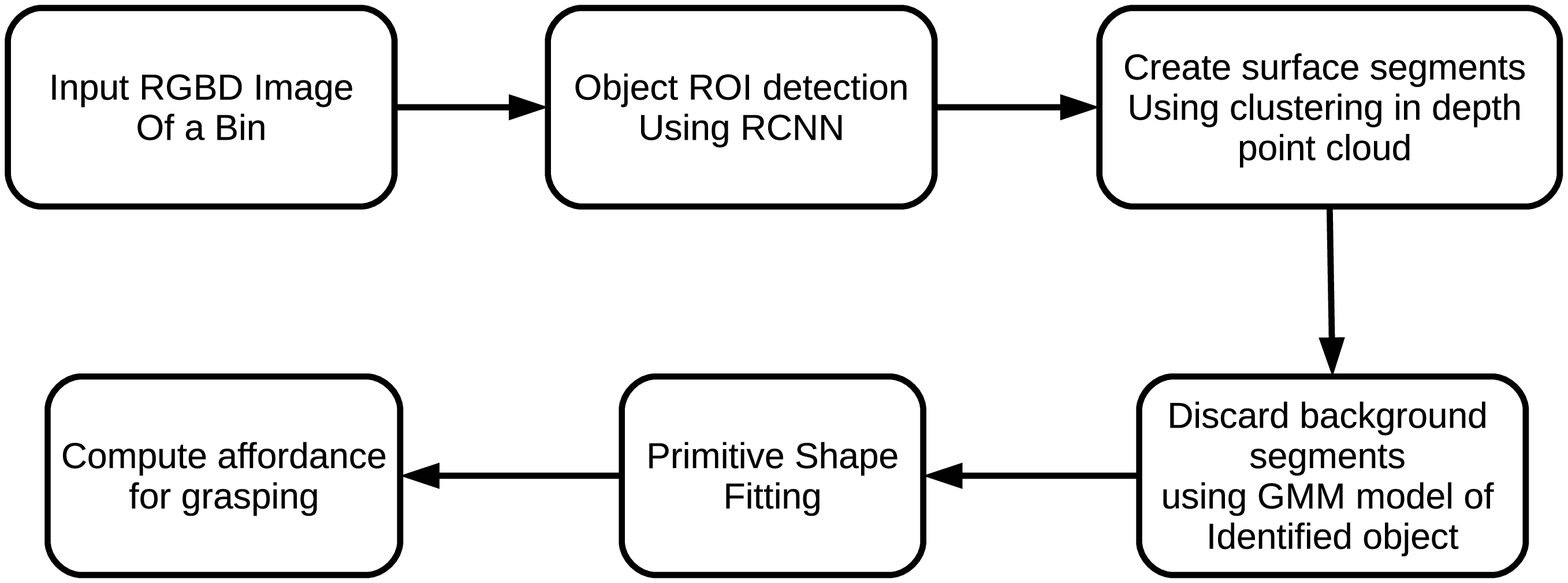}
  \caption{Schematic block diagram for computing grasping affordances
  for objects using RGBD images obtained from a Kinect Camera.}
  \label{fig:grasp-scheme}
\end{figure}

Grasping involves two steps - finding grasp pose for the target object
then making actual motion to make physical contact with the object.
The first part is usually difficult and has attracted a lot of
attention over last couple of decades. There are primarily two
approaches to solve the grasping problem - one of them makes use of
known 3D CAD models \cite{kehoe2013cloud} and the other one which does
not require these CAD models \cite{fischinger2013learning}
\cite{saxena2008robotic} \cite{gualtieri2016high}. The latter method
directly works on the partial depth point cloud obtained from a range
sensor.  Quite recently, researchers are exploring the use of deep
learning networks to detect grasping directly from images
\cite{redmon2015real} \cite{lenz2015deep}.

In this paper we follow the latter approach where we detect the
graspable affordance for the recognized object directly from the RGBD
point cloud obtained from the on-board Kinect camera. Figure
\ref{fig:grasp-scheme} shows the schematic block diagram of the method
employed for grasp pose detection. Input to this scheme is an RGBD
point cloud of the bin viewed by the on-board robot camera. The
bounding box of the query object is obtained by the RCNN based object
recognition system. The bounding box returned by the RCNN module may
have a bigger size than the object itself depending on the amount of
training of the network used. This bounding box acts as the region of
interest (ROI) for finding graspable regions. This bounding box may
contain parts of the background as well other objects in the vicinity.
Within this ROI, a clustering method combined with region growing
algorithm \cite{otto1989region} \cite{schnabel2007efficient} is used
to create several surface segments by identifying discontinuity in the
space of surface normals \cite{mitra2003estimating}
\cite{kovesi2005shapelets} \cite{rabbani2006segmentation}. Apart from
having different surfaces for different objects and backgrounds, there
can be multiple surface segments for the same object. Then the
background segments are separated from the foreground target segments
using a Gaussian Mixture Model (GMM) \cite{zivkovic2004improved}
\cite{zivkovic2004improved} of the identified object using both color
(RGB) and depth curvature information. Once the background segments
are discarded, a primitive shape is identified for the object using
empirical rules based on surface normals, radius of curvature,
alignment of surfaces etc. Once the shape is identified, the best
graspable affordance for the object is computed using a modified
version of the method presented in \cite{zivkovic2004improved}. The
complete details for the method is beyond the scope of this paper and
will be made available as a separate publication. The outcome of this
algorithm is discussed in the experiment section later in this paper. 


\subsection{Motion Planning} \label{sec:motion_plan}

In case of industrial manipulators where one
  does not have access to internal motor controllers, motion planning
  refers to providing suitable joint angle position (or velocity)
  trajectories needed for taking the robot from one pose to another.
  In other words, motion planning becomes a path planning problem
  which is about finding way point poses between the current pose and the
  desired end-effector pose.  The problem of generating collision
  free paths for manipulators with increasingly larger number of links
  is quite complex and has attracted considerable interest over last
  couple of decades. Readers can refer to \cite{siciliano2010robotics}
  for an overview of these methods.  These methods could be primarily
  divided into two categories - local and global. Local methods start
  from a given initial configuration and step towards final
  configuration by using local information of the workspace. 
  Artificial potential field-based methods \cite{koditschek1989robot}
  \cite{barraquand1992numerical} \cite{hwang1992potential} are one
  such category of methods where the search is guided along the
  negative gradient of  artificially created vector fields.  On the
  other hand, global methods use search algorithms over the entire
  workspace to find suitable paths. Some of the examples of global
  methods are probabilistic roadmaps (PRM) \cite{malone2014efficient}
  \cite{kavraki1996probabilistic}and Cell-decomposition based C-Space
  methods \cite{avnaim1988practical} \cite{lingelbach2004path}.
  Rapidly exploring random tree (RRT) \cite{lavalle2000rapidly} is one
  of the most popular PRM method used for path planning.  Many of
  these state-of-the-art algorithms are available in the form of the
  open motion planning library (OMPL) \cite{sucan2012open} which has
  been integrated into several easy-to-use software packages like
  Moveit!  \cite{moveit}, Kautham \cite{rosell2014kautham} and
  OpenRave \cite{diankov2008openrave}.

\begin{figure}[!h]
  \centering
  \begin{tabular}{cc}
    \subfigure[Home Pose]{\label{f:home}\includegraphics[scale=0.25]{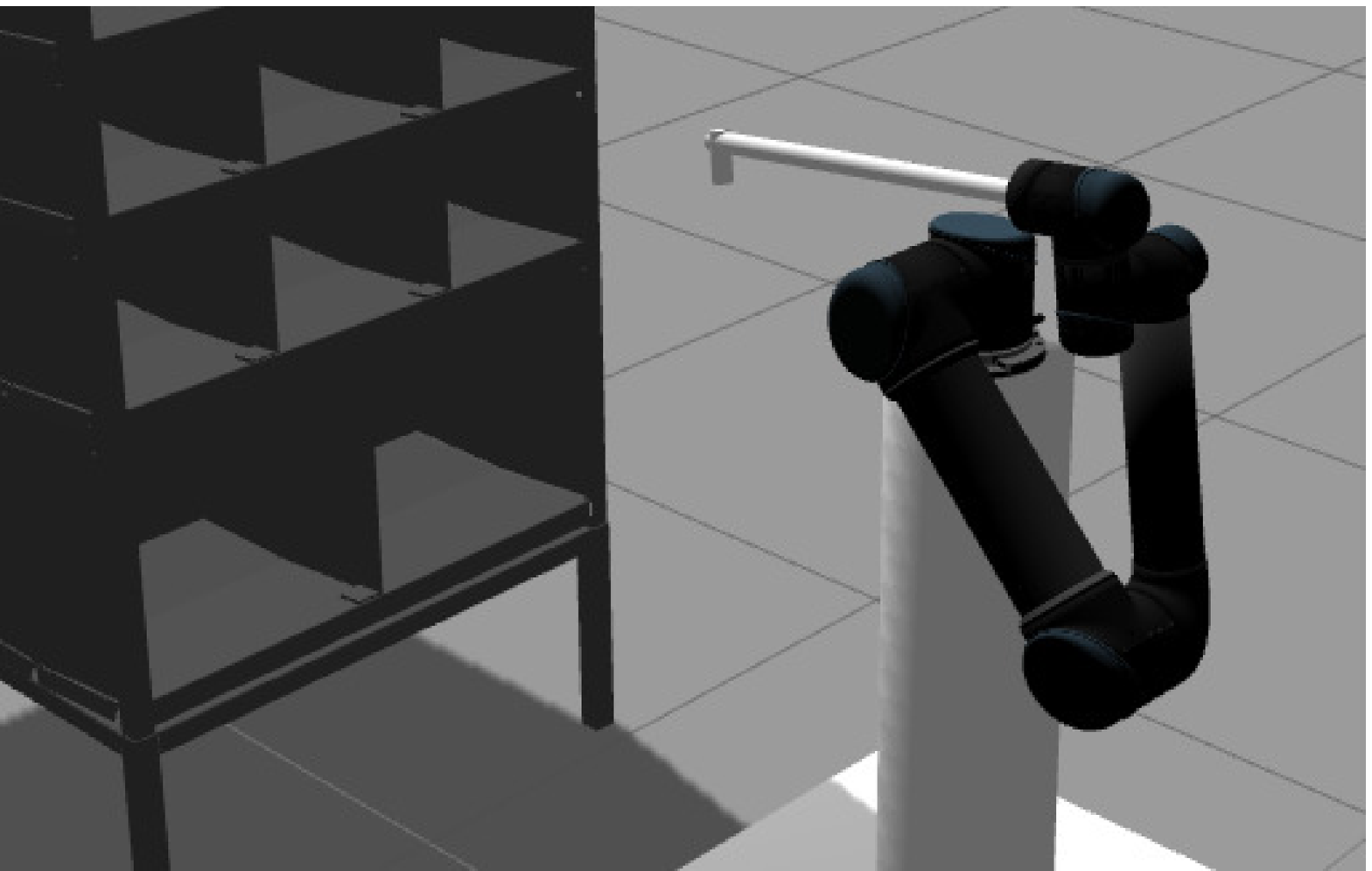}} & 
    \subfigure[Bin View Pose]{\label{f:bvp}\includegraphics[scale=0.2]{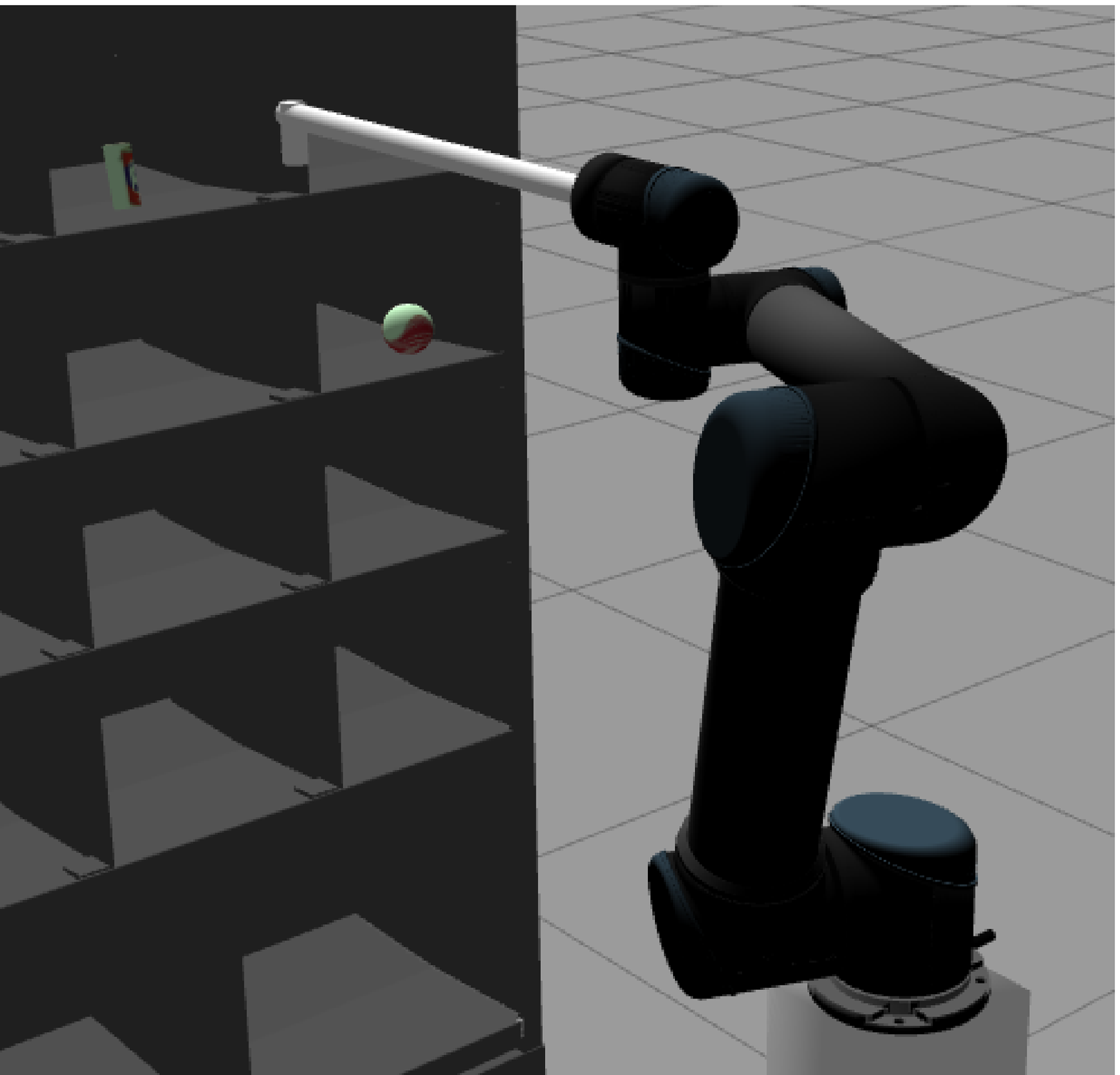}} \\ 
    \subfigure[Desired Pose]{\label{f:desired}\includegraphics[scale=0.2]{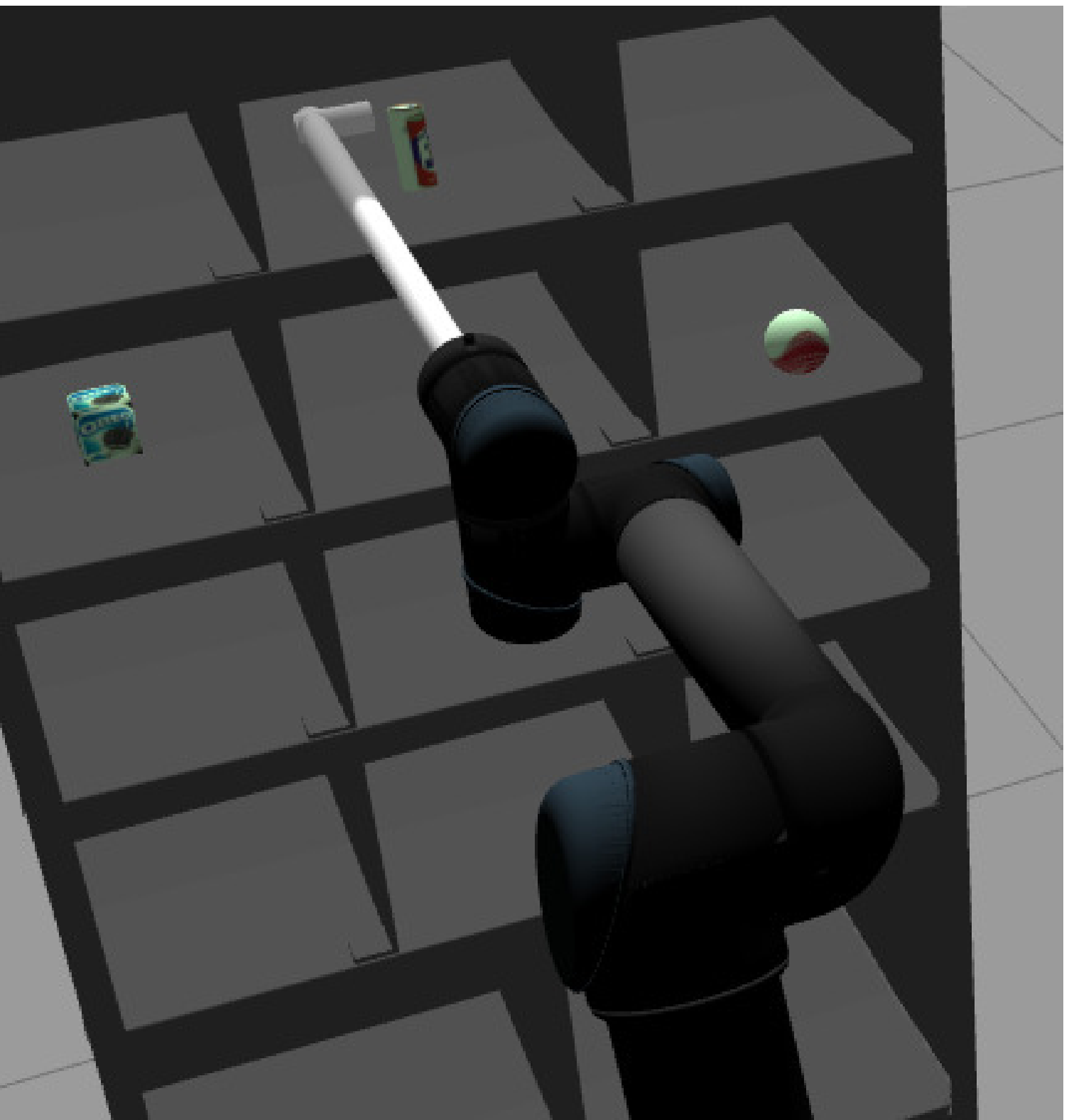}} & 
    \subfigure[End-effector pose traversal]{\label{f:motion}\includegraphics[scale=0.09]{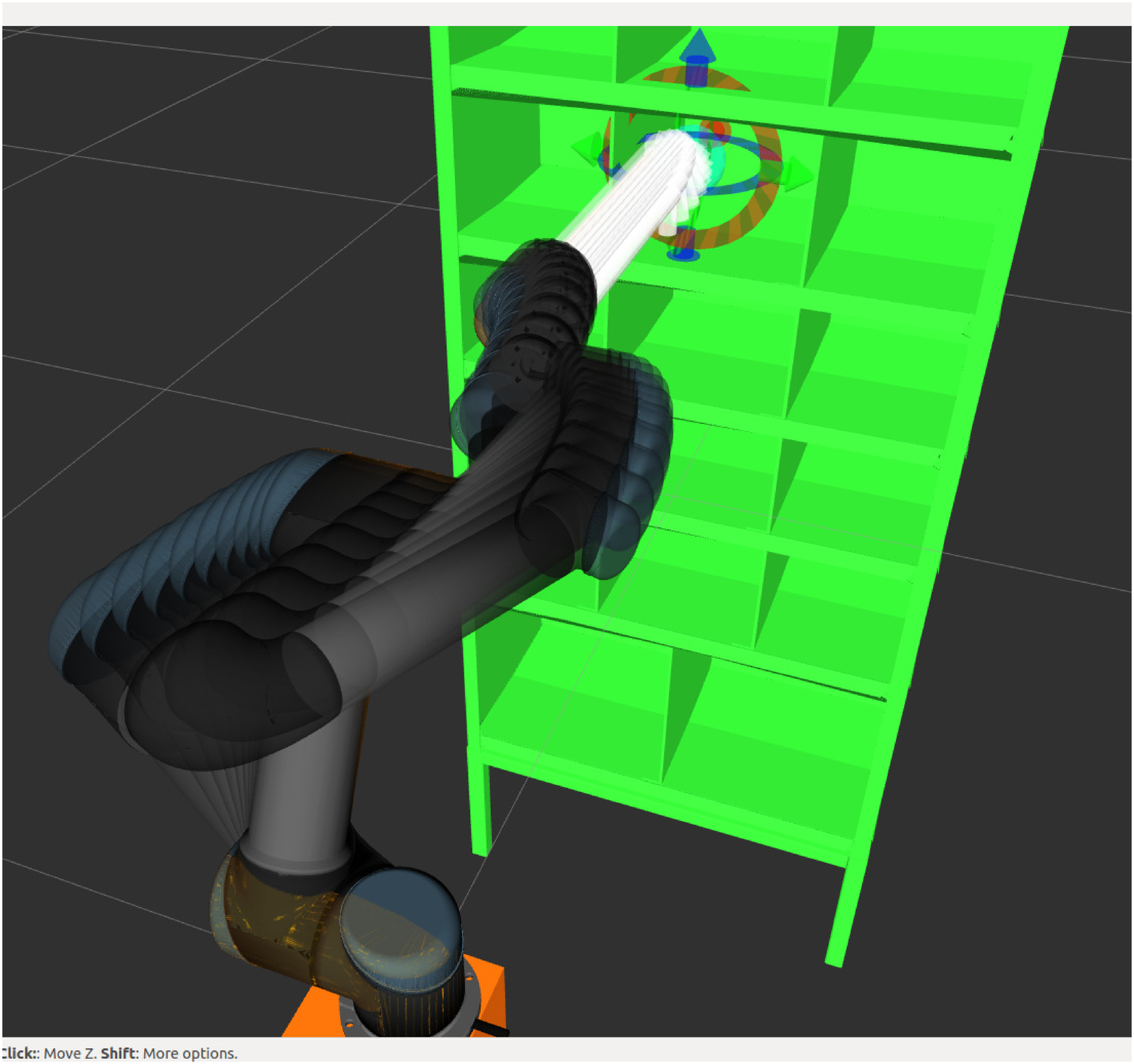}} 
  \end{tabular}
  \caption{Simulating motion planning using Moveit and Gazebo. (a) In
  idle state the robot stays at home pose; (b) On receiving queried
bin number, the robot moves to the Bin View pose where it takes an
image of the bin; (c) the required desired for picking the can in
picture is obtained after processing the image to identify target
item; (d) The end-effector trajectory from bin view pose to the
desired pose is obtained using RRT motion planning algorithm available
with Moveit.}
  \label{fig:gazebosim} 
\end{figure}

In this paper, we have used \emph{Moveit!} package available with ROS
\cite{quigley2009ros} for building motion planning algorithms for UR5
robot manipulator. The simulation is carried out using Gazebo
\cite{roberts2003constructing} environment. Some of the snapshots of
the robot are shown in Figure \ref{fig:gazebosim}. The robot starts
its operation from its home pose which is shown in Figure
\ref{f:home}. The pose is so selected so that the entire rack is
visible from the on-board Kinect camera (not shown in the picture).
This image is used for system calibration process as described in
Section \ref{sec:calib} and \ref{sec:rack_detect} respectively.  Once
the bin number is obtained from the JSON query file, the robot moves
to the bin view pose shown in Figure \ref{f:bvp}. At this pose, a
close up picture of the bin is taken by the Kinect camera mounted on
the wrist of the robot. Every bin has a pre-defined bin view pose
which is selected so as to get a good view of the bin. The desired
pose necessary for picking an item in the bin is obtained from the
object recognition and grasping algorithm. One such desired pose is
shown in Figure \ref{f:desired}. The robot configuration trajectory
generated by the motion planning algorithm is shown in Figure
\ref{f:motion}. It also performs collision avoidance by considering
the rack (shown in green color) as an obstacle.

\begin{figure}[!h]
  \centering
  \includegraphics[scale=0.3]{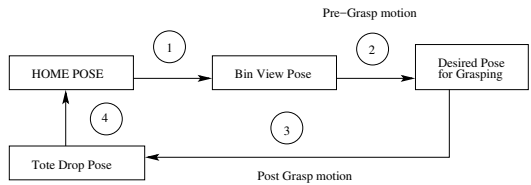}
  \caption{Sequence of steps for motion planning for a picking task. }
  \label{fig:mpseq}
\end{figure}

\begin{figure}[!h]
  \centering
  \begin{tabular}{cc}
    \subfigure[Avoiding collision with Rack]{\label{f:rackob}\includegraphics[scale=0.2]{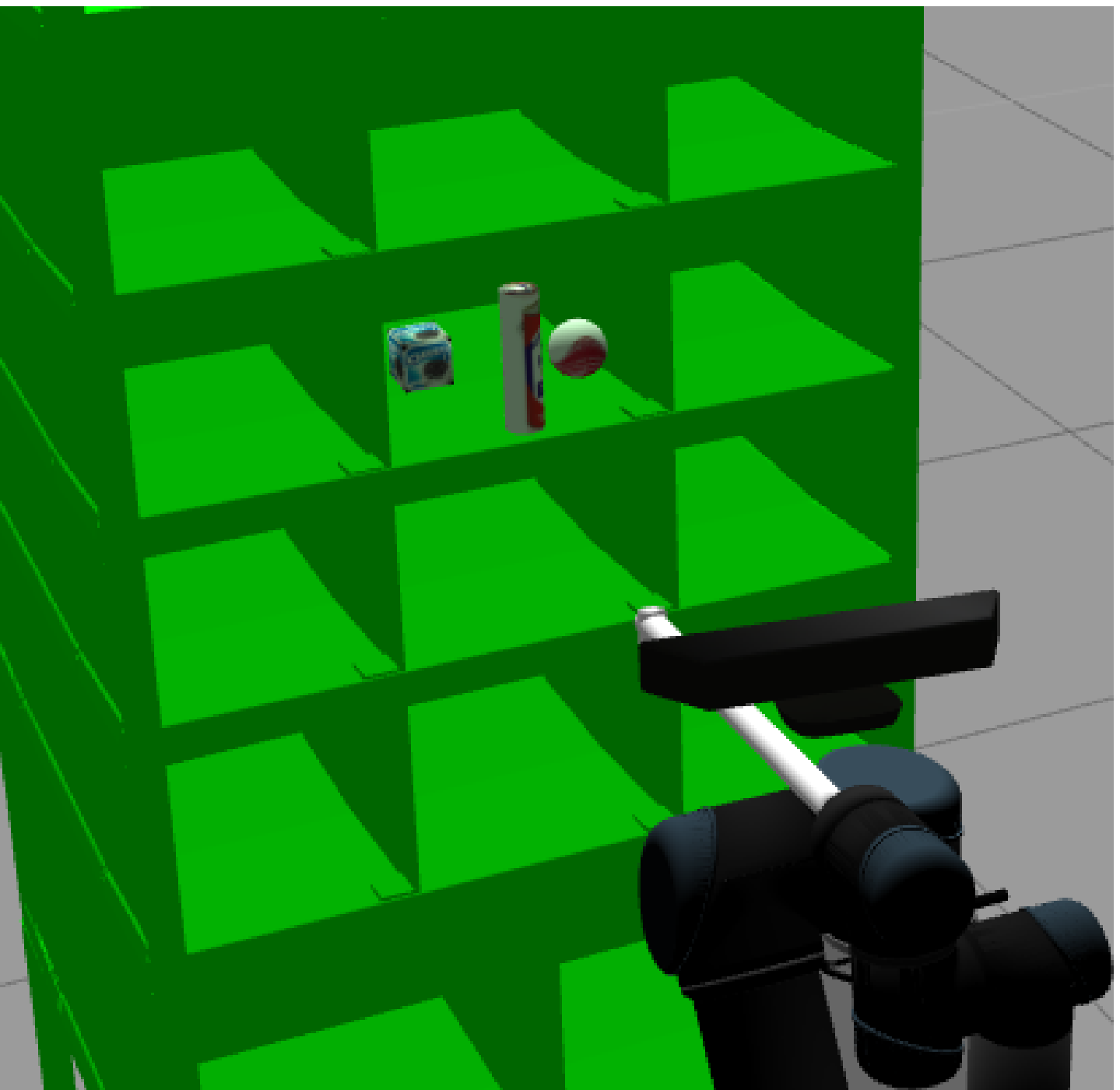}} &
    \subfigure[Avoiding collision with Objects using Octomap]{\label{f:objob}\includegraphics[scale=0.3]{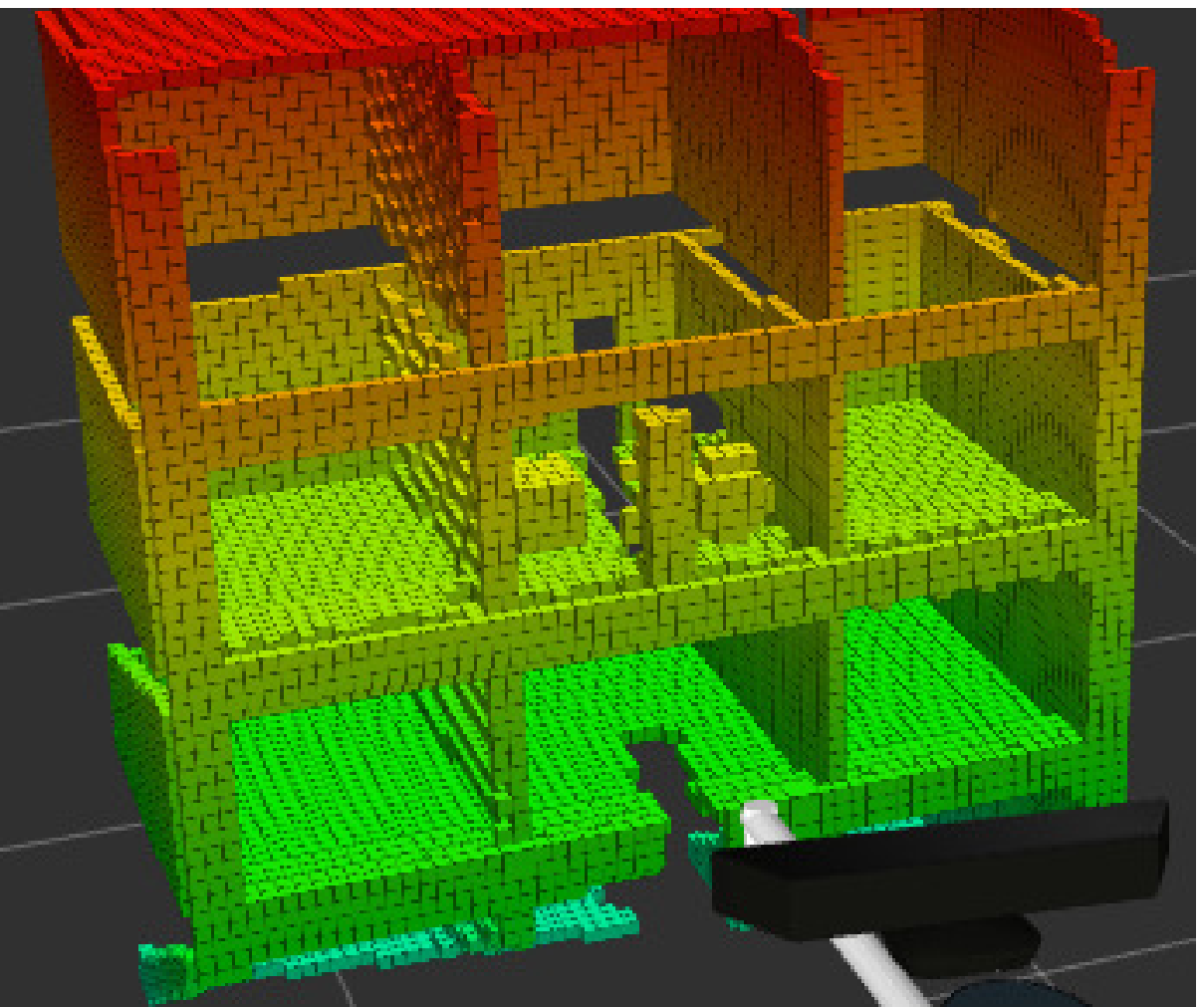}} 
  \end{tabular}
  \caption{Collision avoidance during motion planning. In (a), the
  green color shows the obstacle created using primitive shapes. In
(b) Octomap is used to create 3D voxels for each object which are
considered as obstacles during motion planning.}
  \label{fig:colav}
\end{figure}

The sequence of steps involved in carrying out motion planning for the
pick task is shown in Figure \ref{fig:mpseq}. It primarily involves
four steps. The motion for segments 1 and 4 are executed using
pre-defined joint angles as these poses do not change during the pick
task. However, the motion planning for segment 2 (pre-grasp motion)
and segment 3 (post-grasp motion) is carried out using RRT algorithm
during the run-time. This is because the desired pose required for
grasping the object will vary from one object to another and hence,
the paths are required to be determined in the run-time. The on-line
motion planning uses flexible collision library (FCL)
\cite{pan2012fcl} to generate paths for robot arm that avoid
collision with the rack as well as the objects surrounding
the target item. In order to avoid collision with the rack, the bin
corners obtained from the rack detection module, described in Section
\ref{sec:rack_detect}, is used to define primitive boxes for each wall
of the bin. These primitive boxes, shown in green color in Figure
\ref{f:rackob}, are then treated as obstacles in the motion planning
space. Similarly, the collision with other objects in the bin is
achieved by creating 3D occupancy map called OctoMap
\cite{wurm2010octomap} which converts point cloud into 3D voxels. This
OctoMap feed is added to the Moveit motion planning scene and FCL is
used for avoiding collision with the required objects. The OctoMap
output of a 3D point cloud is shown in Figure \ref{f:objob}.

\subsection{End-effector Design} \label{sec:gripper}

\begin{figure}[!h]
  \centering
  \includegraphics[scale=0.3]{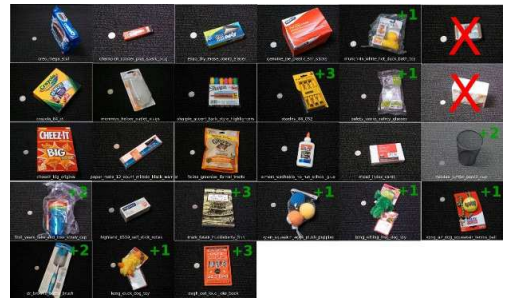}
  \caption{Typical items that were to be picked or stowed in the Amazon Picking Challenge}
  \label{fig:apc_items}
\end{figure}

Amazon Picking Challenge focusses on solving the challenges involved
in automating picking and stowing various kinds of retail goods using
robots. These items include both rigid as well as deformable objects
of varied shape and size. The maximum specified weight was about 1.5
Kgs. A snapshot of typical objects that were specified for the APC
2015 event \cite{wurman2016amazon} is shown in Figure
\ref{fig:apc_items}. The authors in \cite{rennie2016dataset} provide a
rich dataset for these objects which can be used for developing
algorithms for grasping and pose estimation. It was necessary to
design an end-effector which could grasp or pick all kinds of objects.
We designed two kinds of end-effectors to solve this problem which are
described below. 

\subsubsection{Suction-based end-effector}

\begin{figure}[!h]
  \centering
  \begin{tabular}{c}
    \subfigure[Suction-based end-effector]{\label{f:ee}\includegraphics[scale=0.05]{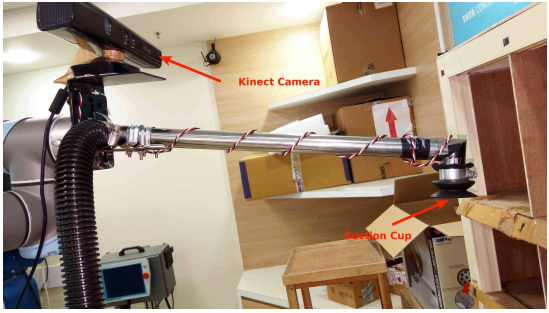}} \\ 
    \subfigure[Close-up view of Suction Cup]{\label{f:cup}\includegraphics[scale=0.3]{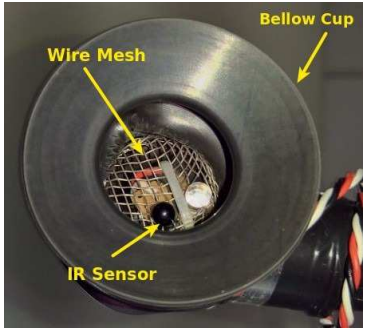}}  
  \end{tabular}
  \caption{The end-effector using suction cup for picking objects. The
    suction cup uses IR Sensor to detect if an object has been picked
    up successfully. The wire mesh prevents smaller or softer items
    getting sucked into the system.} 
    \label{fig:sucker} 
  \end{figure}

This end-effector essentially makes use of a vacuum suction system to
pull the objects towards it and hold it attached to the end-effector.
Such a system was successfully used by the TU-Berlin team
\cite{eppnerlessons} in the APC 2015 event where they came out as
clear winners. A normal household cleaner could be used as the robot
end-effector. It was sufficient only to make the nozzle end of the
vacuum suction to reach any point on the object to be picked
irrespective of its orientation. However, the suction can work only if
it makes contact with the object with sufficient surface area
necessary to block the cross-section of the suction pipe. One such
system designed for our system is shown in Figure \ref{fig:sucker}.
The cross section of the suction pipe should be big enough to generate
necessary force to lift the object. It can not be used for picking
small objects having smaller cross section area, for instance, a pen
or a pencil or a metal dumbbell having narrow cylindrical surface. A
more close-up view of the suction cup is shown in Figure \ref{f:cup}.
A set of IR sensors are used inside the bellow cup in order to detect
the successful pick operation for a given object. A fine mesh is
embedded inside the cup to prevent finer and soft materials like cotton or
clothes from getting sucked into the tube and thereby, damaging the
end-effector.

\subsubsection{Combining Gripping with Suction}

\begin{figure}[!h]
  \centering
  \begin{tabular}{cc}
    \subfigure[Front side]{\label{f:front}\includegraphics[scale=0.06]{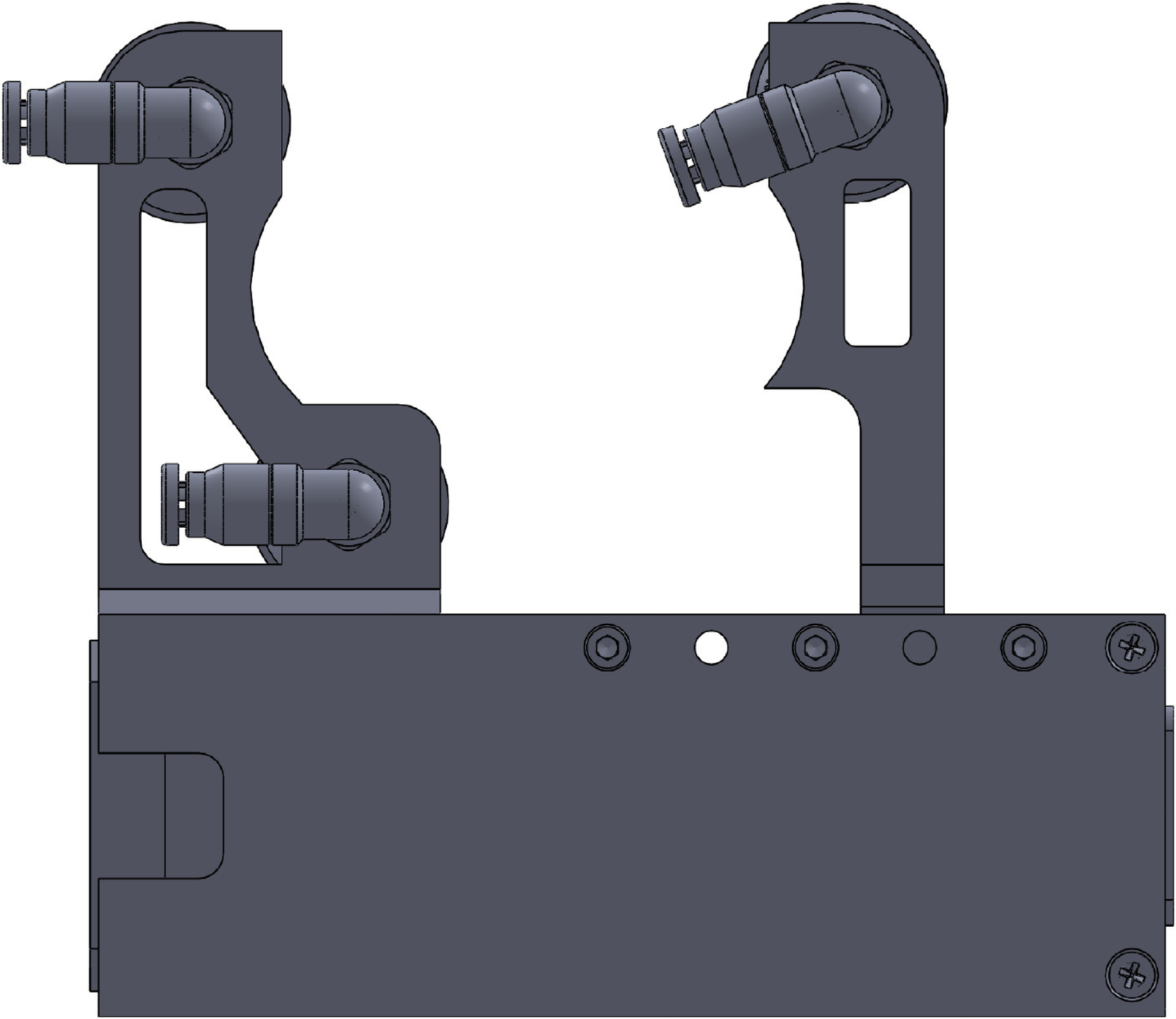}} & 
    \subfigure[Back Side]{\label{f:back}\includegraphics[scale=0.06]{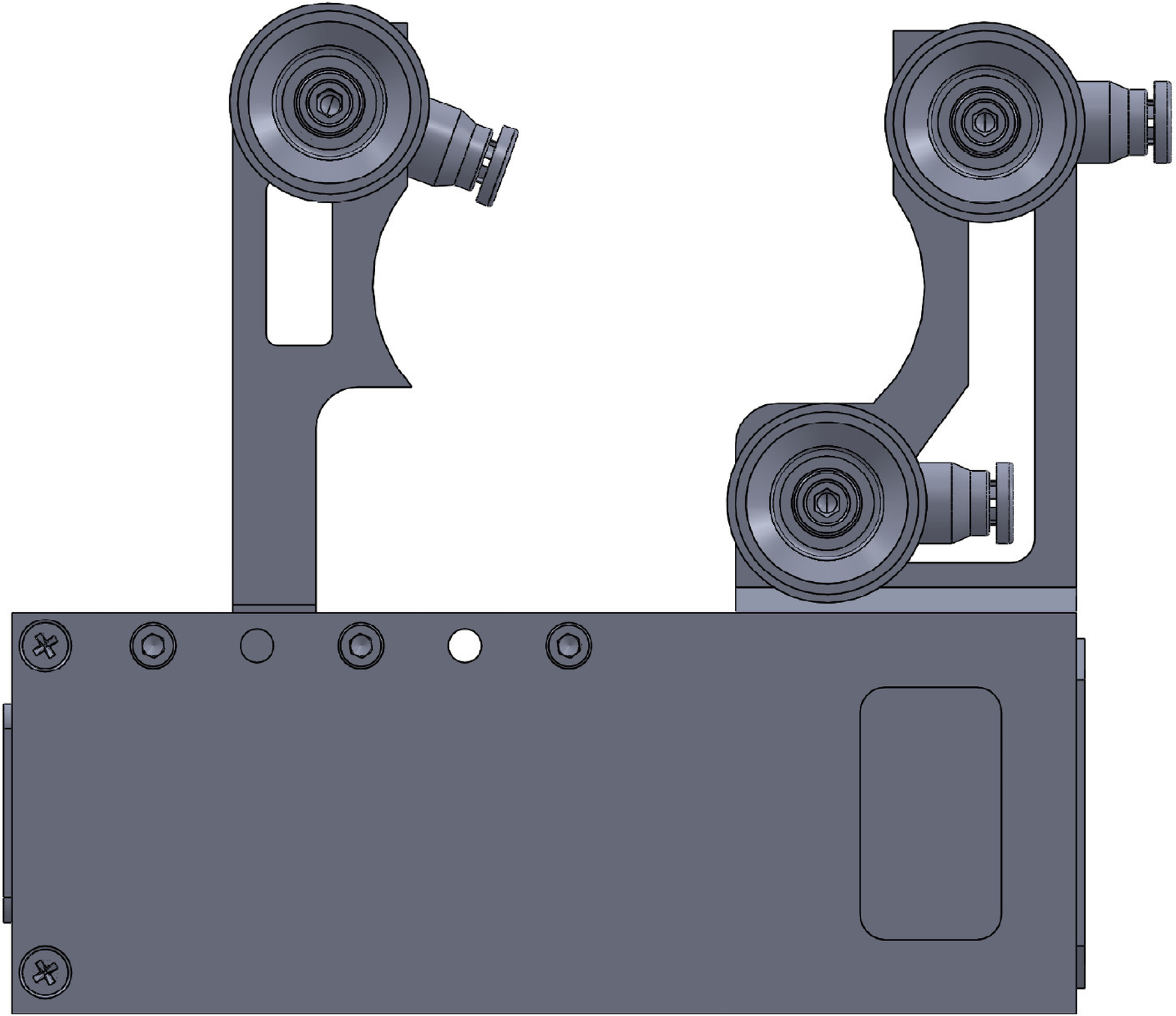}} \\
    \subfigure[Right Side]{\label{f:right}\includegraphics[scale=0.08]{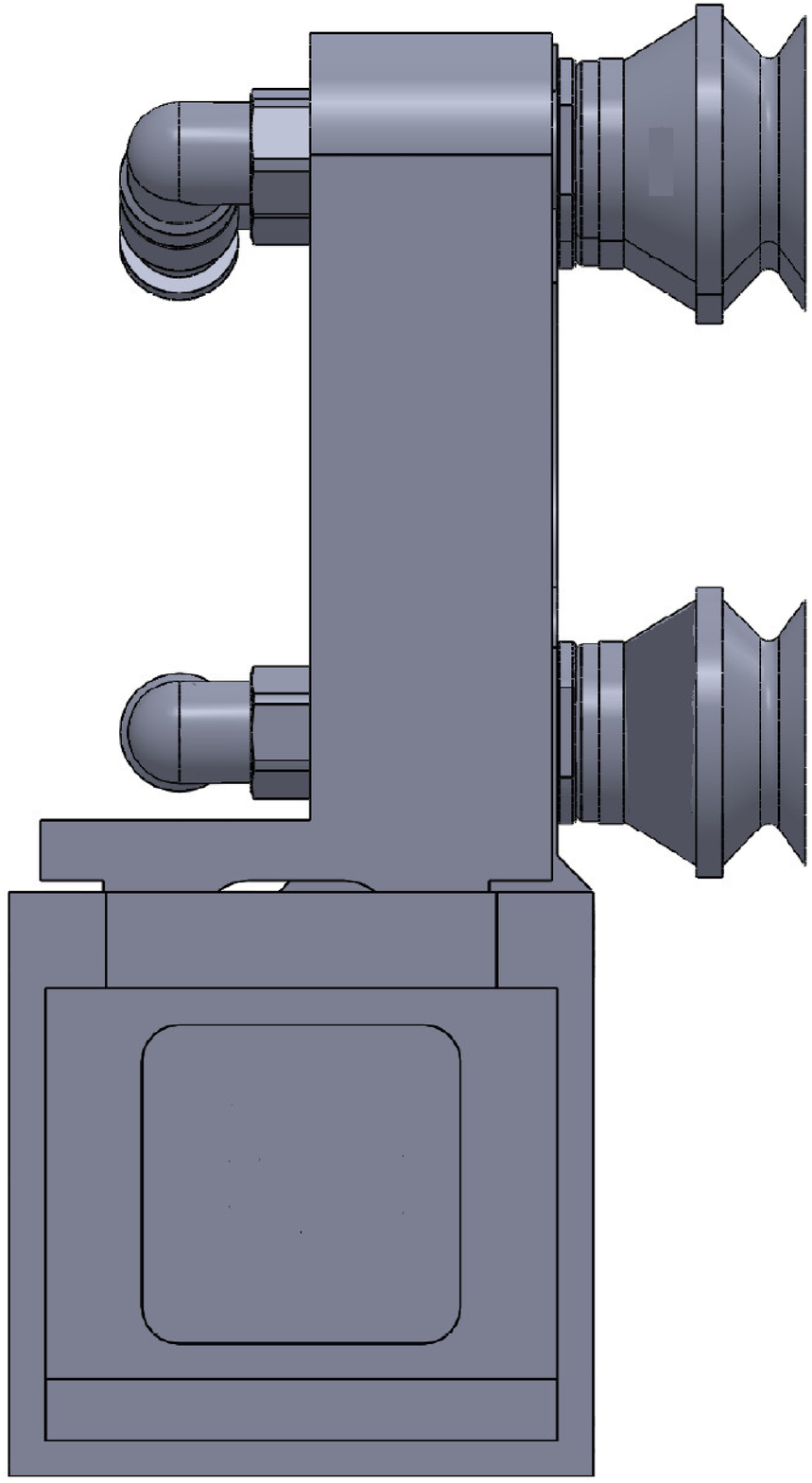}} & 
    \subfigure[Isometric View]{\label{f:iso}\includegraphics[scale=0.08]{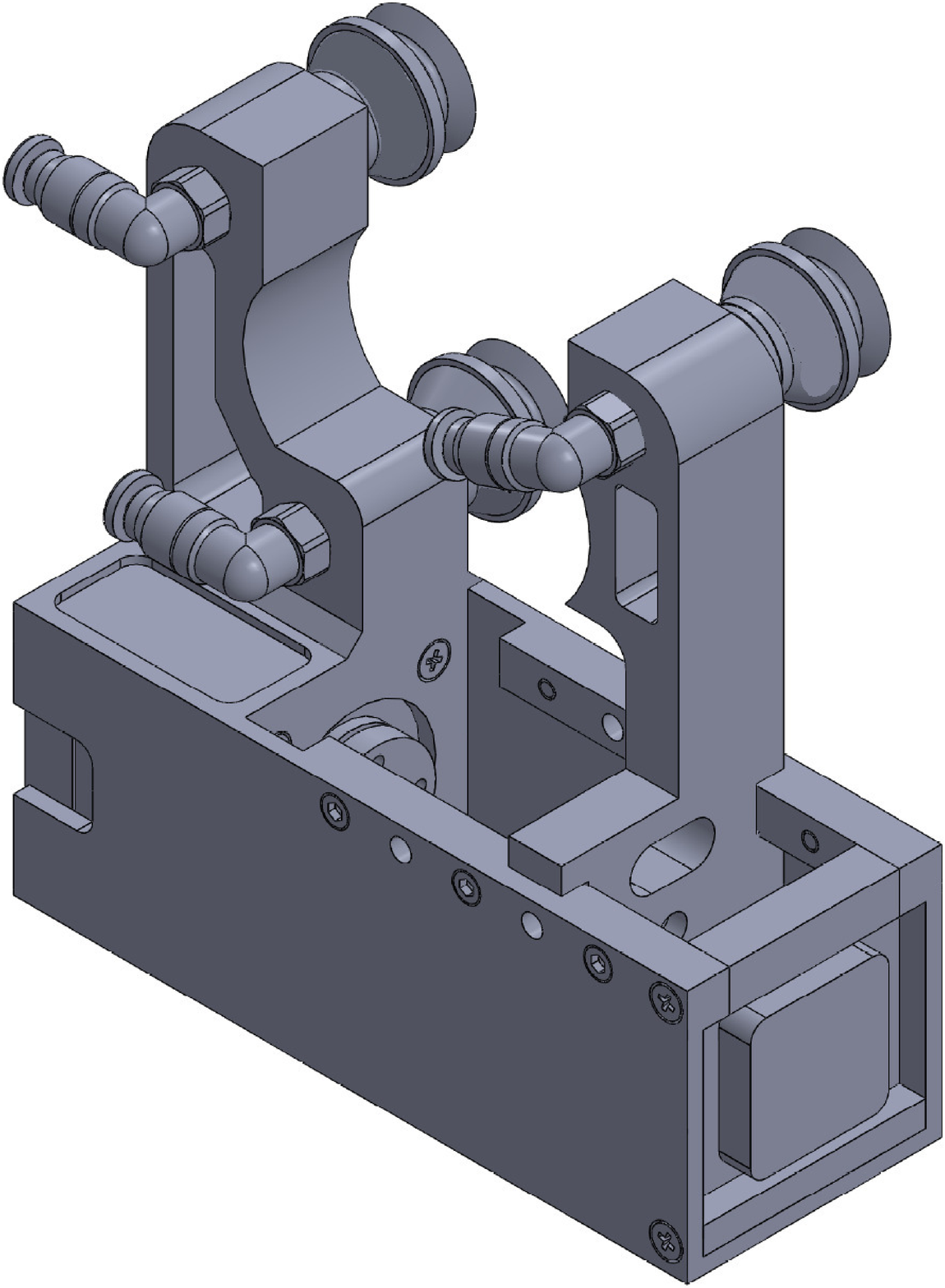}}
  \end{tabular}
  \caption{Novel Gripper design that combines gripping with suction}
  \label{fig:gripper_drwg}
\end{figure}

\begin{figure}[!h]
  \centering
  \begin{tabular}{cc}
    \subfigure[Actual Gripper]{\label{f:grip1}\includegraphics[scale=0.02]{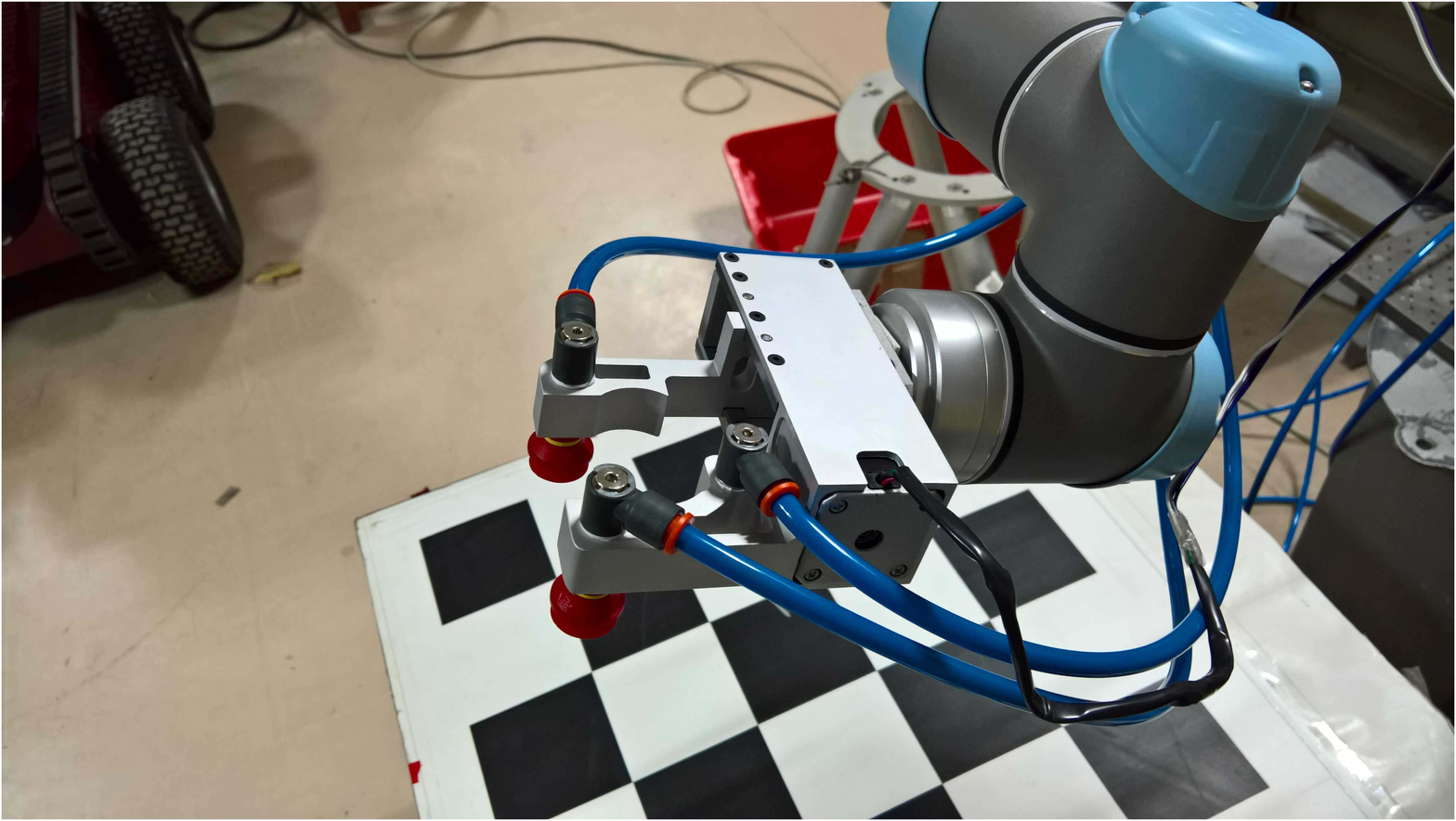}} &
    \subfigure[Gripper with pneumatic valve assembly]{\label{f:grip2}\includegraphics[scale=0.02]{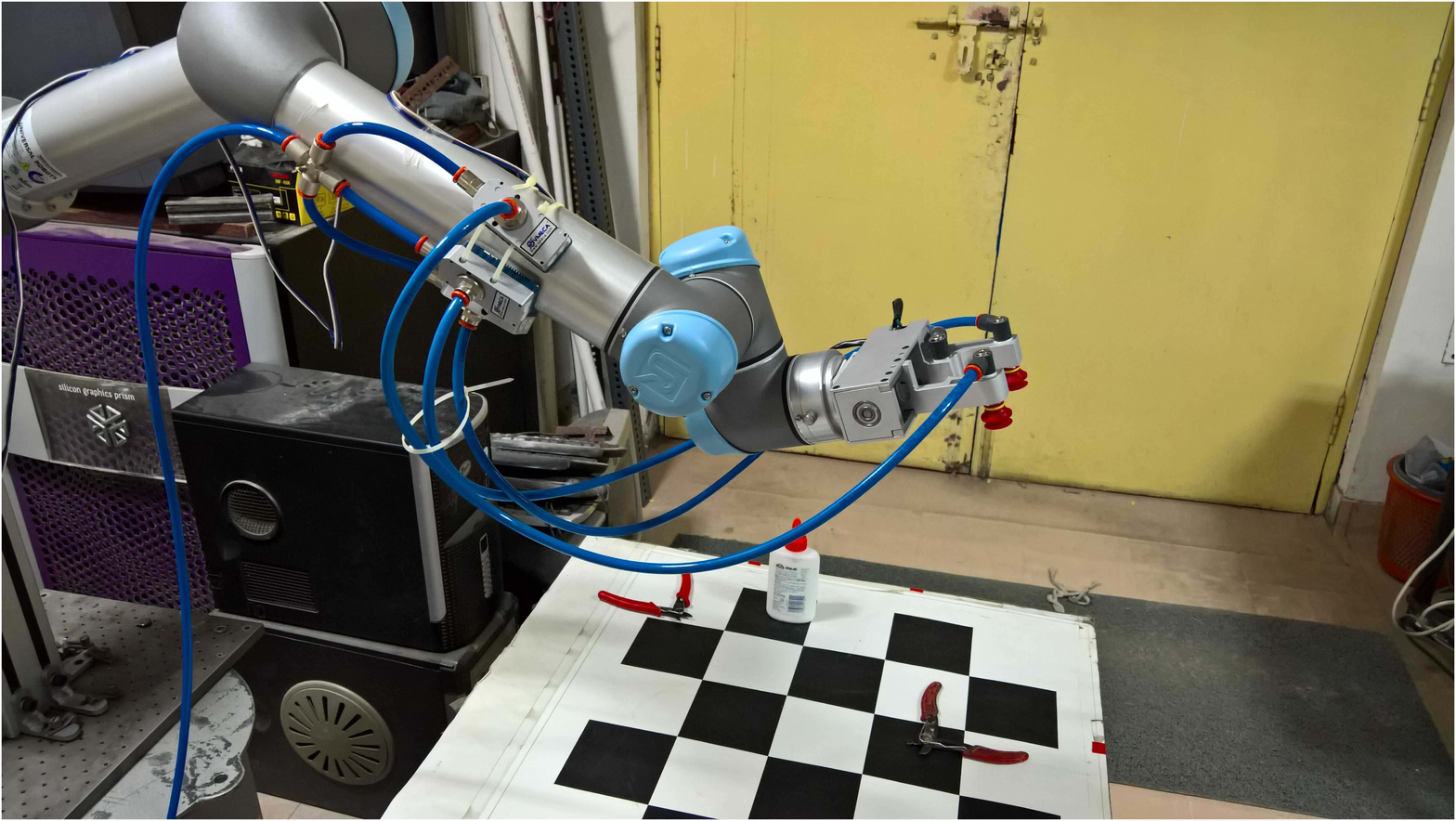}} 
  \end{tabular}
  \caption{Actual Gripper after fabrication and assembly.}
  \label{fig:gripper_actual}
\end{figure}

This particular design was employed by the MIT team
\cite{yu2016summary} during the APC 2015 event. In this design, they
combined a parallel jaw gripper with a suction system. They also used
a compliant spatula to emulate scooping action. In this design,
suction was used for picking only very few items which could not be
picked by the parallel jaw gripper and hence, a employed single bellow
cup capable of picking smaller items.  Inspired by this design, we
developed a similar hybrid gripper by combining suction cups with a
two finger gripper as shown in Figure \ref{fig:gripper_drwg}. This
gripper was designed to lift a weight of around 2 Kgs. It uses a
single actuator with rack pinion mechanism to achieve linear motion
between the fingers. The stationary finger houses two bellow cups
while the moving finger houses one bellow cup. Hence, it is possible
to pick bigger objects through suction by increasing the space between
the fingers. The bellow cups are actuated by pneumatic valves that
create suction by diverting pressurized air through them. The actual
gripper assembly with pneumatic valves and pipes are shown in Figure
\ref{fig:gripper_actual}(a) and (b) respectively.  The working of the
gripper is demonstrated in the experiment section.

\subsection{Robot Manipulator Model} \label{sec:rob_model}

\begin{figure}[!h] \centering \begin{tabular}{c}
    \includegraphics[scale=0.1]{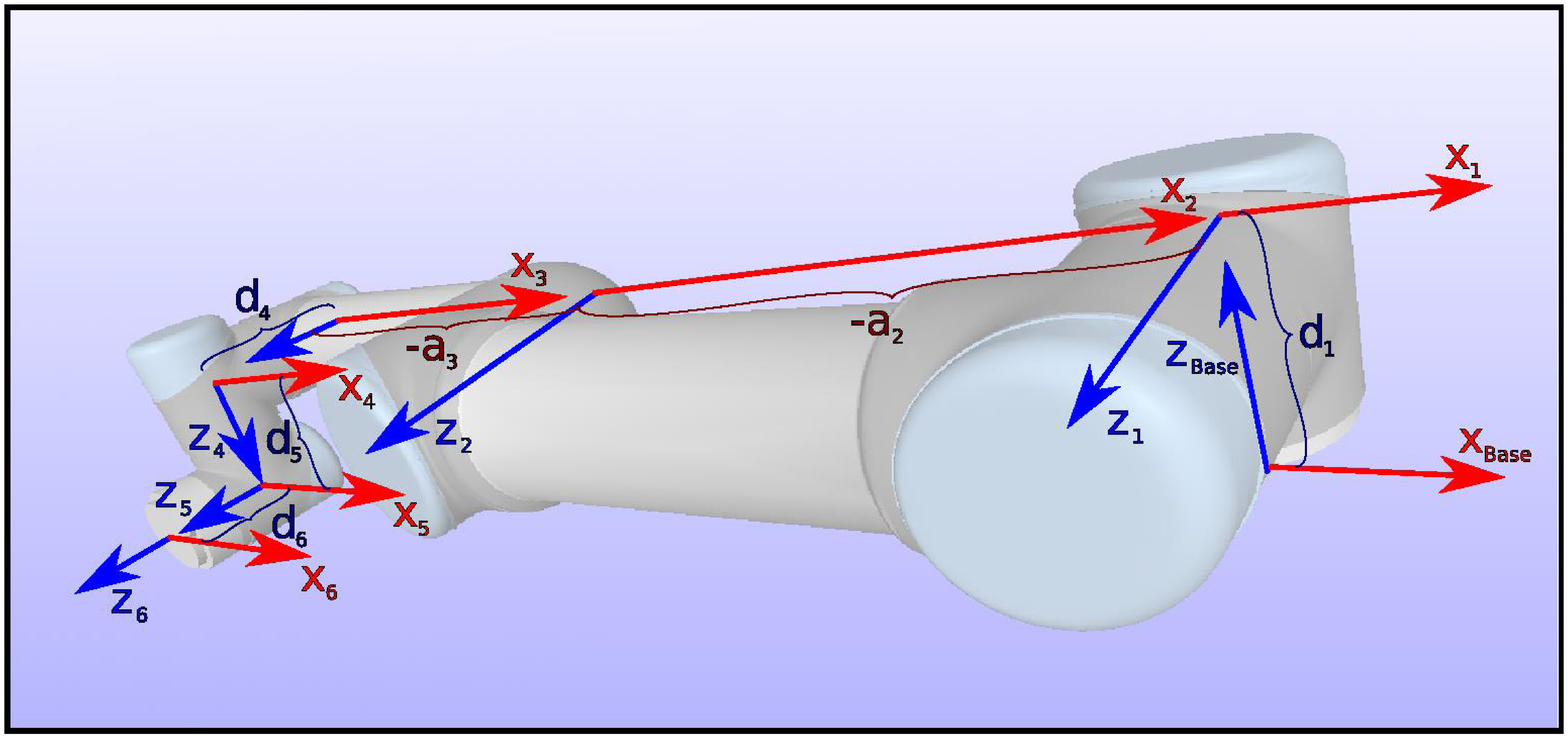} \\
    \includegraphics[scale=0.2]{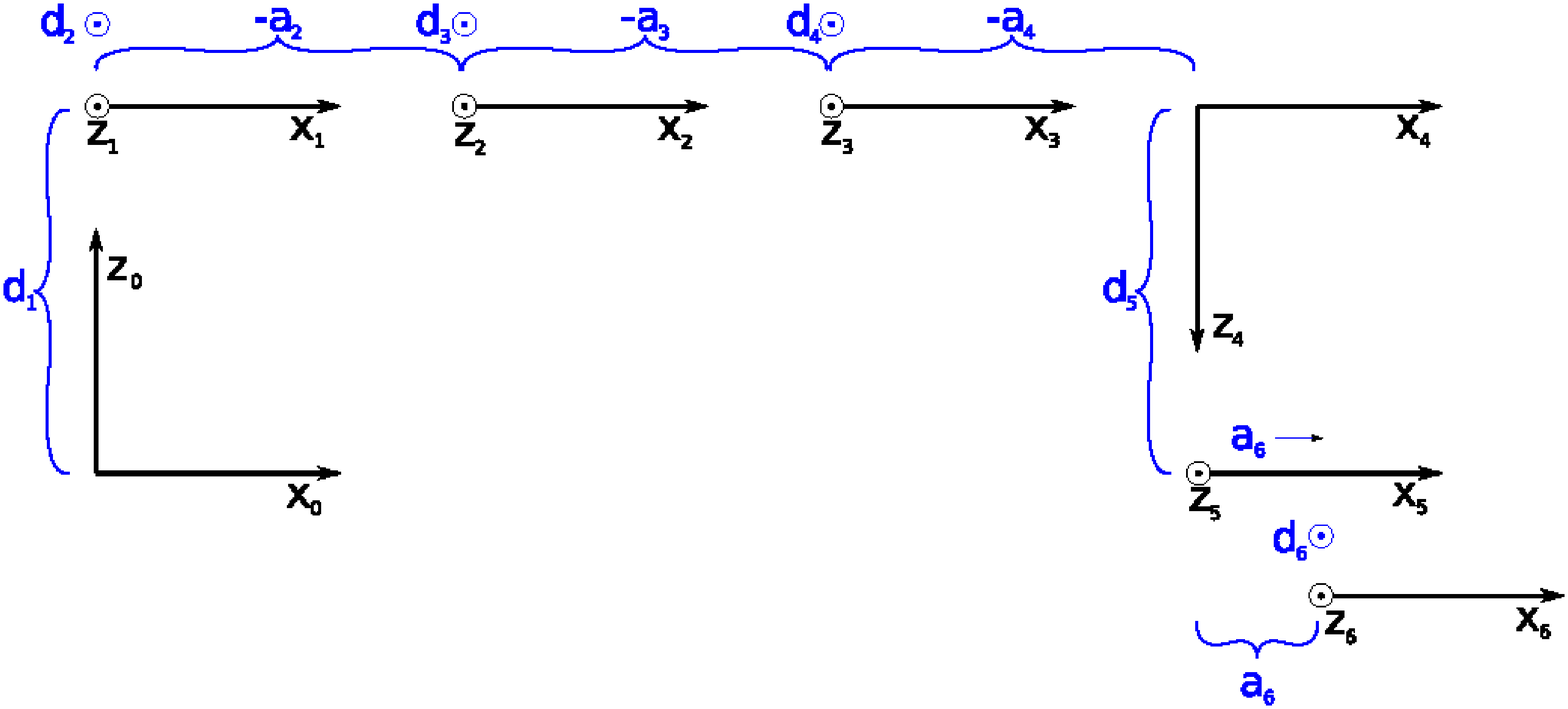} \end{tabular}
  \caption{Axes for computing D-H parameters of UR5 robot manipulator}
  \label{fig:ur5axes} \end{figure}

In order to carry out simulation for the actual system, one may need
the forward kinematic model of the robot being used. This can be
derived using the D-H parameters \cite{craig2005introduction} of the
robot. The D-H parameters for UR5 robot \cite{ur5dhp} is shown in
Table \ref{tab:ur5dhp} and the corresponding axes for deriving these
values are shown in Figure \ref{fig:ur5axes}. The forward kinematic
model thus obtained can be used for solving inverse kinematics of the
robot manipulator, developing visual servoing and other motion
planning algorithms. In the rest of this section, we describe three
popular methods for solving inverse kinematics. The readers are
referred to \cite{craig2005introduction} \cite{spong1991robot} for
more detailed treatment on the subject.
                        
\begin{table}[!h]
  \centering
  \caption{D-H Parameters of UR5 robot}
  \label{tab:ur5dhp}
  \begin{tabular}{|c|c|c|c|}\hline 
    a (m) & d (m) & $\alpha$ (rad) & $\theta$ \\ \hline \hline
    0 & 0.0895 & 1.5708 & $\theta_1$ \\ \hline
    -0.425 & 0 & 0 & $\theta_2$ \\ \hline
    -0.3923 & 0 & 0 & $\theta_3$ \\ \hline
    0 & 0.1092 & 1.5708 & $\theta_4$ \\ \hline
    0 & 0.0947 & -1.5708 & $\theta_5$ \\ \hline
    0 & 0.0823 & 0 &  $\theta_6$ \\ \hline
  \end{tabular}
\end{table}

The forward-kinematic equation is given by the following equation:
\begin{equation}
  \mathbf{x} = \boldsymbol{f}(\mathbf{q})
  \label{eq:fk}
\end{equation}
Let us assume that $\mathbf{q} \in \mathscr{R}^n$ and $\mathbf{x} \in
\mathscr{R}^m$. For a redundant manipulator, $n > m$. By taking
time-derivative on both sides of the above equation, we get
\begin{equation}
  \dot{\mathbf{x}} = J(\mathbf{q}) \dot{\mathbf{q}}
  \label{eq:fkvel}
\end{equation}
where $J$ is the $m \times n$ dimensional Jacobian of the manipulator.
The joint angles for a given end-effector pose $\mathbf{x}_d$ can be
obtained using matrix pseudo-inverse as shown below:
\begin{equation}
  \dot{\mathbf{q}} = J^{\dagger}(\mathbf{q})\dot{\mathbf{x}_d}
  \label{eq:ikvel}
\end{equation}
where $J^{\dagger}(\mathbf{q})$ represents the inverse of the Jacobian
matrix $J$. If $(J^TJ)$ is invertible, the pseudo-inverse is 
given by the Moore-Penrose inverse equation:
\begin{equation}
  J^{\dagger}(\mathbf{q}) = (J^TJ)^{-1}J^T
  \label{eq:mpi}
\end{equation}
This is otherwise known as the least square solution which minimizes
the cost function $\|\dot{\mathbf{x}}-J\dot{\mathbf{q}}\|^2$.  The
equation \eqref{eq:mpi} is considered as a solution for an
over-constrained problem where the number of equations ($m$) is less
than the number of variables $n$ and $\textrm{rank}(J) \le n$.  

If $(JJ^T)$ is invertible, then pseudo-inverse is the \textbf{minimum
norm} solution of the least square problem given by the following
equation:
\begin{equation}
  J^{\dagger} = J^T(JJ^T)^{-1}
  \label{eq:mns}
\end{equation}

The equation \eqref{eq:mns} is considered to be a solution for an
under-constrained problem where the number of equations $m$ is less
than the number of unknown variables $n$. Note that the equation
\eqref{eq:mns} is also said to provide the \emph{right pseudoinverse}
of $J$ as $JJ^{\dagger} = I$. Note that, $J^{\dagger}J \in
\mathscr{R}^{n\times n}$ and in general, $J^{\dagger}J \ne I$.

\subsubsection{Null space optimization}

An other property of the pseudoinverse is that the matrix
$I-J^{\dagger}J$ is a projection of $J$ onto nullspace. Such that for
any vector $\psi$ that satisfies $J(I-J^{\dagger}J)\psi = 0$, the
joint angle velocities could be written as
\begin{equation}
  \dot{\mathbf{q}} = J^{\dagger}\dot{\mathbf{x}} + (I-J^{\dagger}J)\psi
  \label{eq:nullsp}
\end{equation}

In general, for $m < n$, $(I-J^{\dagger}J)\ne 0$, and all vectors of
the form $(I-J^{\dagger}J)\psi$ lie in the null space of $J$, i.e.,
$J(I-J^{\dagger}J)\psi = 0$. By substituting $\psi =
\dot{\mathbf{q}}_0$ in the above equation, the general inverse
kinematic solution may be written as 
\begin{equation}
  \dot{\mathbf{q}} = J^{\dagger} \dot{\mathbf{x}} + (I-J^{\dagger}J)\dot{\mathbf{q}}_0
  \label{eq:gik}
\end{equation}
where $(I-J^{\dagger}J)$ is a projector of the joint velocity vector
$\dot{\mathbf{q}}_0$ onto $\mathscr{N}(J)$. The typical choice of the null space joint velocity vector is 
\begin{equation}
  \dot{\mathbf{q}}_0 = k_0 \left(\frac{\partial w(\mathbf{q})}{\partial \mathbf{q}} \right)^T
  \label{eq:nspq}
\end{equation}
with $k_0 > 0$ and $w(\mathbf{q})$ is a scalar objective function of
the joint variables and $\left(\frac{\partial w(\mathbf{q})}{\partial
  \mathbf{q}} \right)^T $ represents the gradient of $w$. A number of
  constraints could be imposed by using this objective function. For
  instance, the joint limit avoidance can be achieved by selecting the
  objective function as 
  \begin{equation}
    w(q) = \frac{1}{n}\sum_i^n \left(\frac{q_i - \bar{q}_i}{q_{iM} - q_{im}}\right)^2
    \label{eq:jla_cost}
  \end{equation}
  where $\bar{q}_i$ is the middle value of joint angles while $q_{iM}$
  ($q_{im}$) represent maximum (minimum) value of joint angles. The
  effect of the null space optimizing on joint angle norm is shown in
  Figure \ref{f:thnorm}. As one can see from this figure, the null
  space optimization for joint limit avoidance leads to a solution
  with smaller joint angle norm compared to the case when self motion
  is not used. 


\subsubsection{Inverse Kinematics as a control problem} \label{sec:ik_ctrl}

The inverse kinematic problem may also be formulated as a closed-loop
control problem as described in \cite{wang2010inverse}. Consider the
end-effector pose error and its time derivative be give as follows:
\begin{equation}
  \mathbf{e} = \mathbf{x_d} - \mathbf{x}; \; \dot{\mathbf{e}} = \dot{\mathbf{x_d}} - \dot{\mathbf{x}} 
  = \dot{\mathbf{x}}_{\mathbf{d}} - J\dot{\mathbf{q}}
  \label{eq:efferror}
\end{equation}

By selecting the joint velocities as 
\begin{equation}
  \dot{\mathbf{q}} = J^{\dagger} (\dot{\mathbf{x}}_{\mathbf{d}} + K_p (\mathbf{x}_d - \mathbf{x}))
  \label{eq:control2}
\end{equation}
the closed loop error dynamics becomes
\[
  \dot{\mathbf{e}} + K_p \mathbf{e} = 0
\]

Hence the control law \eqref{eq:control2} stabilizes the closed loop
error dynamics and the error will converge to zero if $K_p$ is
positive definite. The homogeneous part of the inverse kinematic
solution in \eqref{eq:nullsp} could be combined with
\eqref{eq:control2} in order to obtained a generalized closed loop
inverse kinematic solution. 

\subsubsection{Damped Least Square Method} \label{sec:dls}

The pseudo-inverse method for inverse kinematics is given by
\begin{equation}
  \Delta \mathbf{q} = J^{\dagger}\mathbf{e}
  \label{eq:pinv2}
\end{equation}

In damped least square method, the $\Delta \mathbf{q}$ is selected so
as to minimize the following cost function
\begin{equation}
  V = \|J\Delta \mathbf{q} - \mathbf{e}\|^2 + \lambda^2 \|\Delta \mathbf{q}\|^2
  \label{eq:dls-cost}
\end{equation}
This gives us the following expression for joint angle velocities:
\begin{equation}
  \Delta \mathbf{q} = J^T(JJ^T+\lambda^2I)^{-1}\mathbf{e}
  \label{eq:dls}
\end{equation}
 
\begin{figure}[!h]
  \centering
  \begin{tabular}{cc}
    \subfigure[Initial and Final Robot Pose]{\label{f:config}\includegraphics[scale=0.3]{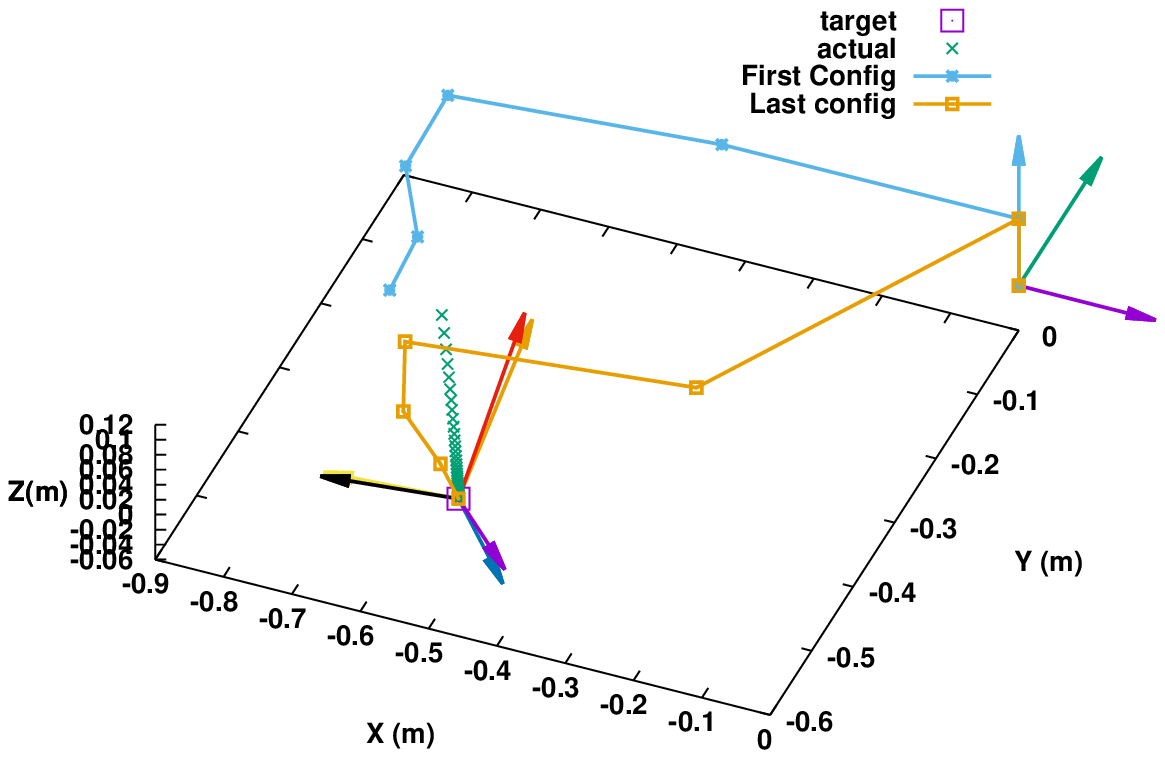}} & 
    \subfigure[Joint angle values are within physical limits of the robot]{\label{f:angle}\includegraphics[scale=0.3]{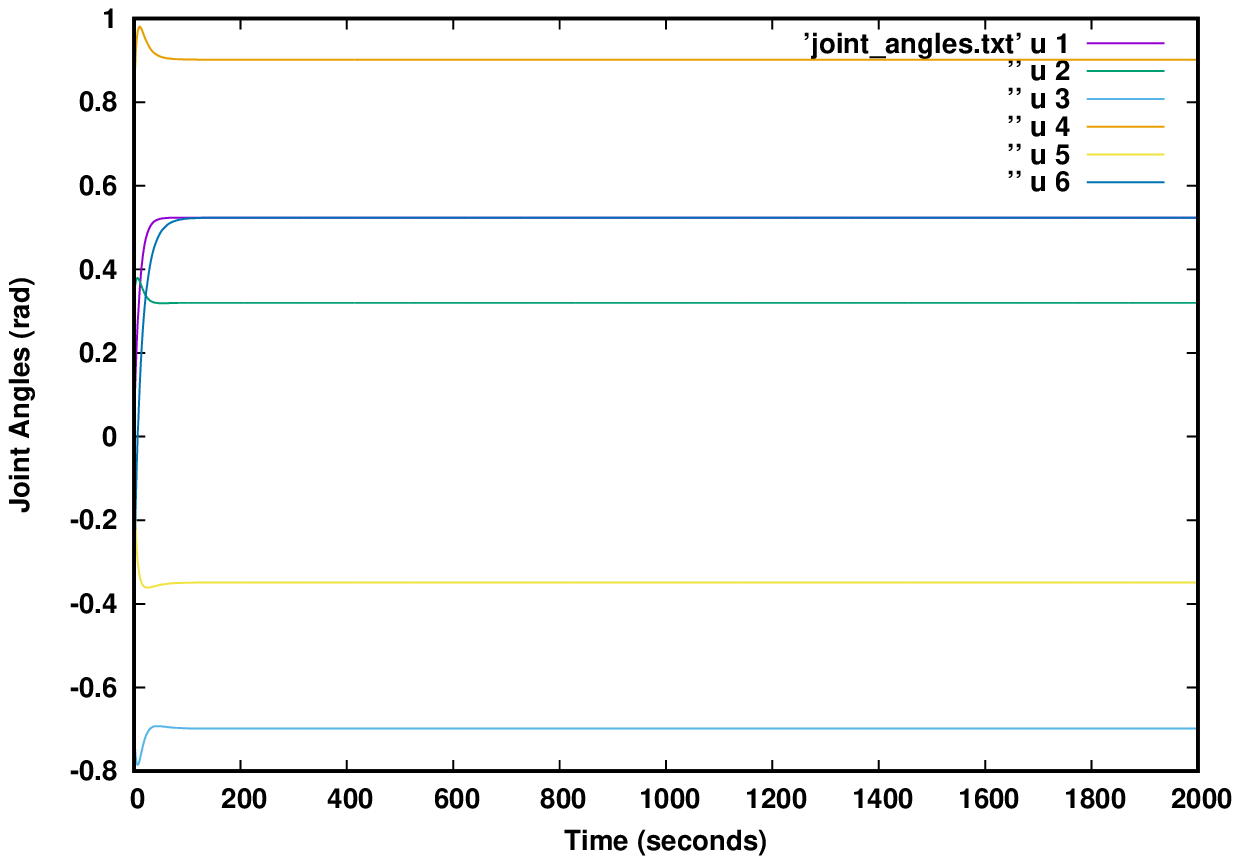}} \\ 
    \subfigure[End-effector position error over time]{\label{f:error}\includegraphics[scale=0.3]{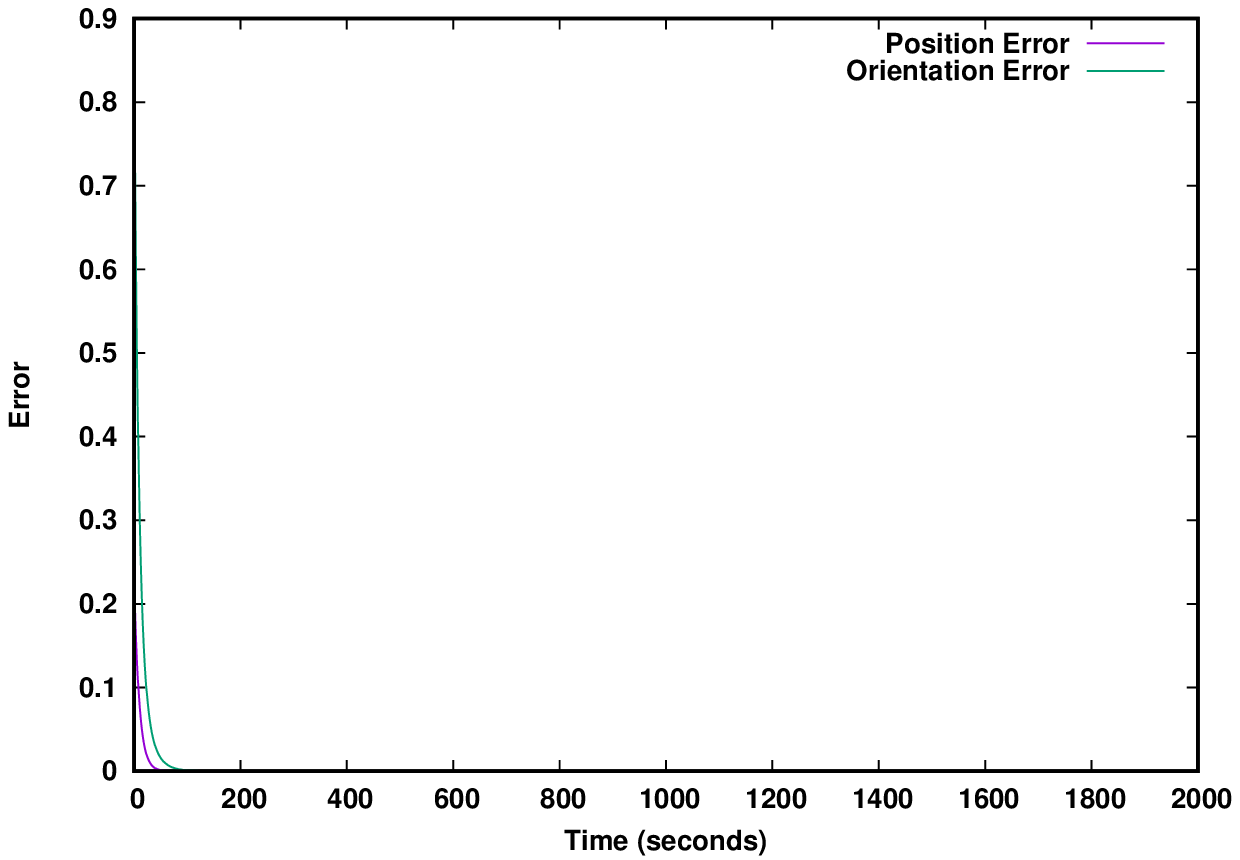}} &  
    \subfigure[Effect of Null Space Optimization on Joint Angle Norm]{\label{f:thnorm}\scalebox{0.3}{\input{./fig/thnorm.tex}}} 
  \end{tabular}
  \caption{Computing Inverse kinematics for a given target pose using
  conventional methods. Figure (a) shows the initial and final robot
pose along with the end-effector trajectory. The frame coordinates for
desired and actual end-effector pose is also shown.}
  \label{fig:invkin1}
\end{figure}

\begin{figure}[!h]
  \centering
  \scalebox{0.8}{\input{./fig/bincfg.tex}}
  \caption{Robot pose for the bin centres of the rack obtained by
  solving inverse kinematics of the robot manipulator. The average
error over 12 points is about 6 mm. }
  \label{fig:bin_config}
\end{figure}
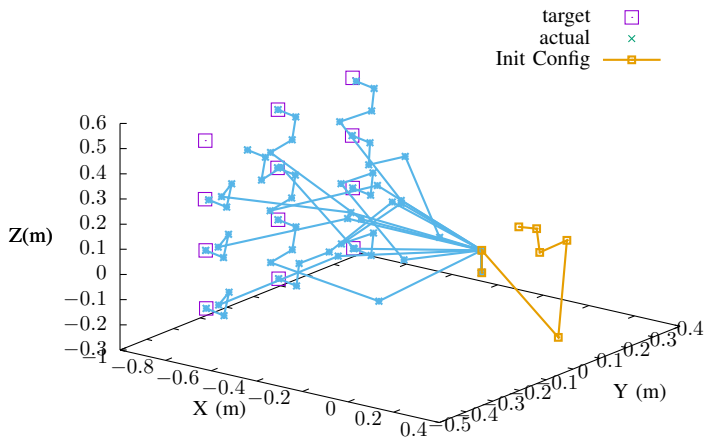

The inverse kinematic solutions computed using these conventional
methods are shown in Figure \ref{fig:invkin1} \ref{fig:bin_config}
respectively. Figure \ref{f:config} shows the inverse kinematic
solution obtained for a given pose using null space optimization
method that avoids joint limits as explained above. The corresponding
joint angles are within their physical limits as shown in Figure
\ref{f:angle}. Figure \ref{fig:bin_config} shows the joint
configurations for reaching all bin centres of the rack. 

\section{Experimental Results} \label{sec:expt}

\begin{figure}[htbp]
  \centering
  \includegraphics[scale=0.3]{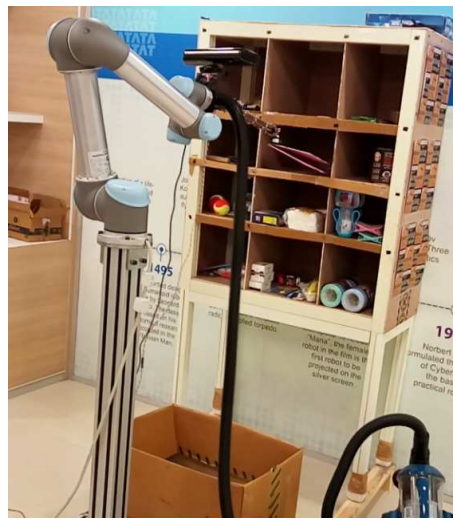}
  \caption{Experimental setup for Automated Pick and Stow System}
  \label{fig:apc_setup}
\end{figure}

The actual system developed for accomplishing the task of automated
picking and stowing is shown in Figure \ref{fig:apc_setup}. The system
comprises of a 6 DOF UR5 robot manipulator with a suction based
end-effector, a rack with 12 bins in a $3\times4$ grid. The
end-effector is powered by a household vacuum cleaner. It uses a
Kinect RGBD sensor in an eye-in-hand configuration for carrying out
all perception tasks.  As explained in Section \ref{sec:sysarch}, the
entire system runs on three laptops connected to each other through
ethernet cables. One of these laptops is a Dell Mobile Precision 7710
workstation with a NVIDIA Quadro M5000M GPU process with 8GB of GPU
RAM. This laptop is used for running the RCNN network for object
detection. The other two laptops have a normal Intel i7 processor with
16 GB of system RAM. The distribution of various nodes on the machines
are shown in Figure \ref{fig:apc_ros_arch}. It is also possible to run
the whole system on a single system having necessary CPU and GPU
configuration required for the task. The videos showing the operation
of the entire system using a suction end-effector \cite{demoVideo1}
\cite{iitkapc2016} and a two-finger gripper \cite{demoVideo2} is made
available on internet for the convenience of the readers. The readers
can also use the source codes \cite{apcdemocode} made publicly
available under MIT license for their own use.

\subsection{Response time}

The computation time for different modules of the robotic pick \&
place system is provided in Table \ref{tab:time}. As one can see the
majority of time is spent in image processing as well as in executing
robot motions. Our loop time for picking each object is about 24
seconds which leads to a pick rate of approximately 2.5 objects per
minute. The rack detection and system calibration is carried out only
once during the whole operation and does not contribute towards the
loop time. 

\begin{table}[!h]
  \caption{Computation time for various modules of the robotic pick \&
    place system. }
  \label{tab:time} 
  \centering
  \begin{tabular}{|c|L{1.5cm}|L{2.5cm}|C{1.5cm}|} \hline
    S. No. & Component & Description & Time (seconds) \\ \hline
    1 & Reading JSON file & For ID extraction & 0.01 \\ \hline
    2 & Motion 1 & Home position to Bin View Position &   3.5 \\ \hline
    3 & Object recognition & using trained RCNN model & 2.32 \\ \hline
    4 & Motion 2 & Pre-grasp motion & 9.6  \\ \hline
    5 & Motion 3 & Post-grasp motion & 4.97 \\ \hline
    6 & Motion 4 & Motion from Tote drop to home position & 3.41 \\ \hline  \hline
    & \multicolumn{2}{|l|}{Total loop time for each object } & 23.81 \\ \hline \hline
    7 & Rack Detection &                    & 2.1  \\ \hline
    8 & Calibration &                       & 13.1 \\ \hline
  \end{tabular}
\end{table}

\subsection{Grasping and Suction} \label{sec:exp_grasp}

\begin{figure}[htbp]
  \centering
  \begin{tabular}{cc}
    \subfigure[Gripping Action]{\label{f:grip}\includegraphics[scale=0.25]{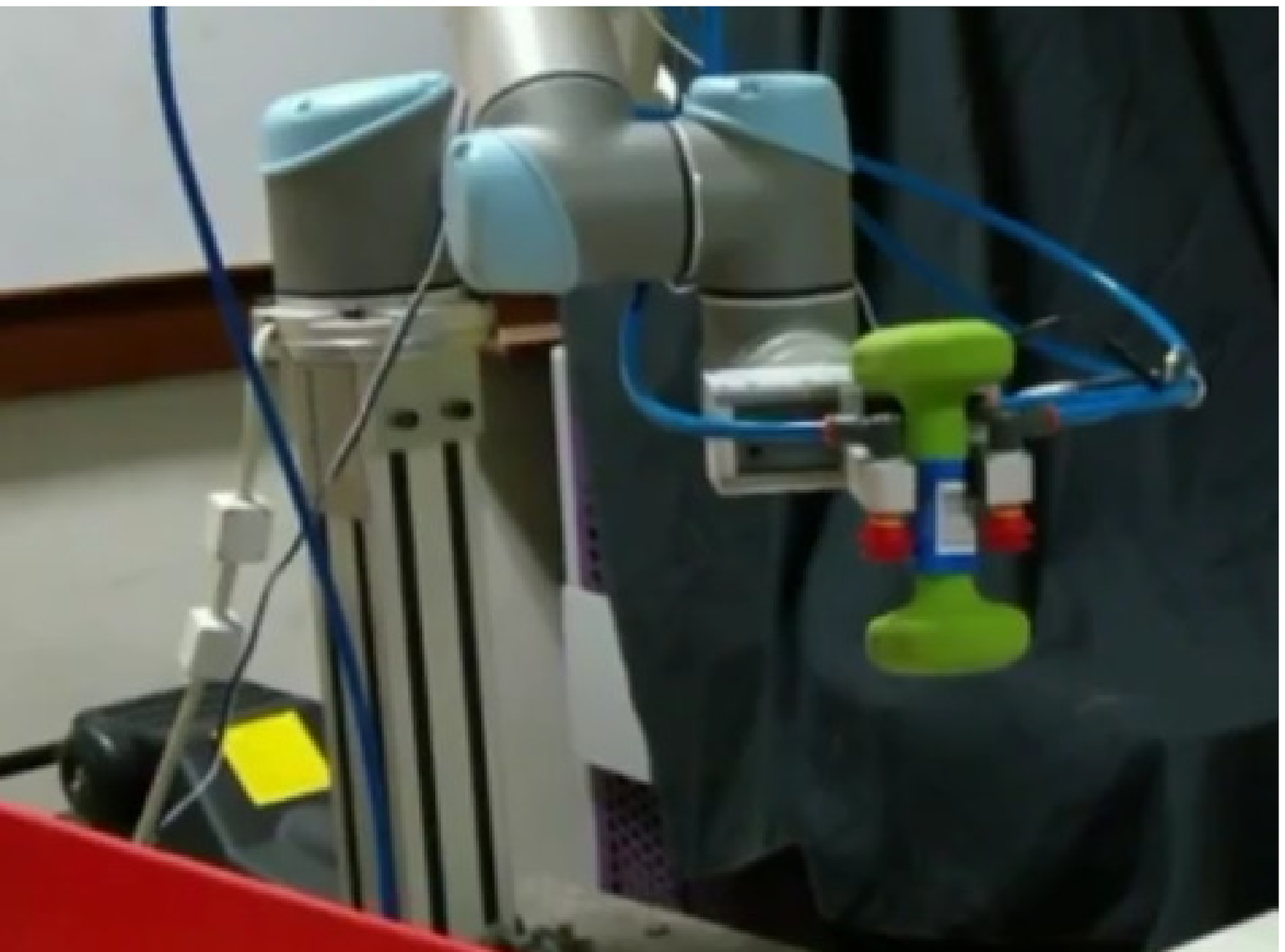}} 
    \subfigure[Suction Action]{\label{f:suck}\includegraphics[scale=0.25]{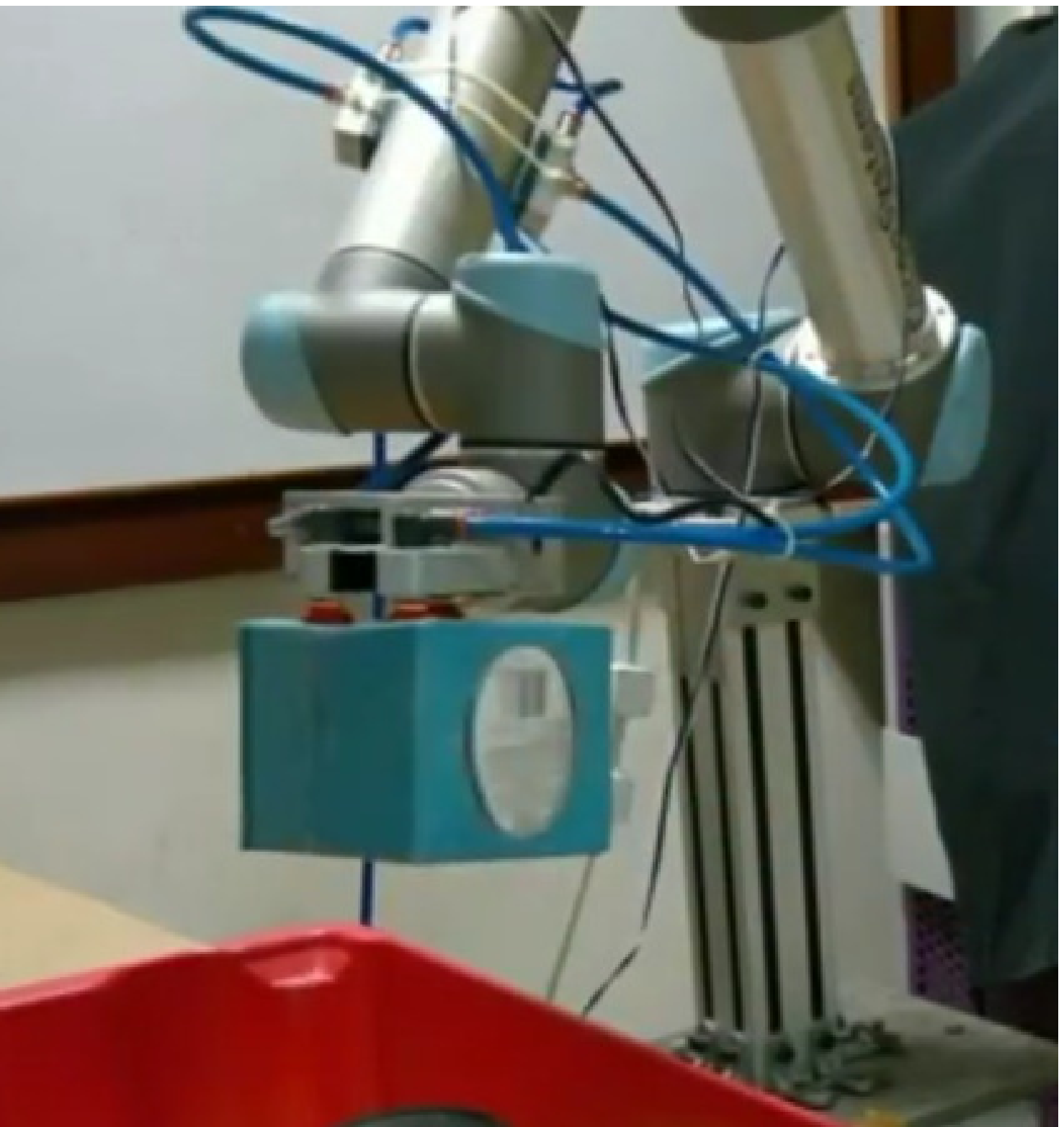}} 
  \end{tabular}
  \caption{The hybrid gripper in action. It uses two-finger gripper to
  pick objects that can fit into its finger span. Suction is used for
picking bigger objects with flat surfaces.}
  \label{fig:grip_action}
\end{figure}

\begin{figure}[!t]
  \centering
  \begin{tabular}{c}
    \subfigure[Segmenting `Fevicol' tube from the clutter]{\label{f:g1}\includegraphics[scale=0.3]{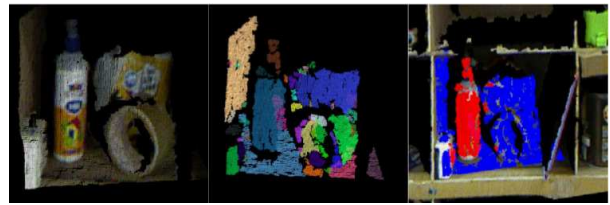}} \\
    \subfigure[Use of GMM model using both color and depth curvature information]{\label{f:g3}\includegraphics[scale=0.25]{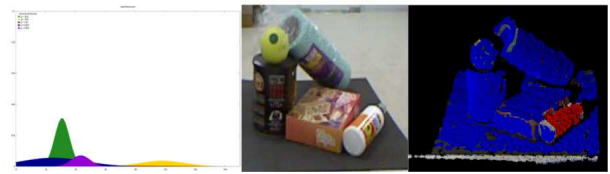}} \\
    \subfigure[Primitive Shape fitting and identifying best graspable
    affordance for objects with different
  shapes]{\label{f:g5}\includegraphics[scale=0.25]{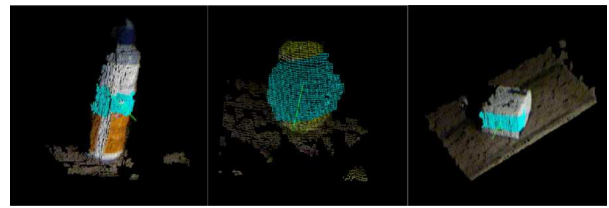}} \\
  \subfigure[Identifying shapes and computing graspable affordance in
  a clutter]{\label{f:g6}\includegraphics[scale=0.25]{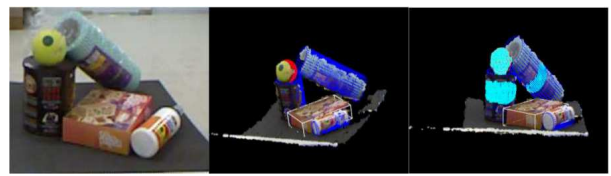}}
\end{tabular} \caption{Computing Graspable Affordance of target object
  in a Cluttered Environment. (a) Shows the use of GMM model
  comprising of RGB and depth curvature information in segmenting the
  target object from clutter. (b) Shows the GMM model used in (a). It
  shows the Gaussian corresponding to depth curvature provides better
  discrimination compared to colors in identifying the target. (c)
  Shows the detection of shape and best graspable affordance for
  isolated objects. (d) Shows the detection of shape and graspable
affordance in a clutter. } 
\label{fig:graspexpt} 
\end{figure}

The working of our custom gripper is shown in Figure
\ref{fig:grip_action}. The maximum clearance between the fingers is
about 7 cm and it has been designed to pick up a payload of 2 Kgs. The
gripper can grasp things using an antipodal configuration
\cite{pas2015using} as shown in Figure \ref{f:grip}. The suction is
applied whenever it is not possible to locate grasping affordances on
the object. The bellow cups are positioned normal to the surface of
the object being picked as shown in Figure \ref{f:suck}. For grasping,
it is necessary to detect the grasp pose and compute the best
graspable affordance for a given object. This is done by using the
method as described in Section \ref{sec:grasp}. Some of the results
corresponding to the grasping algorithm is shown in Figure
\ref{fig:graspexpt}. As explained before, a GMM model comprising of
color (RGB) information and depth curvature information is effective
in segmenting the target object from its background as shown in Figure
\ref{f:g1} and \ref{f:g3} respectively. The outcome of the grasping
algorithm is shown in Figure \ref{f:g5} and \ref{f:g6} respectively.
The figure \ref{f:g5} shows the best graspable affordance for objects
with different shapes while the Figure \ref{f:g6} shows the graspable
affordance of objects in a clutter.

\subsection{Object Recognition}

\begin{figure}[!t]
  \centering
  \begin{tabular}{cccc}
    \includegraphics[scale=0.08]{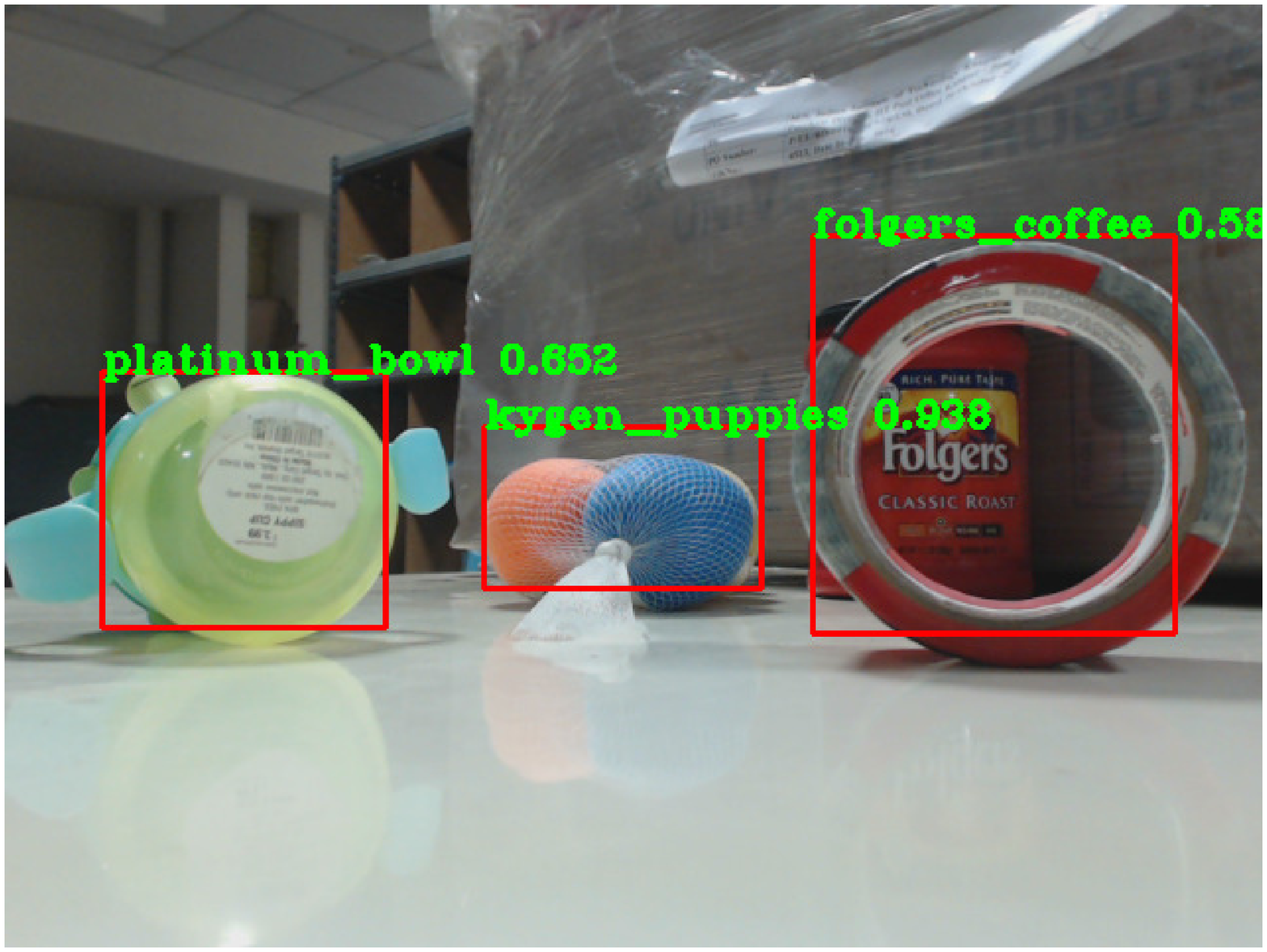} &
    \includegraphics[scale=0.08]{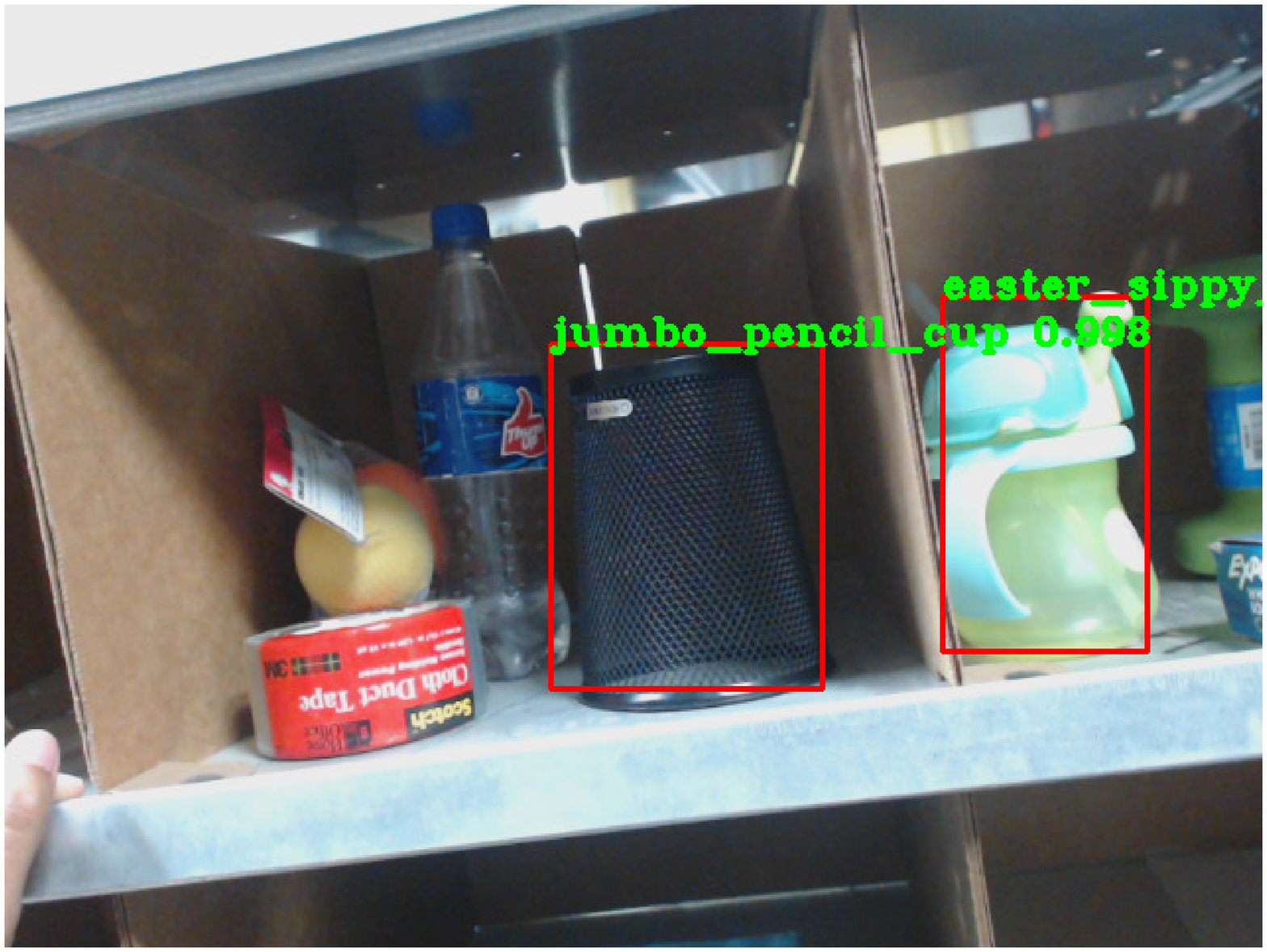} &
    \includegraphics[scale=0.08]{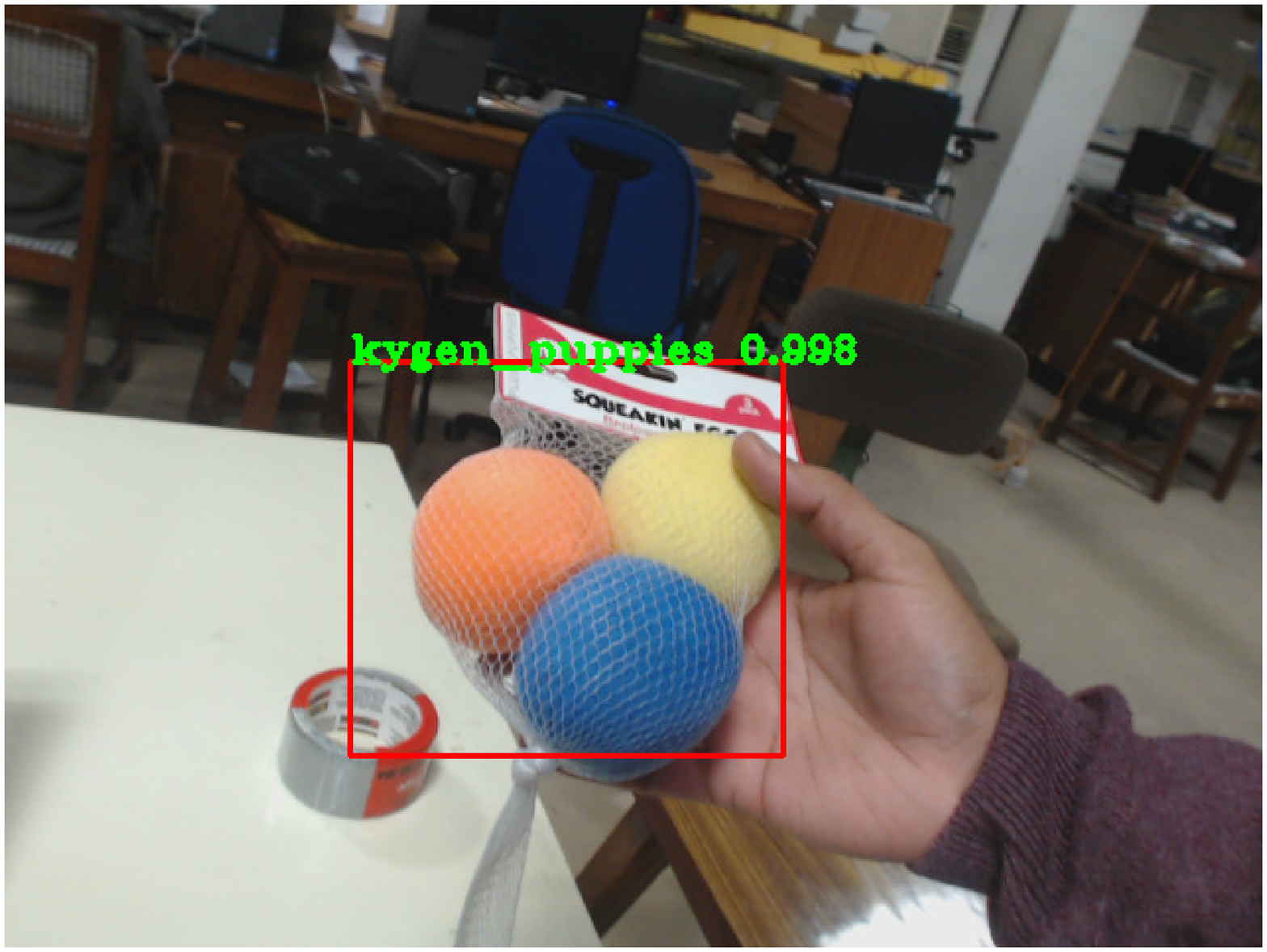} &
    \includegraphics[scale=0.08]{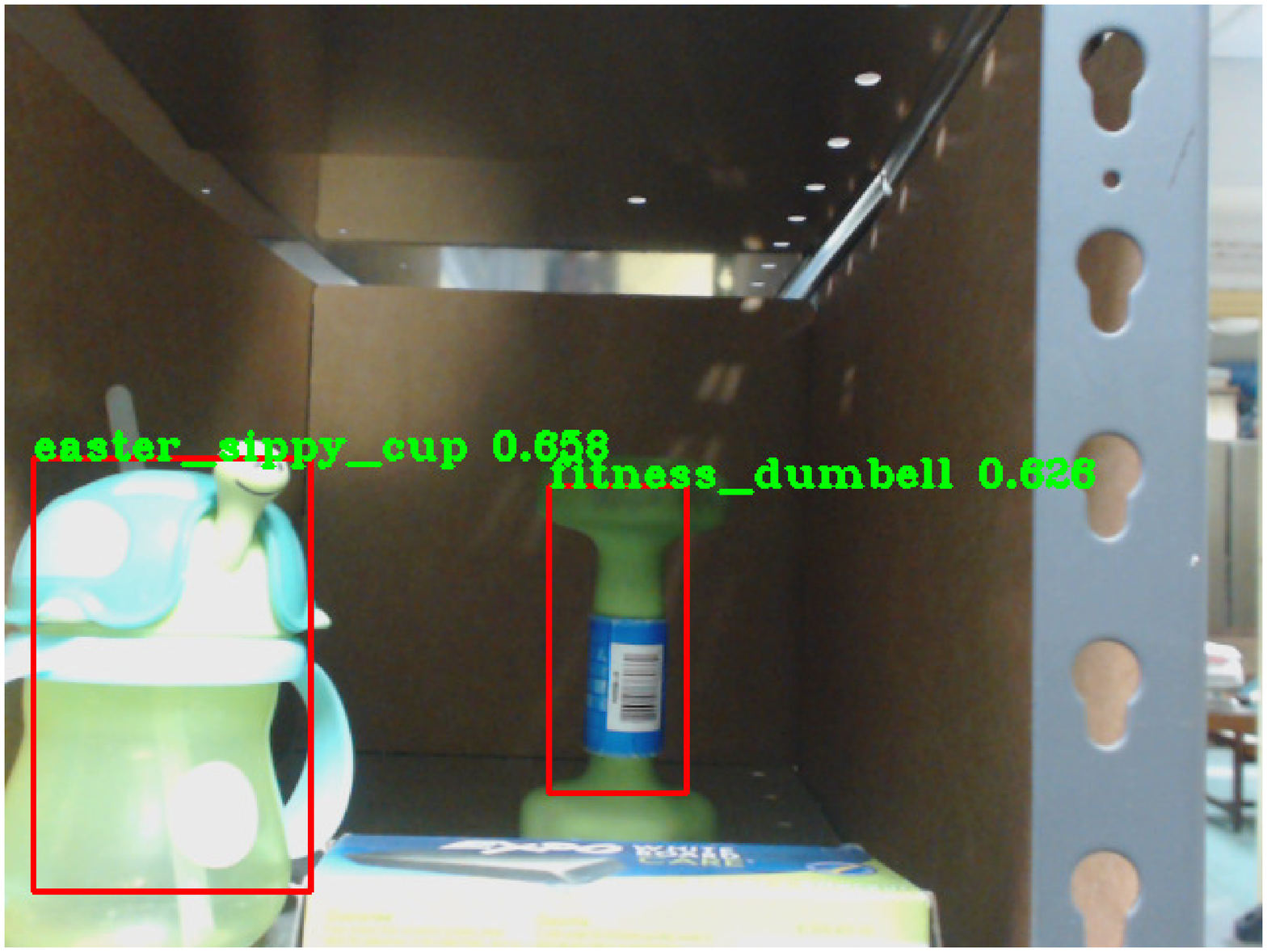} \\
    \includegraphics[scale=0.08]{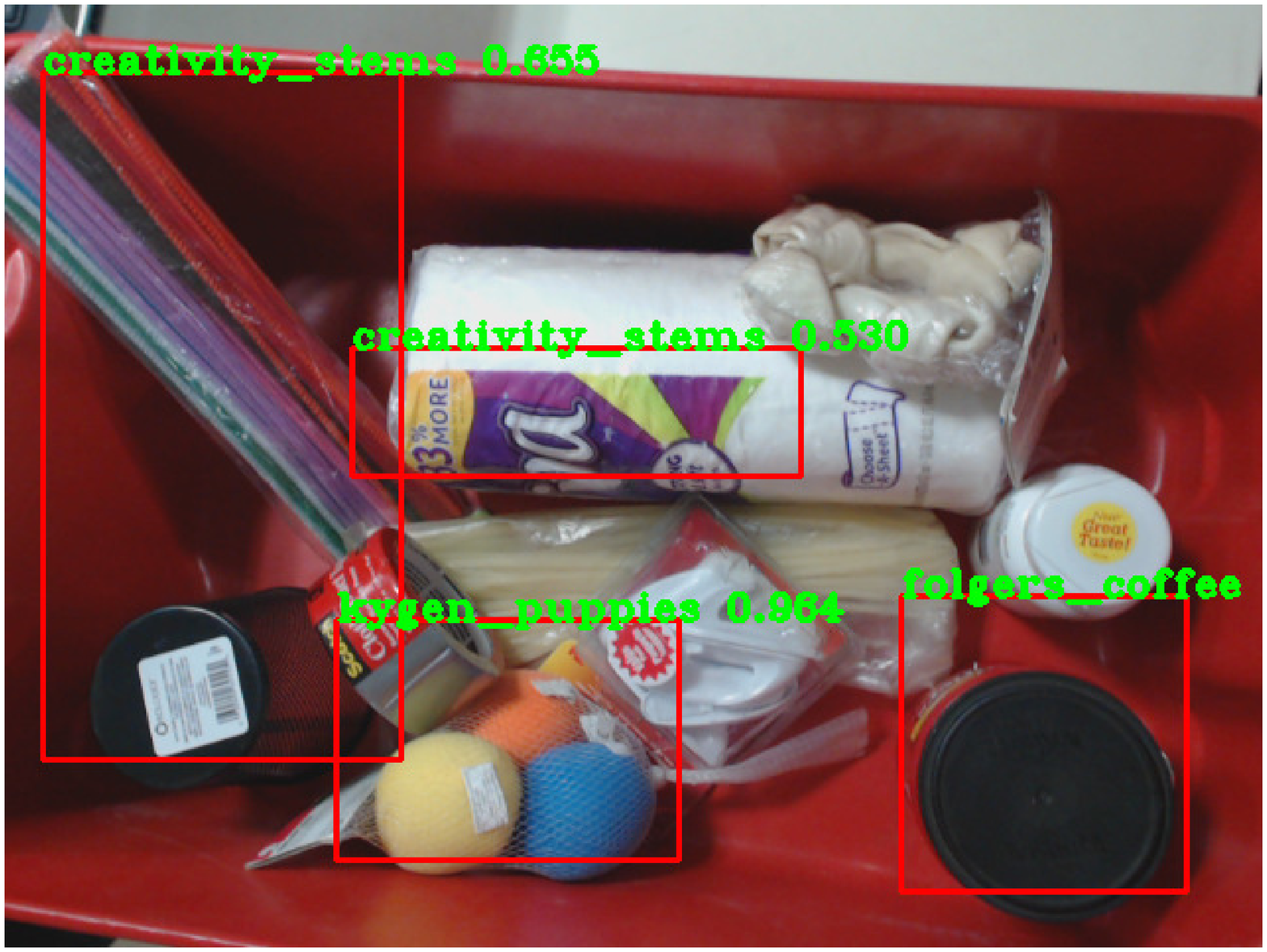} &
    \includegraphics[scale=0.08]{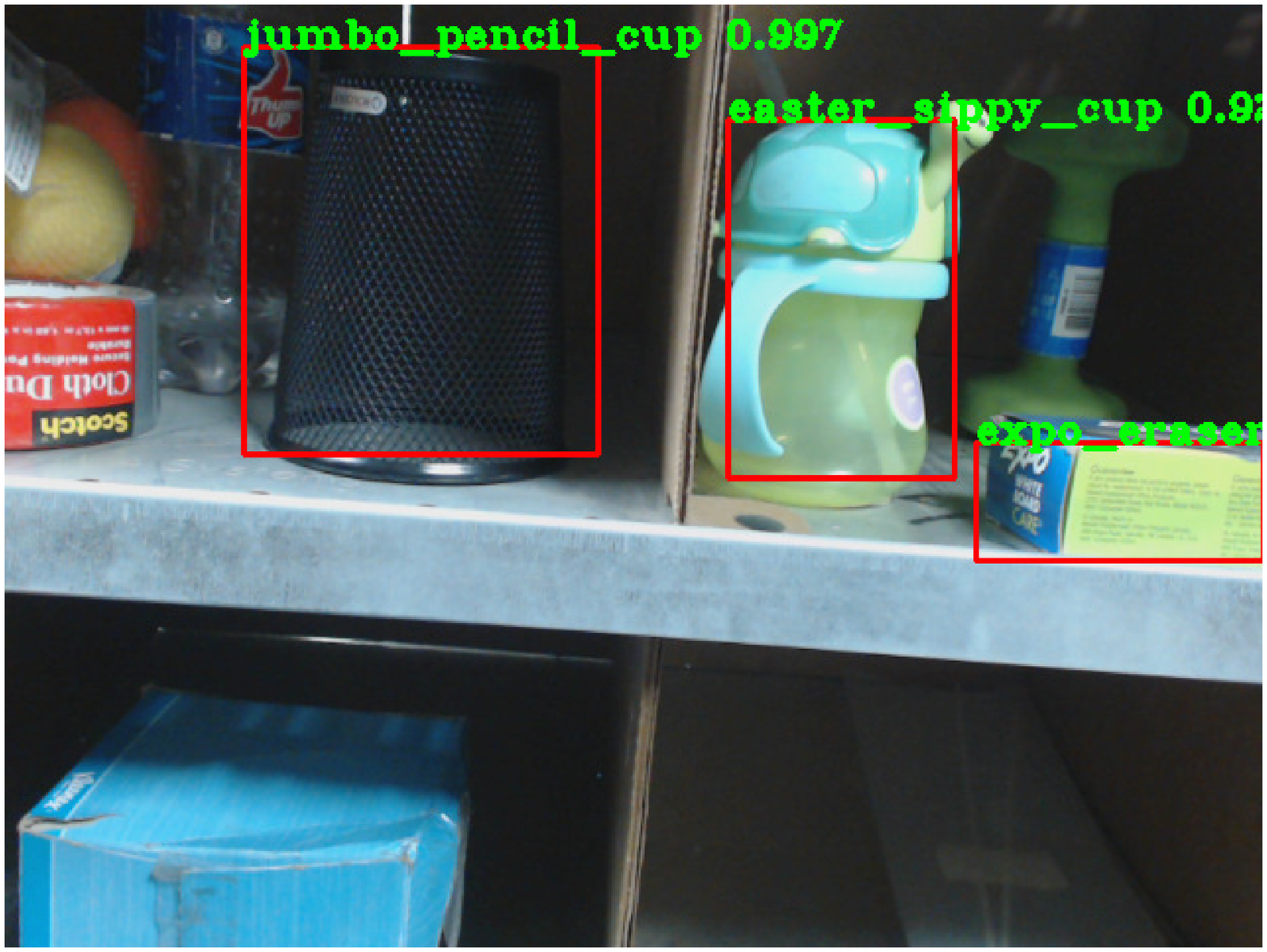} &
    \includegraphics[scale=0.08]{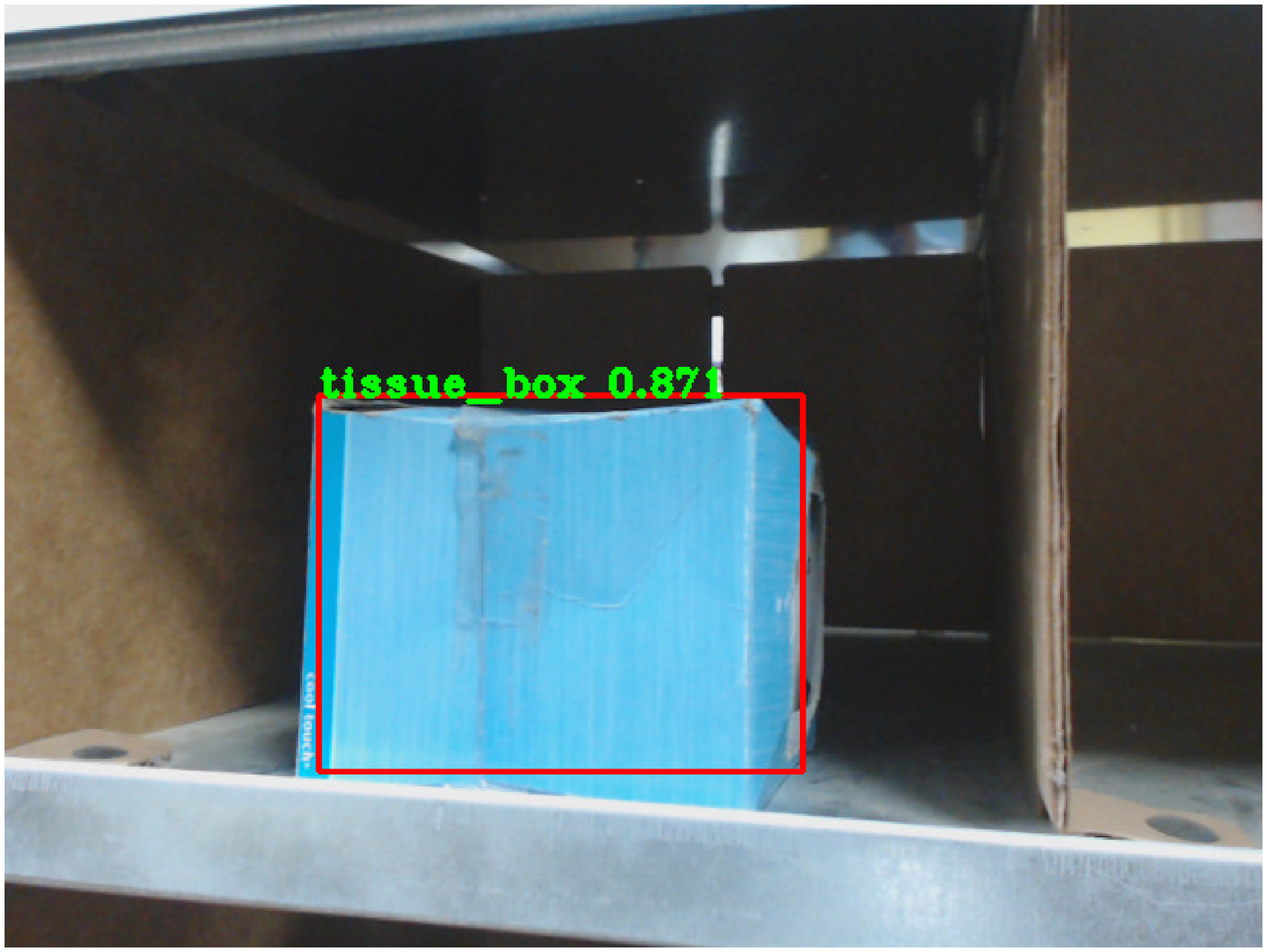} &
    \includegraphics[scale=0.08]{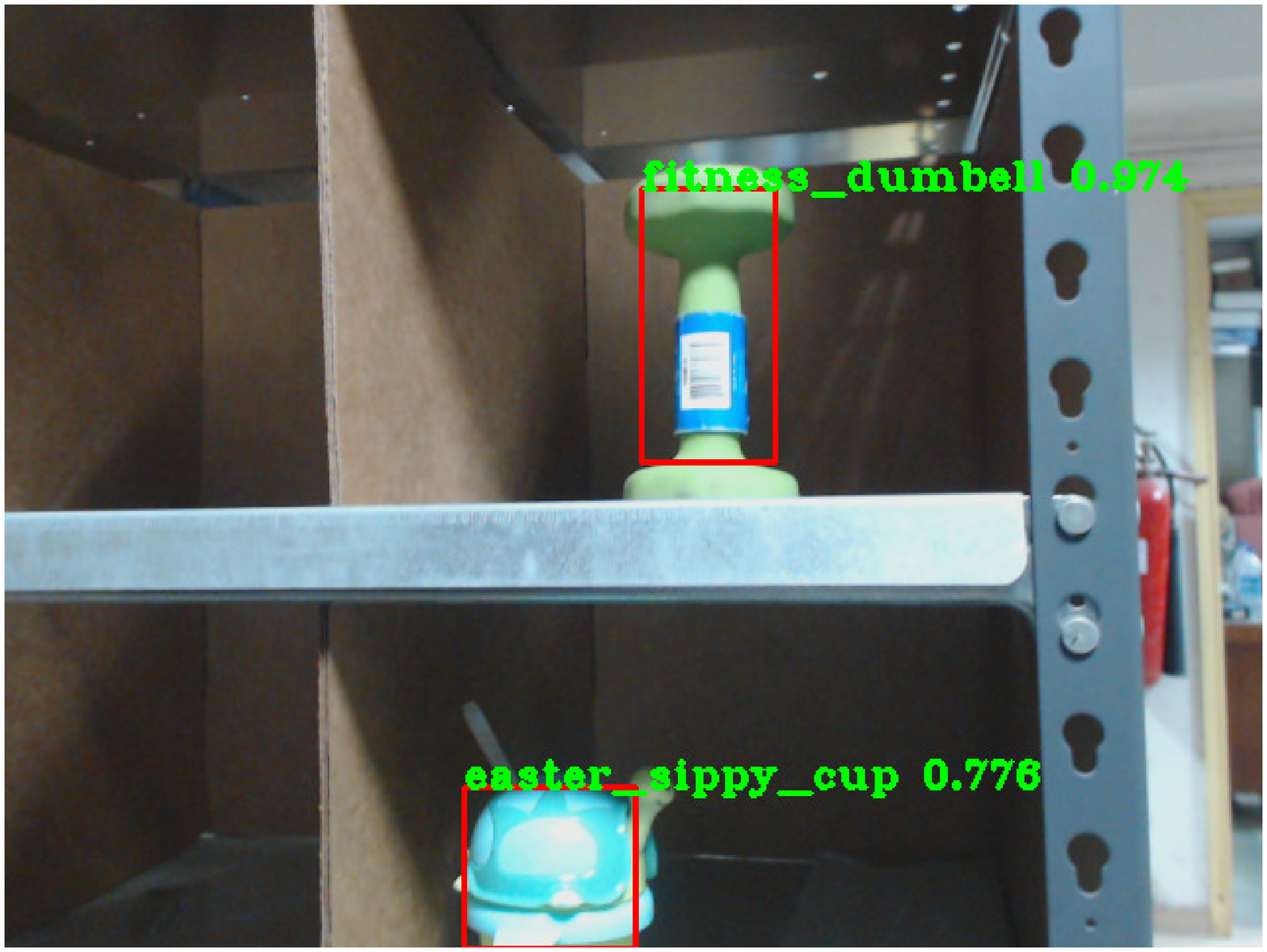} \\
    \includegraphics[scale=0.08]{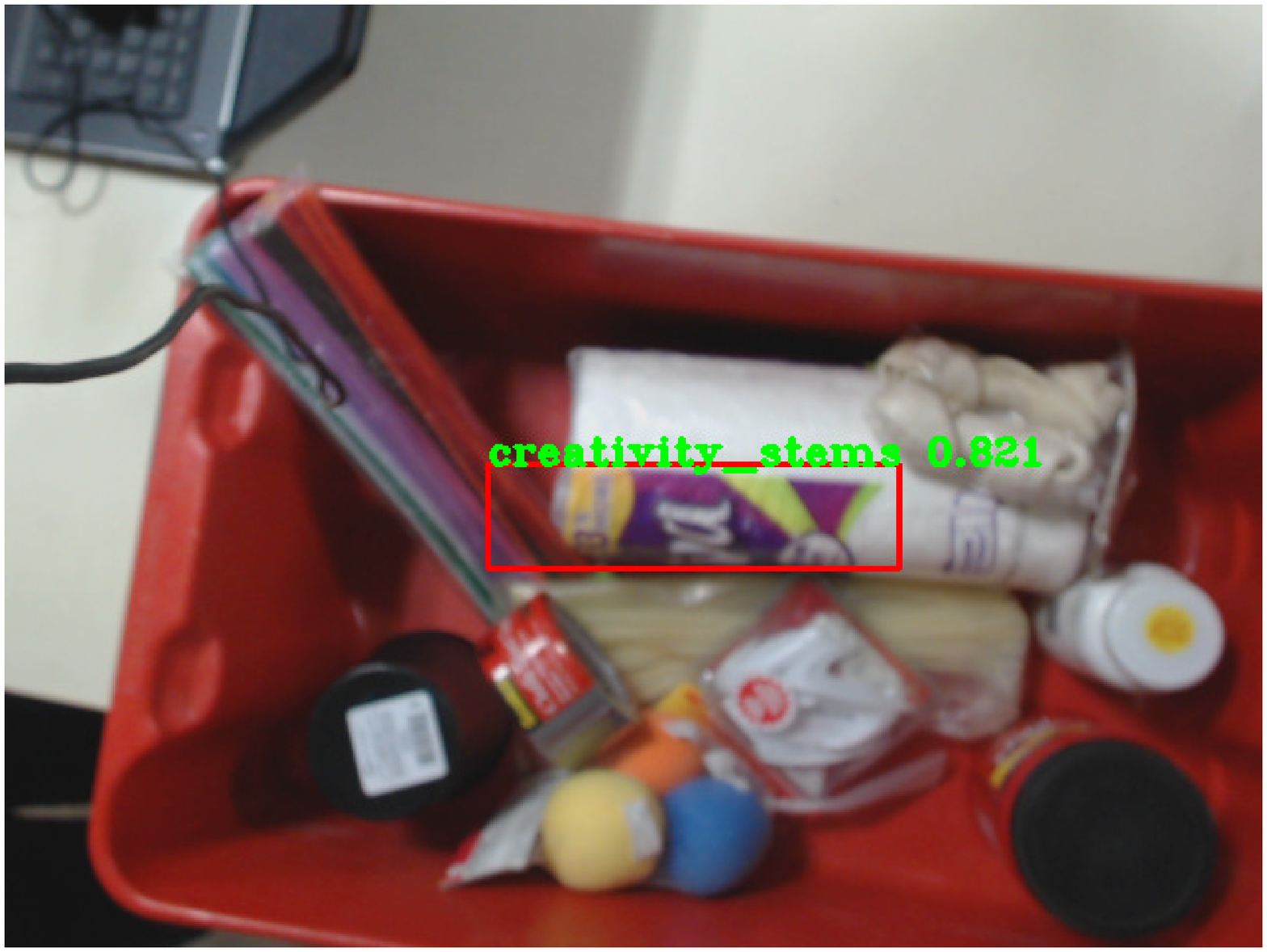} &
    \includegraphics[scale=0.08]{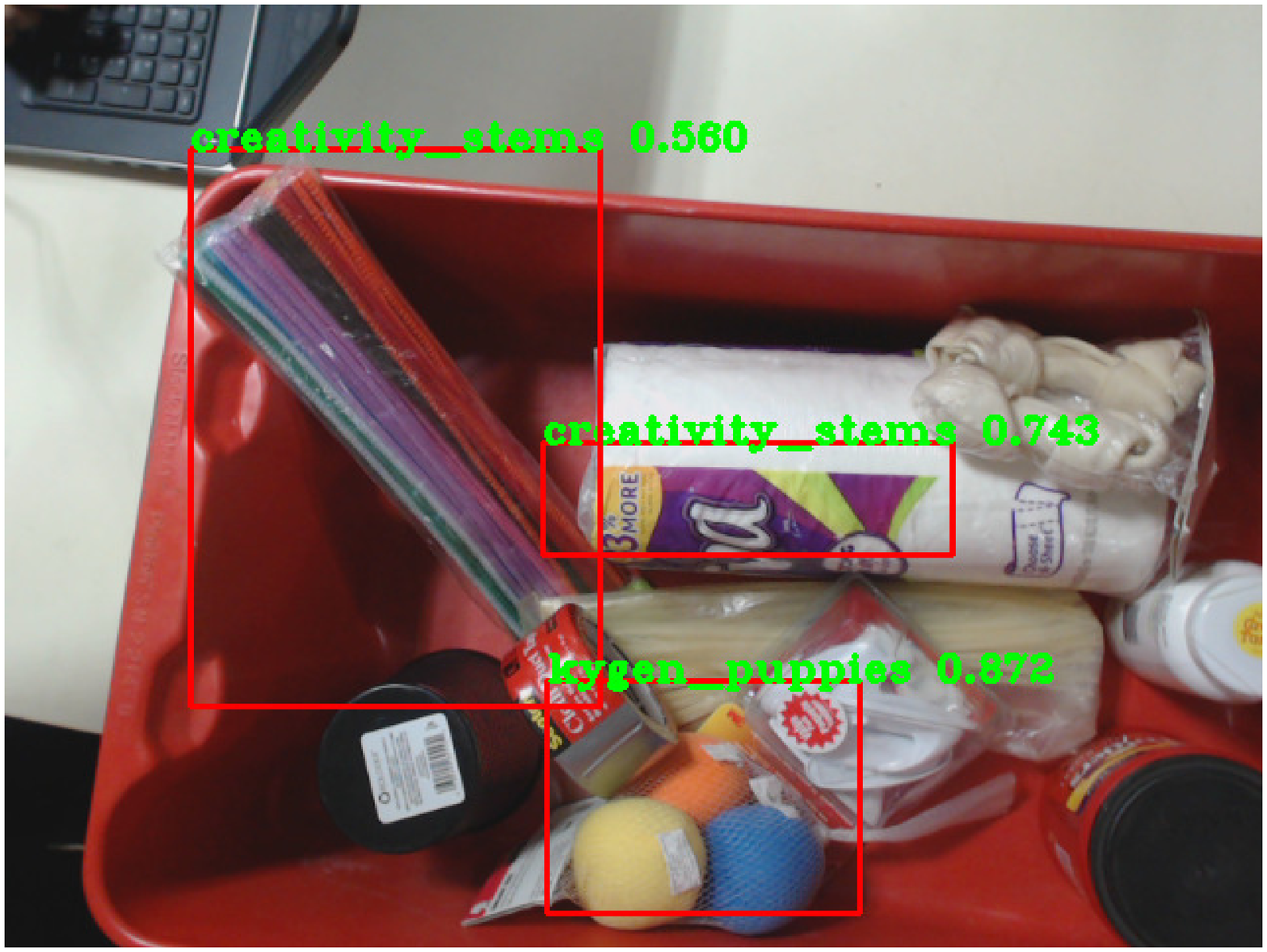} &
    \includegraphics[scale=0.08]{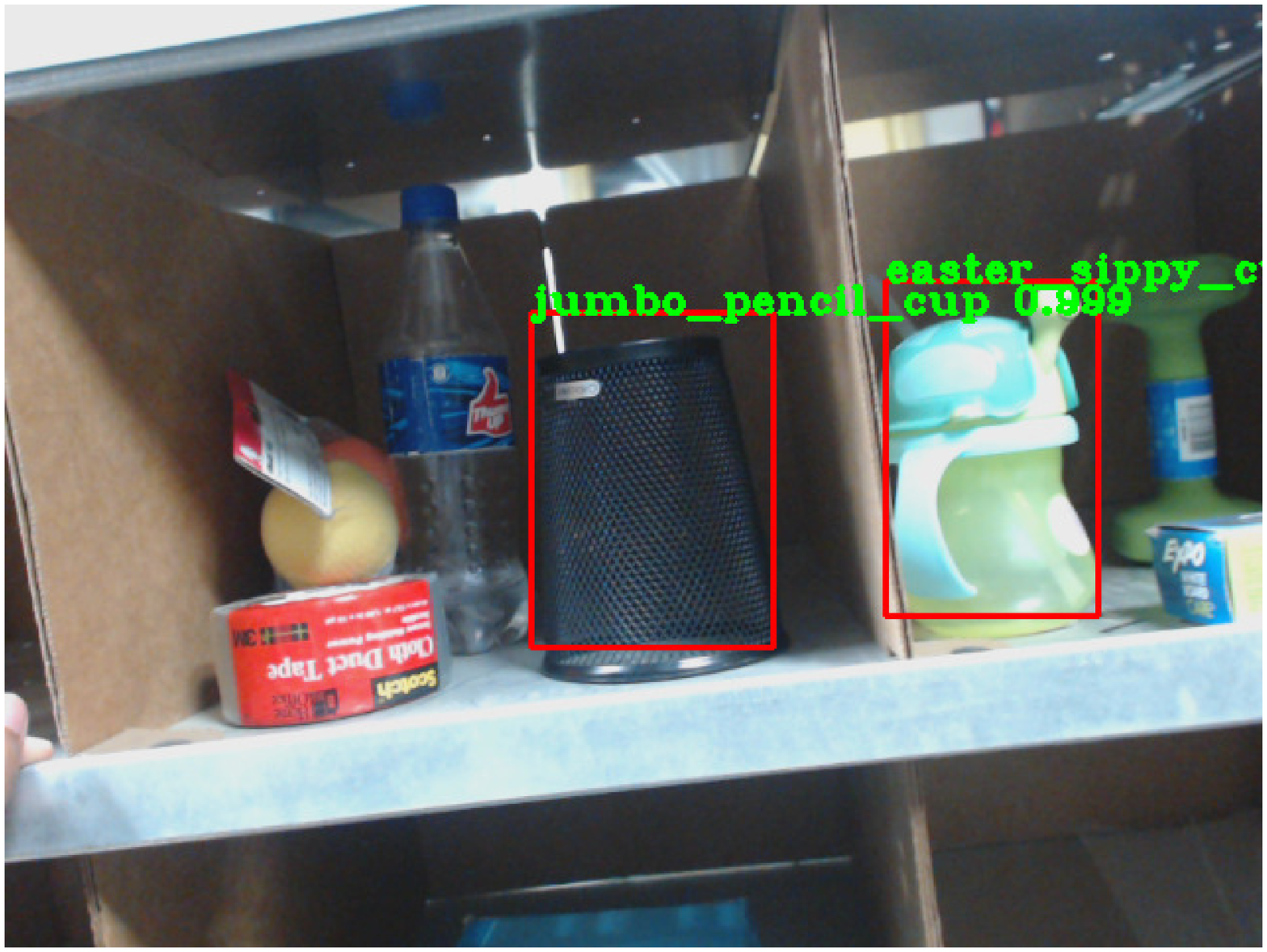} &
    \includegraphics[scale=0.08]{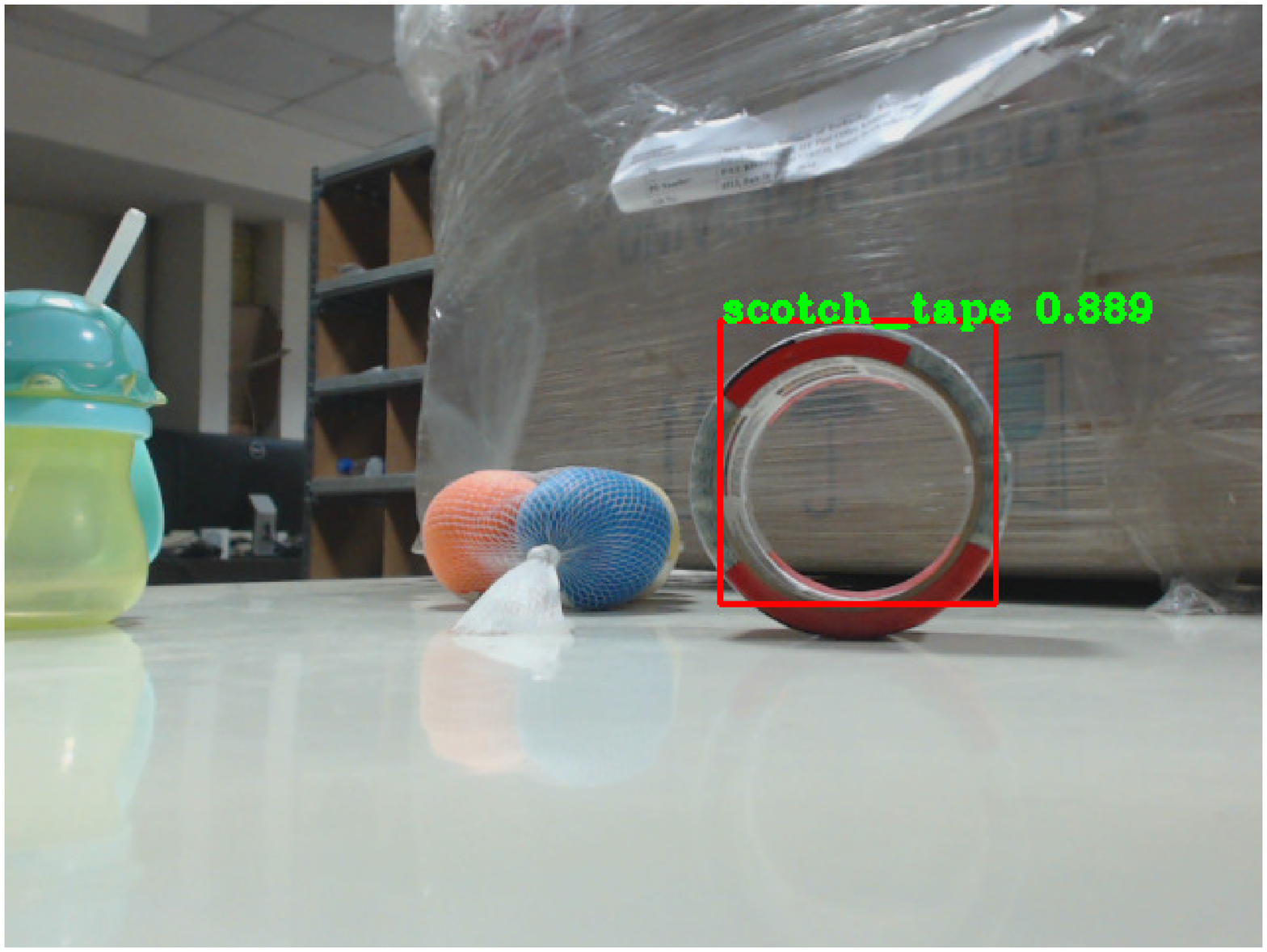} \\
    \includegraphics[scale=0.08]{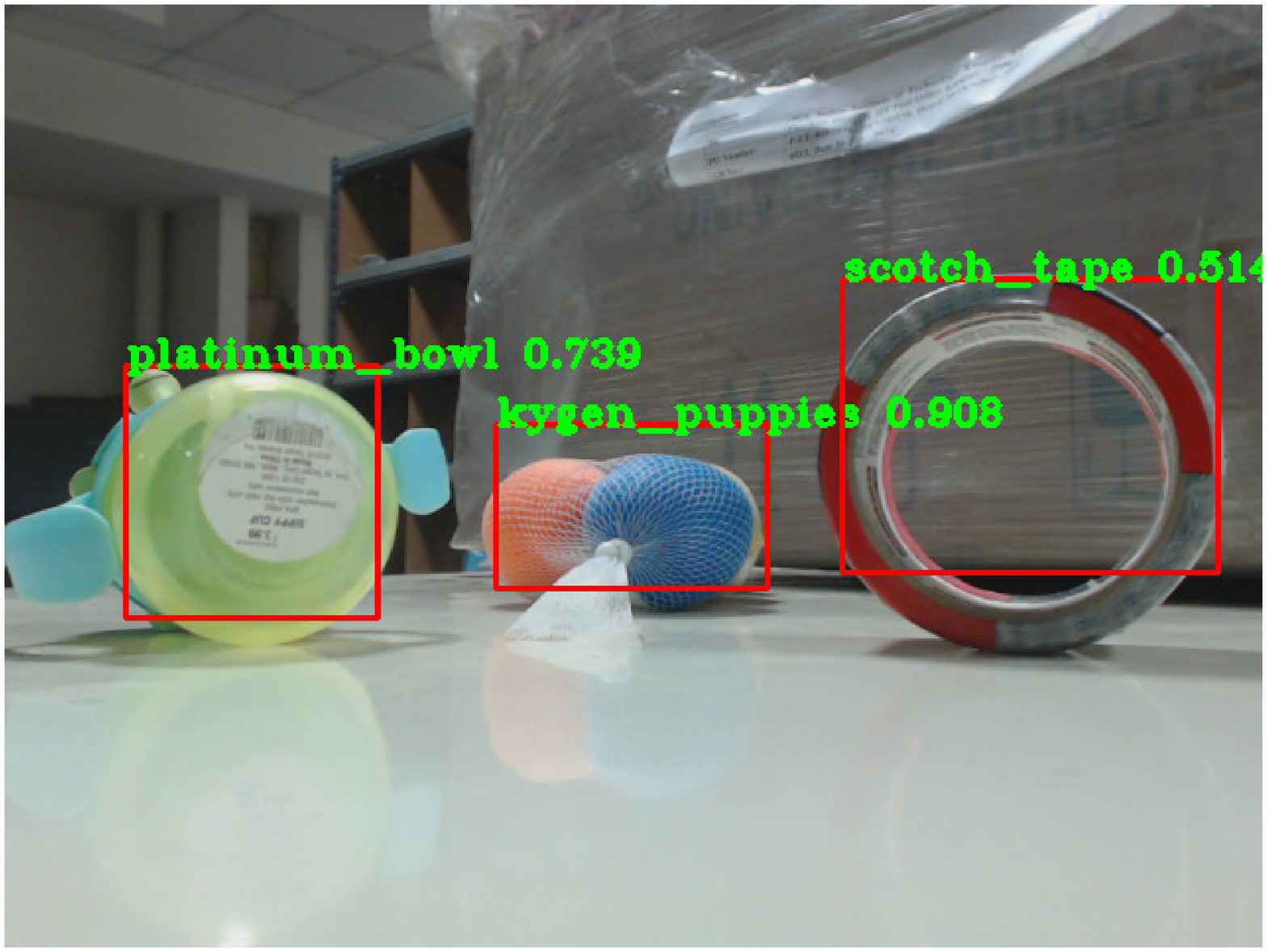} &
    \includegraphics[scale=0.08]{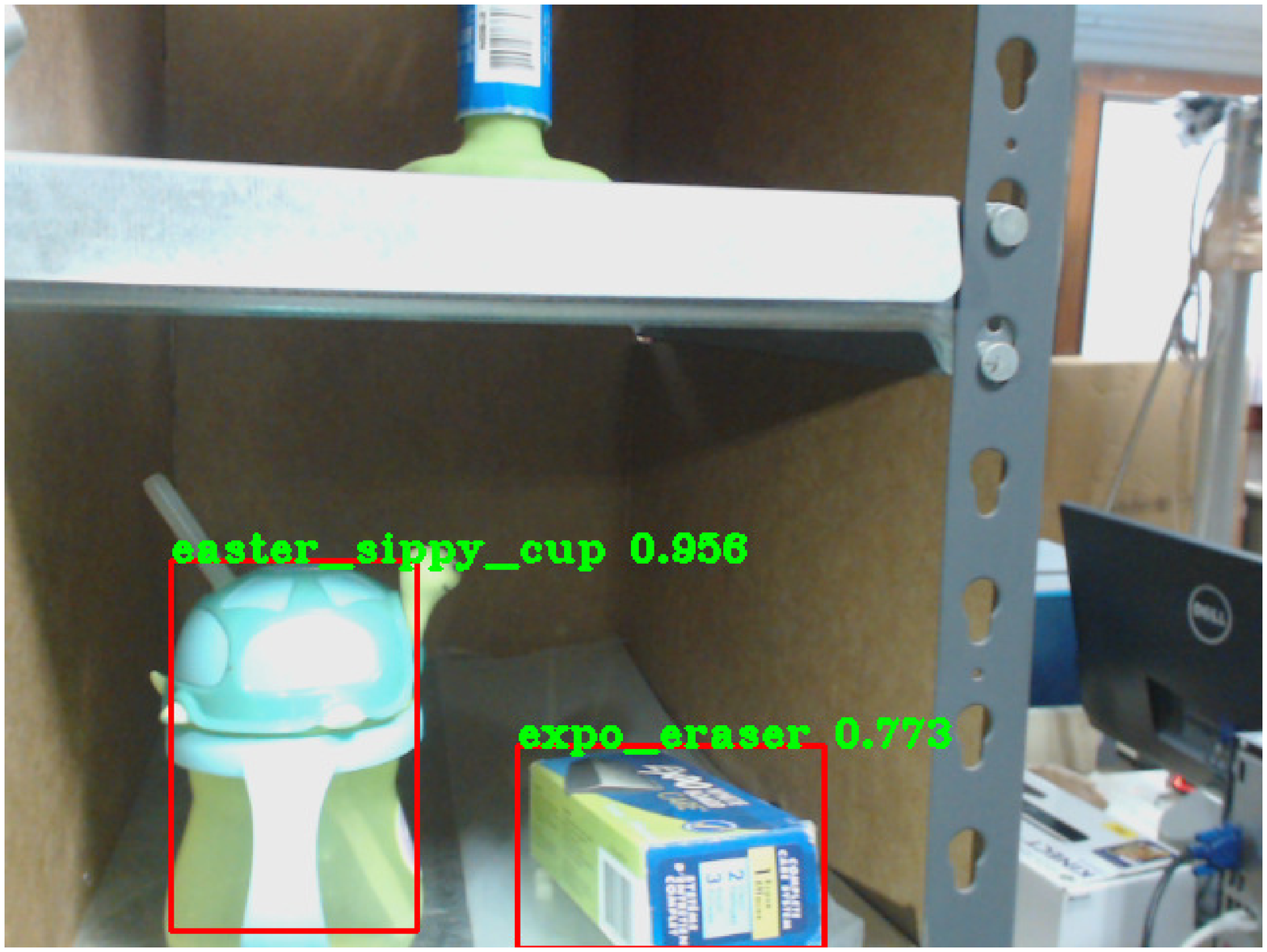} &
    \includegraphics[scale=0.08]{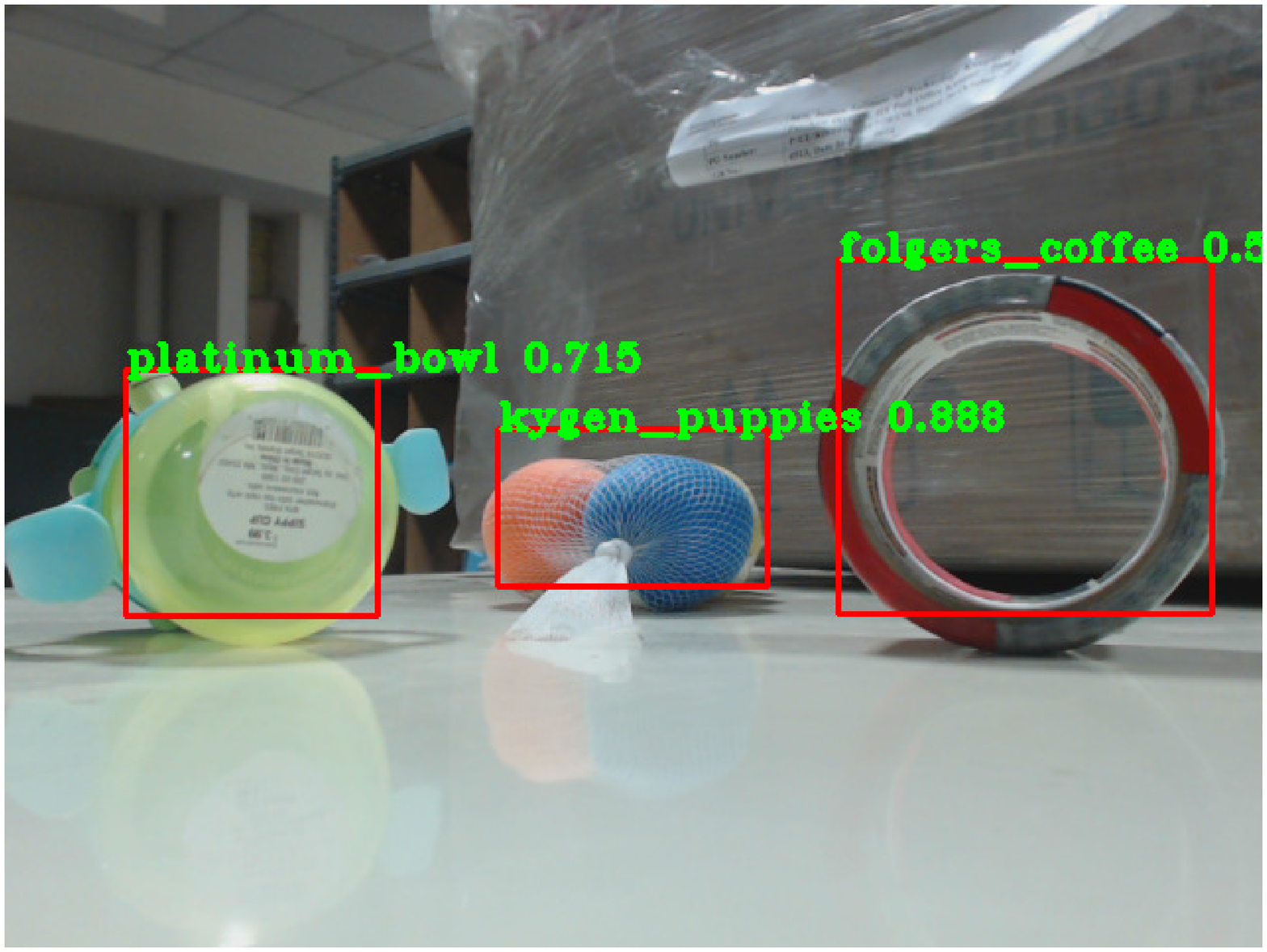} &
    \includegraphics[scale=0.08]{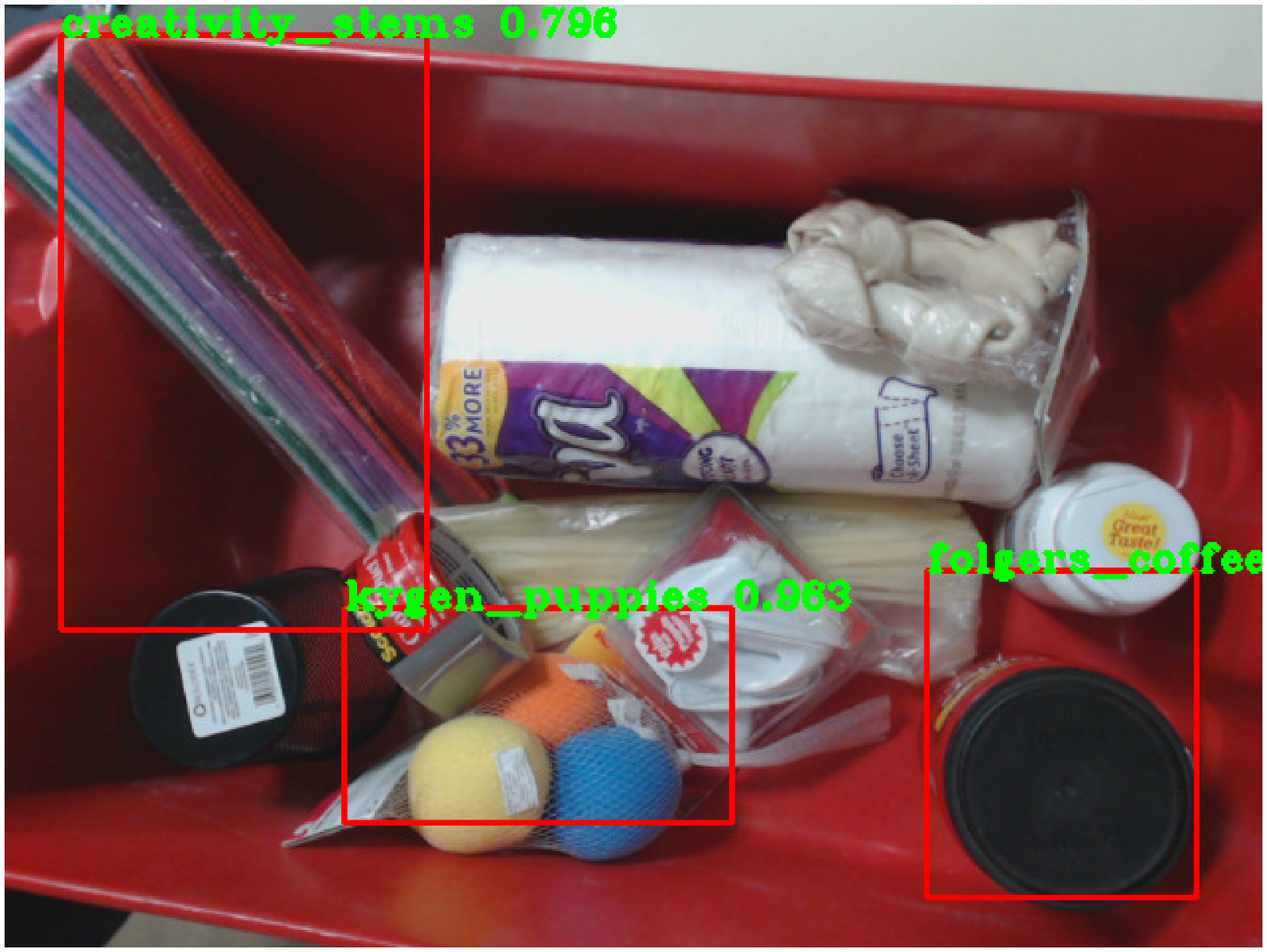} 
\end{tabular}
  \caption{Output of RCNN after training. The objects are detected in
  different environments (different background). Each recognized
object is provided a label and a bounding box. }
  \label{fig:rcnn_output}
\end{figure}

\begin{figure}
  \centering
  \includegraphics[scale=0.75]{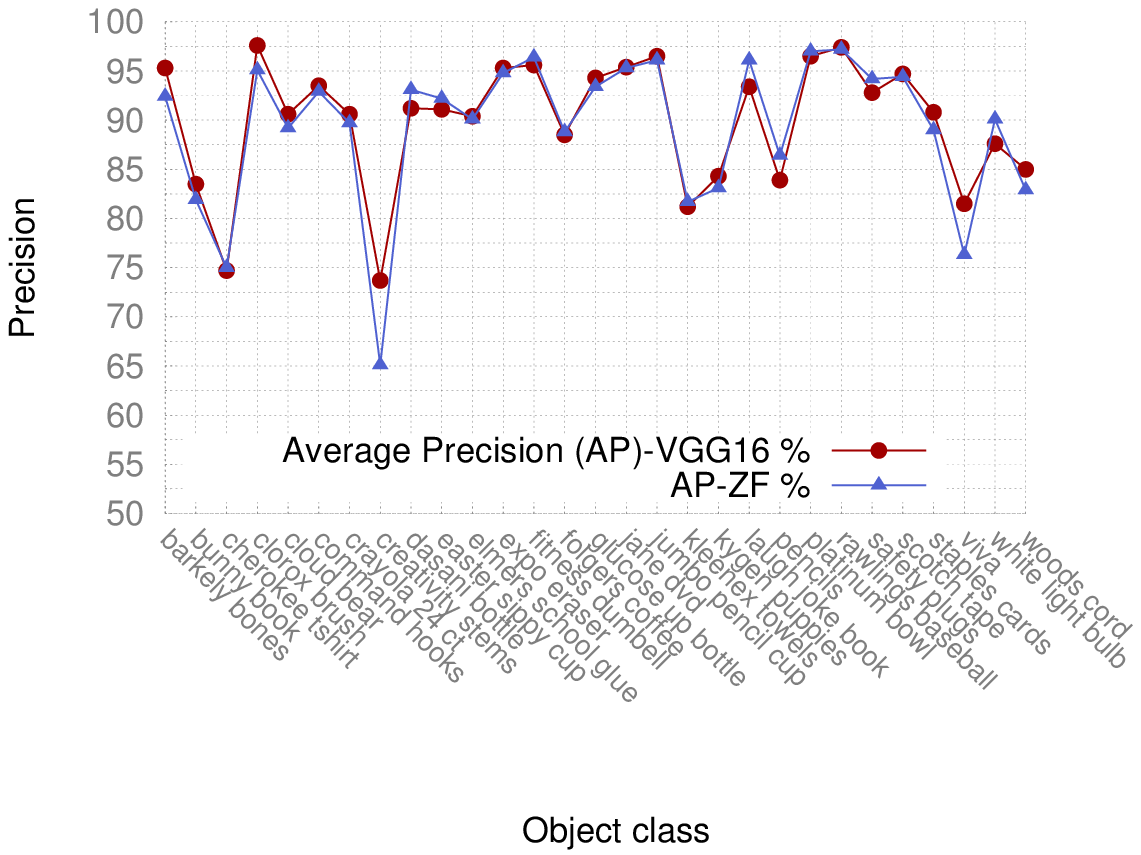} 
  \caption{Plot showing average precision of individual objects obtained using Faster RCNN for both VGG-16 and ZF model}
  \label{fig:prec_plot}
\end{figure}

Experiments on object recognition are performed using our APC dataset
with 6000 images for 40 different objects. The images are taken at
different lighting conditions with various background.  Pretrained
VGG-16 model of the Faster R-CNN is fine tuned using $80\%$ of the
whole dataset and remaining $20\%$ is used to validate the recognition
performance. Figure \ref{fig:rcnn_output} presents some object
recognition results when tested with new images. Statistical analysis
have been carried out on the validation set. We have achieved a mean
Average Precision (mAP) of $89.9\%$ for our validation set, which is a
pretty good performance for such an unconstrained and challenging
environment. The individual precision of randomly picked $29$ objects
and their mAP are shown in Table \ref{tab:map-ind_p}.  Observation
shows that, when the objects are deformable, such as cherokee tshirt
and creativity stems, the precision are reasonably lower. In our case,
the precisions are $74.7\%$ and $73.65\%$ respectively.  The
performance can be boosted if the size of the dataset is increased
with new set of images.  Detailed information of the experimental
setup are given in the Table \ref{tab:exptsum}. GPU system NVIDIA
Quadro M5000M is used to train the Faster R-CNN VGG-16 model. Objects
in an image are detected in just $0.125$ second, which is in
real-time. In order to compare the recognition performance of VGG-16, we 
trained and validated the given dataset using ZF model.  
Object recognition results using VGG-16 is observed to be 
slightly better than that of ZF model (mAP is $89.3\%$ in case of ZF model). 
Average precision of individual objects for both the VGG-16 and the ZF model are shown in the Figure \ref{fig:prec_plot}.


\begin{table}[!h] 
  \centering
  \caption{Experimental details for object recognition task using Faster- RCNN}
  \label{tab:exptsum}
  \begin{tabular}{ccccc}\toprule
    System  & Training & Validation & Testing time & mAP\\
    configuration & data size  & data size & & \\ 
    \midrule
    GPU \cellcolor[gray]{0.8}\textbf{NVIDIA}& 4800  & 1200 & 0.125  & $89.9 \%$\\
    \cellcolor[gray]{0.8}\textbf{Quadro M5000M} & samples & samples & second &\\
    \bottomrule
  \end{tabular}
\end{table}

\begin{table}[!h] \setlength{\tabcolsep}{5pt}
  \centering
 \caption{Mean average precision and per-class average precision}
  \label{tab:map-ind_p}
\tiny{
\begin{tabular}{cccccc}\toprule
\small{mAP} &\multicolumn{4}{c}{\small{per-class average precision}}    \\ \cmidrule[1.0pt]{2-6}
   & barkely bones & bunny book & cherokee tshirt & clorox brush & cloud bear  \\

\multirow{8}{*}{\textbf{89.9}} & 95.31  & 83.51 & 74.70 & 97.63& 90.58 \\

& command hooks & crayola 24 ct & creativity stems  & dasani bottle & easter sippy cup  \\ 

&93.52& 90.57 &73.65  & 91.21 &91.13 \\
&elmers school glue & expo eraser &fitness dumbell  & folgers coffee & glucose up bottle \\
& 90.36& 95.27 &95.64  &88.45  & 94.34 \\
& jane dvd&jumbo pencil cup & kleenex towels & kygen puppies &laugh joke book   \\
&95.43 &96.53 & 81.24 & 84.35 &  93.41 \\
&pencils  &platinum bowl &rawlings baseball &safety plugs & scotch tape \\
&  83.93& 96.54& 97.39& 92.77 & 94.75 \\

&staples cards  &viva  &white lightbulb &woods cord & \\
&  90.84&  81.46& 87.62& 85.01 & \\
\bottomrule
\end{tabular}}
\end{table}

\subsection{Direction for future research }\label{sec:future}

While the current system can carry out the picking and stowing tasks
with reasonable accuracy and speed, a lot of work still needs to be
done before it can be deployed in real world scenarios. Improving the
system further forms the direction of our future work. Some of the
ways of improving the system is as follows:

\begin{itemize}
  \item The performance of the system relies on the performance of
    individual modules, particularly, perception module for object
    recognition and grasping. One of the future direction would be
    to carry out research towards improving the performance of the
    perception module.

  \item One of the challenges of deep learning based approaches for
    perception and grasping is the amount of samples required for
    training such models. Most of these training samples are created
    manually which is laborious and slow. One of our future direction
    would be to automate the process of data generation and explore
    deep learning models that can be trained incrementally
    \cite{xiao2014error} or through transfer learning
    \cite{shin2016deep}\cite{yosinski2014transferable}. 

  \item The design of custom grippers that can pick all types of
    objects, including soft and deformable objects, still remains a
    challenge. Picking items from a closely packed stack of objects is
    another challenge which will be looked into as a part of our
    future research. 

  \item The real-time performance of the system needs to be improved
    further without increasing the infrastructure cost. This can be
    done by parallelizing several modules, improving CPU utilization and
    reducing network latency.  The use of state machine based
    software architecture \cite{herrero2015state} such as ROS SMACH
    \cite{bohren2010smach}\cite{field2011smach} will be explored as a
    part of our future work. 

  \item Even though the existing motion planning algorithms are quite
    mature, it is still not possible to deal with all possible cases
    of failure. One possible way to deal with these extreme cases
    would be to have a human in loop that intervenes only when a
    failure is reported by the system. The human operator can then
    teach the robot through demonstration \cite{calinon2007active}
    \cite{maeda2002teaching} to deal with such situations. The robot,
    in turn, can learn from such human inputs over a period of time to
    deal with such complex situations through methods such as deep
    reinforcement learning \cite{lillicrap2015continuous}
    \cite{levine2016learning} \cite{zhang2015towards}. Some of these
    directions will be explored in the future. 
    
    \item The real-world deployment of such systems will be explored
      through the use of Cloud Robotics platforms like Rapyuta
      \cite{mohanarajah2015rapyuta}.  
   
   \end{itemize}

\section{Conclusions} \label{sec:conc}

This paper presents the details of a robot-arm based automatic pick
and place system for a retail warehouse. The two major tasks which are
considered here include (1) \emph{picking} items from a rack into a
tote and, (2) \emph{stowing} items from a tote to a rack. These two
tasks are currently done manually by employing a large number of
people in a big warehouse. This work was carried out as a part of our
preparation for participating in the Amazon Picking Challenge 2016
event held in Leipzig, Germany. The problem is challenging from
several perspectives. Identifying objects from visual images under
conditions of varying illumination, occlusion, scaling, rotation and
change of shape (for deformable objects) is an open problem in the
computer vision literature. This problem is solved in this work by the
use of deep learning networks (RCNN) which gives reasonable accuracy
suitable for real world operation. The average recognition accuracy is
about $90\pm 5\%$ for 29 objects which is obtained by training the
RCNN network using 4800 training samples. This is quite small compared
what was used by other teams at the competition. The lesser accuracy
(resulting in bigger bounding boxes for objects) of the recognition
module is compensated by the proposed grasping algorithm that uses
surface normals, curvatures in the depth space to segment the target
object from background, identify its shape and find suitable graspable
affordance which could be used for picking the object either using a
suction-based end-effector or a two-finger gripper. Apart from these
issues, the constraints on real-time performance also pose significant
challenge to real-world deployment of such systems. Currently, we are
able to achieve a pick or stow rate of 2.5 objects per minute under
normal conditions.

\bibliographystyle{ieeetr}
\bibliography{apc_sk}

\end{document}

%% file: fig/FlowchartStyle.tex
\tikzstyle{every node}=[font=\footnotesize]
\tikzstyle{io} = [trapezium, trapezium left angle=70, trapezium right angle=110, minimum width=3cm, minimum height=.5cm, text centered, draw=black, fill=blue!30]
\tikzstyle{process} = [rectangle, minimum width=3cm, minimum height=.5cm, text centered, draw=black, fill=orange!30]
\tikzstyle{arrow} = [thick,->,>=stealth]
\tikzstyle{cloud} = [draw, ellipse,fill=red!20, node distance=3cm, minimum height=.5cm]
\tikzstyle{decision} = [diamond, draw, fill=blue!20,
    text width=4em,text badly centered, node distance=3cm, inner sep=0pt]
\tikzstyle{block} = [rectangle, draw, fill=blue!20, 
    text width=5em, text centered, rounded corners, minimum height=.5cm]
\tikzstyle{line} = [draw, -latex']
\tikzstyle{cloud} = [draw, ellipse,fill=red!20, node distance=3cm, minimum height=.5cm]
\tikzstyle{blank} = [node distance=.5cm]
 \tikzset{line/.style={draw, thick, -latex'}}

\tikzstyle{sensor}=[draw, fill=blue!20, text width=5em, 
    text centered, minimum height=2.5em]
    \tikzstyle{process}=[draw, fill=blue!20, text width=5em, 
    text centered, minimum height=2.5em]
\tikzstyle{io}=[draw, fill=blue!20, text width=5em, 
    text centered, minimum height=2.5em]

\tikzstyle{ann} = [above, text width=5em]
\tikzstyle{naveqs} = [sensor, text width=6em, fill=red!20, 
    minimum height=20em, rounded corners]

%% file: fig/apc_ros_arch2.pstex_t
\begin{picture}(0,0)%
\includegraphics{apc_ros_arch2.pstex}%
\end{picture}%
\setlength{\unitlength}{4144sp}%
\begingroup\makeatletter\ifx\SetFigFont\undefined%
\gdef\SetFigFont#1#2#3#4#5{%
  \reset@font\fontsize{#1}{#2pt}%
  \fontfamily{#3}\fontseries{#4}\fontshape{#5}%
  \selectfont}%
\fi\endgroup%
\begin{picture}(21194,13319)(-4521,-10883)
\put(2026,-7486){\makebox(0,0)[lb]{\smash{{\SetFigFont{17}{20.4}{\familydefault}{\mddefault}{\updefault}{\color[rgb]{0,0,0}PC3}%
}}}}
\put(1801,-8161){\makebox(0,0)[lb]{\smash{{\SetFigFont{17}{20.4}{\familydefault}{\mddefault}{\updefault}{\color[rgb]{0,0,0}IR\_Read}%
}}}}
\put(14626,-3661){\makebox(0,0)[lb]{\smash{{\SetFigFont{17}{20.4}{\familydefault}{\mddefault}{\updefault}{\color[rgb]{0,0,0}Service for}%
}}}}
\put(14626,-4021){\makebox(0,0)[lb]{\smash{{\SetFigFont{17}{20.4}{\familydefault}{\mddefault}{\updefault}{\color[rgb]{0,0,0}writing object}%
}}}}
\put(14626,-4381){\makebox(0,0)[lb]{\smash{{\SetFigFont{17}{20.4}{\familydefault}{\mddefault}{\updefault}{\color[rgb]{0,0,0}status}%
}}}}
\put(8551,-7711){\makebox(0,0)[lb]{\smash{{\SetFigFont{17}{20.4}{\familydefault}{\mddefault}{\updefault}{\color[rgb]{0,0,0}PC3}%
}}}}
\put(8101,-8386){\makebox(0,0)[lb]{\smash{{\SetFigFont{17}{20.4}{\familydefault}{\mddefault}{\updefault}{\color[rgb]{0,0,0}Toggle\_LED}%
}}}}
\put(1576,-736){\makebox(0,0)[lb]{\smash{{\SetFigFont{17}{20.4}{\familydefault}{\mddefault}{\updefault}{\color[rgb]{0,0,0}PC3}%
}}}}
\put(1126,-1411){\makebox(0,0)[lb]{\smash{{\SetFigFont{17}{20.4}{\familydefault}{\mddefault}{\updefault}{\color[rgb]{0,0,0}Vacuum\_Pump}%
}}}}
\put(-449,-3436){\makebox(0,0)[lb]{\smash{{\SetFigFont{17}{20.4}{\familydefault}{\mddefault}{\updefault}{\color[rgb]{0,0,0}UR5\_Controller}%
}}}}
\put(  1,-2761){\makebox(0,0)[lb]{\smash{{\SetFigFont{17}{20.4}{\familydefault}{\mddefault}{\updefault}{\color[rgb]{0,0,0}PC1}%
}}}}
\put(-674,-5911){\makebox(0,0)[lb]{\smash{{\SetFigFont{17}{20.4}{\familydefault}{\mddefault}{\updefault}{\color[rgb]{0,0,0}Kinect\_Read}%
}}}}
\put(-224,-5236){\makebox(0,0)[lb]{\smash{{\SetFigFont{17}{20.4}{\familydefault}{\mddefault}{\updefault}{\color[rgb]{0,0,0}PC3}%
}}}}
\put(-4049,-5686){\makebox(0,0)[lb]{\smash{{\SetFigFont{17}{20.4}{\familydefault}{\mddefault}{\updefault}{\color[rgb]{0,0,0}Topic for}%
}}}}
\put(-4049,-6046){\makebox(0,0)[lb]{\smash{{\SetFigFont{17}{20.4}{\familydefault}{\mddefault}{\updefault}{\color[rgb]{0,0,0}reading}%
}}}}
\put(-4049,-6406){\makebox(0,0)[lb]{\smash{{\SetFigFont{17}{20.4}{\familydefault}{\mddefault}{\updefault}{\color[rgb]{0,0,0}point cloud}%
}}}}
\put(-4049,-6766){\makebox(0,0)[lb]{\smash{{\SetFigFont{17}{20.4}{\familydefault}{\mddefault}{\updefault}{\color[rgb]{0,0,0}data}%
}}}}
\put(13951,-1996){\makebox(0,0)[lb]{\smash{{\SetFigFont{17}{20.4}{\familydefault}{\mddefault}{\updefault}{\color[rgb]{0,0,0}object name}%
}}}}
\put(13951,-1636){\makebox(0,0)[lb]{\smash{{\SetFigFont{17}{20.4}{\familydefault}{\mddefault}{\updefault}{\color[rgb]{0,0,0}Service getting}%
}}}}
\put(11476,614){\makebox(0,0)[lb]{\smash{{\SetFigFont{17}{20.4}{\familydefault}{\mddefault}{\updefault}{\color[rgb]{0,0,0}motion planning}%
}}}}
\put(11926,839){\makebox(0,0)[lb]{\smash{{\SetFigFont{17}{20.4}{\familydefault}{\mddefault}{\updefault}{\color[rgb]{0,0,0}Service for}%
}}}}
\put(4726,-10186){\makebox(0,0)[lb]{\smash{{\SetFigFont{17}{20.4}{\familydefault}{\mddefault}{\updefault}{\color[rgb]{0,0,0}Service for}%
}}}}
\put(4726,-10546){\makebox(0,0)[lb]{\smash{{\SetFigFont{17}{20.4}{\familydefault}{\mddefault}{\updefault}{\color[rgb]{0,0,0}Rack Registration}%
}}}}
\put(11251,-2311){\makebox(0,0)[lb]{\smash{{\SetFigFont{17}{20.4}{\familydefault}{\mddefault}{\updefault}{\color[rgb]{0,0,0}PC3}%
}}}}
\put(10801,-2986){\makebox(0,0)[lb]{\smash{{\SetFigFont{17}{20.4}{\familydefault}{\mddefault}{\updefault}{\color[rgb]{0,0,0}JSON\_RD\_WR}%
}}}}
\put(11476,-3886){\makebox(0,0)[lb]{\smash{{\SetFigFont{17}{20.4}{\familydefault}{\mddefault}{\updefault}{\color[rgb]{0,0,0}PC3}%
}}}}
\put(11026,-4561){\makebox(0,0)[lb]{\smash{{\SetFigFont{17}{20.4}{\familydefault}{\mddefault}{\updefault}{\color[rgb]{0,0,0}JSON\_RD\_WR}%
}}}}
\put(10801,-6136){\makebox(0,0)[lb]{\smash{{\SetFigFont{17}{20.4}{\familydefault}{\mddefault}{\updefault}{\color[rgb]{0,0,0}PC2}%
}}}}
\put(9901,-6811){\makebox(0,0)[lb]{\smash{{\SetFigFont{17}{20.4}{\familydefault}{\mddefault}{\updefault}{\color[rgb]{0,0,0}RCNN\_Object\_Detect}%
}}}}
\put(-3374,-511){\makebox(0,0)[lb]{\smash{{\SetFigFont{17}{20.4}{\familydefault}{\mddefault}{\updefault}{\color[rgb]{0,0,0}Service for}%
}}}}
\put(-3374,-871){\makebox(0,0)[lb]{\smash{{\SetFigFont{17}{20.4}{\familydefault}{\mddefault}{\updefault}{\color[rgb]{0,0,0}sending joint}%
}}}}
\put(-3374,-1231){\makebox(0,0)[lb]{\smash{{\SetFigFont{17}{20.4}{\familydefault}{\mddefault}{\updefault}{\color[rgb]{0,0,0}angles to robot}%
}}}}
\put(-3374,-1636){\makebox(0,0)[lb]{\smash{{\SetFigFont{17}{20.4}{\familydefault}{\mddefault}{\updefault}{\color[rgb]{0,0,0}arm}%
}}}}
\put(-1799,-9871){\makebox(0,0)[lb]{\smash{{\SetFigFont{17}{20.4}{\familydefault}{\mddefault}{\updefault}{\color[rgb]{0,0,0}reading IR}%
}}}}
\put(-1799,-10231){\makebox(0,0)[lb]{\smash{{\SetFigFont{17}{20.4}{\familydefault}{\mddefault}{\updefault}{\color[rgb]{0,0,0}values}%
}}}}
\put(-1799,-9511){\makebox(0,0)[lb]{\smash{{\SetFigFont{17}{20.4}{\familydefault}{\mddefault}{\updefault}{\color[rgb]{0,0,0}Service for }%
}}}}
\put(-449,1604){\makebox(0,0)[lb]{\smash{{\SetFigFont{17}{20.4}{\familydefault}{\mddefault}{\updefault}{\color[rgb]{0,0,0}switching on}%
}}}}
\put(-449,1244){\makebox(0,0)[lb]{\smash{{\SetFigFont{17}{20.4}{\familydefault}{\mddefault}{\updefault}{\color[rgb]{0,0,0}and off}%
}}}}
\put(-449,1964){\makebox(0,0)[lb]{\smash{{\SetFigFont{17}{20.4}{\familydefault}{\mddefault}{\updefault}{\color[rgb]{0,0,0}Service for }%
}}}}
\put(-449,884){\makebox(0,0)[lb]{\smash{{\SetFigFont{17}{20.4}{\familydefault}{\mddefault}{\updefault}{\color[rgb]{0,0,0}vacuum pump}%
}}}}
\put(11251,-9736){\makebox(0,0)[lb]{\smash{{\SetFigFont{17}{20.4}{\familydefault}{\mddefault}{\updefault}{\color[rgb]{0,0,0}Topic to}%
}}}}
\put(11251,-10096){\makebox(0,0)[lb]{\smash{{\SetFigFont{17}{20.4}{\familydefault}{\mddefault}{\updefault}{\color[rgb]{0,0,0}toggle LED}%
}}}}
\put(14176,-7711){\makebox(0,0)[lb]{\smash{{\SetFigFont{17}{20.4}{\familydefault}{\mddefault}{\updefault}{\color[rgb]{0,0,0}Service for}%
}}}}
\put(14176,-8071){\makebox(0,0)[lb]{\smash{{\SetFigFont{17}{20.4}{\familydefault}{\mddefault}{\updefault}{\color[rgb]{0,0,0}localizing}%
}}}}
\put(14176,-8431){\makebox(0,0)[lb]{\smash{{\SetFigFont{17}{20.4}{\familydefault}{\mddefault}{\updefault}{\color[rgb]{0,0,0}object}%
}}}}
\put(5626,-3886){\makebox(0,0)[lb]{\smash{{\SetFigFont{17}{20.4}{\familydefault}{\mddefault}{\updefault}{\color[rgb]{0,0,0}PC3}%
}}}}
\put(4951,-4561){\makebox(0,0)[lb]{\smash{{\SetFigFont{17}{20.4}{\familydefault}{\mddefault}{\updefault}{\color[rgb]{0,0,0}apc\_controller}%
}}}}
\put(6076,-1861){\makebox(0,0)[lb]{\smash{{\SetFigFont{17}{20.4}{\familydefault}{\mddefault}{\updefault}{\color[rgb]{0,0,0}/bin\_corner}%
}}}}
\put(5401,-1186){\makebox(0,0)[lb]{\smash{{\SetFigFont{17}{20.4}{\familydefault}{\mddefault}{\updefault}{\color[rgb]{0,0,0}RVIZ}%
}}}}
\put(4726,1514){\makebox(0,0)[lb]{\smash{{\SetFigFont{17}{20.4}{\familydefault}{\mddefault}{\updefault}{\color[rgb]{0,0,0}Service to show bin}%
}}}}
\put(4726,1154){\makebox(0,0)[lb]{\smash{{\SetFigFont{17}{20.4}{\familydefault}{\mddefault}{\updefault}{\color[rgb]{0,0,0}markers in rviz}%
}}}}
\put(9901,-511){\makebox(0,0)[lb]{\smash{{\SetFigFont{17}{20.4}{\familydefault}{\mddefault}{\updefault}{\color[rgb]{0,0,0}PC3}%
}}}}
\put(5626,-511){\makebox(0,0)[lb]{\smash{{\SetFigFont{17}{20.4}{\familydefault}{\mddefault}{\updefault}{\color[rgb]{0,0,0}PC3}%
}}}}
\put(5626,-7936){\makebox(0,0)[lb]{\smash{{\SetFigFont{17}{20.4}{\familydefault}{\mddefault}{\updefault}{\color[rgb]{0,0,0}PC3}%
}}}}
\put(4951,-8611){\makebox(0,0)[lb]{\smash{{\SetFigFont{17}{20.4}{\familydefault}{\mddefault}{\updefault}{\color[rgb]{0,0,0}Rack\_Registration}%
}}}}
\put(9226,-1186){\makebox(0,0)[lb]{\smash{{\SetFigFont{17}{20.4}{\familydefault}{\mddefault}{\updefault}{\color[rgb]{0,0,0}Trajectory\_RPT}%
}}}}
\put(8101,-3661){\makebox(0,0)[lb]{\smash{{\SetFigFont{17}{20.4}{\familydefault}{\mddefault}{\updefault}{\color[rgb]{0,0,0}/pick\_object}%
}}}}
\put(8101,-4561){\makebox(0,0)[lb]{\smash{{\SetFigFont{17}{20.4}{\familydefault}{\mddefault}{\updefault}{\color[rgb]{0,0,0}/pick\_object\_status}%
}}}}
\put(8776,-2536){\makebox(0,0)[lb]{\smash{{\SetFigFont{17}{20.4}{\familydefault}{\mddefault}{\updefault}{\color[rgb]{0,0,0}/trajectory\_rpt}%
}}}}
\put(8776,-5461){\makebox(0,0)[lb]{\smash{{\SetFigFont{17}{20.4}{\familydefault}{\mddefault}{\updefault}{\color[rgb]{0,0,0}/detect\_object}%
}}}}
\put(7651,-6811){\makebox(0,0)[lb]{\smash{{\SetFigFont{17}{20.4}{\familydefault}{\mddefault}{\updefault}{\color[rgb]{0,0,0}/toggle\_led}%
}}}}
\put(6076,-7036){\makebox(0,0)[lb]{\smash{{\SetFigFont{17}{20.4}{\familydefault}{\mddefault}{\updefault}{\color[rgb]{0,0,0}/rack\_regn}%
}}}}
\put(3826,-7036){\makebox(0,0)[lb]{\smash{{\SetFigFont{17}{20.4}{\familydefault}{\mddefault}{\updefault}{\color[rgb]{0,0,0}/ir\_topic}%
}}}}
\put(3826,-2086){\makebox(0,0)[lb]{\smash{{\SetFigFont{17}{20.4}{\familydefault}{\mddefault}{\updefault}{\color[rgb]{0,0,0}/vacuum\_topic}%
}}}}
\put(1801,-5686){\makebox(0,0)[lb]{\smash{{\SetFigFont{17}{20.4}{\familydefault}{\mddefault}{\updefault}{\color[rgb]{0,0,0}/camera/depth\_registration}%
}}}}
\put(2026,-3211){\makebox(0,0)[lb]{\smash{{\SetFigFont{17}{20.4}{\familydefault}{\mddefault}{\updefault}{\color[rgb]{0,0,0}/ur5\_single\_goal}%
}}}}
\end{picture}%

%% file: fig/thnorm.tex
\begingroup%
\makeatletter%
\newcommand{\GNUPLOTspecial}{%
  \@sanitize\catcode`\%=14\relax\special}%
\setlength{\unitlength}{0.0500bp}%
\begin{picture}(7200,5040)(0,0)%
  {\GNUPLOTspecial{"
/gnudict 256 dict def
gnudict begin
%
%
/Color true def
/Blacktext true def
/Solid true def
/Dashlength 1 def
/Landscape false def
/Level1 false def
/Rounded false def
/ClipToBoundingBox false def
/SuppressPDFMark false def
/TransparentPatterns false def
/gnulinewidth 10.000 def
/userlinewidth gnulinewidth def
/Gamma 1.0 def
/BackgroundColor {-1.000 -1.000 -1.000} def
/vshift -66 def
/dl1 {
  10.0 Dashlength mul mul
  Rounded { currentlinewidth 0.75 mul sub dup 0 le { pop 0.01 } if } if
} def
/dl2 {
  10.0 Dashlength mul mul
  Rounded { currentlinewidth 0.75 mul add } if
} def
/hpt_ 31.5 def
/vpt_ 31.5 def
/hpt hpt_ def
/vpt vpt_ def
/doclip {
  ClipToBoundingBox {
    newpath 0 0 moveto 360 0 lineto 360 252 lineto 0 252 lineto closepath
    clip
  } if
} def
%
%
%
/M {moveto} bind def
/L {lineto} bind def
/R {rmoveto} bind def
/V {rlineto} bind def
/N {newpath moveto} bind def
/Z {closepath} bind def
/C {setrgbcolor} bind def
/f {rlineto fill} bind def
/g {setgray} bind def
/Gshow {show} def   
/vpt2 vpt 2 mul def
/hpt2 hpt 2 mul def
/Lshow {currentpoint stroke M 0 vshift R 
	Blacktext {gsave 0 setgray show grestore} {show} ifelse} def
/Rshow {currentpoint stroke M dup stringwidth pop neg vshift R
	Blacktext {gsave 0 setgray show grestore} {show} ifelse} def
/Cshow {currentpoint stroke M dup stringwidth pop -2 div vshift R 
	Blacktext {gsave 0 setgray show grestore} {show} ifelse} def
/UP {dup vpt_ mul /vpt exch def hpt_ mul /hpt exch def
  /hpt2 hpt 2 mul def /vpt2 vpt 2 mul def} def
/DL {Color {setrgbcolor Solid {pop []} if 0 setdash}
 {pop pop pop 0 setgray Solid {pop []} if 0 setdash} ifelse} def
/BL {stroke userlinewidth 2 mul setlinewidth
	Rounded {1 setlinejoin 1 setlinecap} if} def
/AL {stroke userlinewidth 2 div setlinewidth
	Rounded {1 setlinejoin 1 setlinecap} if} def
/UL {dup gnulinewidth mul /userlinewidth exch def
	dup 1 lt {pop 1} if 10 mul /udl exch def} def
/PL {stroke userlinewidth setlinewidth
	Rounded {1 setlinejoin 1 setlinecap} if} def
3.8 setmiterlimit
/LCw {1 1 1} def
/LCb {0 0 0} def
/LCa {0 0 0} def
/LC0 {1 0 0} def
/LC1 {0 1 0} def
/LC2 {0 0 1} def
/LC3 {1 0 1} def
/LC4 {0 1 1} def
/LC5 {1 1 0} def
/LC6 {0 0 0} def
/LC7 {1 0.3 0} def
/LC8 {0.5 0.5 0.5} def
/LTw {PL [] 1 setgray} def
/LTb {BL [] LCb DL} def
/LTa {AL [1 udl mul 2 udl mul] 0 setdash LCa setrgbcolor} def
/LT0 {PL [] LC0 DL} def
/LT1 {PL [4 dl1 2 dl2] LC1 DL} def
/LT2 {PL [2 dl1 3 dl2] LC2 DL} def
/LT3 {PL [1 dl1 1.5 dl2] LC3 DL} def
/LT4 {PL [6 dl1 2 dl2 1 dl1 2 dl2] LC4 DL} def
/LT5 {PL [3 dl1 3 dl2 1 dl1 3 dl2] LC5 DL} def
/LT6 {PL [2 dl1 2 dl2 2 dl1 6 dl2] LC6 DL} def
/LT7 {PL [1 dl1 2 dl2 6 dl1 2 dl2 1 dl1 2 dl2] LC7 DL} def
/LT8 {PL [2 dl1 2 dl2 2 dl1 2 dl2 2 dl1 2 dl2 2 dl1 4 dl2] LC8 DL} def
/Pnt {stroke [] 0 setdash gsave 1 setlinecap M 0 0 V stroke grestore} def
/Dia {stroke [] 0 setdash 2 copy vpt add M
  hpt neg vpt neg V hpt vpt neg V
  hpt vpt V hpt neg vpt V closepath stroke
  Pnt} def
/Pls {stroke [] 0 setdash vpt sub M 0 vpt2 V
  currentpoint stroke M
  hpt neg vpt neg R hpt2 0 V stroke
 } def
/Box {stroke [] 0 setdash 2 copy exch hpt sub exch vpt add M
  0 vpt2 neg V hpt2 0 V 0 vpt2 V
  hpt2 neg 0 V closepath stroke
  Pnt} def
/Crs {stroke [] 0 setdash exch hpt sub exch vpt add M
  hpt2 vpt2 neg V currentpoint stroke M
  hpt2 neg 0 R hpt2 vpt2 V stroke} def
/TriU {stroke [] 0 setdash 2 copy vpt 1.12 mul add M
  hpt neg vpt -1.62 mul V
  hpt 2 mul 0 V
  hpt neg vpt 1.62 mul V closepath stroke
  Pnt} def
/Star {2 copy Pls Crs} def
/BoxF {stroke [] 0 setdash exch hpt sub exch vpt add M
  0 vpt2 neg V hpt2 0 V 0 vpt2 V
  hpt2 neg 0 V closepath fill} def
/TriUF {stroke [] 0 setdash vpt 1.12 mul add M
  hpt neg vpt -1.62 mul V
  hpt 2 mul 0 V
  hpt neg vpt 1.62 mul V closepath fill} def
/TriD {stroke [] 0 setdash 2 copy vpt 1.12 mul sub M
  hpt neg vpt 1.62 mul V
  hpt 2 mul 0 V
  hpt neg vpt -1.62 mul V closepath stroke
  Pnt} def
/TriDF {stroke [] 0 setdash vpt 1.12 mul sub M
  hpt neg vpt 1.62 mul V
  hpt 2 mul 0 V
  hpt neg vpt -1.62 mul V closepath fill} def
/DiaF {stroke [] 0 setdash vpt add M
  hpt neg vpt neg V hpt vpt neg V
  hpt vpt V hpt neg vpt V closepath fill} def
/Pent {stroke [] 0 setdash 2 copy gsave
  translate 0 hpt M 4 {72 rotate 0 hpt L} repeat
  closepath stroke grestore Pnt} def
/PentF {stroke [] 0 setdash gsave
  translate 0 hpt M 4 {72 rotate 0 hpt L} repeat
  closepath fill grestore} def
/Circle {stroke [] 0 setdash 2 copy
  hpt 0 360 arc stroke Pnt} def
/CircleF {stroke [] 0 setdash hpt 0 360 arc fill} def
/C0 {BL [] 0 setdash 2 copy moveto vpt 90 450 arc} bind def
/C1 {BL [] 0 setdash 2 copy moveto
	2 copy vpt 0 90 arc closepath fill
	vpt 0 360 arc closepath} bind def
/C2 {BL [] 0 setdash 2 copy moveto
	2 copy vpt 90 180 arc closepath fill
	vpt 0 360 arc closepath} bind def
/C3 {BL [] 0 setdash 2 copy moveto
	2 copy vpt 0 180 arc closepath fill
	vpt 0 360 arc closepath} bind def
/C4 {BL [] 0 setdash 2 copy moveto
	2 copy vpt 180 270 arc closepath fill
	vpt 0 360 arc closepath} bind def
/C5 {BL [] 0 setdash 2 copy moveto
	2 copy vpt 0 90 arc
	2 copy moveto
	2 copy vpt 180 270 arc closepath fill
	vpt 0 360 arc} bind def
/C6 {BL [] 0 setdash 2 copy moveto
	2 copy vpt 90 270 arc closepath fill
	vpt 0 360 arc closepath} bind def
/C7 {BL [] 0 setdash 2 copy moveto
	2 copy vpt 0 270 arc closepath fill
	vpt 0 360 arc closepath} bind def
/C8 {BL [] 0 setdash 2 copy moveto
	2 copy vpt 270 360 arc closepath fill
	vpt 0 360 arc closepath} bind def
/C9 {BL [] 0 setdash 2 copy moveto
	2 copy vpt 270 450 arc closepath fill
	vpt 0 360 arc closepath} bind def
/C10 {BL [] 0 setdash 2 copy 2 copy moveto vpt 270 360 arc closepath fill
	2 copy moveto
	2 copy vpt 90 180 arc closepath fill
	vpt 0 360 arc closepath} bind def
/C11 {BL [] 0 setdash 2 copy moveto
	2 copy vpt 0 180 arc closepath fill
	2 copy moveto
	2 copy vpt 270 360 arc closepath fill
	vpt 0 360 arc closepath} bind def
/C12 {BL [] 0 setdash 2 copy moveto
	2 copy vpt 180 360 arc closepath fill
	vpt 0 360 arc closepath} bind def
/C13 {BL [] 0 setdash 2 copy moveto
	2 copy vpt 0 90 arc closepath fill
	2 copy moveto
	2 copy vpt 180 360 arc closepath fill
	vpt 0 360 arc closepath} bind def
/C14 {BL [] 0 setdash 2 copy moveto
	2 copy vpt 90 360 arc closepath fill
	vpt 0 360 arc} bind def
/C15 {BL [] 0 setdash 2 copy vpt 0 360 arc closepath fill
	vpt 0 360 arc closepath} bind def
/Rec {newpath 4 2 roll moveto 1 index 0 rlineto 0 exch rlineto
	neg 0 rlineto closepath} bind def
/Square {dup Rec} bind def
/Bsquare {vpt sub exch vpt sub exch vpt2 Square} bind def
/S0 {BL [] 0 setdash 2 copy moveto 0 vpt rlineto BL Bsquare} bind def
/S1 {BL [] 0 setdash 2 copy vpt Square fill Bsquare} bind def
/S2 {BL [] 0 setdash 2 copy exch vpt sub exch vpt Square fill Bsquare} bind def
/S3 {BL [] 0 setdash 2 copy exch vpt sub exch vpt2 vpt Rec fill Bsquare} bind def
/S4 {BL [] 0 setdash 2 copy exch vpt sub exch vpt sub vpt Square fill Bsquare} bind def
/S5 {BL [] 0 setdash 2 copy 2 copy vpt Square fill
	exch vpt sub exch vpt sub vpt Square fill Bsquare} bind def
/S6 {BL [] 0 setdash 2 copy exch vpt sub exch vpt sub vpt vpt2 Rec fill Bsquare} bind def
/S7 {BL [] 0 setdash 2 copy exch vpt sub exch vpt sub vpt vpt2 Rec fill
	2 copy vpt Square fill Bsquare} bind def
/S8 {BL [] 0 setdash 2 copy vpt sub vpt Square fill Bsquare} bind def
/S9 {BL [] 0 setdash 2 copy vpt sub vpt vpt2 Rec fill Bsquare} bind def
/S10 {BL [] 0 setdash 2 copy vpt sub vpt Square fill 2 copy exch vpt sub exch vpt Square fill
	Bsquare} bind def
/S11 {BL [] 0 setdash 2 copy vpt sub vpt Square fill 2 copy exch vpt sub exch vpt2 vpt Rec fill
	Bsquare} bind def
/S12 {BL [] 0 setdash 2 copy exch vpt sub exch vpt sub vpt2 vpt Rec fill Bsquare} bind def
/S13 {BL [] 0 setdash 2 copy exch vpt sub exch vpt sub vpt2 vpt Rec fill
	2 copy vpt Square fill Bsquare} bind def
/S14 {BL [] 0 setdash 2 copy exch vpt sub exch vpt sub vpt2 vpt Rec fill
	2 copy exch vpt sub exch vpt Square fill Bsquare} bind def
/S15 {BL [] 0 setdash 2 copy Bsquare fill Bsquare} bind def
/D0 {gsave translate 45 rotate 0 0 S0 stroke grestore} bind def
/D1 {gsave translate 45 rotate 0 0 S1 stroke grestore} bind def
/D2 {gsave translate 45 rotate 0 0 S2 stroke grestore} bind def
/D3 {gsave translate 45 rotate 0 0 S3 stroke grestore} bind def
/D4 {gsave translate 45 rotate 0 0 S4 stroke grestore} bind def
/D5 {gsave translate 45 rotate 0 0 S5 stroke grestore} bind def
/D6 {gsave translate 45 rotate 0 0 S6 stroke grestore} bind def
/D7 {gsave translate 45 rotate 0 0 S7 stroke grestore} bind def
/D8 {gsave translate 45 rotate 0 0 S8 stroke grestore} bind def
/D9 {gsave translate 45 rotate 0 0 S9 stroke grestore} bind def
/D10 {gsave translate 45 rotate 0 0 S10 stroke grestore} bind def
/D11 {gsave translate 45 rotate 0 0 S11 stroke grestore} bind def
/D12 {gsave translate 45 rotate 0 0 S12 stroke grestore} bind def
/D13 {gsave translate 45 rotate 0 0 S13 stroke grestore} bind def
/D14 {gsave translate 45 rotate 0 0 S14 stroke grestore} bind def
/D15 {gsave translate 45 rotate 0 0 S15 stroke grestore} bind def
/DiaE {stroke [] 0 setdash vpt add M
  hpt neg vpt neg V hpt vpt neg V
  hpt vpt V hpt neg vpt V closepath stroke} def
/BoxE {stroke [] 0 setdash exch hpt sub exch vpt add M
  0 vpt2 neg V hpt2 0 V 0 vpt2 V
  hpt2 neg 0 V closepath stroke} def
/TriUE {stroke [] 0 setdash vpt 1.12 mul add M
  hpt neg vpt -1.62 mul V
  hpt 2 mul 0 V
  hpt neg vpt 1.62 mul V closepath stroke} def
/TriDE {stroke [] 0 setdash vpt 1.12 mul sub M
  hpt neg vpt 1.62 mul V
  hpt 2 mul 0 V
  hpt neg vpt -1.62 mul V closepath stroke} def
/PentE {stroke [] 0 setdash gsave
  translate 0 hpt M 4 {72 rotate 0 hpt L} repeat
  closepath stroke grestore} def
/CircE {stroke [] 0 setdash 
  hpt 0 360 arc stroke} def
/Opaque {gsave closepath 1 setgray fill grestore 0 setgray closepath} def
/DiaW {stroke [] 0 setdash vpt add M
  hpt neg vpt neg V hpt vpt neg V
  hpt vpt V hpt neg vpt V Opaque stroke} def
/BoxW {stroke [] 0 setdash exch hpt sub exch vpt add M
  0 vpt2 neg V hpt2 0 V 0 vpt2 V
  hpt2 neg 0 V Opaque stroke} def
/TriUW {stroke [] 0 setdash vpt 1.12 mul add M
  hpt neg vpt -1.62 mul V
  hpt 2 mul 0 V
  hpt neg vpt 1.62 mul V Opaque stroke} def
/TriDW {stroke [] 0 setdash vpt 1.12 mul sub M
  hpt neg vpt 1.62 mul V
  hpt 2 mul 0 V
  hpt neg vpt -1.62 mul V Opaque stroke} def
/PentW {stroke [] 0 setdash gsave
  translate 0 hpt M 4 {72 rotate 0 hpt L} repeat
  Opaque stroke grestore} def
/CircW {stroke [] 0 setdash 
  hpt 0 360 arc Opaque stroke} def
/BoxFill {gsave Rec 1 setgray fill grestore} def
/Density {
  /Fillden exch def
  currentrgbcolor
  /ColB exch def /ColG exch def /ColR exch def
  /ColR ColR Fillden mul Fillden sub 1 add def
  /ColG ColG Fillden mul Fillden sub 1 add def
  /ColB ColB Fillden mul Fillden sub 1 add def
  ColR ColG ColB setrgbcolor} def
/BoxColFill {gsave Rec PolyFill} def
/PolyFill {gsave Density fill grestore grestore} def
/h {rlineto rlineto rlineto gsave closepath fill grestore} bind def
%
%
/PatternFill {gsave /PFa [ 9 2 roll ] def
  PFa 0 get PFa 2 get 2 div add PFa 1 get PFa 3 get 2 div add translate
  PFa 2 get -2 div PFa 3 get -2 div PFa 2 get PFa 3 get Rec
  TransparentPatterns {} {gsave 1 setgray fill grestore} ifelse
  clip
  currentlinewidth 0.5 mul setlinewidth
  /PFs PFa 2 get dup mul PFa 3 get dup mul add sqrt def
  0 0 M PFa 5 get rotate PFs -2 div dup translate
  0 1 PFs PFa 4 get div 1 add floor cvi
	{PFa 4 get mul 0 M 0 PFs V} for
  0 PFa 6 get ne {
	0 1 PFs PFa 4 get div 1 add floor cvi
	{PFa 4 get mul 0 2 1 roll M PFs 0 V} for
 } if
  stroke grestore} def
/languagelevel where
 {pop languagelevel} {1} ifelse
 2 lt
	{/InterpretLevel1 true def}
	{/InterpretLevel1 Level1 def}
 ifelse
%
%
/Level2PatternFill {
/Tile8x8 {/PaintType 2 /PatternType 1 /TilingType 1 /BBox [0 0 8 8] /XStep 8 /YStep 8}
	bind def
/KeepColor {currentrgbcolor [/Pattern /DeviceRGB] setcolorspace} bind def
<< Tile8x8
 /PaintProc {0.5 setlinewidth pop 0 0 M 8 8 L 0 8 M 8 0 L stroke} 
>> matrix makepattern
/Pat1 exch def
<< Tile8x8
 /PaintProc {0.5 setlinewidth pop 0 0 M 8 8 L 0 8 M 8 0 L stroke
	0 4 M 4 8 L 8 4 L 4 0 L 0 4 L stroke}
>> matrix makepattern
/Pat2 exch def
<< Tile8x8
 /PaintProc {0.5 setlinewidth pop 0 0 M 0 8 L
	8 8 L 8 0 L 0 0 L fill}
>> matrix makepattern
/Pat3 exch def
<< Tile8x8
 /PaintProc {0.5 setlinewidth pop -4 8 M 8 -4 L
	0 12 M 12 0 L stroke}
>> matrix makepattern
/Pat4 exch def
<< Tile8x8
 /PaintProc {0.5 setlinewidth pop -4 0 M 8 12 L
	0 -4 M 12 8 L stroke}
>> matrix makepattern
/Pat5 exch def
<< Tile8x8
 /PaintProc {0.5 setlinewidth pop -2 8 M 4 -4 L
	0 12 M 8 -4 L 4 12 M 10 0 L stroke}
>> matrix makepattern
/Pat6 exch def
<< Tile8x8
 /PaintProc {0.5 setlinewidth pop -2 0 M 4 12 L
	0 -4 M 8 12 L 4 -4 M 10 8 L stroke}
>> matrix makepattern
/Pat7 exch def
<< Tile8x8
 /PaintProc {0.5 setlinewidth pop 8 -2 M -4 4 L
	12 0 M -4 8 L 12 4 M 0 10 L stroke}
>> matrix makepattern
/Pat8 exch def
<< Tile8x8
 /PaintProc {0.5 setlinewidth pop 0 -2 M 12 4 L
	-4 0 M 12 8 L -4 4 M 8 10 L stroke}
>> matrix makepattern
/Pat9 exch def
/Pattern1 {PatternBgnd KeepColor Pat1 setpattern} bind def
/Pattern2 {PatternBgnd KeepColor Pat2 setpattern} bind def
/Pattern3 {PatternBgnd KeepColor Pat3 setpattern} bind def
/Pattern4 {PatternBgnd KeepColor Landscape {Pat5} {Pat4} ifelse setpattern} bind def
/Pattern5 {PatternBgnd KeepColor Landscape {Pat4} {Pat5} ifelse setpattern} bind def
/Pattern6 {PatternBgnd KeepColor Landscape {Pat9} {Pat6} ifelse setpattern} bind def
/Pattern7 {PatternBgnd KeepColor Landscape {Pat8} {Pat7} ifelse setpattern} bind def
} def
%
%
%
/PatternBgnd {
  TransparentPatterns {} {gsave 1 setgray fill grestore} ifelse
} def
%
%
/Level1PatternFill {
/Pattern1 {0.250 Density} bind def
/Pattern2 {0.500 Density} bind def
/Pattern3 {0.750 Density} bind def
/Pattern4 {0.125 Density} bind def
/Pattern5 {0.375 Density} bind def
/Pattern6 {0.625 Density} bind def
/Pattern7 {0.875 Density} bind def
} def
%
%
Level1 {Level1PatternFill} {Level2PatternFill} ifelse
/Symbol-Oblique /Symbol findfont [1 0 .167 1 0 0] makefont
dup length dict begin {1 index /FID eq {pop pop} {def} ifelse} forall
currentdict end definefont pop
Level1 SuppressPDFMark or 
{} {
/SDict 10 dict def
systemdict /pdfmark known not {
  userdict /pdfmark systemdict /cleartomark get put
} if
SDict begin [
  /Title (thnorm.tex)
  /Subject (gnuplot plot)
  /Creator (gnuplot 4.6 patchlevel 4)
  /Author (swagat)
  /CreationDate (Wed Apr  8 12:33:03 2015)
  /DOCINFO pdfmark
end
} ifelse
end
gnudict begin
gsave
doclip
0 0 translate
0.050 0.050 scale
0 setgray
newpath
BackgroundColor 0 lt 3 1 roll 0 lt exch 0 lt or or not {BackgroundColor C 1.000 0 0 7200.00 5040.00 BoxColFill} if
1.000 UL
LTb
LCb setrgbcolor
860 640 M
63 0 V
5916 0 R
-63 0 V
stroke
LTb
LCb setrgbcolor
860 1056 M
63 0 V
5916 0 R
-63 0 V
stroke
LTb
LCb setrgbcolor
860 1472 M
63 0 V
5916 0 R
-63 0 V
stroke
LTb
LCb setrgbcolor
860 1888 M
63 0 V
5916 0 R
-63 0 V
stroke
LTb
LCb setrgbcolor
860 2304 M
63 0 V
5916 0 R
-63 0 V
stroke
LTb
LCb setrgbcolor
860 2720 M
63 0 V
5916 0 R
-63 0 V
stroke
LTb
LCb setrgbcolor
860 3135 M
63 0 V
5916 0 R
-63 0 V
stroke
LTb
LCb setrgbcolor
860 3551 M
63 0 V
5916 0 R
-63 0 V
stroke
LTb
LCb setrgbcolor
860 3967 M
63 0 V
5916 0 R
-63 0 V
stroke
LTb
LCb setrgbcolor
860 4383 M
63 0 V
5916 0 R
-63 0 V
stroke
LTb
LCb setrgbcolor
860 4799 M
63 0 V
5916 0 R
-63 0 V
stroke
LTb
LCb setrgbcolor
860 640 M
0 63 V
0 4096 R
0 -63 V
stroke
LTb
LCb setrgbcolor
1458 640 M
0 63 V
0 4096 R
0 -63 V
stroke
LTb
LCb setrgbcolor
2056 640 M
0 63 V
0 4096 R
0 -63 V
stroke
LTb
LCb setrgbcolor
2654 640 M
0 63 V
0 4096 R
0 -63 V
stroke
LTb
LCb setrgbcolor
3252 640 M
0 63 V
0 4096 R
0 -63 V
stroke
LTb
LCb setrgbcolor
3850 640 M
0 63 V
0 4096 R
0 -63 V
stroke
LTb
LCb setrgbcolor
4447 640 M
0 63 V
0 4096 R
0 -63 V
stroke
LTb
LCb setrgbcolor
5045 640 M
0 63 V
0 4096 R
0 -63 V
stroke
LTb
LCb setrgbcolor
5643 640 M
0 63 V
0 4096 R
0 -63 V
stroke
LTb
LCb setrgbcolor
6241 640 M
0 63 V
0 4096 R
0 -63 V
stroke
LTb
LCb setrgbcolor
6839 640 M
0 63 V
0 4096 R
0 -63 V
stroke
LTb
LCb setrgbcolor
1.000 UL
LTb
LCb setrgbcolor
860 4799 N
860 640 L
5979 0 V
0 4159 V
-5979 0 V
Z stroke
LCb setrgbcolor
LTb
LCb setrgbcolor
LTb
1.000 UP
1.000 UL
LTb
LCb setrgbcolor
2.000 UL
LT0
LC0 setrgbcolor
LCb setrgbcolor
LT0
6056 2819 M
543 0 V
860 646 M
3 2269 V
3 -979 V
3 -192 V
3 -48 V
3 -23 V
3 -18 V
3 -18 V
3 -19 V
3 -18 V
3 -19 V
3 -18 V
3 -18 V
3 -18 V
3 -17 V
3 -17 V
3 -16 V
3 -16 V
3 -15 V
3 -15 V
3 -14 V
3 -14 V
3 -13 V
3 -12 V
3 -12 V
3 -12 V
3 -11 V
3 -11 V
3 -10 V
3 -10 V
3 -9 V
3 -9 V
3 -9 V
3 -8 V
3 -8 V
3 -7 V
3 -7 V
3 -6 V
3 -7 V
3 -6 V
3 -5 V
3 -6 V
3 -5 V
3 -4 V
3 -5 V
3 -4 V
3 -4 V
3 -3 V
2 -4 V
3 -3 V
3 -3 V
3 -3 V
3 -2 V
3 -2 V
3 -2 V
3 -2 V
3 -2 V
3 -1 V
3 -2 V
3 -1 V
3 -1 V
3 -1 V
3 -1 V
3 0 V
3 -1 V
3 0 V
3 0 V
3 0 V
3 0 V
3 0 V
3 0 V
3 0 V
3 1 V
3 0 V
3 1 V
3 1 V
3 0 V
3 1 V
3 1 V
3 1 V
3 1 V
3 1 V
3 1 V
3 2 V
3 1 V
3 1 V
3 1 V
3 2 V
3 1 V
3 2 V
3 1 V
3 2 V
3 2 V
3 1 V
3 2 V
3 2 V
3 2 V
3 1 V
3 2 V
3 2 V
3 2 V
3 2 V
3 2 V
stroke 1165 1211 M
3 2 V
3 2 V
3 2 V
3 2 V
3 2 V
3 2 V
3 2 V
3 2 V
3 2 V
3 2 V
3 3 V
3 2 V
3 2 V
3 2 V
3 2 V
3 3 V
3 2 V
3 2 V
3 2 V
3 3 V
3 2 V
3 2 V
3 2 V
3 3 V
3 2 V
3 2 V
3 3 V
3 2 V
3 2 V
3 3 V
3 2 V
3 2 V
3 3 V
3 2 V
3 3 V
3 2 V
3 2 V
3 3 V
3 2 V
3 2 V
2 3 V
3 2 V
3 3 V
3 2 V
3 2 V
3 3 V
3 2 V
3 2 V
3 3 V
3 2 V
3 3 V
3 2 V
3 2 V
3 3 V
3 2 V
3 2 V
3 3 V
3 2 V
3 2 V
3 3 V
3 2 V
3 3 V
3 2 V
3 2 V
3 3 V
3 2 V
3 2 V
3 3 V
3 2 V
3 2 V
3 2 V
3 3 V
3 2 V
3 2 V
3 3 V
3 2 V
3 2 V
3 2 V
3 3 V
3 2 V
3 2 V
3 2 V
3 3 V
3 2 V
3 2 V
3 2 V
3 2 V
3 3 V
3 2 V
3 2 V
3 2 V
3 2 V
3 2 V
3 3 V
3 2 V
3 2 V
3 2 V
3 2 V
3 2 V
3 2 V
3 2 V
3 2 V
3 2 V
3 2 V
stroke 1476 1445 M
3 2 V
3 2 V
3 2 V
3 2 V
3 2 V
3 2 V
3 2 V
3 2 V
3 2 V
3 2 V
3 2 V
3 2 V
3 2 V
3 2 V
3 2 V
3 1 V
3 2 V
3 2 V
3 2 V
3 2 V
3 2 V
3 1 V
3 2 V
3 2 V
3 1 V
3 2 V
3 2 V
3 2 V
3 1 V
3 2 V
3 1 V
3 2 V
2 2 V
3 1 V
3 2 V
3 1 V
3 2 V
3 1 V
3 2 V
3 1 V
3 2 V
3 1 V
3 2 V
3 1 V
3 1 V
3 2 V
3 1 V
3 1 V
3 2 V
3 1 V
3 1 V
3 1 V
3 2 V
3 1 V
3 1 V
3 1 V
3 1 V
3 1 V
3 2 V
3 1 V
3 1 V
3 1 V
3 1 V
3 1 V
3 1 V
3 1 V
3 0 V
3 1 V
3 1 V
3 1 V
3 1 V
3 1 V
3 1 V
3 0 V
3 1 V
3 1 V
3 0 V
3 1 V
3 1 V
3 0 V
3 1 V
3 1 V
3 0 V
3 1 V
3 0 V
3 1 V
3 0 V
3 1 V
3 0 V
3 1 V
3 0 V
3 0 V
3 1 V
3 0 V
3 0 V
3 0 V
3 1 V
3 0 V
3 0 V
3 0 V
3 0 V
3 1 V
3 0 V
3 0 V
stroke 1787 1567 M
3 0 V
3 0 V
3 0 V
3 0 V
3 0 V
3 0 V
3 0 V
3 0 V
3 -1 V
3 0 V
3 0 V
3 0 V
3 0 V
3 -1 V
3 0 V
3 0 V
3 -1 V
3 0 V
3 0 V
3 -1 V
3 0 V
3 -1 V
3 0 V
2 -1 V
3 0 V
3 -1 V
3 0 V
3 -1 V
3 -1 V
3 0 V
3 -1 V
3 -1 V
3 0 V
3 -1 V
3 -1 V
3 0 V
3 -1 V
3 -1 V
3 -1 V
3 -1 V
3 -1 V
3 -1 V
3 0 V
3 -1 V
3 -1 V
3 -1 V
3 -1 V
3 -1 V
3 -1 V
3 -2 V
3 -1 V
3 -1 V
3 -1 V
3 -1 V
3 -1 V
3 -1 V
3 -2 V
3 -1 V
3 -1 V
3 -1 V
3 -2 V
3 -1 V
3 -1 V
3 -1 V
3 -2 V
3 -1 V
3 -2 V
3 -1 V
3 -1 V
3 -2 V
3 -1 V
3 -2 V
3 -1 V
3 -2 V
3 -1 V
3 -2 V
3 -2 V
3 -1 V
3 -2 V
3 -1 V
3 -2 V
3 -2 V
3 -1 V
3 -2 V
3 -2 V
3 -1 V
3 -2 V
3 -2 V
3 -2 V
3 -1 V
3 -2 V
3 -2 V
3 -2 V
3 -2 V
3 -2 V
3 -1 V
3 -2 V
3 -2 V
3 -2 V
3 -2 V
3 -2 V
3 -2 V
3 -2 V
3 -2 V
stroke 2098 1456 M
3 -2 V
3 -2 V
3 -2 V
3 -2 V
3 -2 V
3 -2 V
3 -2 V
3 -2 V
3 -2 V
3 -2 V
3 -2 V
3 -2 V
3 -2 V
3 -2 V
2 -2 V
3 -3 V
3 -2 V
3 -2 V
3 -2 V
3 -2 V
3 -2 V
3 -2 V
3 -3 V
3 -2 V
3 -2 V
3 -2 V
3 -2 V
3 -3 V
3 -2 V
3 -2 V
3 -2 V
3 -3 V
3 -2 V
3 -2 V
3 -2 V
3 -3 V
3 -2 V
3 -2 V
3 -2 V
3 -3 V
3 -2 V
3 -2 V
3 -3 V
3 -2 V
3 -2 V
3 -2 V
3 -3 V
3 -2 V
3 -2 V
3 -3 V
3 -2 V
3 -2 V
3 -3 V
3 -2 V
3 -2 V
3 -3 V
3 -2 V
3 -2 V
3 -3 V
3 -2 V
3 -2 V
3 -3 V
3 -2 V
3 -3 V
3 -2 V
3 -2 V
3 -3 V
3 -2 V
3 -2 V
3 -3 V
3 -2 V
3 -2 V
3 -3 V
3 -2 V
3 -3 V
3 -2 V
3 -2 V
3 -3 V
3 -2 V
3 -2 V
3 -3 V
3 -2 V
3 -2 V
3 -3 V
3 -2 V
3 -2 V
3 -3 V
3 -2 V
3 -2 V
3 -3 V
3 -2 V
3 -2 V
3 -3 V
3 -2 V
3 -2 V
3 -3 V
3 -2 V
3 -2 V
3 -3 V
3 -2 V
3 -2 V
3 -2 V
3 -3 V
3 -2 V
stroke 2409 1221 M
3 -2 V
3 -2 V
3 -3 V
3 -2 V
3 -2 V
2 -2 V
3 -2 V
3 -3 V
3 -2 V
3 -2 V
3 -2 V
3 -2 V
3 -2 V
3 -3 V
3 -2 V
3 -2 V
3 -2 V
3 -2 V
3 -2 V
3 -2 V
3 -2 V
3 -2 V
3 -2 V
3 -2 V
3 -2 V
3 -2 V
3 -2 V
3 -2 V
3 -2 V
3 -2 V
3 -2 V
3 -2 V
3 -2 V
3 -2 V
3 -2 V
3 -1 V
3 -2 V
3 -2 V
3 -2 V
3 -2 V
3 -1 V
3 -2 V
3 -2 V
3 -1 V
3 -2 V
3 -2 V
3 -1 V
3 -2 V
3 -2 V
3 -1 V
3 -2 V
3 -1 V
3 -2 V
3 -1 V
3 -2 V
3 -1 V
3 -2 V
3 -1 V
3 -2 V
3 -1 V
3 -1 V
3 -2 V
3 -1 V
3 -1 V
3 -2 V
3 -1 V
3 -1 V
3 -1 V
3 -1 V
3 -1 V
3 -2 V
3 -1 V
3 -1 V
3 -1 V
3 -1 V
3 -1 V
3 -1 V
3 -1 V
3 -1 V
3 0 V
3 -1 V
3 -1 V
3 -1 V
3 -1 V
3 0 V
3 -1 V
3 -1 V
3 0 V
3 -1 V
3 -1 V
3 0 V
3 -1 V
3 0 V
3 -1 V
3 0 V
3 -1 V
3 0 V
3 0 V
3 -1 V
3 0 V
3 0 V
2 -1 V
3 0 V
3 0 V
stroke 2719 1073 M
3 0 V
3 0 V
3 0 V
3 0 V
3 0 V
3 0 V
3 0 V
3 0 V
3 0 V
3 0 V
3 0 V
3 0 V
3 0 V
3 1 V
3 0 V
3 0 V
3 1 V
3 0 V
3 0 V
3 1 V
3 0 V
3 1 V
3 0 V
3 1 V
3 0 V
3 1 V
3 1 V
3 0 V
3 1 V
3 1 V
3 0 V
3 1 V
3 1 V
3 1 V
3 1 V
3 0 V
3 1 V
3 1 V
3 1 V
3 1 V
3 1 V
3 1 V
3 1 V
3 1 V
3 1 V
3 2 V
3 1 V
3 1 V
3 1 V
3 1 V
3 2 V
3 1 V
3 1 V
3 2 V
3 1 V
3 1 V
3 2 V
3 1 V
3 2 V
3 1 V
3 1 V
3 2 V
3 1 V
3 2 V
3 2 V
3 1 V
3 2 V
3 1 V
3 2 V
3 2 V
3 1 V
3 2 V
3 2 V
3 2 V
3 1 V
3 2 V
3 2 V
3 2 V
3 2 V
3 1 V
3 2 V
3 2 V
3 2 V
3 2 V
3 2 V
3 2 V
3 2 V
3 2 V
3 2 V
3 2 V
3 2 V
3 2 V
2 2 V
3 2 V
3 2 V
3 2 V
3 2 V
3 2 V
3 2 V
3 2 V
3 2 V
3 3 V
3 2 V
3 2 V
stroke 3030 1197 M
3 2 V
3 2 V
3 2 V
3 3 V
3 2 V
3 2 V
3 2 V
3 3 V
3 2 V
3 2 V
3 2 V
3 3 V
3 2 V
3 2 V
3 3 V
3 2 V
3 2 V
3 2 V
3 3 V
3 2 V
3 2 V
3 3 V
3 2 V
3 2 V
3 3 V
3 2 V
3 3 V
3 2 V
3 2 V
3 3 V
3 2 V
3 2 V
3 3 V
3 2 V
3 3 V
3 2 V
3 2 V
3 3 V
3 2 V
3 3 V
3 2 V
3 2 V
3 3 V
3 2 V
3 3 V
3 2 V
3 2 V
3 3 V
3 2 V
3 3 V
3 2 V
3 2 V
3 3 V
3 2 V
3 3 V
3 2 V
3 2 V
3 3 V
3 2 V
3 2 V
3 3 V
3 2 V
3 3 V
3 2 V
3 2 V
3 3 V
3 2 V
3 2 V
3 3 V
3 2 V
3 2 V
3 3 V
3 2 V
3 2 V
3 3 V
3 2 V
3 2 V
3 3 V
3 2 V
3 2 V
3 2 V
3 3 V
3 2 V
2 2 V
3 2 V
3 3 V
3 2 V
3 2 V
3 2 V
3 2 V
3 3 V
3 2 V
3 2 V
3 2 V
3 2 V
3 3 V
3 2 V
3 2 V
3 2 V
3 2 V
3 2 V
3 2 V
3 2 V
3 2 V
stroke 3341 1436 M
3 3 V
3 2 V
3 2 V
3 2 V
3 2 V
3 2 V
3 2 V
3 2 V
3 2 V
3 2 V
3 2 V
3 2 V
3 2 V
3 2 V
3 1 V
3 2 V
3 2 V
3 2 V
3 2 V
3 2 V
3 2 V
3 2 V
3 1 V
3 2 V
3 2 V
3 2 V
3 1 V
3 2 V
3 2 V
3 2 V
3 1 V
3 2 V
3 2 V
3 1 V
3 2 V
3 2 V
3 1 V
3 2 V
3 1 V
3 2 V
3 1 V
3 2 V
3 1 V
3 2 V
3 1 V
3 2 V
3 1 V
3 2 V
3 1 V
3 1 V
3 2 V
3 1 V
3 1 V
3 2 V
3 1 V
3 1 V
3 1 V
3 1 V
3 2 V
3 1 V
3 1 V
3 1 V
3 1 V
3 1 V
3 1 V
3 1 V
3 1 V
3 1 V
3 1 V
3 1 V
3 1 V
3 1 V
3 1 V
3 1 V
2 1 V
3 0 V
3 1 V
3 1 V
3 1 V
3 1 V
3 0 V
3 1 V
3 1 V
3 0 V
3 1 V
3 0 V
3 1 V
3 1 V
3 0 V
3 1 V
3 0 V
3 1 V
3 0 V
3 0 V
3 1 V
3 0 V
3 0 V
3 1 V
3 0 V
3 0 V
3 1 V
3 0 V
3 0 V
3 0 V
stroke 3652 1566 M
3 0 V
3 1 V
3 0 V
3 0 V
3 0 V
3 0 V
3 0 V
3 0 V
3 0 V
3 0 V
3 0 V
3 0 V
3 -1 V
3 0 V
3 0 V
3 0 V
3 0 V
3 -1 V
3 0 V
3 0 V
3 -1 V
3 0 V
3 0 V
3 -1 V
3 0 V
3 -1 V
3 0 V
3 -1 V
3 0 V
3 -1 V
3 0 V
3 -1 V
3 0 V
3 -1 V
3 -1 V
3 0 V
3 -1 V
3 -1 V
3 0 V
3 -1 V
3 -1 V
3 -1 V
3 -1 V
3 -1 V
3 0 V
3 -1 V
3 -1 V
3 -1 V
3 -1 V
3 -1 V
3 -1 V
3 -1 V
3 -1 V
3 -1 V
3 -1 V
3 -1 V
3 -1 V
3 -2 V
3 -1 V
3 -1 V
3 -1 V
3 -1 V
3 -2 V
3 -1 V
3 -1 V
3 -1 V
2 -2 V
3 -1 V
3 -1 V
3 -2 V
3 -1 V
3 -2 V
3 -1 V
3 -1 V
3 -2 V
3 -1 V
3 -2 V
3 -1 V
3 -2 V
3 -1 V
3 -2 V
3 -2 V
3 -1 V
3 -2 V
3 -1 V
3 -2 V
3 -2 V
3 -1 V
3 -2 V
3 -2 V
3 -2 V
3 -1 V
3 -2 V
3 -2 V
3 -2 V
3 -1 V
3 -2 V
3 -2 V
3 -2 V
3 -2 V
3 -2 V
3 -2 V
3 -1 V
3 -2 V
stroke 3963 1465 M
3 -2 V
3 -2 V
3 -2 V
3 -2 V
3 -2 V
3 -2 V
3 -2 V
3 -2 V
3 -2 V
3 -2 V
3 -2 V
3 -2 V
3 -2 V
3 -2 V
3 -2 V
3 -2 V
3 -3 V
3 -2 V
3 -2 V
3 -2 V
3 -2 V
3 -2 V
3 -2 V
3 -2 V
3 -3 V
3 -2 V
3 -2 V
3 -2 V
3 -2 V
3 -2 V
3 -3 V
3 -2 V
3 -2 V
3 -2 V
3 -3 V
3 -2 V
3 -2 V
3 -2 V
3 -3 V
3 -2 V
3 -2 V
3 -2 V
3 -3 V
3 -2 V
3 -2 V
3 -2 V
3 -3 V
3 -2 V
3 -2 V
3 -3 V
3 -2 V
3 -2 V
3 -3 V
3 -2 V
3 -2 V
3 -3 V
3 -2 V
2 -2 V
3 -3 V
3 -2 V
3 -2 V
3 -3 V
3 -2 V
3 -2 V
3 -3 V
3 -2 V
3 -2 V
3 -3 V
3 -2 V
3 -3 V
3 -2 V
3 -2 V
3 -3 V
3 -2 V
3 -2 V
3 -3 V
3 -2 V
3 -2 V
3 -3 V
3 -2 V
3 -2 V
3 -3 V
3 -2 V
3 -3 V
3 -2 V
3 -2 V
3 -3 V
3 -2 V
3 -2 V
3 -3 V
3 -2 V
3 -2 V
3 -3 V
3 -2 V
3 -2 V
3 -3 V
3 -2 V
3 -2 V
3 -3 V
3 -2 V
3 -2 V
3 -3 V
3 -2 V
3 -2 V
stroke 4274 1231 M
3 -2 V
3 -3 V
3 -2 V
3 -2 V
3 -2 V
3 -3 V
3 -2 V
3 -2 V
3 -2 V
3 -3 V
3 -2 V
3 -2 V
3 -2 V
3 -2 V
3 -2 V
3 -3 V
3 -2 V
3 -2 V
3 -2 V
3 -2 V
3 -2 V
3 -2 V
3 -2 V
3 -3 V
3 -2 V
3 -2 V
3 -2 V
3 -2 V
3 -2 V
3 -2 V
3 -2 V
3 -2 V
3 -2 V
3 -1 V
3 -2 V
3 -2 V
3 -2 V
3 -2 V
3 -2 V
3 -2 V
3 -2 V
3 -1 V
3 -2 V
3 -2 V
3 -2 V
3 -2 V
3 -1 V
3 -2 V
2 -2 V
3 -1 V
3 -2 V
3 -2 V
3 -1 V
3 -2 V
3 -1 V
3 -2 V
3 -1 V
3 -2 V
3 -1 V
3 -2 V
3 -1 V
3 -2 V
3 -1 V
3 -2 V
3 -1 V
3 -1 V
3 -2 V
3 -1 V
3 -1 V
3 -1 V
3 -2 V
3 -1 V
3 -1 V
3 -1 V
3 -1 V
3 -1 V
3 -1 V
3 -1 V
3 -1 V
3 -1 V
3 -1 V
3 -1 V
3 -1 V
3 -1 V
3 -1 V
3 -1 V
3 -1 V
3 0 V
3 -1 V
3 -1 V
3 -1 V
3 0 V
3 -1 V
3 0 V
3 -1 V
3 -1 V
3 0 V
3 -1 V
3 0 V
3 0 V
3 -1 V
3 0 V
3 -1 V
3 0 V
stroke 4585 1074 M
3 0 V
3 0 V
3 -1 V
3 0 V
3 0 V
3 0 V
3 0 V
3 0 V
3 0 V
3 0 V
3 0 V
3 0 V
3 0 V
3 0 V
3 0 V
3 0 V
3 0 V
3 1 V
3 0 V
3 0 V
3 0 V
3 1 V
3 0 V
3 1 V
3 0 V
3 1 V
3 0 V
3 1 V
3 0 V
3 1 V
3 0 V
3 1 V
3 1 V
3 0 V
3 1 V
3 1 V
3 1 V
3 0 V
3 1 V
2 1 V
3 1 V
3 1 V
3 1 V
3 1 V
3 1 V
3 1 V
3 1 V
3 1 V
3 1 V
3 1 V
3 1 V
3 2 V
3 1 V
3 1 V
3 1 V
3 1 V
3 2 V
3 1 V
3 1 V
3 2 V
3 1 V
3 2 V
3 1 V
3 2 V
3 1 V
3 1 V
3 2 V
3 2 V
3 1 V
3 2 V
3 1 V
3 2 V
3 2 V
3 1 V
3 2 V
3 2 V
3 1 V
3 2 V
3 2 V
3 2 V
3 1 V
3 2 V
3 2 V
3 2 V
3 2 V
3 2 V
3 1 V
3 2 V
3 2 V
3 2 V
3 2 V
3 2 V
3 2 V
3 2 V
3 2 V
3 2 V
3 2 V
3 2 V
3 2 V
3 2 V
3 2 V
3 2 V
3 3 V
3 2 V
stroke 4896 1188 M
3 2 V
3 2 V
3 2 V
3 2 V
3 2 V
3 3 V
3 2 V
3 2 V
3 2 V
3 2 V
3 3 V
3 2 V
3 2 V
3 2 V
3 3 V
3 2 V
3 2 V
3 3 V
3 2 V
3 2 V
3 2 V
3 3 V
3 2 V
3 2 V
3 3 V
3 2 V
3 2 V
3 3 V
3 2 V
3 2 V
2 3 V
3 2 V
3 3 V
3 2 V
3 2 V
3 3 V
3 2 V
3 2 V
3 3 V
3 2 V
3 3 V
3 2 V
3 2 V
3 3 V
3 2 V
3 3 V
3 2 V
3 2 V
3 3 V
3 2 V
3 3 V
3 2 V
3 2 V
3 3 V
3 2 V
3 3 V
3 2 V
3 2 V
3 3 V
3 2 V
3 3 V
3 2 V
3 2 V
3 3 V
3 2 V
3 2 V
3 3 V
3 2 V
3 3 V
3 2 V
3 2 V
3 3 V
3 2 V
3 2 V
3 3 V
3 2 V
3 2 V
3 3 V
3 2 V
3 2 V
3 2 V
3 3 V
3 2 V
3 2 V
3 3 V
3 2 V
3 2 V
3 2 V
3 3 V
3 2 V
3 2 V
3 2 V
3 3 V
3 2 V
3 2 V
3 2 V
3 2 V
3 3 V
3 2 V
3 2 V
3 2 V
3 2 V
3 2 V
3 2 V
stroke 5207 1427 M
3 3 V
3 2 V
3 2 V
3 2 V
3 2 V
3 2 V
3 2 V
3 2 V
3 2 V
3 2 V
3 2 V
3 2 V
3 2 V
3 2 V
3 2 V
3 2 V
3 2 V
3 2 V
3 2 V
3 2 V
3 2 V
3 2 V
2 1 V
3 2 V
3 2 V
3 2 V
3 2 V
3 2 V
3 1 V
3 2 V
3 2 V
3 2 V
3 1 V
3 2 V
3 2 V
3 1 V
3 2 V
3 2 V
3 1 V
3 2 V
3 2 V
3 1 V
3 2 V
3 1 V
3 2 V
3 1 V
3 2 V
3 1 V
3 2 V
3 1 V
3 2 V
3 1 V
3 2 V
3 1 V
3 1 V
3 2 V
3 1 V
3 1 V
3 1 V
3 2 V
3 1 V
3 1 V
3 1 V
3 1 V
3 2 V
3 1 V
3 1 V
3 1 V
3 1 V
3 1 V
3 1 V
3 1 V
3 1 V
3 1 V
3 1 V
3 1 V
3 1 V
3 1 V
3 0 V
3 1 V
3 1 V
3 1 V
3 1 V
3 0 V
3 1 V
3 1 V
3 0 V
3 1 V
3 1 V
3 0 V
3 1 V
3 0 V
3 1 V
3 0 V
3 1 V
3 0 V
3 1 V
3 0 V
3 1 V
3 0 V
3 0 V
3 1 V
3 0 V
3 0 V
stroke 5518 1565 M
3 1 V
3 0 V
3 0 V
3 0 V
3 0 V
3 1 V
3 0 V
3 0 V
3 0 V
3 0 V
3 0 V
3 0 V
3 0 V
2 0 V
3 0 V
3 0 V
3 -1 V
3 0 V
3 0 V
3 0 V
3 0 V
3 -1 V
3 0 V
3 0 V
3 0 V
3 -1 V
3 0 V
3 -1 V
3 0 V
3 0 V
3 -1 V
3 0 V
3 -1 V
3 0 V
3 -1 V
3 -1 V
3 0 V
3 -1 V
3 0 V
3 -1 V
3 -1 V
3 0 V
3 -1 V
3 -1 V
3 -1 V
3 -1 V
3 0 V
3 -1 V
3 -1 V
3 -1 V
3 -1 V
3 -1 V
3 -1 V
3 -1 V
3 -1 V
3 -1 V
3 -1 V
3 -1 V
3 -1 V
3 -1 V
3 -1 V
3 -1 V
3 -1 V
3 -2 V
3 -1 V
3 -1 V
3 -1 V
3 -1 V
3 -2 V
3 -1 V
3 -1 V
3 -2 V
3 -1 V
3 -1 V
3 -2 V
3 -1 V
3 -2 V
3 -1 V
3 -1 V
3 -2 V
3 -1 V
3 -2 V
3 -1 V
3 -2 V
3 -2 V
3 -1 V
3 -2 V
3 -1 V
3 -2 V
3 -2 V
3 -1 V
3 -2 V
3 -2 V
3 -1 V
3 -2 V
3 -2 V
3 -2 V
3 -1 V
3 -2 V
3 -2 V
3 -2 V
3 -2 V
3 -1 V
3 -2 V
stroke 5829 1473 M
3 -2 V
3 -2 V
3 -2 V
3 -2 V
2 -2 V
3 -2 V
3 -2 V
3 -2 V
3 -1 V
3 -2 V
3 -2 V
3 -2 V
3 -2 V
3 -2 V
3 -2 V
3 -3 V
3 -2 V
3 -2 V
3 -2 V
3 -2 V
3 -2 V
3 -2 V
3 -2 V
3 -2 V
3 -2 V
3 -2 V
3 -3 V
3 -2 V
3 -2 V
3 -2 V
3 -2 V
3 -2 V
3 -3 V
3 -2 V
3 -2 V
3 -2 V
3 -2 V
3 -3 V
3 -2 V
3 -2 V
3 -2 V
3 -3 V
3 -2 V
3 -2 V
3 -2 V
3 -3 V
3 -2 V
3 -2 V
3 -2 V
3 -3 V
3 -2 V
3 -2 V
3 -3 V
3 -2 V
3 -2 V
3 -3 V
3 -2 V
3 -2 V
3 -3 V
3 -2 V
3 -2 V
3 -3 V
3 -2 V
3 -2 V
3 -3 V
3 -2 V
3 -2 V
3 -3 V
3 -2 V
3 -2 V
3 -3 V
3 -2 V
3 -2 V
3 -3 V
3 -2 V
3 -2 V
3 -3 V
3 -2 V
3 -3 V
3 -2 V
3 -2 V
3 -3 V
3 -2 V
3 -2 V
3 -3 V
3 -2 V
3 -2 V
3 -3 V
3 -2 V
3 -3 V
3 -2 V
3 -2 V
3 -3 V
3 -2 V
3 -2 V
3 -3 V
3 -2 V
3 -2 V
3 -3 V
2 -2 V
3 -2 V
3 -3 V
3 -2 V
3 -2 V
stroke 6139 1241 M
3 -2 V
3 -3 V
3 -2 V
3 -2 V
3 -3 V
3 -2 V
3 -2 V
3 -2 V
3 -3 V
3 -2 V
3 -2 V
3 -2 V
3 -3 V
3 -2 V
3 -2 V
3 -2 V
3 -2 V
3 -3 V
3 -2 V
3 -2 V
3 -2 V
3 -2 V
3 -2 V
3 -2 V
3 -3 V
3 -2 V
3 -2 V
3 -2 V
3 -2 V
3 -2 V
3 -2 V
3 -2 V
3 -2 V
3 -2 V
3 -2 V
3 -2 V
3 -2 V
3 -2 V
3 -2 V
3 -2 V
3 -2 V
3 -1 V
3 -2 V
3 -2 V
3 -2 V
3 -2 V
3 -2 V
3 -1 V
3 -2 V
3 -2 V
3 -2 V
3 -1 V
3 -2 V
3 -2 V
3 -1 V
3 -2 V
3 -2 V
3 -1 V
3 -2 V
3 -1 V
3 -2 V
3 -1 V
3 -2 V
3 -1 V
3 -2 V
3 -1 V
3 -2 V
3 -1 V
3 -1 V
3 -2 V
3 -1 V
3 -1 V
3 -2 V
3 -1 V
3 -1 V
3 -1 V
3 -2 V
3 -1 V
3 -1 V
3 -1 V
3 -1 V
3 -1 V
3 -1 V
3 -1 V
3 -1 V
3 -1 V
3 -1 V
3 -1 V
3 -1 V
3 -1 V
3 0 V
2 -1 V
3 -1 V
3 -1 V
3 0 V
3 -1 V
3 -1 V
3 0 V
3 -1 V
3 0 V
3 -1 V
3 0 V
3 -1 V
3 0 V
stroke 6450 1076 M
3 -1 V
3 0 V
3 0 V
3 -1 V
3 0 V
3 0 V
3 -1 V
3 0 V
3 0 V
3 0 V
3 0 V
3 0 V
3 0 V
3 0 V
3 0 V
3 0 V
3 0 V
3 0 V
3 0 V
3 0 V
3 0 V
3 1 V
3 0 V
3 0 V
3 0 V
3 1 V
3 0 V
3 1 V
3 0 V
3 0 V
3 1 V
3 0 V
3 1 V
3 1 V
3 0 V
3 1 V
3 0 V
3 1 V
3 1 V
3 1 V
3 0 V
3 1 V
3 1 V
3 1 V
3 1 V
3 1 V
3 1 V
3 1 V
3 1 V
3 1 V
3 1 V
3 1 V
3 1 V
3 1 V
3 1 V
3 1 V
3 1 V
3 2 V
3 1 V
3 1 V
3 1 V
3 2 V
3 1 V
3 1 V
3 2 V
3 1 V
3 2 V
3 1 V
3 2 V
3 1 V
3 2 V
3 1 V
3 2 V
3 1 V
3 2 V
3 1 V
3 2 V
3 2 V
3 1 V
3 2 V
3 2 V
3 1 V
2 2 V
3 2 V
3 2 V
3 2 V
3 1 V
3 2 V
3 2 V
3 2 V
3 2 V
3 2 V
3 2 V
3 2 V
3 1 V
3 2 V
3 2 V
3 2 V
3 2 V
3 2 V
3 2 V
3 2 V
3 3 V
3 2 V
stroke 6761 1179 M
3 2 V
3 2 V
3 2 V
3 2 V
3 2 V
3 2 V
3 2 V
3 3 V
3 2 V
3 2 V
3 2 V
3 2 V
3 2 V
3 3 V
3 2 V
3 2 V
3 2 V
3 3 V
3 2 V
3 2 V
3 2 V
3 3 V
3 2 V
3 2 V
3 3 V
stroke
LT1
LC1 setrgbcolor
LCb setrgbcolor
LT1
6056 2619 M
543 0 V
860 646 M
3 2269 V
3 -55 V
3 10 V
3 -6 V
3 -55 V
3 -43 V
3 -29 V
3 -14 V
3 4 V
3 23 V
3 41 V
3 54 V
3 59 V
3 62 V
3 61 V
3 61 V
3 60 V
3 59 V
3 60 V
3 59 V
3 58 V
3 57 V
3 54 V
3 47 V
3 41 V
3 33 V
3 29 V
3 25 V
3 21 V
3 19 V
3 17 V
3 15 V
3 14 V
3 13 V
3 12 V
3 11 V
3 11 V
3 9 V
3 9 V
3 9 V
3 8 V
3 7 V
3 7 V
3 7 V
3 6 V
3 6 V
3 5 V
2 5 V
3 5 V
3 5 V
3 4 V
3 4 V
3 4 V
3 3 V
3 3 V
3 3 V
3 3 V
3 3 V
3 2 V
3 2 V
3 2 V
3 2 V
3 2 V
3 2 V
3 1 V
3 1 V
3 2 V
3 1 V
3 1 V
3 0 V
3 1 V
3 1 V
3 0 V
3 1 V
3 0 V
3 0 V
3 1 V
3 0 V
3 0 V
3 0 V
3 0 V
3 0 V
3 0 V
3 -1 V
3 0 V
3 0 V
3 -1 V
3 0 V
3 0 V
3 -1 V
3 0 V
3 -1 V
3 -1 V
3 0 V
3 -1 V
3 0 V
3 -1 V
3 -1 V
3 -1 V
3 0 V
3 -1 V
3 -1 V
stroke 1165 3929 M
3 -1 V
3 0 V
3 -1 V
3 -1 V
3 -1 V
3 -1 V
3 -1 V
3 -1 V
3 0 V
3 -1 V
3 -1 V
3 -1 V
3 -1 V
3 -1 V
3 -1 V
3 -1 V
3 -1 V
3 -1 V
3 -1 V
3 0 V
3 -1 V
3 -1 V
3 -1 V
3 -1 V
3 -1 V
3 -1 V
3 -1 V
3 -1 V
3 -1 V
3 -1 V
3 0 V
3 -1 V
3 -1 V
3 -1 V
3 -1 V
3 -1 V
3 -1 V
3 0 V
3 -1 V
3 -1 V
2 -1 V
3 -1 V
3 0 V
3 -1 V
3 -1 V
3 -1 V
3 -1 V
3 0 V
3 -1 V
3 -1 V
3 0 V
3 -1 V
3 -1 V
3 0 V
3 -1 V
3 -1 V
3 0 V
3 -1 V
3 0 V
3 -1 V
3 -1 V
3 0 V
3 -1 V
3 0 V
3 -1 V
3 0 V
3 -1 V
3 0 V
3 0 V
3 -1 V
3 0 V
3 -1 V
3 0 V
3 0 V
3 -1 V
3 0 V
3 0 V
3 0 V
3 -1 V
3 0 V
3 0 V
3 0 V
3 0 V
3 0 V
3 -1 V
3 0 V
3 0 V
3 0 V
3 0 V
3 0 V
3 0 V
3 0 V
3 0 V
3 0 V
3 1 V
3 0 V
3 0 V
3 0 V
3 0 V
3 0 V
3 1 V
3 0 V
3 0 V
3 0 V
stroke 1476 3873 M
3 1 V
3 0 V
3 0 V
3 1 V
3 0 V
3 1 V
3 0 V
3 1 V
3 0 V
3 1 V
3 0 V
3 1 V
3 0 V
3 1 V
3 0 V
3 1 V
3 1 V
3 0 V
3 1 V
3 1 V
3 0 V
3 1 V
3 1 V
3 1 V
3 0 V
3 1 V
3 1 V
3 1 V
3 0 V
3 1 V
3 1 V
3 1 V
2 1 V
3 1 V
3 1 V
3 0 V
3 1 V
3 1 V
3 1 V
3 1 V
3 1 V
3 1 V
3 1 V
3 1 V
3 1 V
3 1 V
3 1 V
3 1 V
3 1 V
3 0 V
3 1 V
3 1 V
3 1 V
3 1 V
3 1 V
3 1 V
3 1 V
3 1 V
3 1 V
3 1 V
3 1 V
3 1 V
3 1 V
3 1 V
3 1 V
3 1 V
3 0 V
3 1 V
3 1 V
3 1 V
3 1 V
3 1 V
3 1 V
3 1 V
3 0 V
3 1 V
3 1 V
3 1 V
3 1 V
3 0 V
3 1 V
3 1 V
3 1 V
3 0 V
3 1 V
3 1 V
3 0 V
3 1 V
3 1 V
3 0 V
3 1 V
3 1 V
3 0 V
3 1 V
3 0 V
3 1 V
3 0 V
3 1 V
3 0 V
3 1 V
3 0 V
3 1 V
3 0 V
3 1 V
stroke 1787 3951 M
3 0 V
3 0 V
3 1 V
3 0 V
3 0 V
3 1 V
3 0 V
3 0 V
3 0 V
3 0 V
3 1 V
3 0 V
3 0 V
3 0 V
3 0 V
3 0 V
3 0 V
3 0 V
3 0 V
3 0 V
3 0 V
3 0 V
3 0 V
2 0 V
3 0 V
3 0 V
3 0 V
3 0 V
3 0 V
3 -1 V
3 0 V
3 0 V
3 0 V
3 0 V
3 -1 V
3 0 V
3 0 V
3 -1 V
3 0 V
3 0 V
3 -1 V
3 0 V
3 0 V
3 -1 V
3 0 V
3 -1 V
3 0 V
3 -1 V
3 0 V
3 0 V
3 -1 V
3 0 V
3 -1 V
3 0 V
3 -1 V
3 0 V
3 -1 V
3 -1 V
3 0 V
3 -1 V
3 0 V
3 -1 V
3 0 V
3 -1 V
3 -1 V
3 0 V
3 -1 V
3 0 V
3 -1 V
3 0 V
3 -1 V
3 -1 V
3 0 V
3 -1 V
3 -1 V
3 0 V
3 -1 V
3 0 V
3 -1 V
3 -1 V
3 0 V
3 -1 V
3 0 V
3 -1 V
3 0 V
3 -1 V
3 -1 V
3 0 V
3 -1 V
3 0 V
3 -1 V
3 0 V
3 -1 V
3 0 V
3 -1 V
3 0 V
3 -1 V
3 0 V
3 -1 V
3 0 V
3 -1 V
3 0 V
3 -1 V
3 0 V
stroke 2098 3917 M
3 -1 V
3 0 V
3 -1 V
3 0 V
3 0 V
3 -1 V
3 0 V
3 0 V
3 -1 V
3 0 V
3 0 V
3 -1 V
3 0 V
3 0 V
2 -1 V
3 0 V
3 0 V
3 0 V
3 0 V
3 -1 V
3 0 V
3 0 V
3 0 V
3 0 V
3 0 V
3 -1 V
3 0 V
3 0 V
3 0 V
3 0 V
3 0 V
3 0 V
3 0 V
3 0 V
3 0 V
3 0 V
3 0 V
3 0 V
3 0 V
3 0 V
3 0 V
3 0 V
3 0 V
3 1 V
3 0 V
3 0 V
3 0 V
3 0 V
3 0 V
3 0 V
3 1 V
3 0 V
3 0 V
3 0 V
3 1 V
3 0 V
3 0 V
3 0 V
3 1 V
3 0 V
3 0 V
3 1 V
3 0 V
3 0 V
3 1 V
3 0 V
3 0 V
3 1 V
3 0 V
3 1 V
3 0 V
3 0 V
3 1 V
3 0 V
3 1 V
3 0 V
3 1 V
3 0 V
3 1 V
3 0 V
3 1 V
3 0 V
3 1 V
3 0 V
3 1 V
3 0 V
3 1 V
3 0 V
3 1 V
3 0 V
3 1 V
3 0 V
3 1 V
3 0 V
3 1 V
3 0 V
3 1 V
3 1 V
3 0 V
3 1 V
3 0 V
3 1 V
3 0 V
3 1 V
stroke 2409 3934 M
3 1 V
3 0 V
3 1 V
3 0 V
3 1 V
2 0 V
3 1 V
3 1 V
3 0 V
3 1 V
3 0 V
3 1 V
3 0 V
3 1 V
3 0 V
3 1 V
3 1 V
3 0 V
3 1 V
3 0 V
3 1 V
3 0 V
3 1 V
3 0 V
3 1 V
3 0 V
3 1 V
3 0 V
3 1 V
3 0 V
3 1 V
3 0 V
3 1 V
3 0 V
3 0 V
3 1 V
3 0 V
3 1 V
3 0 V
3 1 V
3 0 V
3 0 V
3 1 V
3 0 V
3 0 V
3 1 V
3 0 V
3 0 V
3 1 V
3 0 V
3 0 V
3 0 V
3 1 V
3 0 V
3 0 V
3 0 V
3 1 V
3 0 V
3 0 V
3 0 V
3 0 V
3 0 V
3 0 V
3 1 V
3 0 V
3 0 V
3 0 V
3 0 V
3 0 V
3 0 V
3 0 V
3 0 V
3 0 V
3 0 V
3 0 V
3 0 V
3 0 V
3 -1 V
3 0 V
3 0 V
3 0 V
3 0 V
3 0 V
3 -1 V
3 0 V
3 0 V
3 0 V
3 -1 V
3 0 V
3 0 V
3 0 V
3 -1 V
3 0 V
3 0 V
3 -1 V
3 0 V
3 0 V
3 -1 V
3 0 V
3 -1 V
3 0 V
2 -1 V
3 0 V
3 -1 V
stroke 2719 3952 M
3 0 V
3 -1 V
3 0 V
3 -1 V
3 0 V
3 -1 V
3 0 V
3 -1 V
3 -1 V
3 0 V
3 -1 V
3 -1 V
3 0 V
3 -1 V
3 -1 V
3 0 V
3 -1 V
3 -1 V
3 0 V
3 -1 V
3 -1 V
3 -1 V
3 -1 V
3 0 V
3 -1 V
3 -1 V
3 -1 V
3 -1 V
3 -1 V
3 0 V
3 -1 V
3 -1 V
3 -1 V
3 -1 V
3 -1 V
3 -1 V
3 -1 V
3 -1 V
3 -1 V
3 -1 V
3 -1 V
3 -1 V
3 -1 V
3 -1 V
3 -1 V
3 -2 V
3 -1 V
3 -1 V
3 -1 V
3 -1 V
3 -1 V
3 -2 V
3 -1 V
3 -1 V
3 -1 V
3 -2 V
3 -1 V
3 -1 V
3 -1 V
3 -2 V
3 -1 V
3 -1 V
3 -2 V
3 -1 V
3 -2 V
3 -1 V
3 -1 V
3 -2 V
3 -1 V
3 -2 V
3 -1 V
3 -2 V
3 -1 V
3 -2 V
3 -1 V
3 -2 V
3 -2 V
3 -1 V
3 -2 V
3 -1 V
3 -2 V
3 -2 V
3 -1 V
3 -2 V
3 -2 V
3 -1 V
3 -2 V
3 -2 V
3 -2 V
3 -1 V
3 -2 V
3 -2 V
2 -2 V
3 -1 V
3 -2 V
3 -2 V
3 -2 V
3 -2 V
3 -2 V
3 -1 V
3 -2 V
3 -2 V
3 -2 V
3 -2 V
stroke 3030 3826 M
3 -2 V
3 -2 V
3 -2 V
3 -2 V
3 -1 V
3 -2 V
3 -2 V
3 -2 V
3 -2 V
3 -2 V
3 -2 V
3 -2 V
3 -2 V
3 -2 V
3 -2 V
3 -2 V
3 -2 V
3 -2 V
3 -2 V
3 -2 V
3 -2 V
3 -3 V
3 -2 V
3 -2 V
3 -2 V
3 -2 V
3 -2 V
3 -2 V
3 -2 V
3 -2 V
3 -2 V
3 -2 V
3 -3 V
3 -2 V
3 -2 V
3 -2 V
3 -2 V
3 -2 V
3 -3 V
3 -2 V
3 -2 V
3 -2 V
3 -2 V
3 -2 V
3 -3 V
3 -2 V
3 -2 V
3 -2 V
3 -2 V
3 -3 V
3 -2 V
3 -2 V
3 -2 V
3 -3 V
3 -2 V
3 -2 V
3 -3 V
3 -2 V
3 -2 V
3 -2 V
3 -3 V
3 -2 V
3 -2 V
3 -3 V
3 -2 V
3 -2 V
3 -3 V
3 -2 V
3 -2 V
3 -3 V
3 -2 V
3 -2 V
3 -3 V
3 -2 V
3 -3 V
3 -2 V
3 -2 V
3 -3 V
3 -2 V
3 -3 V
3 -2 V
3 -2 V
3 -3 V
2 -2 V
3 -3 V
3 -2 V
3 -3 V
3 -2 V
3 -3 V
3 -2 V
3 -3 V
3 -2 V
3 -3 V
3 -2 V
3 -3 V
3 -2 V
3 -3 V
3 -2 V
3 -3 V
3 -3 V
3 -2 V
3 -3 V
3 -2 V
3 -3 V
stroke 3341 3592 M
3 -2 V
3 -3 V
3 -3 V
3 -2 V
3 -3 V
3 -3 V
3 -2 V
3 -3 V
3 -3 V
3 -2 V
3 -3 V
3 -3 V
3 -2 V
3 -3 V
3 -3 V
3 -2 V
3 -3 V
3 -3 V
3 -3 V
3 -2 V
3 -3 V
3 -3 V
3 -3 V
3 -2 V
3 -3 V
3 -3 V
3 -3 V
3 -2 V
3 -3 V
3 -3 V
3 -3 V
3 -3 V
3 -2 V
3 -3 V
3 -3 V
3 -3 V
3 -3 V
3 -3 V
3 -2 V
3 -3 V
3 -3 V
3 -3 V
3 -3 V
3 -3 V
3 -2 V
3 -3 V
3 -3 V
3 -3 V
3 -3 V
3 -3 V
3 -2 V
3 -3 V
3 -3 V
3 -3 V
3 -3 V
3 -3 V
3 -2 V
3 -3 V
3 -3 V
3 -3 V
3 -2 V
3 -3 V
3 -3 V
3 -3 V
3 -2 V
3 -3 V
3 -3 V
3 -2 V
3 -3 V
3 -2 V
3 -3 V
3 -2 V
3 -3 V
3 -2 V
2 -3 V
3 -2 V
3 -3 V
3 -2 V
3 -2 V
3 -2 V
3 -3 V
3 -2 V
3 -2 V
3 -2 V
3 -2 V
3 -2 V
3 -2 V
3 -2 V
3 -2 V
3 -1 V
3 -2 V
3 -2 V
3 -1 V
3 -2 V
3 -2 V
3 -1 V
3 -1 V
3 -2 V
3 -1 V
3 -1 V
3 -1 V
3 -1 V
3 -1 V
3 -1 V
stroke 3652 3337 M
3 -1 V
3 -1 V
3 -1 V
3 -1 V
3 0 V
3 -1 V
3 -1 V
3 0 V
3 0 V
3 -1 V
3 0 V
3 0 V
3 0 V
3 0 V
3 0 V
3 0 V
3 0 V
3 0 V
3 1 V
3 0 V
3 0 V
3 1 V
3 1 V
3 0 V
3 1 V
3 1 V
3 1 V
3 1 V
3 1 V
3 1 V
3 1 V
3 1 V
3 2 V
3 1 V
3 1 V
3 2 V
3 1 V
3 2 V
3 2 V
3 1 V
3 2 V
3 2 V
3 2 V
3 2 V
3 2 V
3 2 V
3 2 V
3 2 V
3 2 V
3 2 V
3 2 V
3 3 V
3 2 V
3 2 V
3 3 V
3 2 V
3 2 V
3 3 V
3 2 V
3 3 V
3 2 V
3 3 V
3 2 V
3 3 V
3 2 V
3 3 V
2 3 V
3 2 V
3 3 V
3 3 V
3 2 V
3 3 V
3 3 V
3 2 V
3 3 V
3 3 V
3 2 V
3 3 V
3 3 V
3 3 V
3 2 V
3 3 V
3 3 V
3 3 V
3 2 V
3 3 V
3 3 V
3 3 V
3 3 V
3 2 V
3 3 V
3 3 V
3 3 V
3 3 V
3 2 V
3 3 V
3 3 V
3 3 V
3 3 V
3 3 V
3 2 V
3 3 V
3 3 V
3 3 V
stroke 3963 3517 M
3 3 V
3 3 V
3 2 V
3 3 V
3 3 V
3 3 V
3 3 V
3 3 V
3 3 V
3 2 V
3 3 V
3 3 V
3 3 V
3 3 V
3 3 V
3 3 V
3 3 V
3 2 V
3 3 V
3 3 V
3 3 V
3 3 V
3 3 V
3 3 V
3 3 V
3 2 V
3 3 V
3 3 V
3 3 V
3 3 V
3 3 V
3 3 V
3 3 V
3 3 V
3 3 V
3 2 V
3 3 V
3 3 V
3 3 V
3 3 V
3 3 V
3 3 V
3 3 V
3 3 V
3 3 V
3 3 V
3 3 V
3 3 V
3 3 V
3 3 V
3 3 V
3 3 V
3 3 V
3 3 V
3 3 V
3 3 V
3 3 V
2 3 V
3 3 V
3 3 V
3 3 V
3 3 V
3 3 V
3 3 V
3 3 V
3 3 V
3 3 V
3 3 V
3 3 V
3 3 V
3 3 V
3 3 V
3 3 V
3 3 V
3 3 V
3 4 V
3 3 V
3 3 V
3 3 V
3 3 V
3 3 V
3 3 V
3 3 V
3 3 V
3 3 V
3 4 V
3 3 V
3 3 V
3 3 V
3 3 V
3 3 V
3 3 V
3 3 V
3 4 V
3 3 V
3 3 V
3 3 V
3 3 V
3 3 V
3 3 V
3 3 V
3 3 V
3 4 V
3 3 V
stroke 4274 3828 M
3 3 V
3 3 V
3 3 V
3 3 V
3 3 V
3 3 V
3 3 V
3 3 V
3 3 V
3 3 V
3 3 V
3 3 V
3 3 V
3 3 V
3 2 V
3 3 V
3 3 V
3 3 V
3 3 V
3 3 V
3 2 V
3 3 V
3 3 V
3 2 V
3 3 V
3 3 V
3 2 V
3 3 V
3 2 V
3 3 V
3 3 V
3 2 V
3 2 V
3 3 V
3 2 V
3 3 V
3 2 V
3 2 V
3 3 V
3 2 V
3 2 V
3 2 V
3 3 V
3 2 V
3 2 V
3 2 V
3 2 V
3 2 V
2 2 V
3 2 V
3 2 V
3 2 V
3 2 V
3 2 V
3 1 V
3 2 V
3 2 V
3 2 V
3 1 V
3 2 V
3 2 V
3 1 V
3 2 V
3 2 V
3 1 V
3 2 V
3 1 V
3 2 V
3 1 V
3 2 V
3 1 V
3 1 V
3 2 V
3 1 V
3 1 V
3 2 V
3 1 V
3 1 V
3 1 V
3 2 V
3 1 V
3 1 V
3 1 V
3 1 V
3 1 V
3 1 V
3 1 V
3 1 V
3 1 V
3 1 V
3 1 V
3 1 V
3 1 V
3 1 V
3 1 V
3 1 V
3 1 V
3 0 V
3 1 V
3 1 V
3 1 V
3 1 V
3 0 V
3 1 V
stroke 4585 4027 M
3 1 V
3 0 V
3 1 V
3 1 V
3 1 V
3 0 V
3 1 V
3 0 V
3 1 V
3 1 V
3 0 V
3 1 V
3 1 V
3 0 V
3 1 V
3 0 V
3 1 V
3 0 V
3 1 V
3 1 V
3 0 V
3 1 V
3 0 V
3 1 V
3 0 V
3 1 V
3 0 V
3 1 V
3 0 V
3 1 V
3 1 V
3 0 V
3 1 V
3 0 V
3 1 V
3 0 V
3 1 V
3 1 V
3 0 V
2 1 V
3 0 V
3 1 V
3 1 V
3 0 V
3 1 V
3 1 V
3 0 V
3 1 V
3 0 V
3 1 V
3 1 V
3 1 V
3 0 V
3 1 V
3 1 V
3 0 V
3 1 V
3 1 V
3 1 V
3 0 V
3 1 V
3 1 V
3 1 V
3 1 V
3 0 V
3 1 V
3 1 V
3 1 V
3 1 V
3 1 V
3 1 V
3 1 V
3 0 V
3 1 V
3 1 V
3 1 V
3 1 V
3 1 V
3 1 V
3 1 V
3 1 V
3 1 V
3 1 V
3 1 V
3 1 V
3 1 V
3 1 V
3 1 V
3 1 V
3 1 V
3 1 V
3 1 V
3 1 V
3 1 V
3 2 V
3 1 V
3 1 V
3 1 V
3 1 V
3 1 V
3 1 V
3 1 V
3 1 V
3 1 V
stroke 4896 4107 M
3 2 V
3 1 V
3 1 V
3 1 V
3 1 V
3 1 V
3 1 V
3 1 V
3 2 V
3 1 V
3 1 V
3 1 V
3 1 V
3 1 V
3 1 V
3 2 V
3 1 V
3 1 V
3 1 V
3 1 V
3 1 V
3 2 V
3 1 V
3 1 V
3 1 V
3 1 V
3 1 V
3 2 V
3 1 V
3 1 V
2 1 V
3 1 V
3 1 V
3 1 V
3 2 V
3 1 V
3 1 V
3 1 V
3 1 V
3 1 V
3 2 V
3 1 V
3 1 V
3 1 V
3 1 V
3 1 V
3 2 V
3 1 V
3 1 V
3 1 V
3 1 V
3 1 V
3 2 V
3 1 V
3 1 V
3 1 V
3 1 V
3 1 V
3 2 V
3 1 V
3 1 V
3 1 V
3 1 V
3 2 V
3 1 V
3 1 V
3 1 V
3 1 V
3 1 V
3 2 V
3 1 V
3 1 V
3 1 V
3 1 V
3 2 V
3 1 V
3 1 V
3 1 V
3 2 V
3 1 V
3 1 V
3 1 V
3 1 V
3 2 V
3 1 V
3 1 V
3 1 V
3 2 V
3 1 V
3 1 V
3 1 V
3 2 V
3 1 V
3 1 V
3 1 V
3 2 V
3 1 V
3 1 V
3 2 V
3 1 V
3 1 V
3 1 V
3 2 V
3 1 V
stroke 5207 4231 M
3 1 V
3 2 V
3 1 V
3 1 V
3 2 V
3 1 V
3 1 V
3 2 V
3 1 V
3 1 V
3 2 V
3 1 V
3 2 V
3 1 V
3 1 V
3 2 V
3 1 V
3 2 V
3 1 V
3 1 V
3 2 V
3 1 V
2 2 V
3 1 V
3 2 V
3 1 V
3 2 V
3 1 V
3 2 V
3 1 V
3 2 V
3 1 V
3 2 V
3 1 V
3 2 V
3 1 V
3 2 V
3 2 V
3 1 V
3 2 V
3 1 V
3 2 V
3 2 V
3 1 V
3 2 V
3 1 V
3 2 V
3 2 V
3 2 V
3 1 V
3 2 V
3 2 V
3 1 V
3 2 V
3 2 V
3 2 V
3 1 V
3 2 V
3 2 V
3 2 V
3 2 V
3 1 V
3 2 V
3 2 V
3 2 V
3 2 V
3 2 V
3 2 V
3 2 V
3 2 V
3 1 V
3 2 V
3 2 V
3 2 V
3 2 V
3 2 V
3 2 V
3 2 V
3 2 V
3 2 V
3 3 V
3 2 V
3 2 V
3 2 V
3 2 V
3 2 V
3 2 V
3 2 V
3 2 V
3 2 V
3 3 V
3 2 V
3 2 V
3 2 V
3 2 V
3 2 V
3 2 V
3 3 V
3 2 V
3 2 V
3 2 V
3 2 V
3 2 V
3 2 V
stroke 5518 4412 M
3 3 V
3 2 V
3 2 V
3 2 V
3 2 V
3 2 V
3 2 V
3 3 V
3 2 V
3 2 V
3 2 V
3 2 V
3 2 V
2 2 V
3 2 V
3 2 V
3 2 V
3 2 V
3 2 V
3 2 V
3 2 V
3 2 V
3 2 V
3 2 V
3 2 V
3 2 V
3 1 V
3 2 V
3 2 V
3 2 V
3 2 V
3 2 V
3 1 V
3 2 V
3 2 V
3 2 V
3 1 V
3 2 V
3 2 V
3 1 V
3 2 V
3 2 V
3 1 V
3 2 V
3 1 V
3 2 V
3 1 V
3 2 V
3 1 V
3 2 V
3 1 V
3 2 V
3 1 V
3 2 V
3 1 V
3 1 V
3 2 V
3 1 V
3 1 V
3 2 V
3 1 V
3 1 V
3 1 V
3 2 V
3 1 V
3 1 V
3 1 V
3 1 V
3 2 V
3 1 V
3 1 V
3 1 V
3 1 V
3 1 V
3 1 V
3 1 V
3 1 V
3 1 V
3 1 V
3 1 V
3 1 V
3 1 V
3 1 V
3 1 V
3 1 V
3 1 V
3 1 V
3 1 V
3 1 V
3 1 V
3 0 V
3 1 V
3 1 V
3 1 V
3 1 V
3 0 V
3 1 V
3 1 V
3 1 V
3 0 V
3 1 V
3 1 V
3 1 V
3 0 V
stroke 5829 4562 M
3 1 V
3 1 V
3 0 V
3 1 V
2 1 V
3 0 V
3 1 V
3 0 V
3 1 V
3 1 V
3 0 V
3 1 V
3 0 V
3 1 V
3 1 V
3 0 V
3 1 V
3 0 V
3 1 V
3 0 V
3 1 V
3 0 V
3 1 V
3 0 V
3 1 V
3 0 V
3 1 V
3 0 V
3 0 V
3 1 V
3 0 V
3 1 V
3 0 V
3 0 V
3 1 V
3 0 V
3 1 V
3 0 V
3 0 V
3 1 V
3 0 V
3 0 V
3 1 V
3 0 V
3 0 V
3 1 V
3 0 V
3 0 V
3 1 V
3 0 V
3 0 V
3 1 V
3 0 V
3 0 V
3 0 V
3 1 V
3 0 V
3 0 V
3 0 V
3 1 V
3 0 V
3 0 V
3 0 V
3 1 V
3 0 V
3 0 V
3 0 V
3 1 V
3 0 V
3 0 V
3 0 V
3 0 V
3 1 V
3 0 V
3 0 V
3 0 V
3 0 V
3 0 V
3 1 V
3 0 V
3 0 V
3 0 V
3 0 V
3 0 V
3 0 V
3 1 V
3 0 V
3 0 V
3 0 V
3 0 V
3 0 V
3 0 V
3 1 V
3 0 V
3 0 V
3 0 V
3 0 V
3 0 V
3 0 V
2 0 V
3 0 V
3 0 V
3 1 V
3 0 V
stroke 6139 4596 M
3 0 V
3 0 V
3 0 V
3 0 V
3 0 V
3 0 V
3 0 V
3 0 V
3 0 V
3 0 V
3 0 V
3 0 V
3 1 V
3 0 V
3 0 V
3 0 V
3 0 V
3 0 V
3 0 V
3 0 V
3 0 V
3 0 V
3 0 V
3 0 V
3 0 V
3 0 V
3 0 V
3 0 V
3 0 V
3 0 V
3 0 V
3 0 V
3 0 V
3 0 V
3 1 V
3 0 V
3 0 V
3 0 V
3 0 V
3 0 V
3 0 V
3 0 V
3 0 V
3 0 V
3 0 V
3 0 V
3 0 V
3 0 V
3 0 V
3 0 V
3 0 V
3 0 V
3 0 V
3 0 V
3 0 V
3 0 V
3 0 V
3 0 V
3 0 V
3 0 V
3 1 V
3 0 V
3 0 V
3 0 V
3 0 V
3 0 V
3 0 V
3 0 V
3 0 V
3 0 V
3 0 V
3 0 V
3 0 V
3 0 V
3 0 V
3 0 V
3 0 V
3 0 V
3 1 V
3 0 V
3 0 V
3 0 V
3 0 V
3 0 V
3 0 V
3 0 V
3 0 V
3 0 V
3 0 V
3 0 V
3 1 V
2 0 V
3 0 V
3 0 V
3 0 V
3 0 V
3 0 V
3 0 V
3 0 V
3 0 V
3 1 V
3 0 V
3 0 V
3 0 V
stroke 6450 4602 M
3 0 V
3 0 V
3 0 V
3 0 V
3 1 V
3 0 V
3 0 V
3 0 V
3 0 V
3 0 V
3 0 V
3 0 V
3 1 V
3 0 V
3 0 V
3 0 V
3 0 V
3 0 V
3 0 V
3 1 V
3 0 V
3 0 V
3 0 V
3 0 V
3 0 V
3 1 V
3 0 V
3 0 V
3 0 V
3 0 V
3 0 V
3 0 V
3 1 V
3 0 V
3 0 V
3 0 V
3 0 V
3 0 V
3 1 V
3 0 V
3 0 V
3 0 V
3 0 V
3 0 V
3 0 V
3 1 V
3 0 V
3 0 V
3 0 V
3 0 V
3 0 V
3 0 V
3 1 V
3 0 V
3 0 V
3 0 V
3 0 V
3 0 V
3 0 V
3 0 V
3 0 V
3 1 V
3 0 V
3 0 V
3 0 V
3 0 V
3 0 V
3 0 V
3 0 V
3 0 V
3 0 V
3 0 V
3 0 V
3 0 V
3 0 V
3 0 V
3 0 V
3 0 V
3 0 V
3 0 V
3 0 V
3 0 V
2 0 V
3 0 V
3 0 V
3 0 V
3 0 V
3 0 V
3 0 V
3 0 V
3 0 V
3 0 V
3 0 V
3 0 V
3 -1 V
3 0 V
3 0 V
3 0 V
3 0 V
3 0 V
3 -1 V
3 0 V
3 0 V
3 0 V
stroke 6761 4609 M
3 0 V
3 -1 V
3 0 V
3 0 V
3 -1 V
3 0 V
3 0 V
3 0 V
3 -1 V
3 0 V
3 0 V
3 -1 V
3 0 V
3 -1 V
3 0 V
3 0 V
3 -1 V
3 0 V
3 -1 V
3 0 V
3 -1 V
3 0 V
3 -1 V
3 0 V
3 -1 V
stroke
1.000 UL
LTb
LCb setrgbcolor
860 4799 N
860 640 L
5979 0 V
0 4159 V
-5979 0 V
Z stroke
1.000 UP
1.000 UL
LTb
LCb setrgbcolor
stroke
grestore
end
showpage
  }}%
  \put(5936,2619){\makebox(0,0)[r]{\strut{}Without}}%
  \put(5936,2819){\makebox(0,0)[r]{\strut{}With Joint Limit Avoidance}}%
  \put(3849,140){\makebox(0,0){\strut{}Time (seconds)}}%
  \put(160,2719){%
  \special{ps: gsave currentpoint currentpoint translate
630 rotate neg exch neg exch translate}%
  \makebox(0,0){$\|\boldsymbol{\theta}\|$}%
  \special{ps: currentpoint grestore moveto}%
  }%
  \put(6839,440){\makebox(0,0){\strut{} 2000}}%
  \put(6241,440){\makebox(0,0){\strut{} 1800}}%
  \put(5643,440){\makebox(0,0){\strut{} 1600}}%
  \put(5045,440){\makebox(0,0){\strut{} 1400}}%
  \put(4447,440){\makebox(0,0){\strut{} 1200}}%
  \put(3850,440){\makebox(0,0){\strut{} 1000}}%
  \put(3252,440){\makebox(0,0){\strut{} 800}}%
  \put(2654,440){\makebox(0,0){\strut{} 600}}%
  \put(2056,440){\makebox(0,0){\strut{} 400}}%
  \put(1458,440){\makebox(0,0){\strut{} 200}}%
  \put(860,440){\makebox(0,0){\strut{} 0}}%
  \put(740,4799){\makebox(0,0)[r]{\strut{} 10}}%
  \put(740,4383){\makebox(0,0)[r]{\strut{} 9}}%
  \put(740,3967){\makebox(0,0)[r]{\strut{} 8}}%
  \put(740,3551){\makebox(0,0)[r]{\strut{} 7}}%
  \put(740,3135){\makebox(0,0)[r]{\strut{} 6}}%
  \put(740,2720){\makebox(0,0)[r]{\strut{} 5}}%
  \put(740,2304){\makebox(0,0)[r]{\strut{} 4}}%
  \put(740,1888){\makebox(0,0)[r]{\strut{} 3}}%
  \put(740,1472){\makebox(0,0)[r]{\strut{} 2}}%
  \put(740,1056){\makebox(0,0)[r]{\strut{} 1}}%
  \put(740,640){\makebox(0,0)[r]{\strut{} 0}}%
\end{picture}%
\endgroup
 

%% file: fig/bincfg.tex
\begingroup%
\makeatletter%
\newcommand{\GNUPLOTspecial}{%
  \@sanitize\catcode`\%=14\relax\special}%
\setlength{\unitlength}{0.0500bp}%
\begin{picture}(7200,5040)(0,0)%
  {\GNUPLOTspecial{"
/gnudict 256 dict def
gnudict begin
%
%
/Color true def
/Blacktext true def
/Solid false def
/Dashlength 1 def
/Landscape false def
/Level1 false def
/Level3 false def
/Rounded false def
/ClipToBoundingBox false def
/SuppressPDFMark false def
/TransparentPatterns false def
/gnulinewidth 10.000 def
/userlinewidth gnulinewidth def
/Gamma 1.0 def
/BackgroundColor {-1.000 -1.000 -1.000} def
/vshift -66 def
/dl1 {
  10.0 Dashlength userlinewidth gnulinewidth div mul mul mul
  Rounded { currentlinewidth 0.75 mul sub dup 0 le { pop 0.01 } if } if
} def
/dl2 {
  10.0 Dashlength userlinewidth gnulinewidth div mul mul mul
  Rounded { currentlinewidth 0.75 mul add } if
} def
/hpt_ 31.5 def
/vpt_ 31.5 def
/hpt hpt_ def
/vpt vpt_ def
/doclip {
  ClipToBoundingBox {
    newpath 0 0 moveto 360 0 lineto 360 252 lineto 0 252 lineto closepath
    clip
  } if
} def
%
%
%
/M {moveto} bind def
/L {lineto} bind def
/R {rmoveto} bind def
/V {rlineto} bind def
/N {newpath moveto} bind def
/Z {closepath} bind def
/C {setrgbcolor} bind def
/f {rlineto fill} bind def
/g {setgray} bind def
/Gshow {show} def   
/vpt2 vpt 2 mul def
/hpt2 hpt 2 mul def
/Lshow {currentpoint stroke M 0 vshift R 
	Blacktext {gsave 0 setgray textshow grestore} {textshow} ifelse} def
/Rshow {currentpoint stroke M dup stringwidth pop neg vshift R
	Blacktext {gsave 0 setgray textshow grestore} {textshow} ifelse} def
/Cshow {currentpoint stroke M dup stringwidth pop -2 div vshift R 
	Blacktext {gsave 0 setgray textshow grestore} {textshow} ifelse} def
/UP {dup vpt_ mul /vpt exch def hpt_ mul /hpt exch def
  /hpt2 hpt 2 mul def /vpt2 vpt 2 mul def} def
/DL {Color {setrgbcolor Solid {pop []} if 0 setdash}
 {pop pop pop 0 setgray Solid {pop []} if 0 setdash} ifelse} def
/BL {stroke userlinewidth 2 mul setlinewidth
	Rounded {1 setlinejoin 1 setlinecap} if} def
/AL {stroke userlinewidth 2 div setlinewidth
	Rounded {1 setlinejoin 1 setlinecap} if} def
/UL {dup gnulinewidth mul /userlinewidth exch def
	dup 1 lt {pop 1} if 10 mul /udl exch def} def
/PL {stroke userlinewidth setlinewidth
	Rounded {1 setlinejoin 1 setlinecap} if} def
3.8 setmiterlimit
/LCw {1 1 1} def
/LCb {0 0 0} def
/LCa {0 0 0} def
/LC0 {1 0 0} def
/LC1 {0 1 0} def
/LC2 {0 0 1} def
/LC3 {1 0 1} def
/LC4 {0 1 1} def
/LC5 {1 1 0} def
/LC6 {0 0 0} def
/LC7 {1 0.3 0} def
/LC8 {0.5 0.5 0.5} def
/LTB {BL [] LCb DL} def
/LTw {PL [] 1 setgray} def
/LTb {PL [] LCb DL} def
/LTa {AL [1 udl mul 2 udl mul] 0 setdash LCa setrgbcolor} def
/LT0 {PL [] LC0 DL} def
/LT1 {PL [2 dl1 3 dl2] LC1 DL} def
/LT2 {PL [1 dl1 1.5 dl2] LC2 DL} def
/LT3 {PL [6 dl1 2 dl2 1 dl1 2 dl2] LC3 DL} def
/LT4 {PL [1 dl1 2 dl2 6 dl1 2 dl2 1 dl1 2 dl2] LC4 DL} def
/LT5 {PL [4 dl1 2 dl2] LC5 DL} def
/LT6 {PL [1.5 dl1 1.5 dl2 1.5 dl1 1.5 dl2 1.5 dl1 6 dl2] LC6 DL} def
/LT7 {PL [3 dl1 3 dl2 1 dl1 3 dl2] LC7 DL} def
/LT8 {PL [2 dl1 2 dl2 2 dl1 6 dl2] LC8 DL} def
/SL {[] 0 setdash} def
/Pnt {stroke [] 0 setdash gsave 1 setlinecap M 0 0 V stroke grestore} def
/Dia {stroke [] 0 setdash 2 copy vpt add M
  hpt neg vpt neg V hpt vpt neg V
  hpt vpt V hpt neg vpt V closepath stroke
  Pnt} def
/Pls {stroke [] 0 setdash vpt sub M 0 vpt2 V
  currentpoint stroke M
  hpt neg vpt neg R hpt2 0 V stroke
 } def
/Box {stroke [] 0 setdash 2 copy exch hpt sub exch vpt add M
  0 vpt2 neg V hpt2 0 V 0 vpt2 V
  hpt2 neg 0 V closepath stroke
  Pnt} def
/Crs {stroke [] 0 setdash exch hpt sub exch vpt add M
  hpt2 vpt2 neg V currentpoint stroke M
  hpt2 neg 0 R hpt2 vpt2 V stroke} def
/TriU {stroke [] 0 setdash 2 copy vpt 1.12 mul add M
  hpt neg vpt -1.62 mul V
  hpt 2 mul 0 V
  hpt neg vpt 1.62 mul V closepath stroke
  Pnt} def
/Star {2 copy Pls Crs} def
/BoxF {stroke [] 0 setdash exch hpt sub exch vpt add M
  0 vpt2 neg V hpt2 0 V 0 vpt2 V
  hpt2 neg 0 V closepath fill} def
/TriUF {stroke [] 0 setdash vpt 1.12 mul add M
  hpt neg vpt -1.62 mul V
  hpt 2 mul 0 V
  hpt neg vpt 1.62 mul V closepath fill} def
/TriD {stroke [] 0 setdash 2 copy vpt 1.12 mul sub M
  hpt neg vpt 1.62 mul V
  hpt 2 mul 0 V
  hpt neg vpt -1.62 mul V closepath stroke
  Pnt} def
/TriDF {stroke [] 0 setdash vpt 1.12 mul sub M
  hpt neg vpt 1.62 mul V
  hpt 2 mul 0 V
  hpt neg vpt -1.62 mul V closepath fill} def
/DiaF {stroke [] 0 setdash vpt add M
  hpt neg vpt neg V hpt vpt neg V
  hpt vpt V hpt neg vpt V closepath fill} def
/Pent {stroke [] 0 setdash 2 copy gsave
  translate 0 hpt M 4 {72 rotate 0 hpt L} repeat
  closepath stroke grestore Pnt} def
/PentF {stroke [] 0 setdash gsave
  translate 0 hpt M 4 {72 rotate 0 hpt L} repeat
  closepath fill grestore} def
/Circle {stroke [] 0 setdash 2 copy
  hpt 0 360 arc stroke Pnt} def
/CircleF {stroke [] 0 setdash hpt 0 360 arc fill} def
/C0 {BL [] 0 setdash 2 copy moveto vpt 90 450 arc} bind def
/C1 {BL [] 0 setdash 2 copy moveto
	2 copy vpt 0 90 arc closepath fill
	vpt 0 360 arc closepath} bind def
/C2 {BL [] 0 setdash 2 copy moveto
	2 copy vpt 90 180 arc closepath fill
	vpt 0 360 arc closepath} bind def
/C3 {BL [] 0 setdash 2 copy moveto
	2 copy vpt 0 180 arc closepath fill
	vpt 0 360 arc closepath} bind def
/C4 {BL [] 0 setdash 2 copy moveto
	2 copy vpt 180 270 arc closepath fill
	vpt 0 360 arc closepath} bind def
/C5 {BL [] 0 setdash 2 copy moveto
	2 copy vpt 0 90 arc
	2 copy moveto
	2 copy vpt 180 270 arc closepath fill
	vpt 0 360 arc} bind def
/C6 {BL [] 0 setdash 2 copy moveto
	2 copy vpt 90 270 arc closepath fill
	vpt 0 360 arc closepath} bind def
/C7 {BL [] 0 setdash 2 copy moveto
	2 copy vpt 0 270 arc closepath fill
	vpt 0 360 arc closepath} bind def
/C8 {BL [] 0 setdash 2 copy moveto
	2 copy vpt 270 360 arc closepath fill
	vpt 0 360 arc closepath} bind def
/C9 {BL [] 0 setdash 2 copy moveto
	2 copy vpt 270 450 arc closepath fill
	vpt 0 360 arc closepath} bind def
/C10 {BL [] 0 setdash 2 copy 2 copy moveto vpt 270 360 arc closepath fill
	2 copy moveto
	2 copy vpt 90 180 arc closepath fill
	vpt 0 360 arc closepath} bind def
/C11 {BL [] 0 setdash 2 copy moveto
	2 copy vpt 0 180 arc closepath fill
	2 copy moveto
	2 copy vpt 270 360 arc closepath fill
	vpt 0 360 arc closepath} bind def
/C12 {BL [] 0 setdash 2 copy moveto
	2 copy vpt 180 360 arc closepath fill
	vpt 0 360 arc closepath} bind def
/C13 {BL [] 0 setdash 2 copy moveto
	2 copy vpt 0 90 arc closepath fill
	2 copy moveto
	2 copy vpt 180 360 arc closepath fill
	vpt 0 360 arc closepath} bind def
/C14 {BL [] 0 setdash 2 copy moveto
	2 copy vpt 90 360 arc closepath fill
	vpt 0 360 arc} bind def
/C15 {BL [] 0 setdash 2 copy vpt 0 360 arc closepath fill
	vpt 0 360 arc closepath} bind def
/Rec {newpath 4 2 roll moveto 1 index 0 rlineto 0 exch rlineto
	neg 0 rlineto closepath} bind def
/Square {dup Rec} bind def
/Bsquare {vpt sub exch vpt sub exch vpt2 Square} bind def
/S0 {BL [] 0 setdash 2 copy moveto 0 vpt rlineto BL Bsquare} bind def
/S1 {BL [] 0 setdash 2 copy vpt Square fill Bsquare} bind def
/S2 {BL [] 0 setdash 2 copy exch vpt sub exch vpt Square fill Bsquare} bind def
/S3 {BL [] 0 setdash 2 copy exch vpt sub exch vpt2 vpt Rec fill Bsquare} bind def
/S4 {BL [] 0 setdash 2 copy exch vpt sub exch vpt sub vpt Square fill Bsquare} bind def
/S5 {BL [] 0 setdash 2 copy 2 copy vpt Square fill
	exch vpt sub exch vpt sub vpt Square fill Bsquare} bind def
/S6 {BL [] 0 setdash 2 copy exch vpt sub exch vpt sub vpt vpt2 Rec fill Bsquare} bind def
/S7 {BL [] 0 setdash 2 copy exch vpt sub exch vpt sub vpt vpt2 Rec fill
	2 copy vpt Square fill Bsquare} bind def
/S8 {BL [] 0 setdash 2 copy vpt sub vpt Square fill Bsquare} bind def
/S9 {BL [] 0 setdash 2 copy vpt sub vpt vpt2 Rec fill Bsquare} bind def
/S10 {BL [] 0 setdash 2 copy vpt sub vpt Square fill 2 copy exch vpt sub exch vpt Square fill
	Bsquare} bind def
/S11 {BL [] 0 setdash 2 copy vpt sub vpt Square fill 2 copy exch vpt sub exch vpt2 vpt Rec fill
	Bsquare} bind def
/S12 {BL [] 0 setdash 2 copy exch vpt sub exch vpt sub vpt2 vpt Rec fill Bsquare} bind def
/S13 {BL [] 0 setdash 2 copy exch vpt sub exch vpt sub vpt2 vpt Rec fill
	2 copy vpt Square fill Bsquare} bind def
/S14 {BL [] 0 setdash 2 copy exch vpt sub exch vpt sub vpt2 vpt Rec fill
	2 copy exch vpt sub exch vpt Square fill Bsquare} bind def
/S15 {BL [] 0 setdash 2 copy Bsquare fill Bsquare} bind def
/D0 {gsave translate 45 rotate 0 0 S0 stroke grestore} bind def
/D1 {gsave translate 45 rotate 0 0 S1 stroke grestore} bind def
/D2 {gsave translate 45 rotate 0 0 S2 stroke grestore} bind def
/D3 {gsave translate 45 rotate 0 0 S3 stroke grestore} bind def
/D4 {gsave translate 45 rotate 0 0 S4 stroke grestore} bind def
/D5 {gsave translate 45 rotate 0 0 S5 stroke grestore} bind def
/D6 {gsave translate 45 rotate 0 0 S6 stroke grestore} bind def
/D7 {gsave translate 45 rotate 0 0 S7 stroke grestore} bind def
/D8 {gsave translate 45 rotate 0 0 S8 stroke grestore} bind def
/D9 {gsave translate 45 rotate 0 0 S9 stroke grestore} bind def
/D10 {gsave translate 45 rotate 0 0 S10 stroke grestore} bind def
/D11 {gsave translate 45 rotate 0 0 S11 stroke grestore} bind def
/D12 {gsave translate 45 rotate 0 0 S12 stroke grestore} bind def
/D13 {gsave translate 45 rotate 0 0 S13 stroke grestore} bind def
/D14 {gsave translate 45 rotate 0 0 S14 stroke grestore} bind def
/D15 {gsave translate 45 rotate 0 0 S15 stroke grestore} bind def
/DiaE {stroke [] 0 setdash vpt add M
  hpt neg vpt neg V hpt vpt neg V
  hpt vpt V hpt neg vpt V closepath stroke} def
/BoxE {stroke [] 0 setdash exch hpt sub exch vpt add M
  0 vpt2 neg V hpt2 0 V 0 vpt2 V
  hpt2 neg 0 V closepath stroke} def
/TriUE {stroke [] 0 setdash vpt 1.12 mul add M
  hpt neg vpt -1.62 mul V
  hpt 2 mul 0 V
  hpt neg vpt 1.62 mul V closepath stroke} def
/TriDE {stroke [] 0 setdash vpt 1.12 mul sub M
  hpt neg vpt 1.62 mul V
  hpt 2 mul 0 V
  hpt neg vpt -1.62 mul V closepath stroke} def
/PentE {stroke [] 0 setdash gsave
  translate 0 hpt M 4 {72 rotate 0 hpt L} repeat
  closepath stroke grestore} def
/CircE {stroke [] 0 setdash 
  hpt 0 360 arc stroke} def
/Opaque {gsave closepath 1 setgray fill grestore 0 setgray closepath} def
/DiaW {stroke [] 0 setdash vpt add M
  hpt neg vpt neg V hpt vpt neg V
  hpt vpt V hpt neg vpt V Opaque stroke} def
/BoxW {stroke [] 0 setdash exch hpt sub exch vpt add M
  0 vpt2 neg V hpt2 0 V 0 vpt2 V
  hpt2 neg 0 V Opaque stroke} def
/TriUW {stroke [] 0 setdash vpt 1.12 mul add M
  hpt neg vpt -1.62 mul V
  hpt 2 mul 0 V
  hpt neg vpt 1.62 mul V Opaque stroke} def
/TriDW {stroke [] 0 setdash vpt 1.12 mul sub M
  hpt neg vpt 1.62 mul V
  hpt 2 mul 0 V
  hpt neg vpt -1.62 mul V Opaque stroke} def
/PentW {stroke [] 0 setdash gsave
  translate 0 hpt M 4 {72 rotate 0 hpt L} repeat
  Opaque stroke grestore} def
/CircW {stroke [] 0 setdash 
  hpt 0 360 arc Opaque stroke} def
/BoxFill {gsave Rec 1 setgray fill grestore} def
/Density {
  /Fillden exch def
  currentrgbcolor
  /ColB exch def /ColG exch def /ColR exch def
  /ColR ColR Fillden mul Fillden sub 1 add def
  /ColG ColG Fillden mul Fillden sub 1 add def
  /ColB ColB Fillden mul Fillden sub 1 add def
  ColR ColG ColB setrgbcolor} def
/BoxColFill {gsave Rec PolyFill} def
/PolyFill {gsave Density fill grestore grestore} def
/h {rlineto rlineto rlineto gsave closepath fill grestore} bind def
%
%
/PatternFill {gsave /PFa [ 9 2 roll ] def
  PFa 0 get PFa 2 get 2 div add PFa 1 get PFa 3 get 2 div add translate
  PFa 2 get -2 div PFa 3 get -2 div PFa 2 get PFa 3 get Rec
  TransparentPatterns {} {gsave 1 setgray fill grestore} ifelse
  clip
  currentlinewidth 0.5 mul setlinewidth
  /PFs PFa 2 get dup mul PFa 3 get dup mul add sqrt def
  0 0 M PFa 5 get rotate PFs -2 div dup translate
  0 1 PFs PFa 4 get div 1 add floor cvi
	{PFa 4 get mul 0 M 0 PFs V} for
  0 PFa 6 get ne {
	0 1 PFs PFa 4 get div 1 add floor cvi
	{PFa 4 get mul 0 2 1 roll M PFs 0 V} for
 } if
  stroke grestore} def
/languagelevel where
 {pop languagelevel} {1} ifelse
dup 2 lt
	{/InterpretLevel1 true def
	 /InterpretLevel3 false def}
	{/InterpretLevel1 Level1 def
	 2 gt
	    {/InterpretLevel3 Level3 def}
	    {/InterpretLevel3 false def}
	 ifelse }
 ifelse
%
%
/Level2PatternFill {
/Tile8x8 {/PaintType 2 /PatternType 1 /TilingType 1 /BBox [0 0 8 8] /XStep 8 /YStep 8}
	bind def
/KeepColor {currentrgbcolor [/Pattern /DeviceRGB] setcolorspace} bind def
<< Tile8x8
 /PaintProc {0.5 setlinewidth pop 0 0 M 8 8 L 0 8 M 8 0 L stroke} 
>> matrix makepattern
/Pat1 exch def
<< Tile8x8
 /PaintProc {0.5 setlinewidth pop 0 0 M 8 8 L 0 8 M 8 0 L stroke
	0 4 M 4 8 L 8 4 L 4 0 L 0 4 L stroke}
>> matrix makepattern
/Pat2 exch def
<< Tile8x8
 /PaintProc {0.5 setlinewidth pop 0 0 M 0 8 L
	8 8 L 8 0 L 0 0 L fill}
>> matrix makepattern
/Pat3 exch def
<< Tile8x8
 /PaintProc {0.5 setlinewidth pop -4 8 M 8 -4 L
	0 12 M 12 0 L stroke}
>> matrix makepattern
/Pat4 exch def
<< Tile8x8
 /PaintProc {0.5 setlinewidth pop -4 0 M 8 12 L
	0 -4 M 12 8 L stroke}
>> matrix makepattern
/Pat5 exch def
<< Tile8x8
 /PaintProc {0.5 setlinewidth pop -2 8 M 4 -4 L
	0 12 M 8 -4 L 4 12 M 10 0 L stroke}
>> matrix makepattern
/Pat6 exch def
<< Tile8x8
 /PaintProc {0.5 setlinewidth pop -2 0 M 4 12 L
	0 -4 M 8 12 L 4 -4 M 10 8 L stroke}
>> matrix makepattern
/Pat7 exch def
<< Tile8x8
 /PaintProc {0.5 setlinewidth pop 8 -2 M -4 4 L
	12 0 M -4 8 L 12 4 M 0 10 L stroke}
>> matrix makepattern
/Pat8 exch def
<< Tile8x8
 /PaintProc {0.5 setlinewidth pop 0 -2 M 12 4 L
	-4 0 M 12 8 L -4 4 M 8 10 L stroke}
>> matrix makepattern
/Pat9 exch def
/Pattern1 {PatternBgnd KeepColor Pat1 setpattern} bind def
/Pattern2 {PatternBgnd KeepColor Pat2 setpattern} bind def
/Pattern3 {PatternBgnd KeepColor Pat3 setpattern} bind def
/Pattern4 {PatternBgnd KeepColor Landscape {Pat5} {Pat4} ifelse setpattern} bind def
/Pattern5 {PatternBgnd KeepColor Landscape {Pat4} {Pat5} ifelse setpattern} bind def
/Pattern6 {PatternBgnd KeepColor Landscape {Pat9} {Pat6} ifelse setpattern} bind def
/Pattern7 {PatternBgnd KeepColor Landscape {Pat8} {Pat7} ifelse setpattern} bind def
} def
%
%
%
/PatternBgnd {
  TransparentPatterns {} {gsave 1 setgray fill grestore} ifelse
} def
%
%
/Level1PatternFill {
/Pattern1 {0.250 Density} bind def
/Pattern2 {0.500 Density} bind def
/Pattern3 {0.750 Density} bind def
/Pattern4 {0.125 Density} bind def
/Pattern5 {0.375 Density} bind def
/Pattern6 {0.625 Density} bind def
/Pattern7 {0.875 Density} bind def
} def
%
%
Level1 {Level1PatternFill} {Level2PatternFill} ifelse
/Symbol-Oblique /Symbol findfont [1 0 .167 1 0 0] makefont
dup length dict begin {1 index /FID eq {pop pop} {def} ifelse} forall
currentdict end definefont pop
Level1 SuppressPDFMark or 
{} {
/SDict 10 dict def
systemdict /pdfmark known not {
  userdict /pdfmark systemdict /cleartomark get put
} if
SDict begin [
  /Title (bincfg.tex)
  /Subject (gnuplot plot)
  /Creator (gnuplot 5.0 patchlevel 3)
  /Author (swagat)
  /CreationDate (Wed Dec 14 19:39:23 2016)
  /DOCINFO pdfmark
end
} ifelse
%
%
/InitTextBox { userdict /TBy2 3 -1 roll put userdict /TBx2 3 -1 roll put
           userdict /TBy1 3 -1 roll put userdict /TBx1 3 -1 roll put
	   /Boxing true def } def
/ExtendTextBox { Boxing
    { gsave dup false charpath pathbbox
      dup TBy2 gt {userdict /TBy2 3 -1 roll put} {pop} ifelse
      dup TBx2 gt {userdict /TBx2 3 -1 roll put} {pop} ifelse
      dup TBy1 lt {userdict /TBy1 3 -1 roll put} {pop} ifelse
      dup TBx1 lt {userdict /TBx1 3 -1 roll put} {pop} ifelse
      grestore } if } def
/PopTextBox { newpath TBx1 TBxmargin sub TBy1 TBymargin sub M
               TBx1 TBxmargin sub TBy2 TBymargin add L
	       TBx2 TBxmargin add TBy2 TBymargin add L
	       TBx2 TBxmargin add TBy1 TBymargin sub L closepath } def
/DrawTextBox { PopTextBox stroke /Boxing false def} def
/FillTextBox { gsave PopTextBox 1 1 1 setrgbcolor fill grestore /Boxing false def} def
0 0 0 0 InitTextBox
/TBxmargin 20 def
/TBymargin 20 def
/Boxing false def
/textshow { ExtendTextBox Gshow } def
%
/LTB {BL [] LCb DL} def
/LTb {PL [] LCb DL} def
end
gnudict begin
gsave
doclip
0 0 translate
0.050 0.050 scale
0 setgray
newpath
BackgroundColor 0 lt 3 1 roll 0 lt exch 0 lt or or not {BackgroundColor C 1.000 0 0 7200.00 5040.00 BoxColFill} if
1.000 UL
LTb
LCb setrgbcolor
1.000 UL
LTb
LCb setrgbcolor
962 1440 M
2268 908 V
stroke
LTb
LCb setrgbcolor
6238 1664 M
3230 2348 L
stroke
LTb
LCb setrgbcolor
962 1440 M
0 2136 V
stroke
LTb
LCb setrgbcolor
962 1440 M
59 23 V
stroke
LTb
LCb setrgbcolor
3230 2348 M
-59 -24 V
stroke
LTb
LCb setrgbcolor
1.000 UL
LTb
LCb setrgbcolor
1392 1342 M
59 24 V
stroke
LTb
LCb setrgbcolor
3659 2250 M
-59 -24 V
stroke
LTb
LCb setrgbcolor
1.000 UL
LTb
LCb setrgbcolor
1822 1244 M
59 24 V
stroke
LTb
LCb setrgbcolor
4088 2152 M
-59 -23 V
stroke
LTb
LCb setrgbcolor
1.000 UL
LTb
LCb setrgbcolor
2252 1147 M
59 23 V
stroke
LTb
LCb setrgbcolor
4518 2054 M
-59 -23 V
stroke
LTb
LCb setrgbcolor
1.000 UL
LTb
LCb setrgbcolor
2682 1049 M
59 24 V
stroke
LTb
LCb setrgbcolor
4948 1957 M
-59 -24 V
stroke
LTb
LCb setrgbcolor
1.000 UL
LTb
LCb setrgbcolor
3112 951 M
59 24 V
stroke
LTb
LCb setrgbcolor
5378 1859 M
-59 -24 V
stroke
LTb
LCb setrgbcolor
1.000 UL
LTb
LCb setrgbcolor
3541 854 M
59 23 V
stroke
LTb
LCb setrgbcolor
5808 1761 M
-59 -23 V
stroke
LTb
LCb setrgbcolor
1.000 UL
LTb
LCb setrgbcolor
3970 756 M
59 23 V
stroke
LTb
LCb setrgbcolor
6238 1664 M
-59 -24 V
stroke
LTb
LCb setrgbcolor
1.000 UL
LTb
LCb setrgbcolor
3970 756 M
-50 11 V
stroke
LTb
LCb setrgbcolor
962 1440 M
50 -12 V
stroke
LTb
LCb setrgbcolor
1.000 UL
LTb
LCb setrgbcolor
4222 857 M
-50 11 V
stroke
LTb
LCb setrgbcolor
1214 1541 M
50 -12 V
stroke
LTb
LCb setrgbcolor
1.000 UL
LTb
LCb setrgbcolor
4474 958 M
-50 11 V
stroke
LTb
LCb setrgbcolor
1466 1642 M
50 -12 V
stroke
LTb
LCb setrgbcolor
1.000 UL
LTb
LCb setrgbcolor
4726 1058 M
-50 12 V
stroke
LTb
LCb setrgbcolor
1718 1742 M
50 -11 V
stroke
LTb
LCb setrgbcolor
1.000 UL
LTb
LCb setrgbcolor
4978 1159 M
-50 12 V
stroke
LTb
LCb setrgbcolor
1970 1843 M
50 -11 V
stroke
LTb
LCb setrgbcolor
1.000 UL
LTb
LCb setrgbcolor
5230 1260 M
-50 12 V
stroke
LTb
LCb setrgbcolor
2222 1944 M
50 -11 V
stroke
LTb
LCb setrgbcolor
1.000 UL
LTb
LCb setrgbcolor
5482 1361 M
-50 11 V
stroke
LTb
LCb setrgbcolor
2474 2045 M
50 -11 V
stroke
LTb
LCb setrgbcolor
1.000 UL
LTb
LCb setrgbcolor
5734 1462 M
-50 11 V
stroke
LTb
LCb setrgbcolor
2726 2146 M
50 -12 V
stroke
LTb
LCb setrgbcolor
1.000 UL
LTb
LCb setrgbcolor
5986 1563 M
-50 11 V
stroke
LTb
LCb setrgbcolor
2978 2247 M
50 -12 V
stroke
LTb
LCb setrgbcolor
1.000 UL
LTb
LCb setrgbcolor
6238 1664 M
-50 11 V
stroke
LTb
LCb setrgbcolor
3230 2348 M
50 -12 V
stroke
LTb
LCb setrgbcolor
1.000 UL
LTb
LCb setrgbcolor
962 1440 M
63 0 V
stroke
LTb
LCb setrgbcolor
1.000 UL
LTb
LCb setrgbcolor
962 1677 M
63 0 V
stroke
LTb
LCb setrgbcolor
1.000 UL
LTb
LCb setrgbcolor
962 1915 M
63 0 V
stroke
LTb
LCb setrgbcolor
1.000 UL
LTb
LCb setrgbcolor
962 2152 M
63 0 V
stroke
LTb
LCb setrgbcolor
1.000 UL
LTb
LCb setrgbcolor
962 2390 M
63 0 V
stroke
LTb
LCb setrgbcolor
1.000 UL
LTb
LCb setrgbcolor
962 2626 M
63 0 V
stroke
LTb
LCb setrgbcolor
1.000 UL
LTb
LCb setrgbcolor
962 2864 M
63 0 V
stroke
LTb
LCb setrgbcolor
1.000 UL
LTb
LCb setrgbcolor
962 3101 M
63 0 V
stroke
LTb
LCb setrgbcolor
1.000 UL
LTb
LCb setrgbcolor
962 3339 M
63 0 V
stroke
LTb
LCb setrgbcolor
1.000 UL
LTb
LCb setrgbcolor
962 3576 M
63 0 V
stroke
LTb
LCb setrgbcolor
LCb setrgbcolor
LTb
1.000 UP
LCb setrgbcolor
2.000 UP
1.000 UL
LTb
0.58 0.00 0.83 C 0.58 0.00 0.83 C 3153 4006 Box
2450 3706 Box
1765 3414 Box
3149 3462 Box
2446 3157 Box
1763 2862 Box
3154 2967 Box
2451 2668 Box
1769 2379 Box
3160 2399 Box
2456 2112 Box
1774 1832 Box
0.58 0.00 0.83 C 5783 4576 Box
LCb setrgbcolor
1.000 UP
1.000 UL
LTb
0.00 0.62 0.45 C 0.00 0.62 0.45 C 3187 3974 Crs
2450 3706 Crs
2163 3326 Crs
3149 3462 Crs
2446 3157 Crs
1798 2855 Crs
3154 2967 Crs
2451 2668 Crs
1769 2379 Crs
3160 2399 Crs
2456 2112 Crs
1774 1832 Crs
0.00 0.62 0.45 C 5783 4376 Crs
1.000 UP
2.000 UL
LTb
0.34 0.71 0.91 C 0.34 0.71 0.91 C 4370 2168 M
0 212 V
-759 470 V
-579 742 V
301 102 V
23 212 V
-169 68 V
stroke
0.34 0.71 0.91 C 4370 2168 Star
4370 2380 Star
3611 2850 Star
3032 3592 Star
3333 3694 Star
3356 3906 Star
3187 3974 Star
1.000 UP
2.000 UL
LTb
0.34 0.71 0.91 C 0.34 0.71 0.91 C 4370 2168 M
0 212 V
3235 2674 L
-855 628 V
204 120 V
34 216 V
-168 68 V
stroke
0.34 0.71 0.91 C 4370 2168 Star
4370 2380 Star
3235 2674 Star
2380 3302 Star
2584 3422 Star
2618 3638 Star
2450 3706 Star
1.000 UP
2.000 UL
LTb
0.34 0.71 0.91 C 0.34 0.71 0.91 C 4370 2168 M
0 212 V
-1202 15 V
-686 768 V
2295 3042 L
37 216 V
-169 68 V
stroke
0.34 0.71 0.91 C 4370 2168 Star
4370 2380 Star
3168 2395 Star
2482 3163 Star
2295 3042 Star
2332 3258 Star
2163 3326 Star
1.000 UP
2.000 UL
LTb
0.34 0.71 0.91 C 0.34 0.71 0.91 C 4370 2168 M
0 212 V
-397 122 V
-324 763 V
-344 -79 V
12 208 V
-168 68 V
stroke
0.34 0.71 0.91 C 4370 2168 Star
4370 2380 Star
3973 2502 Star
3649 3265 Star
3305 3186 Star
3317 3394 Star
3149 3462 Star
1.000 UP
2.000 UL
LTb
0.34 0.71 0.91 C 0.34 0.71 0.91 C 4370 2168 M
0 212 V
-983 611 V
2376 2753 L
203 120 V
35 216 V
-168 68 V
stroke
0.34 0.71 0.91 C 4370 2168 Star
4370 2380 Star
3387 2991 Star
2376 2753 Star
2579 2873 Star
2614 3089 Star
2446 3157 Star
1.000 UP
2.000 UL
LTb
0.34 0.71 0.91 C 0.34 0.71 0.91 C 4370 2168 M
0 212 V
3137 2736 L
1918 2886 L
96 121 V
-47 -220 V
-169 68 V
stroke
0.34 0.71 0.91 C 4370 2168 Star
4370 2380 Star
3137 2736 Star
1918 2886 Star
2014 3007 Star
1967 2787 Star
1798 2855 Star
1.000 UP
2.000 UL
LTb
0.34 0.71 0.91 C 0.34 0.71 0.91 C 4370 2168 M
0 212 V
-733 -90 V
-593 718 V
301 102 V
-22 -211 V
-169 68 V
stroke
0.34 0.71 0.91 C 4370 2168 Star
4370 2380 Star
3637 2290 Star
3044 3008 Star
3345 3110 Star
3323 2899 Star
3154 2967 Star
1.000 UP
2.000 UL
LTb
0.34 0.71 0.91 C 0.34 0.71 0.91 C 4370 2168 M
0 212 V
3401 1900 L
2381 2265 L
204 120 V
35 216 V
-169 67 V
stroke
0.34 0.71 0.91 C 4370 2168 Star
4370 2380 Star
3401 1900 Star
2381 2265 Star
2585 2385 Star
2620 2601 Star
2451 2668 Star
1.000 UP
2.000 UL
LTb
0.34 0.71 0.91 C 0.34 0.71 0.91 C 4370 2168 M
0 212 V
3106 2679 L
1887 2410 L
97 121 V
-47 -220 V
-168 68 V
stroke
0.34 0.71 0.91 C 4370 2168 Star
4370 2380 Star
3106 2679 Star
1887 2410 Star
1984 2531 Star
1937 2311 Star
1769 2379 Star
1.000 UP
2.000 UL
LTb
0.34 0.71 0.91 C 0.34 0.71 0.91 C 4370 2168 M
0 212 V
-759 443 V
3048 2440 L
302 103 V
-22 -212 V
-168 68 V
stroke
0.34 0.71 0.91 C 4370 2168 Star
4370 2380 Star
3611 2823 Star
3048 2440 Star
3350 2543 Star
3328 2331 Star
3160 2399 Star
1.000 UP
2.000 UL
LTb
0.34 0.71 0.91 C 0.34 0.71 0.91 C 4370 2168 M
0 212 V
-843 455 V
2933 2364 L
2649 2256 L
-24 -212 V
-169 68 V
stroke
0.34 0.71 0.91 C 4370 2168 Star
4370 2380 Star
3527 2835 Star
2933 2364 Star
2649 2256 Star
2625 2044 Star
2456 2112 Star
1.000 UP
2.000 UL
LTb
0.34 0.71 0.91 C 0.34 0.71 0.91 C 4370 2168 M
0 212 V
3017 2325 L
1891 1864 L
98 121 V
-47 -221 V
-168 68 V
stroke
0.34 0.71 0.91 C 4370 2168 Star
4370 2380 Star
3017 2325 Star
1891 1864 Star
1989 1985 Star
1942 1764 Star
1774 1832 Star
LCb setrgbcolor
1.000 UP
2.000 UL
LTb
0.90 0.62 0.00 C 0.90 0.62 0.00 C 4370 2168 M
0 212 V
721 -821 V
81 915 V
4916 2360 L
-29 224 V
-168 17 V
stroke
0.90 0.62 0.00 C 4370 2168 Box
4370 2380 Box
5091 1559 Box
5172 2474 Box
4916 2360 Box
4887 2584 Box
4719 2601 Box
0.90 0.62 0.00 C 5512 4176 M
543 0 V
stroke
0.90 0.62 0.00 C 5783 4176 Box
1.000 UL
LTb
LCb setrgbcolor
6238 1664 M
3970 756 L
stroke
LTb
LCb setrgbcolor
962 1440 M
3970 756 L
stroke
LCb setrgbcolor
LTb
LCb setrgbcolor
LTb
LCb setrgbcolor
LTb
1.000 UP
stroke
grestore
end
showpage
  }}%
  \put(122,2509){\makebox(0,0){\strut{}Z(m)}}%
  \put(5856,1039){\makebox(0,0){\strut{}Y (m)}}%
  \put(1900,871){\makebox(0,0){\strut{}X (m)}}%
  \put(5392,4176){\makebox(0,0)[r]{\strut{}Init Config}}%
  \put(5392,4376){\makebox(0,0)[r]{\strut{}actual}}%
  \put(5392,4576){\makebox(0,0)[r]{\strut{}target}}%
  \put(122,2509){\makebox(0,0){\strut{}Z(m)}}%
  \put(836,3576){\makebox(0,0)[r]{\strut{}$0.6$}}%
  \put(836,3339){\makebox(0,0)[r]{\strut{}$0.5$}}%
  \put(836,3101){\makebox(0,0)[r]{\strut{}$0.4$}}%
  \put(836,2864){\makebox(0,0)[r]{\strut{}$0.3$}}%
  \put(836,2626){\makebox(0,0)[r]{\strut{}$0.2$}}%
  \put(836,2390){\makebox(0,0)[r]{\strut{}$0.1$}}%
  \put(836,2152){\makebox(0,0)[r]{\strut{}$0$}}%
  \put(836,1915){\makebox(0,0)[r]{\strut{}$-0.1$}}%
  \put(836,1677){\makebox(0,0)[r]{\strut{}$-0.2$}}%
  \put(836,1440){\makebox(0,0)[r]{\strut{}$-0.3$}}%
  \put(6334,1627){\makebox(0,0){\strut{}$0.4$}}%
  \put(6082,1526){\makebox(0,0){\strut{}$0.3$}}%
  \put(5830,1426){\makebox(0,0){\strut{}$0.2$}}%
  \put(5578,1325){\makebox(0,0){\strut{}$0.1$}}%
  \put(5326,1224){\makebox(0,0){\strut{}$0$}}%
  \put(5074,1123){\makebox(0,0){\strut{}$-0.1$}}%
  \put(4822,1022){\makebox(0,0){\strut{}$-0.2$}}%
  \put(4570,921){\makebox(0,0){\strut{}$-0.3$}}%
  \put(4318,820){\makebox(0,0){\strut{}$-0.4$}}%
  \put(4066,720){\makebox(0,0){\strut{}$-0.5$}}%
  \put(3858,681){\makebox(0,0)[r]{\strut{}$0.4$}}%
  \put(3429,779){\makebox(0,0)[r]{\strut{}$0.2$}}%
  \put(2999,876){\makebox(0,0)[r]{\strut{}$0$}}%
  \put(2569,974){\makebox(0,0)[r]{\strut{}$-0.2$}}%
  \put(2139,1072){\makebox(0,0)[r]{\strut{}$-0.4$}}%
  \put(1709,1169){\makebox(0,0)[r]{\strut{}$-0.6$}}%
  \put(1280,1267){\makebox(0,0)[r]{\strut{}$-0.8$}}%
  \put(850,1365){\makebox(0,0)[r]{\strut{}$-1$}}%
\end{picture}%
\endgroup
 